\documentclass[english]{article}
\usepackage[T1]{fontenc}
\usepackage[latin9]{inputenc}
\usepackage{array}
\usepackage{float}
\usepackage{url}
\usepackage{multirow}
\usepackage{amsmath}
\usepackage{amsthm}
\usepackage{amssymb}
\usepackage{graphicx}

\makeatletter

\providecommand{\tabularnewline}{\\}
\newcommand{\lyxdot}{.}

\theoremstyle{plain}
\newtheorem{thm}{\protect\theoremname}
\theoremstyle{definition}
\newtheorem{defn}[thm]{\protect\definitionname}
\theoremstyle{plain}
\newtheorem{fact}[thm]{\protect\factname}
\theoremstyle{plain}
\newtheorem{prop}[thm]{\protect\propositionname}

\usepackage{iclr2020_conference}
\usepackage{hyperref}
\usepackage{grffile}

\usepackage{times}
\usepackage{microtype}
\renewcommand{\cite}{\citep}

\@ifundefined{showcaptionsetup}{}{%
 \PassOptionsToPackage{caption=false}{subfig}}
\usepackage{subfig}
\makeatother

\usepackage{babel}
\providecommand{\definitionname}{Definition}
\providecommand{\factname}{Fact}
\providecommand{\propositionname}{Proposition}
\providecommand{\theoremname}{Theorem}

\begin{document}
\iclrfinalcopy

\title{Theory and Evaluation Metrics for Learning Disentangled Representations}

\author{Kien Do and Truyen Tran\\
Applied AI Institute, Deakin University, Geelong, Australia\\
\texttt{\{k.do,truyen.tran\}@deakin.edu.au}}
\maketitle
\begin{abstract}
We make two theoretical contributions to disentanglement learning
by (a) defining precise semantics of disentangled representations,
and (b) establishing robust metrics for evaluation. First, we characterize
the concept ``disentangled representations'' used in supervised
and unsupervised methods along three dimensions--\emph{informativeness},
\emph{separability} and \emph{interpretability}--which can be expressed
and quantified explicitly using information-theoretic constructs.
This helps explain the behaviors of several well-known disentanglement
learning models. We then propose robust metrics for measuring informativeness,
separability, and interpretability. Through a comprehensive suite
of experiments, we show that our metrics correctly characterize the
representations learned by different methods and are consistent with
qualitative (visual) results. Thus, the metrics allow disentanglement
learning methods to be compared on a fair ground. We also empirically
uncovered new interesting properties of VAE-based methods and interpreted
them with our formulation. These findings are promising and hopefully
will encourage the design of more theoretically driven models for
learning disentangled representations\footnote{Code for our work is avaiable at: \url{https://github.com/clarken92/DisentanglementMetrics}}.

\end{abstract}
\global\long\def\Expect{\mathbb{E}}
\global\long\def\Real{\mathbb{R}}
\global\long\def\Loss{\mathcal{L}}
\global\long\def\Data{\mathcal{D}}
\global\long\def\avg{\text{avg}}
\global\long\def\Expect{\mathbb{E}}
\global\long\def\ELBO{\text{ELBO}}
\global\long\def\VAE{\text{VAE}}
\global\long\def\FactorVAE{\text{FactorVAE}}
\global\long\def\VIB{\text{VIB}}
\global\long\def\IB{\text{IB}}
\global\long\def\Normal{\mathcal{N}}
\global\long\def\adv{\text{adv}}
\global\long\def\factor{\text{f}}
\global\long\def\noise{\text{n}}
\global\long\def\Rec{\text{Rec}}
\global\long\def\Reg{\text{Reg}}
\global\long\def\Adv{\text{Adv}}
\global\long\def\TC{\text{TC}}
\global\long\def\erf{\text{erf}}
\global\long\def\Urm{\mathrm{U}}
\global\long\def\Rrm{\mathrm{R}}
\global\long\def\Irm{\mathrm{I}}
\global\long\def\Lrm{\mathrm{L}}
\global\long\def\Wrm{\mathrm{W}}
\global\long\def\Crm{\mathrm{C}}
\global\long\def\Mrm{\mathrm{M}}
\global\long\def\Scal{\mathcal{S}}
\global\long\def\Ucal{\mathcal{U}}

\section{Introduction}

Disentanglement learning holds the key for understanding the world
from observations, transferring knowledge across different tasks and
domains, generating novel designs, and learning compositional concepts
\cite{bengio2013representation,higgins2017scan,lake2017building,peters2017elements,schmidhuber1992learning}.
Assuming the observation $x$ is generated from latent factors $z$
via $p(x|z)$, the goal of disentanglement learning is to correctly
uncover a set of independent factors $\{z_{i}\}$ that give rise to
the observation. While there has been a considerable progress in recent
years, common assumptions about disentangled representations appear
to be inadequate \cite{locatello2019challenging}.

Unsupervised disentangling methods are highly desirable as they assume
no prior knowledge about the ground truth factors. These methods typically
impose constraints to encourage independence among latent variables.
Examples of constraints include forcing the variational posterior
$q(z|x)$ to be similar to a factorial $p(z)$ \cite{burgess2018understanding,higgins2017beta},
forcing the variational aggregated prior $q(z)$ to be similar to
the prior $p(z)$ \cite{makhzani2015adversarial}, adding total correlation
loss \cite{kim2018disentangling}, forcing the covariance matrix of
$q(z)$ to be close to the identity matrix \cite{kumar2017variational},
and using a kernel-based measure of independence \cite{lopez2018information}.
However, it remains unclear how the independence constraint affects
other properties of representation. Indeed, more independence may
lead to higher reconstruction error in some models \cite{higgins2017beta,kim2018disentangling}.
Worse still, the independent representations may mismatch human's
predefined concepts \cite{locatello2019challenging}. This suggests
that supervised methods -- which associate a representation (or a
group of representations) $z_{i}$ with a particular ground truth
factor $y_{k}$ -- may be more adequate. However, most supervised
methods have only been shown to perform well on toy datasets \cite{harsh2018disentangling,kulkarni2015deep,mathieu2016disentangling}
in which data are generated from multiplicative combination of the
ground truth factors. It is still unclear about their performance
on real datasets.

We believe that there are at least two major reasons for the current
unsatisfying state of disentanglement learning: i) the lack of a formal
notion of disentangled representations to support the design of proper
objective functions \cite{tschannen2018recent,locatello2019challenging},
and ii) the lack of robust evaluation metrics to enable a fair comparison
between models, regardless of their architectures or design purposes.
To that end, we contribute by formally characterizing disentangled
representations along three dimensions, namely \emph{informativeness},
\emph{separability} and \emph{interpretability}, drawing from concepts
in information theory (Section~\ref{sec:Rethinking-Disentanglement}).
We then design robust quantitative metrics for these properties and
argue that an ideal method for disentanglement learning should achieve
high performance on these metrics (Section~\ref{sec:Robust-Evaluation-Metrics}).

We run a series of experiments to demonstrate how to compare different
models using our proposed metrics, showing that the quantitative results
provided by these metrics are consistent with visual results (Section~\ref{sec:Experiments}).
In the process, we gain important insights about some well-known disentanglement
learning methods namely FactorVAE \cite{kim2018disentangling}, $\beta$-VAE
\cite{higgins2017beta}, and AAE \cite{makhzani2015adversarial}.

\section{Rethinking Disentanglement \label{sec:Rethinking-Disentanglement}}

Inspired by \cite{bengio2013representation,ridgeway2016survey}, we
adopt the notion of disentangled representation learning as ``a process\emph{
}of \emph{decorrelating information} in the data into separate \emph{informative}
representations, each of which corresponds to a concept \emph{defined
by humans}''. This suggests three important properties of a disentangled
representation: \emph{informativeness}, \emph{separability} and \emph{interpretability},
which we quantify as follows:

\paragraph{Informativeness}

We formulate the \emph{informativeness} of a particular representation
(or a group of representations) $z_{i}$ w.r.t. the data $x$ as the
mutual information between $z_{i}$ and $x$:
\begin{equation}
I(x,z_{i})=\int_{x}\int_{z}p_{\Data}(x)q(z_{i}|x)\log\frac{q(z_{i}|x)}{q(z_{i})}\ dz\ dx\label{eq:informativeness}
\end{equation}
where $q(z_{i})=\int_{x}p_{\Data}(x)q(z_{i}|x)\ dx$. In order to
represent the data faithfully, a representation $z_{i}$ should be
informative of $x$, meaning $I(x,z_{i})$ should be large. Because
$I(x,z_{i})=H(z_{i})-H(z_{i}|x)$, a large value of $I(x,z_{i})$
means that $H(z_{i}|x)\approx0$ given that $H(z_{i})$ can be chosen
to be relatively fixed. In other words, if $z_{i}$ is informative
w.r.t. $x$, $q(z_{i}|x)$ \emph{usually has small variance}. It is
important to note that $I(x,z_{i})$ in Eq.~\ref{eq:informativeness}
is defined on the variational encoder $q(z_{i}|x)$, and does not
require a decoder. It implies that we do not need to minimize the
reconstruction error over $x$ (e.g., in VAEs) to increase the informativeness
of a particular $z_{i}$.

\paragraph{Separability and Independence}

Two representations $z_{i}$, $z_{j}$ are \emph{separable }w.r.t.
the data $x$ if they do not share common information about $x$,
which can be formulated as follows:
\begin{equation}
I(x,z_{i},z_{j})=0\label{eq:separability}
\end{equation}
where $I(x,z_{i},z_{j})$ denotes the multivariate mutual information
\cite{mcgill1954multivariate} between $x$, $z_{i}$ and $z_{j}$.
$I(x,z_{i},z_{j})$ can be decomposed into standard bivariate mutual
information terms as follows:
\begin{eqnarray*}
I(x,z_{i},z_{j}) & = & I(x,z_{i})+I(x,z_{j})-I(x,(z_{i},z_{j}))=I(z_{i},z_{j})-I(z_{i},z_{j}|x)
\end{eqnarray*}
$I(x,z_{i},z_{j})$ can be either positive or negative. It is positive
if $z_{i}$ and $z_{j}$ contain redundant information about $x$.
The meaning of a negative $I(x,z_{i},z_{j})$ remains elusive \cite{bell2003co}. 

Achieving separability with respect to $x$ does not guarantee that
$z_{i}$ and $z_{j}$ are separable in general. $z_{i}$ and $z_{j}$
are \emph{fully separable} or \emph{statistically independent} if
and only if:
\begin{equation}
I(z_{i},z_{j})=0\label{eq:independence}
\end{equation}
If we have access to all representations $z$, we can generally say
that a representation $z_{i}$ is \emph{fully separable} (from other
representations $z_{\neq i}$) if and only if $I(z_{i},z_{\neq i})=0$.

Note that there is a trade-off between informativeness, independence
and the number of latent variables which we discuss in Appdx.~\ref{subsec:Trade-off-between-informativenes}.

\paragraph{Interpretability}

Obtaining informative and independent representations does not guarantee
interpretability by human \cite{locatello2019challenging}. We argue
that in order to achieve interpretability, we should provide models
with a set of predefined concepts $y$. In this case, a representation
$z_{i}$ is interpretable with respect to $y_{k}$ if it only contains
information about $y_{k}$ (given that $z_{i}$ is separable from
all other $z_{\neq i}$ and all $y_{k}$ are distinct). \emph{Full
interpretability} can be formulated as follows:
\begin{equation}
I(z_{i},y_{k})=H(z_{i})=H(y_{k})\label{eq:full_interpretability}
\end{equation}
Eq.~\ref{eq:full_interpretability} is equivalent to the condition
that $z_{i}$ is an \emph{invertible function} of $y_{k}$. If we
want $z_{i}$ to generalize beyond the observed $y_{k}$ (i.e., $H(z_{i})>H(y_{k})$),
we can change the condition in Eq.~\ref{eq:full_interpretability}
into:
\begin{equation}
I(z_{i},y_{k})=H(y_{k})\ \ \ \text{or}\ \ \ H(y_{k}|z_{i})=0\label{eq:partial_interpretability}
\end{equation}
which suggests that the model should accurately predict $y_{k}$ given
$z_{i}$. If $z_{i}$ satisfies the condition in Eq.~\ref{eq:partial_interpretability},
it is said to be \emph{partially interpretable} w.r.t $y_{k}$.

In real data, underlying factors of variation are usually correlated.
For example, men usually have beard and short hair. Therefore, it
is very difficult to match independent latent variables to different
ground truth factors at the same time. We believe that in order to
achieve good interpretability, we should isolate the factors and learn
one at a time.

\subsection{An information-theoretic definition of disentangled representations\label{subsec:Definition}}

Given a dataset $\Data=\left\{ x_{i}\right\} _{i=1}^{N}$, where each
data point $x$ is associated with a set of $K$ labeled factors of
variation $y=\left\{ y_{1},...,y_{K}\right\} $. Assume that there
exists a mapping from $x$ to $m$ groups of latent representations
$z=\left\{ z_{1},z_{2},...,z_{m}\right\} $ which follows the distribution
$q(z|x)$. Denoting $q(z_{i}|x)=\sum_{z_{\neq i}}q(z|x)$ and $q(z_{i})=\Expect_{p_{\Data}(x)}\left[q(z_{i}|x)\right]$.
We define disentangled representations for\textbf{ unsupervised cases}
as follows:
\begin{defn}
[Unsupervised]A representation or a group of representations $z_{i}$
is said to be \emph{``fully disentangled''} w.r.t a ground truth
factor $y_{k}$ if $z_{i}$ is \emph{fully separable} (from $z_{\neq i}$)
and $z_{i}$ is \emph{fully interpretable} w.r.t $y_{k}$. Mathematically,
this can be written as:
\begin{equation}
I(z_{i},z_{\neq i})=0\ \ \ \text{and\ \ \ }I(z_{i},y_{k})=H(z_{i},y_{k})\label{eq:disentanglement_definition}
\end{equation}

The definition of disentangled representations for \textbf{supervised
cases} is similar as above except that now we model $q(z|x,y)$ instead
of $q(z|x)$ and $q(z)=\sum_{x,y}p_{\Data}(x,y)q(z|x,y)$. 

Recently, there have been several works \cite{eastwood2018framework,higgins2018towards,ridgeway2018learning}
that attempted to define disentangled representations. Higgin et.
al. \cite{higgins2018towards} proposed a definition based on group
theory \cite{cohen2014learning} which is (informally) stated as follows:
``A representation $z$ is disentangled w.r.t a particular subgroup
$y_{k}$ (from a symmetry group $y=\left\{ y_{k}\right\} _{k=1}^{K}$)
if $z$ can be decomposed into different subspaces $\left\{ z_{i}\right\} _{i=1}^{H}$
in which the subspace $z_{i}$ should be independent of all other
representation subspaces $z_{\neq i}$, and $z_{i}$ should only be
affected by the action of a single subgroup $y_{k}$ and not by other
subgroups $y_{\neq k}$.''. Their definition shares similar observation
as ours. However, it is less convenient for designing models and metrics
than our information-theoretic definition.
\end{defn}

Eastwood et. al. \cite{eastwood2018framework} did not provide any
explicit definition of disentangled representations but characterizing
them along three dimensions namely \emph{``disentanglement''}, \emph{``compactness''},
and \emph{``informativeness''} (between $z$ any $y_{k}$). A high
``disentanglement'' score ($\approx1$) for $z_{i}$ indicates that
it captures at most one factor, let's say $y_{k}$. A high ``completeness''
score ($\approx1$) for $y_{k}$ indicates that it is captured by
at most one latent $z_{j}$ and $j$ is likely to be $i$. A high
``informativeness'' score\footnote{In \cite{eastwood2018framework}, the authors consider the prediction
error of $y_{k}$ given $z$ instead. High ``informativeness'' score
means this error should be close to $0$.} for $y_{k}$ indicates that all information of $y_{k}$ is captured
by the representations $z$. Intuitively, when all the three notions
achieve optimal values, there should be only a single representation
$z_{i}$ that captures all information of the factor $y_{k}$ but
no information from other factors $y_{\neq k}$. However, even in
that case, $z_{i}$ is still \emph{not} fully interpretable w.r.t
$y_{k}$ since $z_{i}$ may contain some information in $x$ that
does not appear in $y_{k}$. This makes their notions only applicable
to toy datasets on which we know that the data $x$ are only generated
from predefined ground truth factors $y=\left\{ y_{k}\right\} _{k=1}^{K}$.
Our definition can handle the situation where we only know some but
not all factors of variation in the data. The notions in \cite{ridgeway2018learning}
follow those in \cite{eastwood2018framework}, hence, suffer from
the same disadvantage.

\section{Robust Evaluation Metrics \label{sec:Robust-Evaluation-Metrics}}

We argue that a robust metric for disentanglement should meet the
following criteria: i) it supports both supervised/unsupervised models;
ii) it can be applied for real datasets; iii) it is computationally
straightforward, i.e. not requiring any training procedure; iv) it
provides consistent results across different methods and different
latent representations; and v) it agrees with qualitative (visual)
results. Here we propose information-theoretic metrics to measure
informativeness, independence and interpretability which meet all
of these robustness criteria.

\subsection{Metrics for informativeness\label{subsec:Metrics-for-informativeness}}

We measure the informativeness of a particular representation $z_{i}$
w.r.t. $x$ by computing $I(x,z_{i})$. If $z_{i}$ is discrete, we
can compute $I(x,z_{i})$ exactly by using Eq.~\ref{eq:informativeness}
but with the integral replaced by the sum. If $z_{i}$ is continuous,
we estimate $I(x,z_{i})$ via sampling or quantization. Details about
these estimations are provided in Appdx.~\ref{subsec:Computing-metrics-for}.

If $H(z_{i})$ is estimated via quantization, we will have $0\leq I(x,z_{i})\leq H(z_{i})$.
In this case, we can divide $I(x,z_{i})$ by $H(z_{i})$ to normalize
it to the range {[}0, 1{]}. However, this normalization may change
the interpretation of the metric and lead to a situation where a representation
$z_{i}$ is less informative than $z_{j}$ (i.e., $I(x,z_{i})<I(x,z_{j})$)
but still has a higher rank than $z_{j}$ because $H(z_{i})<H(z_{j})$.
A better way is to divide $I(x,z_{i})$ by $\log(\text{\#bins})$.

\subsection{Metrics for separability and independence\label{subsec:Metrics-for-independence}}

\paragraph{MISJED}

We can characterize the independence between two latent variables
$z_{i}$, $z_{j}$ based on $I(z_{i},z_{j})$. However, a serious
problem of $I(z_{i},z_{j})$ is that it generates the following order
among pairs of representations:
\[
I(z_{\factor,i},z_{\factor,j})>I(z_{\factor,i},z_{\noise,j})>I(z_{\noise,i},z_{\noise,j})\geq0
\]
where $z_{\factor,i}$, $z_{\factor,j}$ are informative representations
and $z_{\noise,i}$, $z_{\noise,j}$ are uninformative (or noisy)
representations. This means if we simply want $z_{i}$, $z_{j}$ to
be independent, the best scenario is that \emph{both are noisy and
independent }(e.g. $q(z_{i}|x)\approx q(z_{j}|x)\approx\Normal(0,\Irm)$).
Therefore, we propose a new metric for independence named \textbf{MISJED}
(which stands for Mutual Information Sums Joint Entropy Difference),
defined as follows:
\begin{eqnarray}
\text{MISJED}(z_{i},z_{j})=\tilde{I}(z_{i},z_{j}) & = & H(z_{i})+H(z_{j})-H(\bar{z}_{i},\bar{z}_{j})\nonumber \\
 & = & H(z_{i})+H(z_{j})-H(z_{i},z_{j})+H(z_{i},z_{j})-H(\bar{z}_{i},\bar{z}_{j})\nonumber \\
 & = & I(z_{i},z_{j})+H(z_{i},z_{j})-H(\bar{z}_{i},\bar{z}_{j})\label{eq:MISJED}
\end{eqnarray}
where $\bar{z}_{i}=\Expect_{q(z_{i}|x)}[z_{i}]$ and $q(\bar{z}_{i})=\Expect_{p_{\Data}(x)}\left[q(\bar{z}_{i}|x)\right]$.
Since $q(\bar{z}_{i})$ and $q(\bar{z}_{j})$ have less variance than
$q(z_{i})$ and $q(z_{j})$, respectively, $H(z_{i},z_{j})-H(\bar{z}_{i},\bar{z}_{j})\geq0$,
making $\tilde{I}(z_{i},z_{j})\geq0$. 

To achieve a small value of $\tilde{I}(z_{i},z_{j})$, two representations
$z_{i}$, $z_{j}$ should be both \emph{independent }and\emph{ informative}
(or, in an extreme case, are deterministic given $x$). Using the
MISJED metric, we can ensure the following order: $0\leq\tilde{I}(z_{\factor,i},z_{\factor,j})<\tilde{I}(z_{\factor,i},z_{\noise,j})<\tilde{I}(z_{\noise,i},z_{\noise,j})$.
If $H(z_{i})$, $H(z_{j})$, and $H(\bar{z}_{i},\bar{z}_{j})$ in
Eq.~\ref{eq:MISJED} are estimated via quantization, we will have
$\tilde{I}(z_{i},z_{j})\leq H(z_{i})+H(z_{j})\leq2\log(\text{\#bins})$.
In this case, we can divide $\tilde{I}(z_{i},z_{j})$ by $2\log(\text{\#bins})$
to normalize it to {[}0, 1{]}.

\paragraph{WSEPIN and WINDIN}

A theoretically correct way to verify that a particular representation
$z_{i}$ is both \emph{separable} from other $z_{\neq i}$ and \emph{informative}
w.r.t $x$ is considering the amount of information in $x$ but not
in $z_{\neq i}$ that $z_{i}$ contains. This quantity is the conditional
mutual information between $x$ and $z_{i}$ given $z_{\neq i}$,
which can be decomposed as follows:
\begin{align}
I(x,z_{i}|z_{\neq i}) & =I(x,z_{i})-I(x,z_{i},z_{\neq i})\nonumber \\
 & =I(x,z_{i})-\left(I(z_{i},z_{\neq i})-I(z_{i},z_{\neq i}|x)\right)\nonumber \\
 & =I(x,z_{i})-I(z_{i},z_{\neq i})+I(z_{i},z_{\neq i}|x)\label{eq:cond_info}
\end{align}

$I(x,z_{i}|z_{\neq i})$ is useful for measuring how disentangled
a representation $z_{i}$ is \emph{in the absence of ground truth
factors}. $I(x,z_{i}|z_{\neq i})$ is close to $0$ if $z_{i}$ is
completely noisy and is high if $z_{i}$ is disentangled\footnote{Note that only informativeness and separability are considered in
this case.}. For models that use factorized encoders, $z_{i}$ and $z_{\neq i}$
are usually assumed to be independent given $x$, hence, $I(z_{i},z_{\neq i}|x)\approx0$
and $I(x,z_{i}|z_{\neq i})\approx I(x,z_{i})-I(z_{i},z_{\neq i})$
which is merely the difference between the \emph{informativeness}
and \emph{full separability} of $z_{i}$. For models that use auto-regressive
encoders, $I(z_{i},z_{\neq i}|x)>0$ which means $z_{i}$ and $z_{\neq i}$
can share information \emph{not} in $x$. 

We can also compute $I(x,z_{i}|z_{\neq i})$ in a different way as
follows:
\begin{align*}
I(x,z_{i}|z_{\neq i}) & =I(x,(z_{i},z_{\neq i}))-I(x,z_{\neq i})\\
 & =I(x,z)-I(x,z_{\neq i})
\end{align*}

If we want $z_{i}$ to be both \emph{independence }of $z_{\neq i}$
and informative w.r.t $x$, we can only use the first two terms in
Eq.~\ref{eq:cond_info} to derive another quantitive measure: 
\begin{align}
\hat{I}(x,z_{i}|z_{\neq i}) & =I(x,z_{i})-I(z_{i},z_{\neq i})\nonumber \\
 & =I(x,z_{i}|z_{\neq i})-I(z_{i},z_{\neq i}|x)\label{eq:cond_info2}
\end{align}
However, unlike $I(x,z_{i}|z_{\neq i})$, $\hat{I}(x,z_{i}|z_{\neq i})$
can be negative.

To normalize $I(x,z_{i}|z_{\neq i})$, we divide it by $H(z_{i})$
($H(z_{i})$ must be estimated via quantization). Note that taking
the average of $I(z_{i},x|z_{\neq i})$ over all representations to
derive a single metric for the whole model is \emph{not appropriate}
because \emph{models with more noisy latent variables will be less
favored}. For example, if model A has 10 latent variables (5 of them
are disentangled and 5 of them are noisy), and model B has 20 latent
variables (5 of them are disentangled and 15 of them are noisy), B
will always be considered worse than A \emph{despite the fact that
both are equivalent} in term of disentanglement (since 5 disentangled
representations are enough to capture all information in $x$ so additional
latent variables should be noisy). We propose two solutions for this
issue. In the first approach, we sort $I(x,z_{i}|z_{\neq i})$ over
all representations in descending order and only take the average
over the top $k$ latents (or groups of latents). This leads to a
metric called SEPIN@$k$\footnote{SEPIN stands for SEParability and INformativeness}
which is similar to Precision@$k$:
\[
\text{SEPIN@}k=\frac{1}{k}\sum_{i=0}^{k-1}I(x,z_{r_{i}}|z_{\neq r_{i}})
\]
where $r_{1},...,r_{L}$ is the rank indices of $L$ latent variables
by sorting $I(x,z_{i}|z_{\neq i})$ ($i=1,...,L$).

In the second approach, we compute the average over all $L$ representations
$z_{0},...,z_{L-1}$ but weighted by their informativeness to derive
a metric called WSEPIN:
\[
\text{WSEPIN}=\sum_{i=0}^{L-1}\rho_{i}I(x,z_{i}|z_{\neq i})
\]
where $\rho_{i}=\frac{I(x,z_{i})}{\sum_{j=0}^{L-1}I(x,z_{j})}$. If
$z_{i}$ is a noisy representation, $I(x,z_{i})\approx0$, thus, $z_{i}$
contributes almost nothing to the final WSEPIN.

Similarly, using the measure $\hat{I}(x,z_{i}|z_{\neq i})$ in Eq.~\ref{eq:cond_info2},
we can derive other two metrics INDIN@$k$\footnote{INDIN stands for INDependence and INformativeness}
and WINDIN as follows:
\[
\text{INDIN@}k=\frac{1}{k}\sum_{i=0}^{k-1}\hat{I}(x,z_{r_{i}}|z_{\neq r_{i}})\ \ \ \text{and}\ \ \ \text{WINDIN}=\sum_{i=0}^{L-1}\rho_{i}\hat{I}(x,z_{i}|z_{\neq i})
\]

\subsection{Metrics for interpretability}

\begin{figure}
\begin{centering}
\subfloat[Unsupervised]{\begin{centering}
\includegraphics[width=0.48\textwidth]{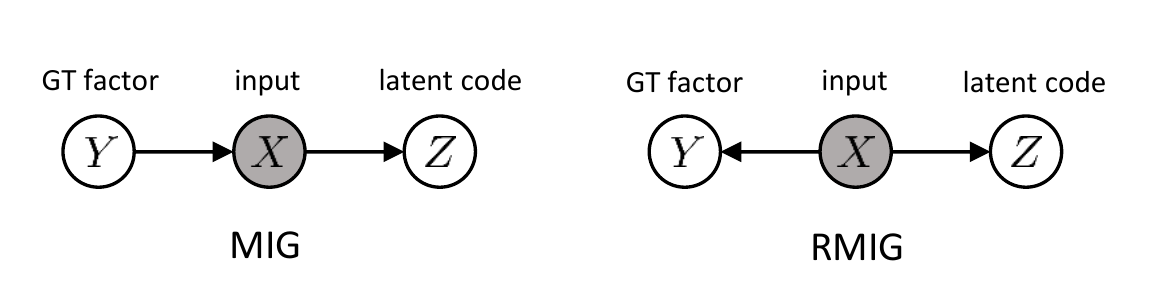}
\par\end{centering}
}~~~\subfloat[Supervised]{\begin{centering}
\includegraphics[width=0.48\textwidth]{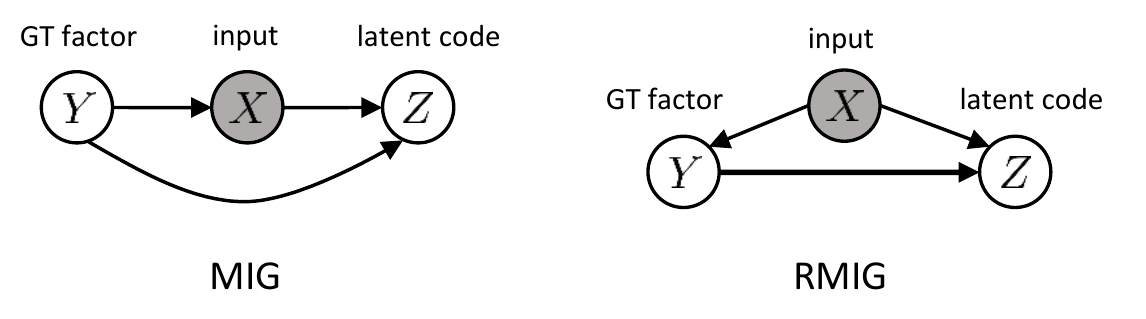}
\par\end{centering}
}
\par\end{centering}
\caption{Differences in probabilistic assumption of MIG and Robust MIG.\label{fig:Differences-in-probabilistic}}
\end{figure}

Recently, several metrics have been proposed to quantitatively evaluate
the interpretability of representations by examining the relationship
between the representations and manually labeled factors of variation.
The most popular ones are Z-diff score \cite{higgins2017beta,kim2018disentangling},
SAP \cite{kumar2017variational}, MIG \cite{chen2018isolating}. Among
them, only MIG is theoretically sound and provides correct computation
of $I(x,z_{i})$. MIG also matches with our formulation of ``interpretability''
in Section~\ref{sec:Rethinking-Disentanglement} to some extent.
However, MIG has only been used for toy datasets like dSprites \cite{dsprites2017}.
The main drawback comes from its probabilistic assumption $p(z_{i},y_{k},x^{(n)})=q(z_{i}|x^{(n)})p(x^{(n)}|y_{k})p(y_{k})$
(see Fig.~\ref{fig:Differences-in-probabilistic}). Note that $p(x^{(n)}|y_{k})$
is a distribution over the high dimensional data space, and is very
hard to robustly estimate but the authors simplified it to be $p(n|y_{k})$
if $x^{(n)}\in\Data_{y_{k}}$ ($\Data_{y_{k}}$is the support set
for a particular value $y_{k}$) and $0$ otherwise. This equation
only holds for toy datasets where we know exactly how $x$ is generated
from $y$. In addition, since $p(n|y_{k})$ depends on the value of
$y_{k}$, it will be problematic if $y_{k}$ is continuous. 

\paragraph{RMIG}

Addressing the drawbacks of MIG, we propose RMIG (which stands for
Robust MIG), formulated as follows:
\begin{eqnarray}
\text{RMIG}(y_{k}) & = & I(z_{i^{*}},y_{k})-I(z_{j^{\circ}},y_{k})\label{eq:RMIG}
\end{eqnarray}
where $I(z_{i^{*}},y_{k})$ and $I(z_{j^{\circ}},y_{k})$ are the
\emph{highest} and the \emph{second highest} mutual information values
computed between every $z_{i}$ and $y_{k}$; $z_{i^{*}}$ and $z_{j^{\circ}}$
are the corresponding latent variables. Like MIG, we can normalize
RMIG($y_{k}$) to {[}0, 1{]} by dividing it by $H(y_{k})$ but it
will favor imbalanced factors (small $H(y_{k})$).

RMIG inherits the idea of MIG but differs in the probabilistic assumption
(and other technicalities). RMIG assumes that $p(z_{i},y_{k},x^{(n)})=q(z_{i}|x^{(n)})p(y_{k}|x^{(n)})p(x^{(n)})$
for unsupervised learning and $p(z_{i},y_{k},x^{(n)})=q(z_{i}|y_{k}^{(n)},x^{(n)})p(y_{k}^{(n)},x^{(n)})$
for supervised learning (see Fig.~\ref{fig:Differences-in-probabilistic}).
Not only this eliminates all the problems of MIG but also provides
additional advantages. First, we can estimate $q(z_{i},y_{k})$ using
Monte Carlo sampling on $p(x^{(n)})$. Second, $p(y_{k}|x^{(n)})$
is well defined for both discrete/continuous $y_{k}$ and deterministic/stochastic
$p(y_{k}|x^{(n)})$. If $y_{k}$ is continuous, we can quantize $p(y_{k}|x^{(n)})$.
If $p(y_{k}|x^{(n)})$ is deterministic (i.e., a Dirac delta function),
we simply set it to $1$ for the value of $y_{k}$ corresponding to
$x^{(n)}$ and $0$ for other values of $y_{k}$. Our metric can also
use $p(y_{k}|x^{(n)})$ from an external expert model. Third, for
any particular value $y_{k}$, we compute $q(z_{i}|x^{(n)})$ for
all $x^{(n)}\in\Data$ rather than just for $x^{(n)}\in\Data_{y_{k}}$,
which gives more accurate results.

\paragraph{JEMMIG}

A high RMIG value of $y_{k}$ means that there is a representation
$z_{i^{*}}$ that captures the factor $y_{k}$. However, $z_{i^{*}}$
may also capture other factors $y_{\neq k}$ of the data. To make
sure that $z_{i^{*}}$ fits exactly to $y_{k}$, we provide another
metric for interpretability named JEMMIG (standing for Joint Entropy
Minuses Mutual Information Gap), computed as follows:
\begin{eqnarray*}
\text{JEMMIG}(y_{k}) & = & H(z_{i^{*}},y_{k})-I(z_{i^{*}},y_{k})+I(z_{j^{\circ}},y_{k})
\end{eqnarray*}
where $I(z_{i^{*}},y_{k})$ and $I(z_{j^{\circ}},y_{k})$ are defined
in Eq.~\ref{eq:RMIG}.

If we estimate $H(z_{i^{*}},y_{k})$ via quantization, we can bound
$\text{JEMMIG}(y_{k})$ between 0 and $H(y_{k})+\log(\text{\#bins})$
(please check Appdx.~\ref{subsec:Normalizing-JEMMIG} for details).
A small $\text{JEMMIG}(y_{k})$ score means that $z_{i^{*}}$ should
match exactly to $y_{k}$ and $z_{j^{\circ}}$ should not be related
to $y_{k}$. Thus, we can use $\text{JEMMIG}(y_{k})$ to \emph{validate
whether a model can learn disentangled representations w.r.t a ground
truth factor} $y_{k}$ \emph{or not} which satisfies the definition
in Section \ref{subsec:Definition}. Note that if we replace $H(z_{i^{*}},y_{k})$
by $H(y_{k})$ to account for the generalization of $z_{i^{*}}$ over
$y_{k}$, we obtain a metric equivalent to RMIG (but in reverse order).

To compute RMIG and JEMMIG for the whole model, we simply take the
average of $\text{RMIG}(y_{k})$ and $\text{JEMMIG}(y_{k})$ over
all $y_{k}$ ($k=1,...,K$) as follows:
\[
\text{RMIG =\ensuremath{\frac{1}{K}\sum_{k=0}^{K-1}\text{RMIG}(y_{k})}}\ \ \ \text{and}\ \ \ \text{JEMMIG}=\frac{1}{K}\sum_{k=0}^{K-1}\text{JEMMIG}(y_{k})
\]

\subsection{Comparison with existing metrics}

In Table \ref{tab:Metrics_Analysis}, we compare our proposed metrics
with existing metrics for learning disentangled representations. For
deeper analysis of these metrics, we refer readers to Appdx.~\ref{subsec:Analysis-of-existing}.
One can easily see that only our metrics satisfy the aforementioned
robustness criteria. Most other metrics (except for MIG and Modularity)
use classifiers, which can cause inconsistent results once the settings
of the classifiers change. Moreover, most other metrics (except for
MIG) use $\Expect_{q(z_{i}|x)}[z_{i}]$ instead of $q(z_{i}|x)$ for
computing mutual information. This can lead to inaccurate evaluation
results since $\Expect_{q(z_{i}|x)}[z_{i}]$ is \emph{theoretically
different} from $z_{i}\sim q(z_{i}|x)$. Among all metrics, JEMMIG
is the only one that can quantify ``disentangled representations''
defined in Section.~\ref{subsec:Definition} on its own.

\begin{table}[H]

\begin{centering}
\begin{tabular}{|c|c|>{\centering}p{0.2\textwidth}|>{\centering}p{0.13\textwidth}|>{\centering}p{0.08\textwidth}|>{\centering}p{0.1\textwidth}|>{\centering}p{0.08\textwidth}|}
\hline 
{\scriptsize{}Metrics} & {\scriptsize{}\#classifiers} & {\scriptsize{}classifier} & {\scriptsize{}nonlinear relationship} & {\scriptsize{}use $q(z_{i}|x)$} & {\scriptsize{}continuous factors} & {\scriptsize{}real data}\tabularnewline
\hline 
\hline 
{\scriptsize{}Z-diff} & {\scriptsize{}$1$} & {\scriptsize{}linear/majority-vote} & {\scriptsize{}$\times$} & {\scriptsize{}$\times$} & {\scriptsize{}$\times$} & {\scriptsize{}$\times$}\tabularnewline
\hline 
{\scriptsize{}SAP} & {\scriptsize{}$L\times K$} & {\scriptsize{}threshold value} & {\scriptsize{}$\times$} & {\scriptsize{}$\times$} & {\scriptsize{}$\checkmark$} & {\scriptsize{}$\checkmark$}\tabularnewline
\hline 
{\scriptsize{}MIG} & {\scriptsize{}$0$} & {\scriptsize{}none} & {\scriptsize{}$\checkmark$} & {\scriptsize{}$\checkmark$} & {\scriptsize{}$\times$} & {\scriptsize{}$\times$}\tabularnewline
\hline 
{\scriptsize{}Disentanglement} & \multirow{3}{*}{{\scriptsize{}$K$}} & \multirow{3}{0.2\textwidth}{\centering{}{\scriptsize{}LASSO/}\\
{\scriptsize{}random forest}} & \multirow{3}{0.13\textwidth}{\centering{}{\scriptsize{}$\times$/$\checkmark$}} & \multirow{3}{0.08\textwidth}{\centering{}{\scriptsize{}$\times$}} & \multirow{3}{0.1\textwidth}{\centering{}{\scriptsize{}$\times$}} & \multirow{3}{0.08\textwidth}{\centering{}{\scriptsize{}$\times$}}\tabularnewline
\cline{1-1} 
{\scriptsize{}Completeness} &  &  &  &  &  & \tabularnewline
\cline{1-1} 
{\scriptsize{}Informativeness} &  &  &  &  &  & \tabularnewline
\hline 
{\scriptsize{}Modularity} & {\scriptsize{}$0$} & {\scriptsize{}none} & {\scriptsize{}$\checkmark$} & {\scriptsize{}$\times$} & {\scriptsize{}$\checkmark$} & {\scriptsize{}$\times$}\tabularnewline
\hline 
{\scriptsize{}Explicitness} & {\scriptsize{}$K$} & {\scriptsize{}one-vs-rest }\\
{\scriptsize{}logistic regressor} & {\scriptsize{}$\times$} & {\scriptsize{}$\times$} & {\scriptsize{}$\times$} & {\scriptsize{}$\times$}\tabularnewline
\hline 
{\scriptsize{}WSEPIN$^{\dagger}$} & {\scriptsize{}$0$} & {\scriptsize{}none} & {\scriptsize{}$\checkmark$} & {\scriptsize{}$\checkmark$} & {\scriptsize{}$\checkmark$} & {\scriptsize{}$\checkmark$}\tabularnewline
\hline 
{\scriptsize{}WINDIN$^{\dagger}$} & {\scriptsize{}$0$} & {\scriptsize{}none} & {\scriptsize{}$\checkmark$} & {\scriptsize{}$\checkmark$} & {\scriptsize{}$\checkmark$} & {\scriptsize{}$\checkmark$}\tabularnewline
\hline 
{\scriptsize{}RMIG} & {\scriptsize{}$0$} & {\scriptsize{}none} & {\scriptsize{}$\checkmark$} & {\scriptsize{}$\checkmark$} & {\scriptsize{}$\checkmark$} & {\scriptsize{}$\checkmark$}\tabularnewline
\hline 
{\scriptsize{}JEMMIG{*}} & {\scriptsize{}$0$} & {\scriptsize{}none} & {\scriptsize{}$\checkmark$} & {\scriptsize{}$\checkmark$} & {\scriptsize{}$\checkmark$} & {\scriptsize{}$\checkmark$}\tabularnewline
\hline 
\end{tabular}
\par\end{centering}
\caption{Analysis of different metrics for disentanglement learning. $L$ and
$K$ are the numbers of latent variables and ground truth factors,
respectively. Metrics marked with {*} are self-contained. Metrics
marked with $^{\dagger}$ do not require ground truth factors of variation.\label{tab:Metrics_Analysis}}
\end{table}

\section{Experiments \label{sec:Experiments}}

We use our proposed metrics to evaluate three representation learning
methods namely FactorVAE \cite{kim2018disentangling}, $\beta$-VAE
\cite{higgins2017beta} and AAE \cite{makhzani2015adversarial} on
\emph{both real and toy datasets} which are CelebA \cite{liu2015faceattributes}
and dSprites \cite{dsprites2017}, respectively. A brief discussion
of these models are given in Appdx.~\ref{subsec:Review-of-FactorVAEs}.
We would like to show the following points: i) how to compare models
based on our metrics; ii) the advantages of our metrics compared to
other metrics; iii) the consistence between qualitative results produced
by our metrics and visual results; and iv) the ablation study of our
metrics.

Due to space limit, we only present experiments for the first two
points. The experiments for points (iii) and (iv) are put in Appdx.~\ref{subsec:Consistence-between-quantiative}
and Appdx.~\ref{subsec:Ablation-study-of}, respectively. Details
about the datasets and model settings are provided in Appdx.~\ref{subsec:Dataset-Details}
and Appdx.~\ref{subsec:Model-settings}, respectively. In all figures
below, ``TC'' refers to the $\gamma$ coefficient of the TC loss
in FactorVAEs \cite{kim2018disentangling}, ``Beta'' refers to the
$\beta$ coefficient in $\beta$-VAEs \cite{higgins2017beta}.

\paragraph{Informativeness}

\begin{figure}
\begin{centering}
\subfloat[mean of $I(z_{i},x)$\label{fig:Mean-of-I_zi_x}]{\begin{centering}
\includegraphics[width=0.33\textwidth]{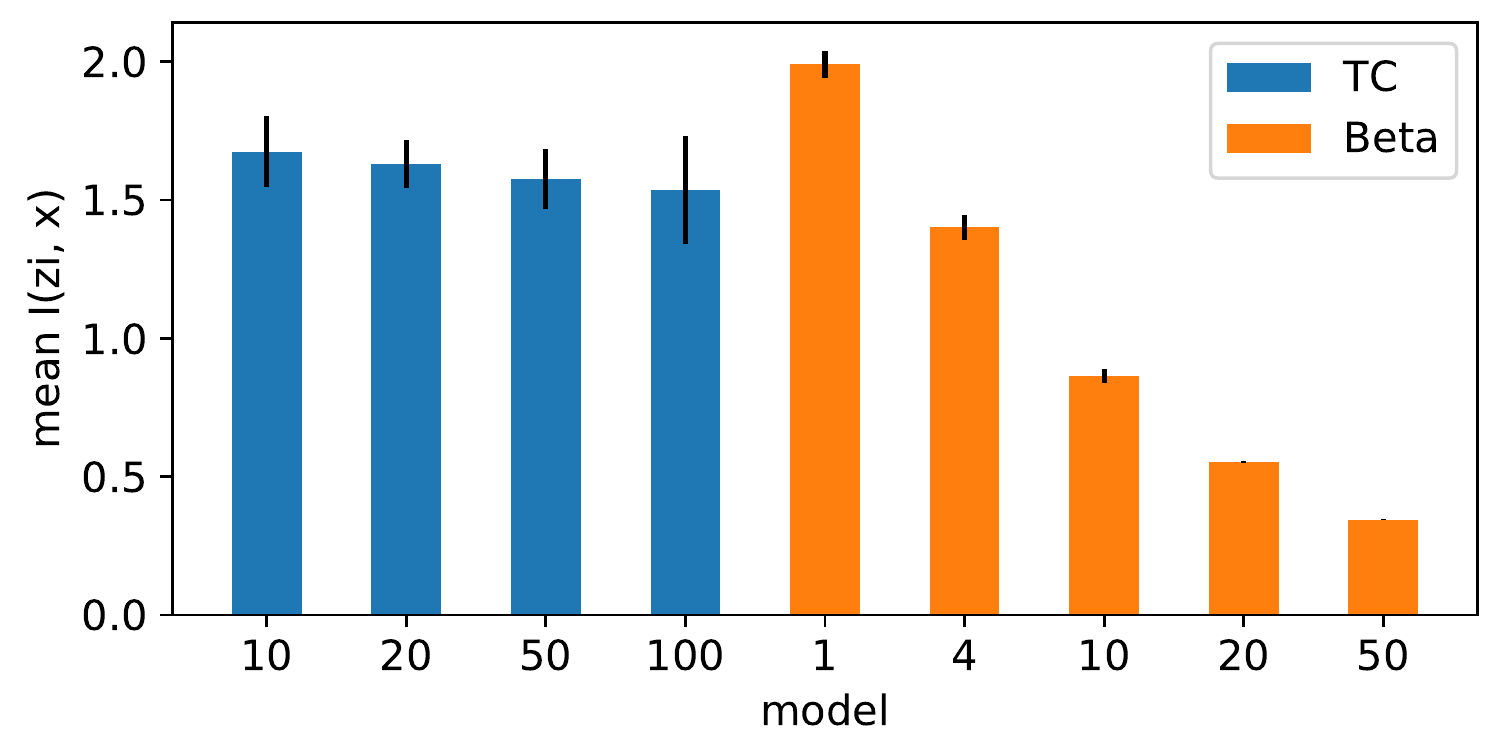}
\par\end{centering}
}\hspace{0.05\textwidth}\subfloat[$I(z,x)$\label{fig:I_z_x}]{\begin{centering}
\includegraphics[width=0.33\textwidth]{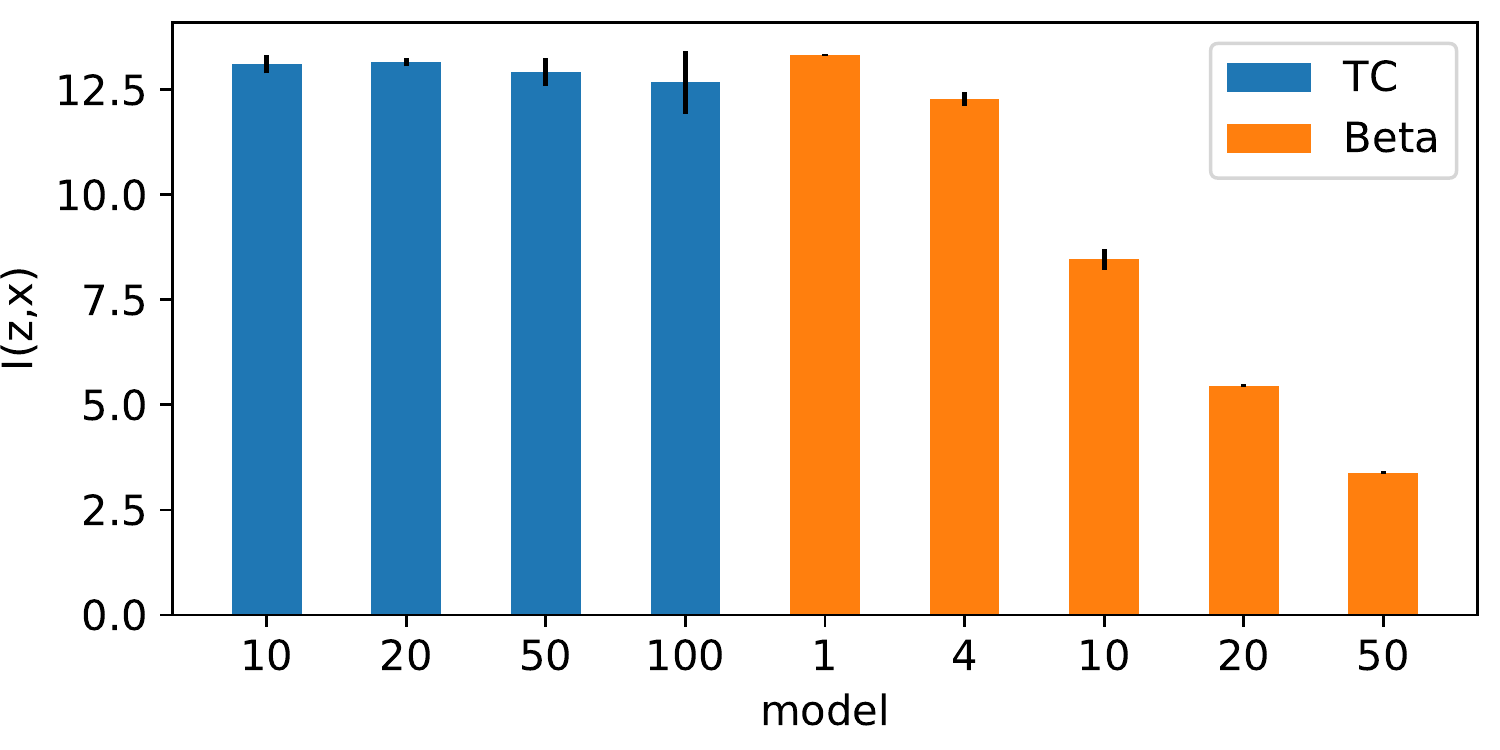}
\par\end{centering}
}
\par\end{centering}
\caption{The informativeness and the total information of some FactorVAE and
$\beta$-VAE models. For each hyperparameter, we report the mean and
the standard error of 4 different runs. }
\end{figure}

In Figs.~\ref{fig:Mean-of-I_zi_x} and \ref{fig:I_z_x}, we show
the average amount of information (of $x$) that a representation
$z_{i}$ contains (the mean of $I(z_{i},x)$) and the total amount
of information that all representations $z$ contain ($I(z,x)$).
It is clear that adding the TC term to the standard VAE loss does
not affect $I(z,x)$ much (Fig.~\ref{fig:I_z_x}). However, because
$z_{i}$ and $z_{j}$ in FactorVAEs are more separable than those
in standard VAEs, FactorVAEs should produce smaller $I(z_{i},x)$
than standard VAEs on average (Fig.~\ref{fig:Mean-of-I_zi_x}). We
also see that the mean of $I(z_{i},x)$ and $I(z,x)$ consistently
decrease for $\beta$-VAEs with higher $\beta$. 

\paragraph{Separability and Independence}

\begin{figure}
\begin{centering}
\subfloat[max/mean/min of $I(z_{i},z_{\protect\neq i})$\label{fig:MaxMeanMinSep}]{\begin{centering}
\includegraphics[width=0.32\textwidth]{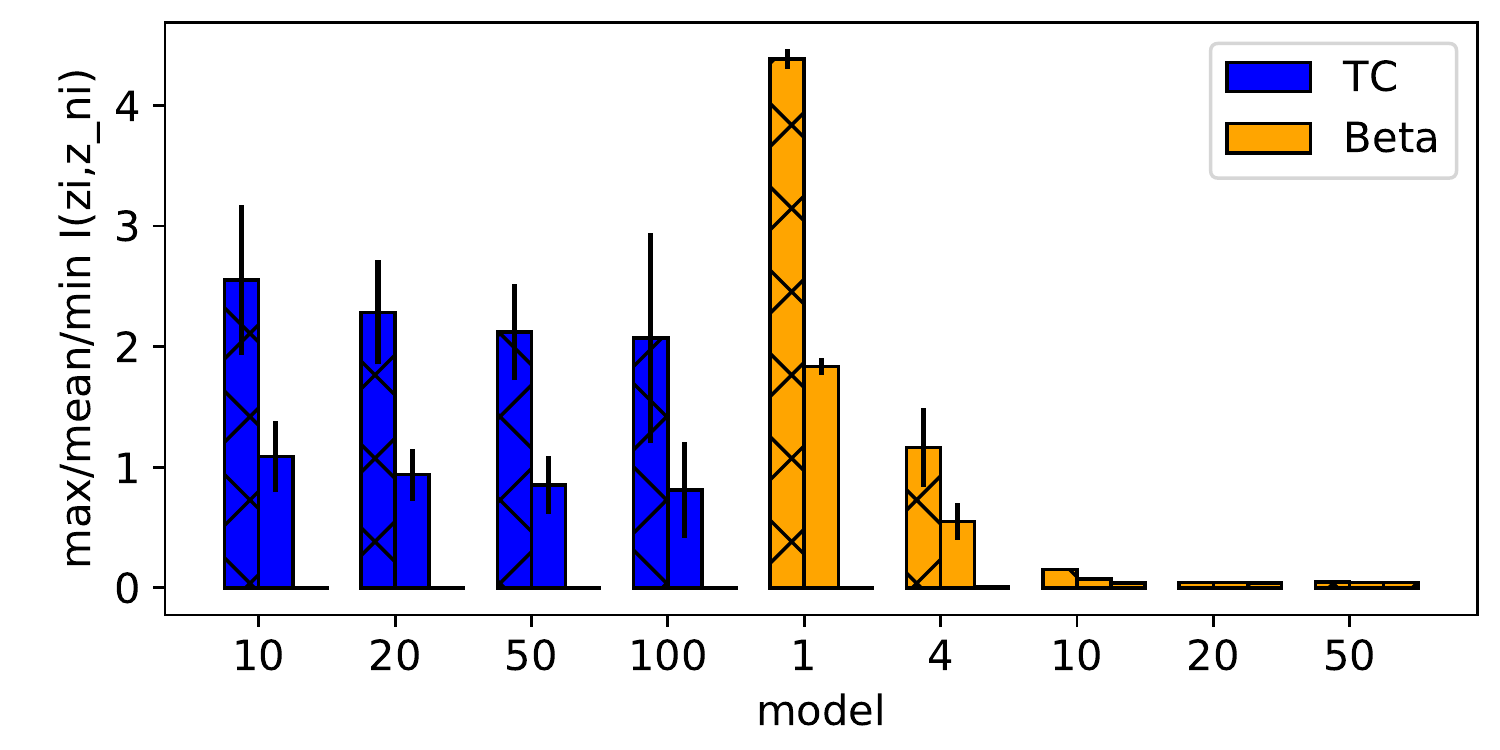}
\par\end{centering}
}\subfloat[WSEPIN\label{fig:WSEPIN}]{\begin{centering}
\includegraphics[width=0.32\textwidth]{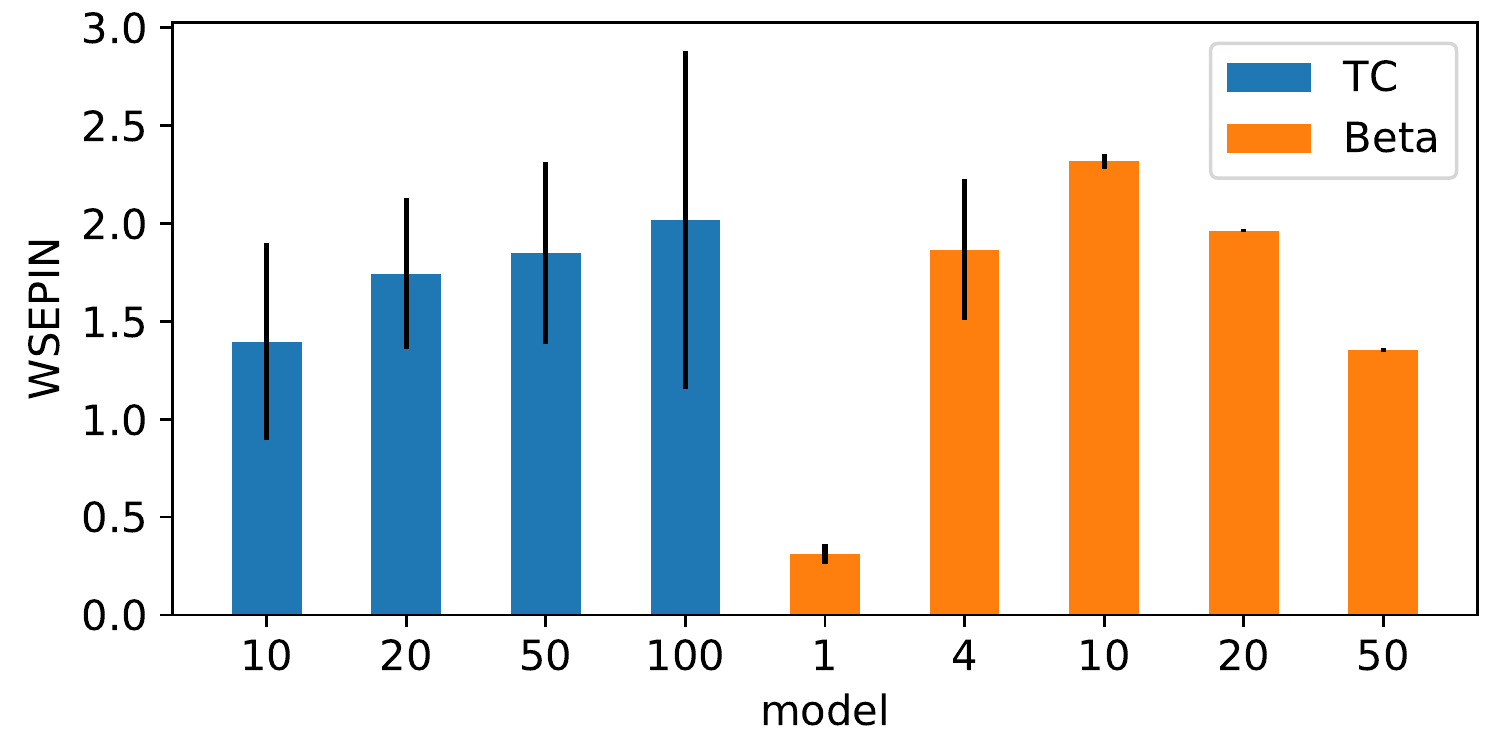}
\par\end{centering}
}\subfloat[SEPIN@3\label{fig:SEPIN@3}]{\begin{centering}
\includegraphics[width=0.32\textwidth]{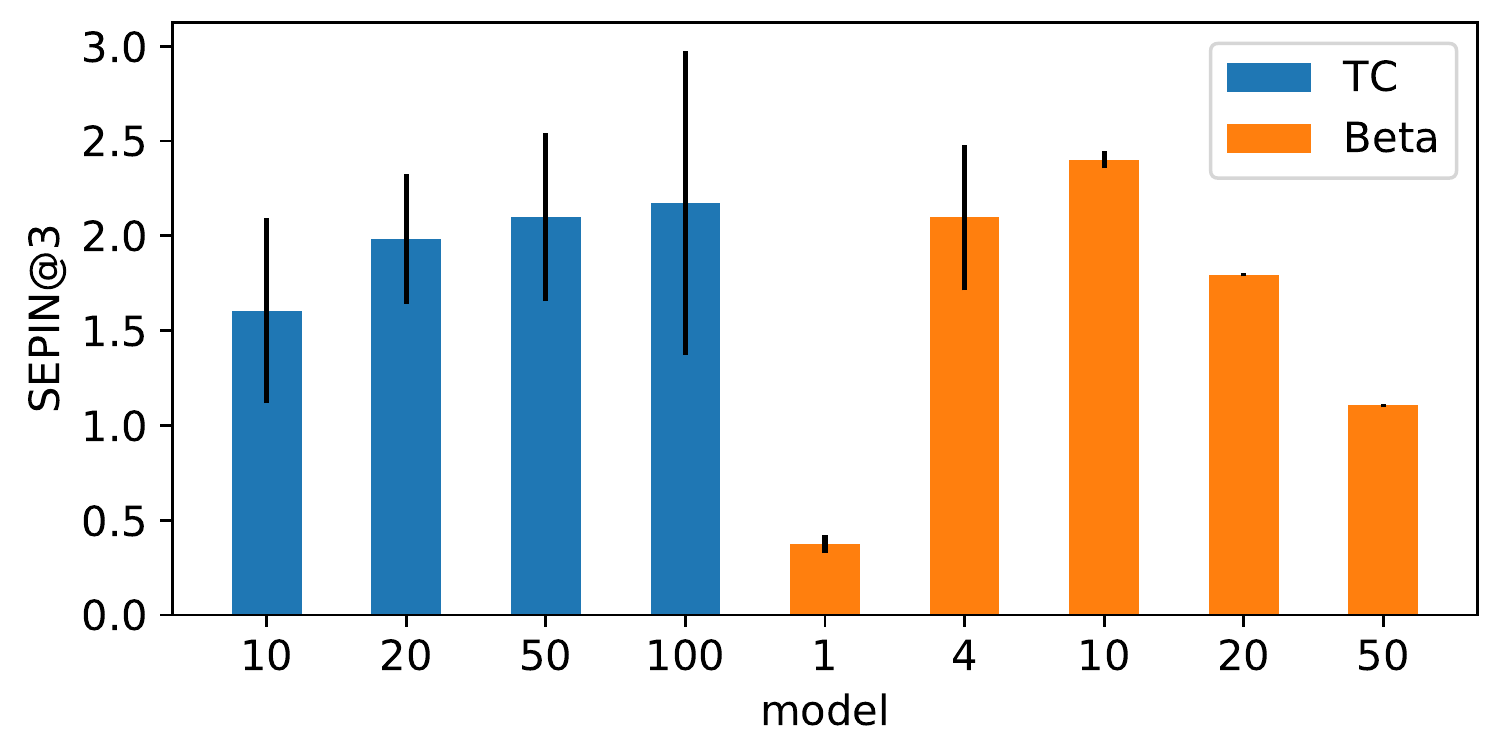}
\par\end{centering}
}
\par\end{centering}
\caption{$I(z_{i},z_{\protect\neq i})$, WSEPIN and SEPIN@3 of some FactorVAE
and $\beta$-VAE models.}
\end{figure}

\begin{figure}[h]
\begin{centering}
\subfloat[FactorVAE (TC=10)\label{fig:FactorVAE_SEP}]{\begin{centering}
\includegraphics[width=0.32\textwidth]{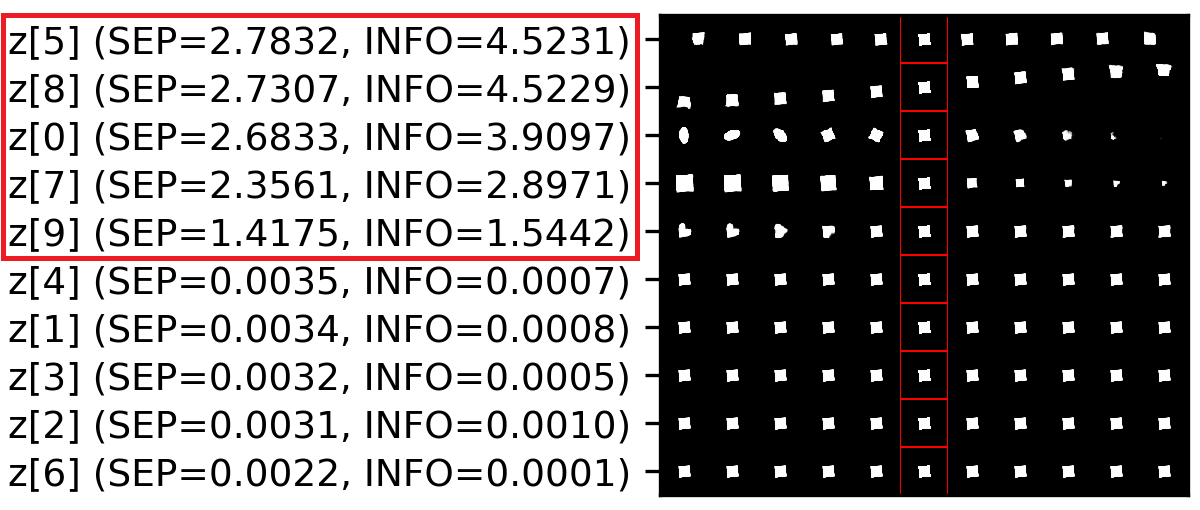}
\par\end{centering}
}\subfloat[VAE\label{fig:VAE_SEP}]{\begin{centering}
\includegraphics[width=0.32\textwidth]{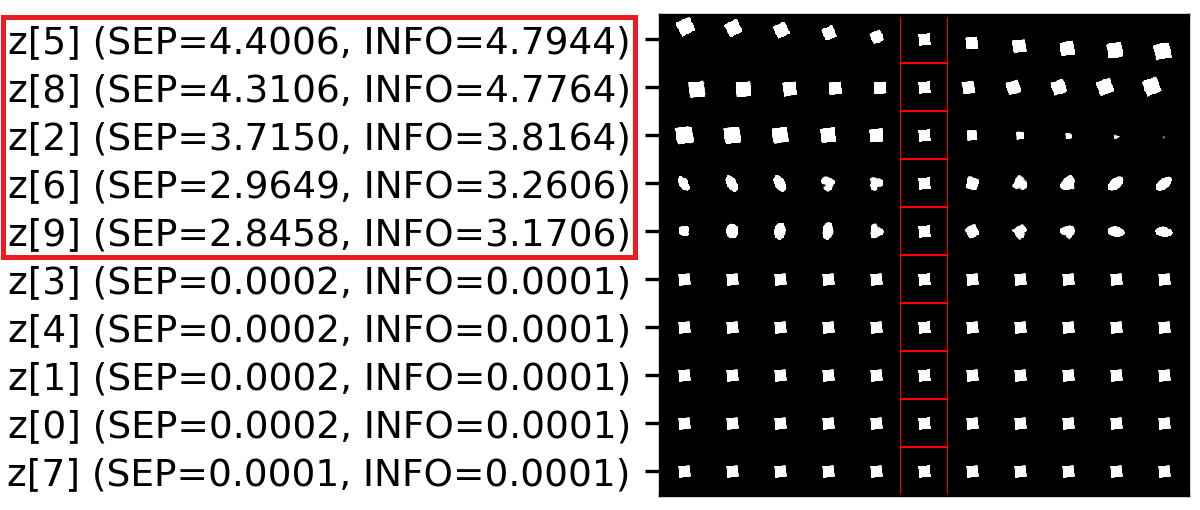}
\par\end{centering}
}\subfloat[$\beta$-VAE ($\beta=10$)\label{fig:BetaVAE_SEP}]{\begin{centering}
\includegraphics[width=0.32\textwidth]{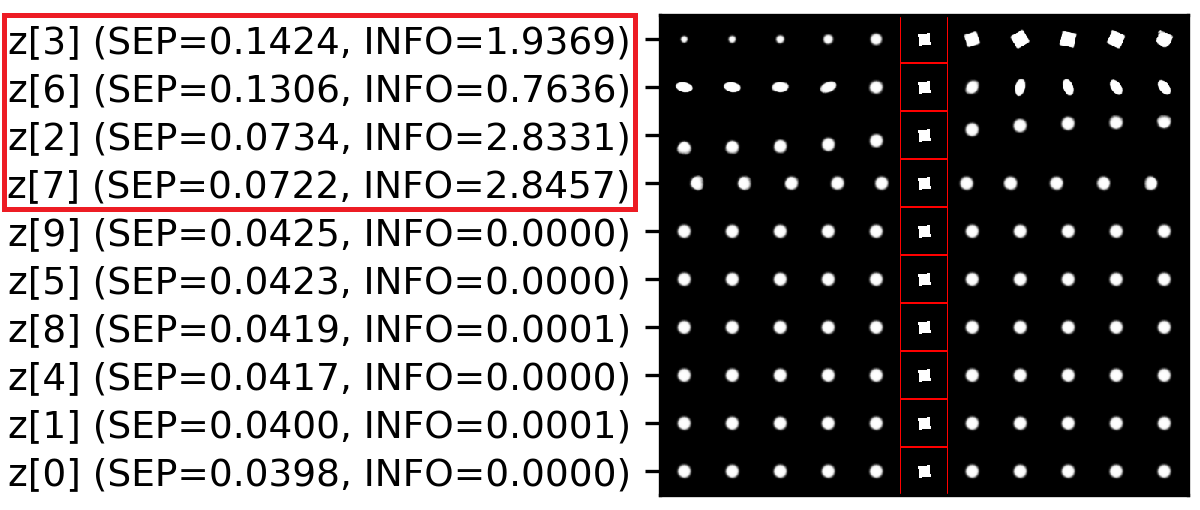}
\par\end{centering}
}
\par\end{centering}
\caption{Visualization of the representations learned by representative FactorVAE,
VAE, and $\beta$-VAE models with separability ($I(z_{i},z_{\protect\neq i})$)
and informativeness ($I(z_{i},x)$) scores. Representations are sorted
by their separability scores.}
\end{figure}

If we only evaluate models based on the separability of representations,
$\beta$-VAE models with large $\beta$ are among the best. These
models force latent representations to be highly separable (as in
Fig.~\ref{fig:MaxMeanMinSep}, we can see that the max/mean/min values
of $I(z_{i},z_{\neq i})$ are equally small for $\beta$-VAEs with
large $\beta$). In FactorVAEs, informative representations usually
have poor separability (large value) and noisy representations usually
have perfect separability ($\approx0$) (Fig.~\ref{fig:FactorVAE_SEP}).
Increasing the weight of the TC loss improves the max and mean of
$I(z_{i},z_{\neq i})$ but not significance (Fig.~\ref{fig:MaxMeanMinSep}).

Using WSEPIN and SEPIN@3 gives us a more reasonable evaluation of
the disentanglement capability of these models. In Fig.~\ref{fig:WSEPIN},
we see that $\beta$-VAE models with $\beta=10$ achieve the highest
WSEPIN and SEPIN@3 scores, which suggests that their informative representations
usually contain large amount of information of $x$ that are \emph{not
shared by other representations}. However, this type of information
may not associate well with the ground truth factors of variation
(e.g., $z_{3},z_{6}$ in Fig.~\ref{fig:BetaVAE_SEP}). The representations
of FactorVAEs, despite containing less information of $x$ \emph{on
their own}, usually reflect the ground truth factors more accurately
(e.g., $z_{5},z_{8},z_{7}$ in Fig.~\ref{fig:FactorVAE_SEP}) than
those of $\beta$-VAEs. These results suggest that ground truth factors
should be used for proper evaluations of disentanglement. 

\paragraph{Interpretability}

Using JEMMIG and RMIG, we see that FactorVAE models can learn representations
that are more interpretable than those learned by $\beta$-VAE models.
Surprisingly, the worst FactorVAE models (with TC=10) clearly outperform
the best $\beta$-VAE models (with $\beta=10$). This result is sensible
because it is accordant with the visualization in Figs.~\ref{fig:FactorVAE_SEP}
and \ref{fig:BetaVAE_SEP}. 

\begin{figure}[h]
\begin{centering}
\subfloat[JEMMIG\label{fig:JEMMIG_dSprite}]{\begin{centering}
\includegraphics[width=0.32\textwidth]{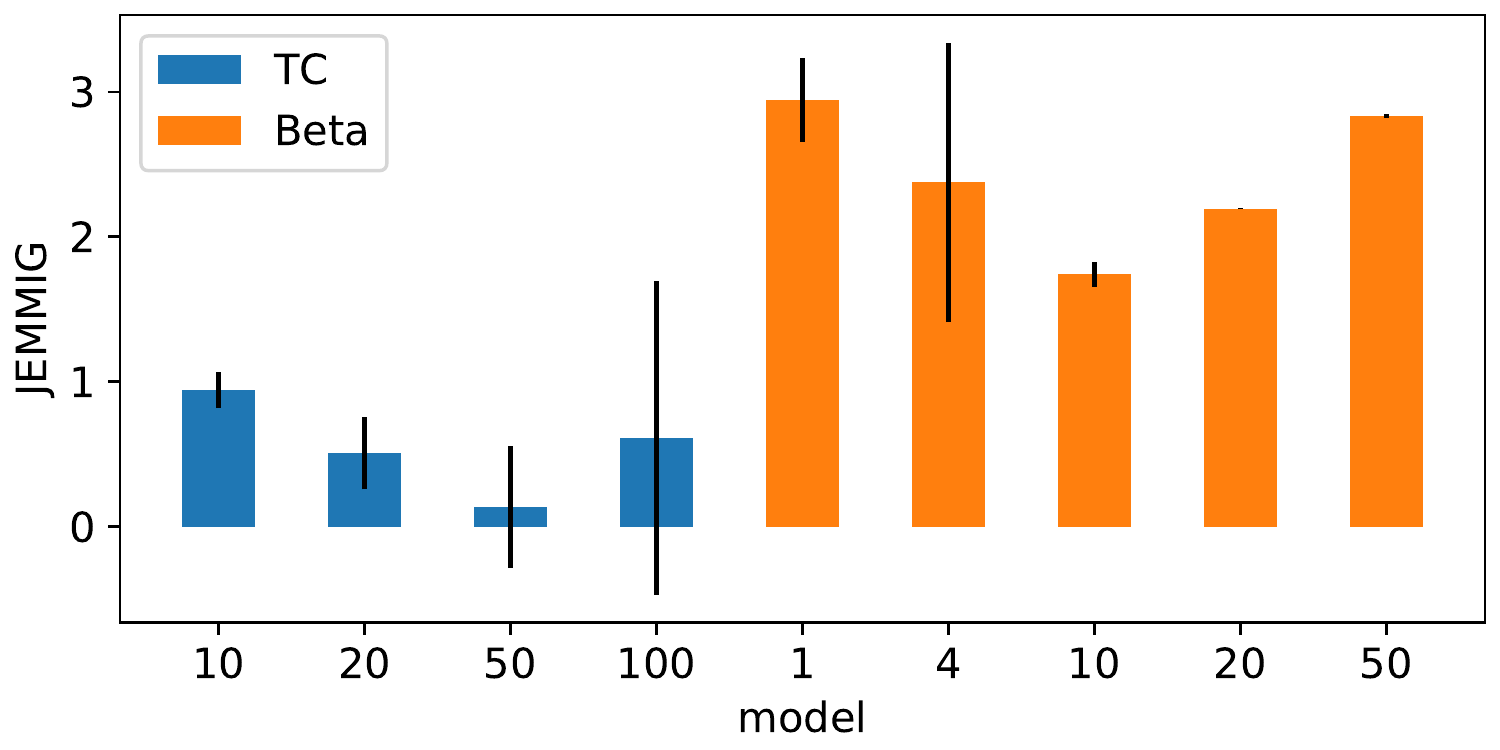}
\par\end{centering}
}\subfloat[RMIG\label{fig:RMIG_dSprites}]{\begin{centering}
\includegraphics[width=0.32\textwidth]{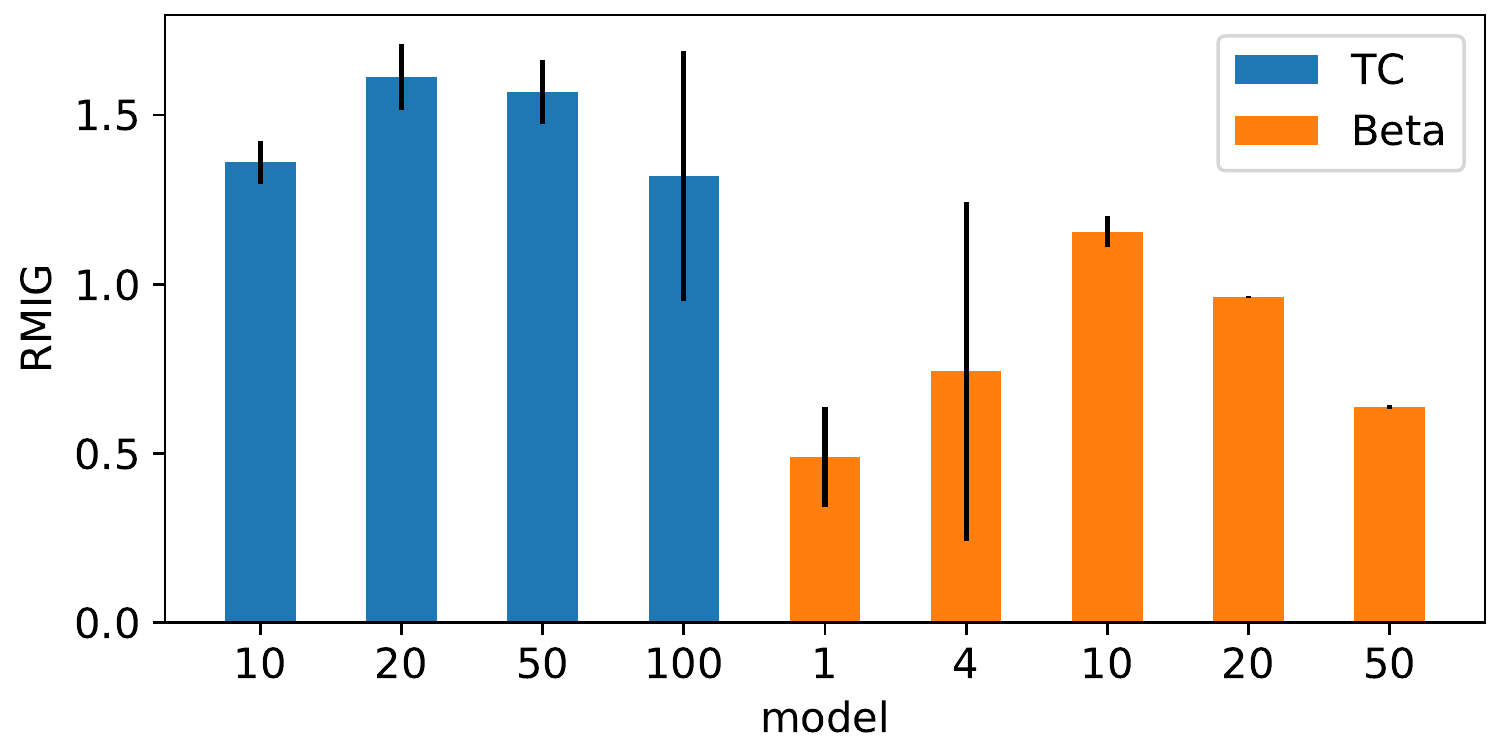}
\par\end{centering}
}\subfloat[JEMMIG vs. RMIG\label{fig:JEMMIG-vs-RMIG}]{\begin{centering}
\includegraphics[width=0.26\textwidth]{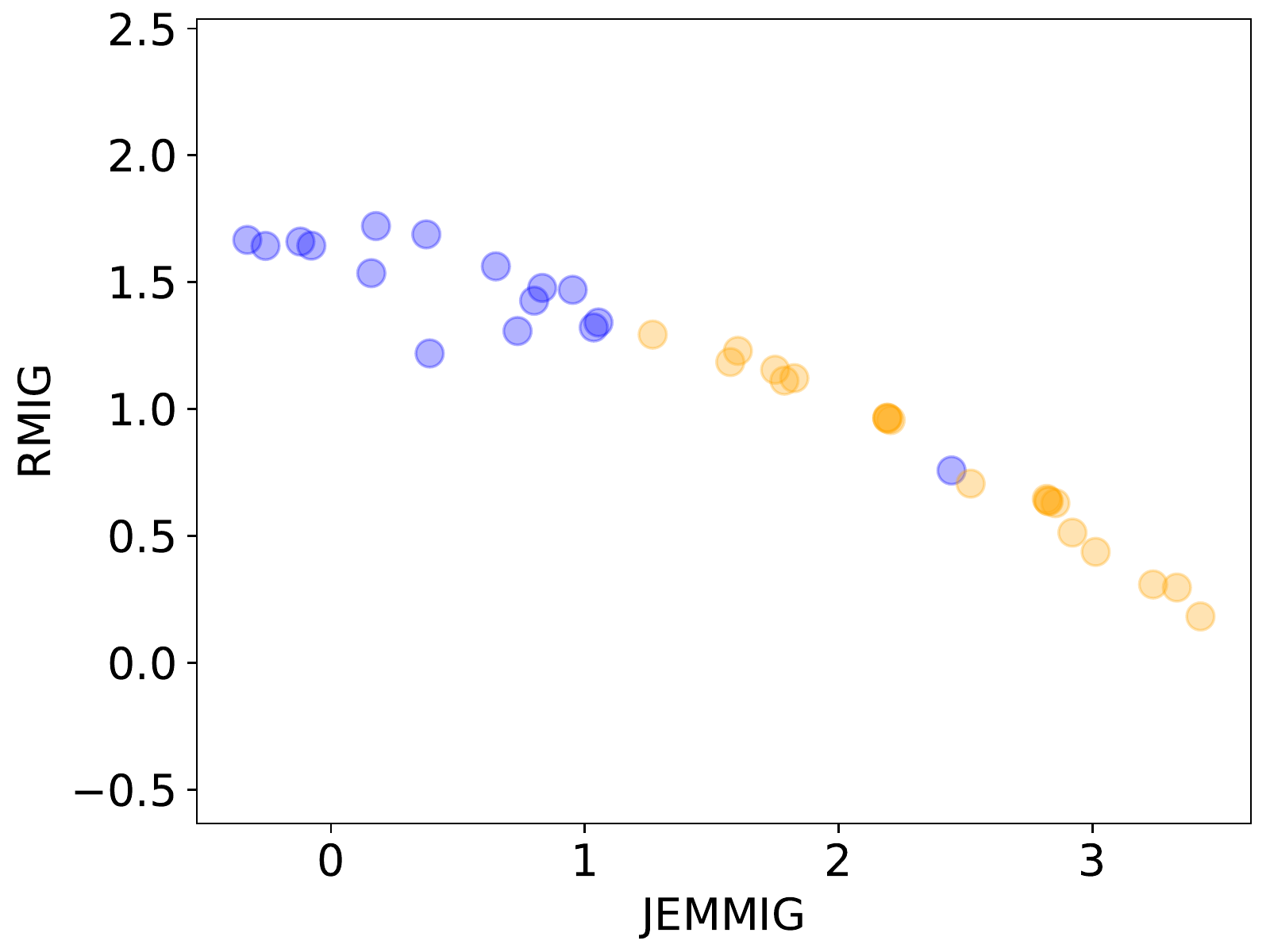}
\par\end{centering}
}
\par\end{centering}
\caption{(a) and (b): Unnormalized JEMMIG and RMIG scores of several FactorVAE
and $\beta$-VAE models. (c): Correlation between JEMMIG and RMIG.}
\end{figure}

\paragraph{Comparison with Z-diff}

In \cite{chen2018isolating}, the authors have already shown that
MIG is more robust than Z-diff \cite{higgins2017beta} so we compare
our metrics with MIG directly.

\paragraph{Comparison with MIG}

On toy datasets like dSprites, RMIG produces similar results as MIG
\cite{chen2018isolating}. Please check Appdx.~\ref{subsec:Comparing-RMIG-with}
for more details.

\paragraph{Comparison with ``disentanglement'', ``completeness'' and ``informativeness''}

\begin{figure}[h]
\begin{centering}
\subfloat[JEMMIG vs. Disentanglement\label{fig:JEMMIG-vs-Disentanglement}]{\begin{centering}
\includegraphics[width=0.31\textwidth]{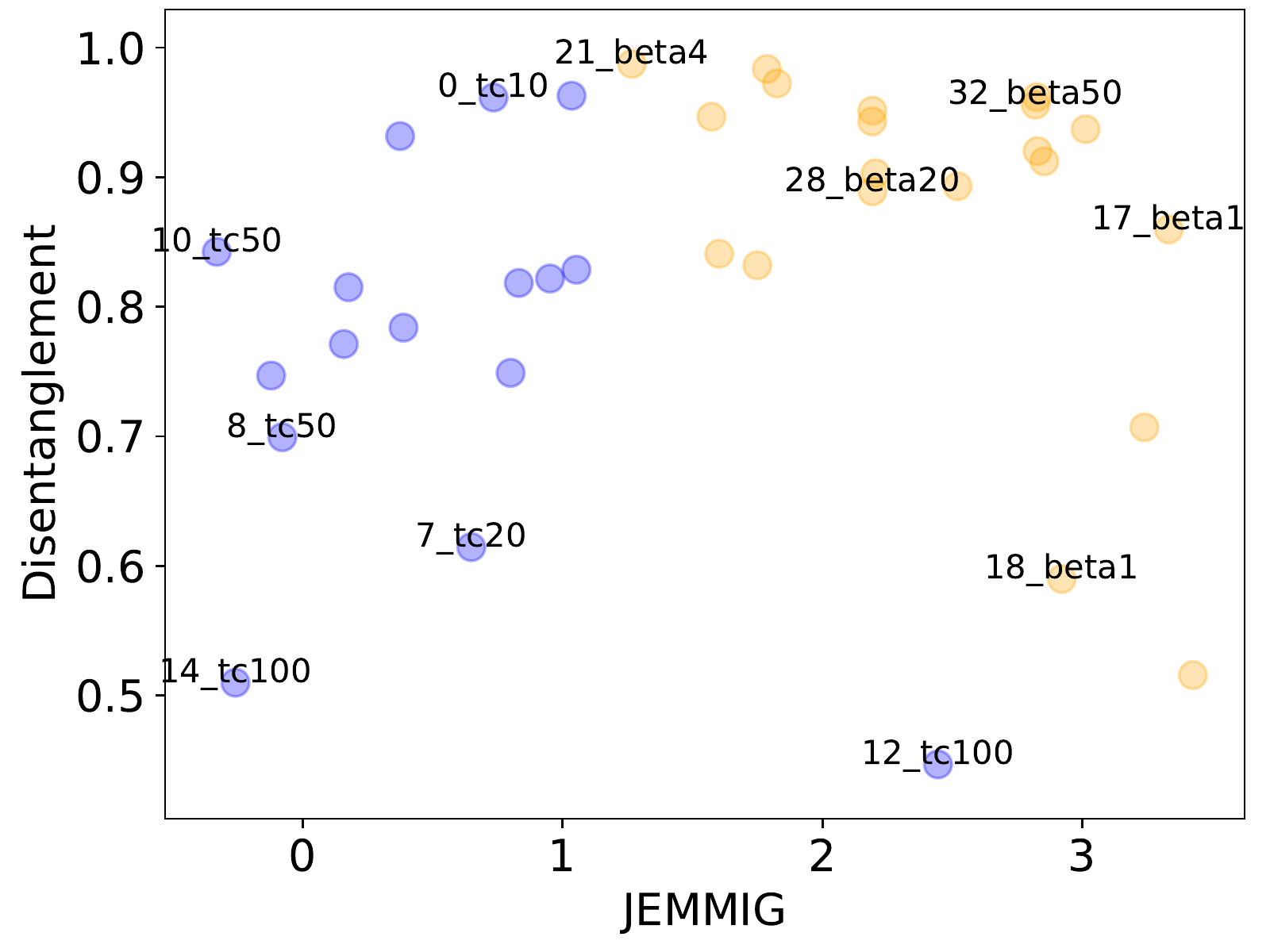}
\par\end{centering}
}\subfloat[JEMMIG vs. Completeness\label{fig:JEMMIG-vs-Completeness}]{\begin{centering}
\includegraphics[width=0.31\textwidth]{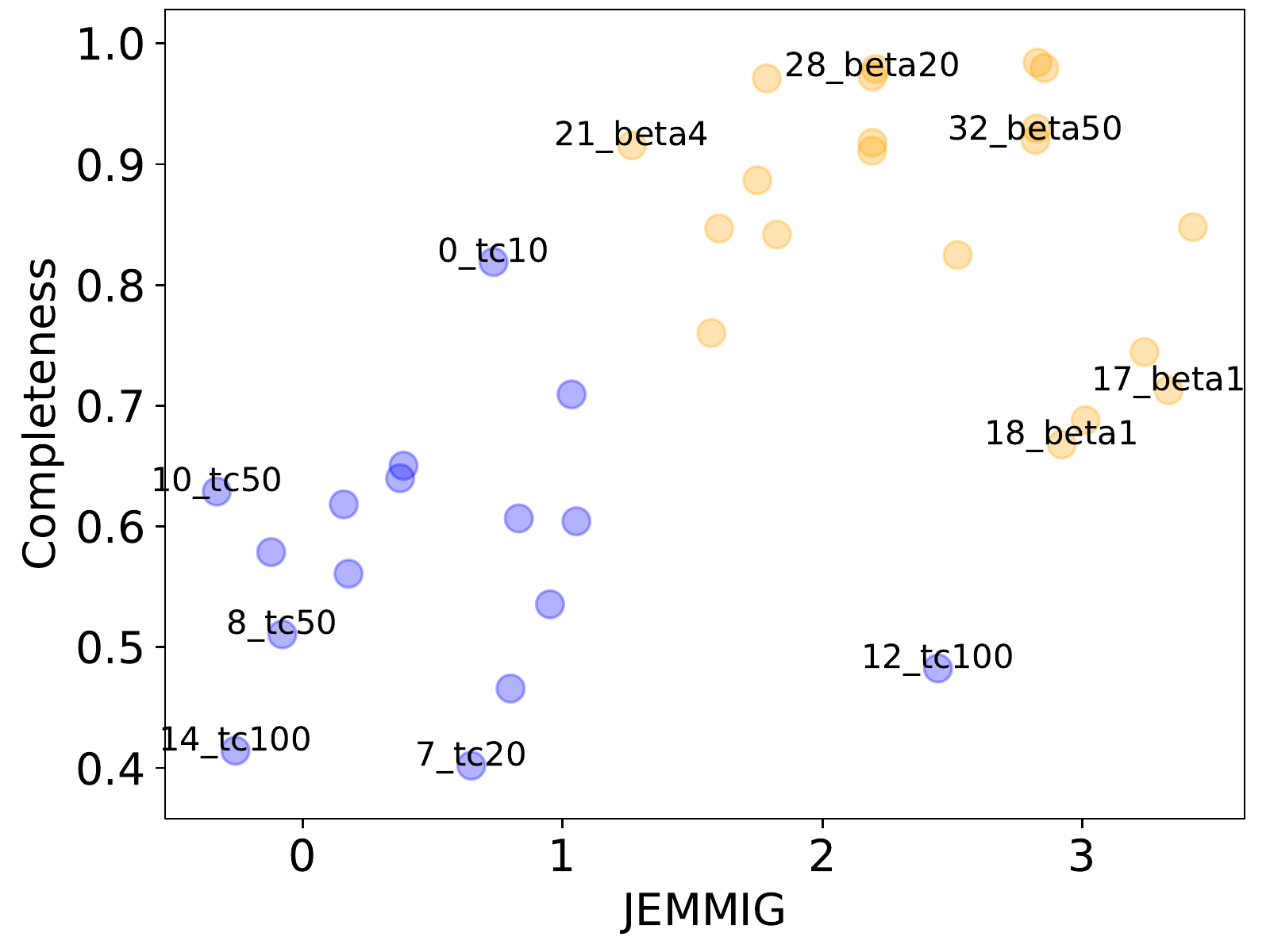}
\par\end{centering}
}\subfloat[JEMMIG vs. Error]{\begin{centering}
\includegraphics[width=0.31\textwidth]{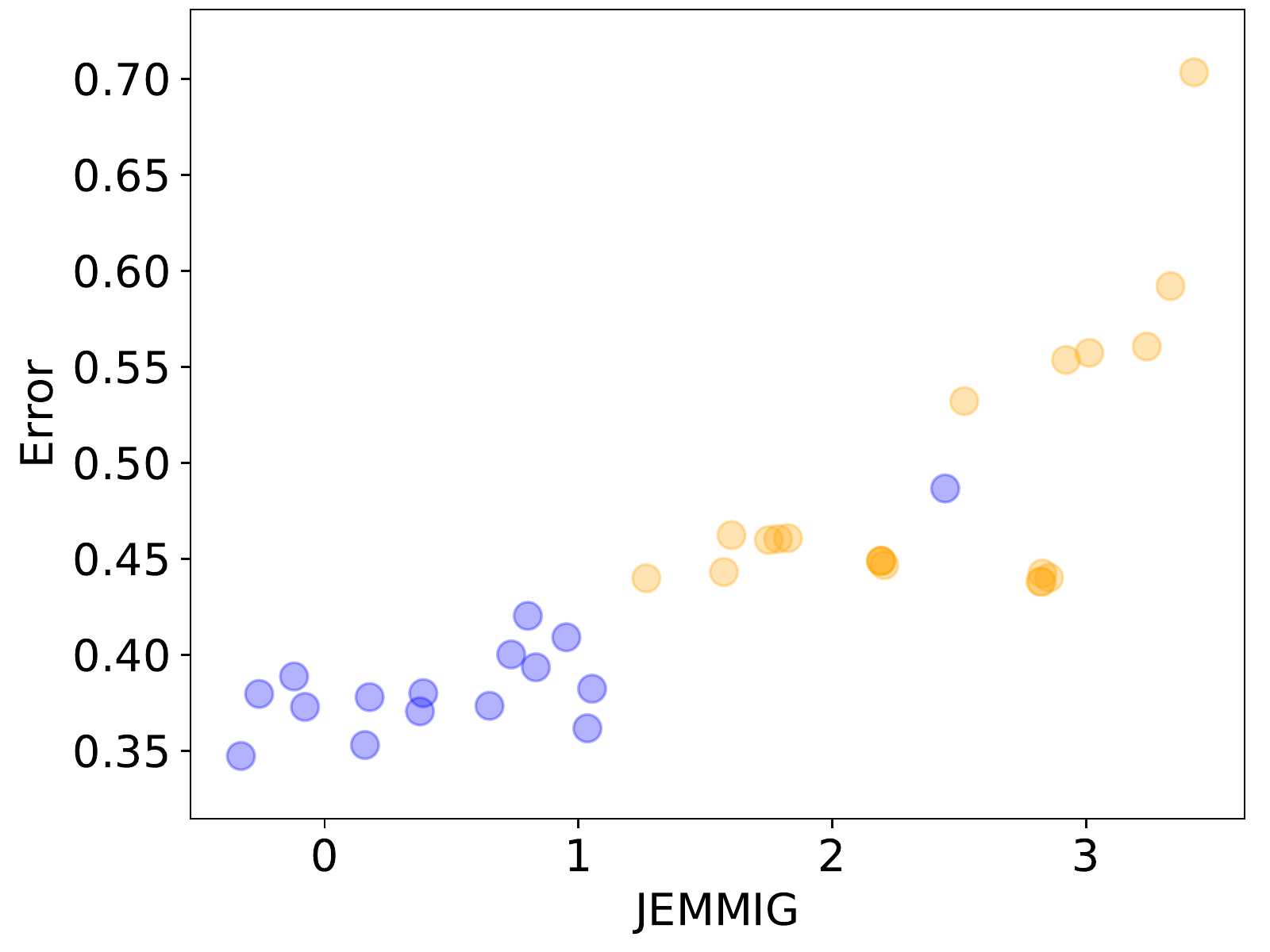}
\par\end{centering}
}
\par\end{centering}
\begin{centering}
\subfloat[RMIG vs. Disentanglement]{\begin{centering}
\includegraphics[width=0.31\textwidth]{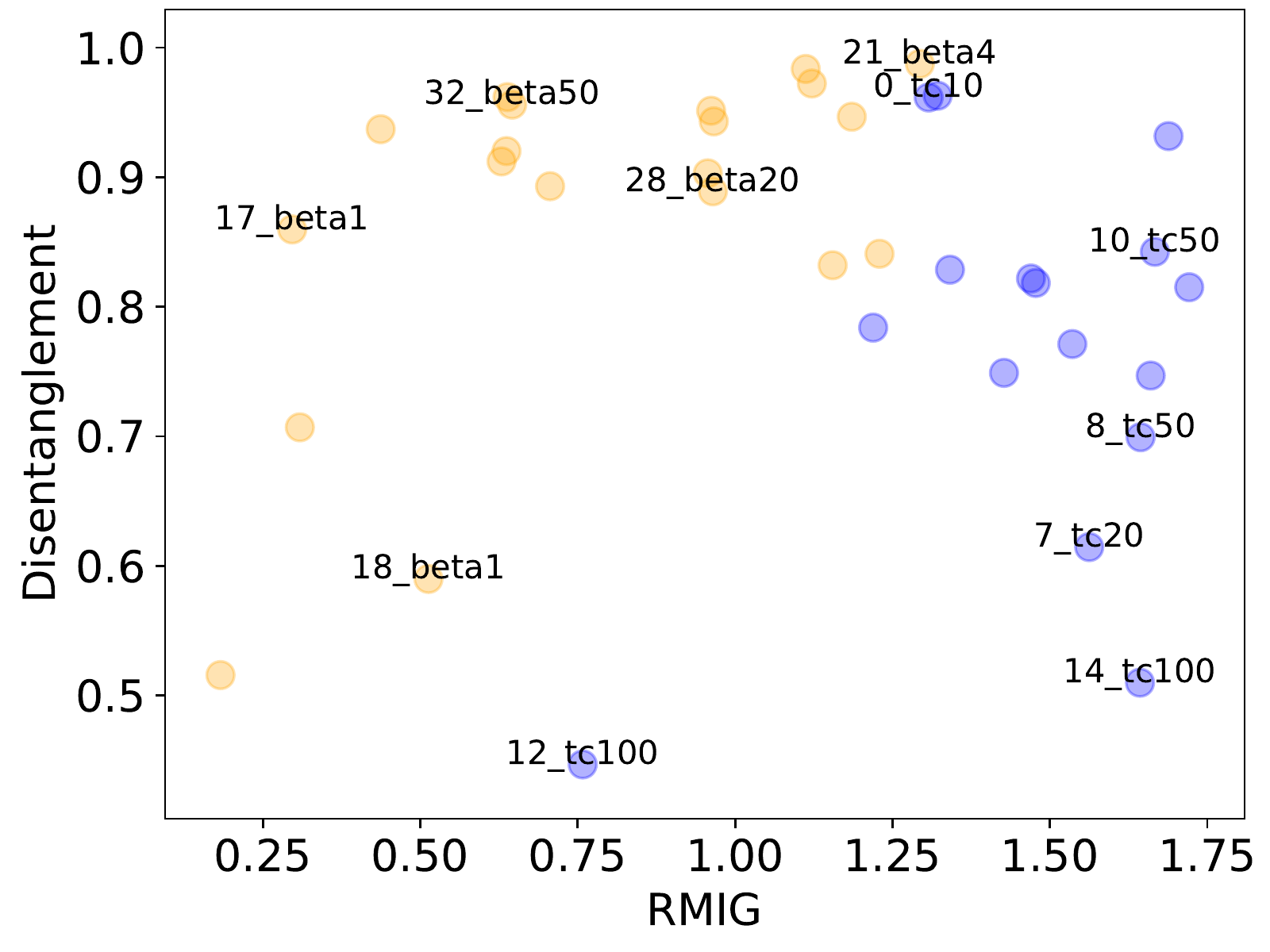}
\par\end{centering}
}\subfloat[RMIG vs. Completeness]{\begin{centering}
\includegraphics[width=0.31\textwidth]{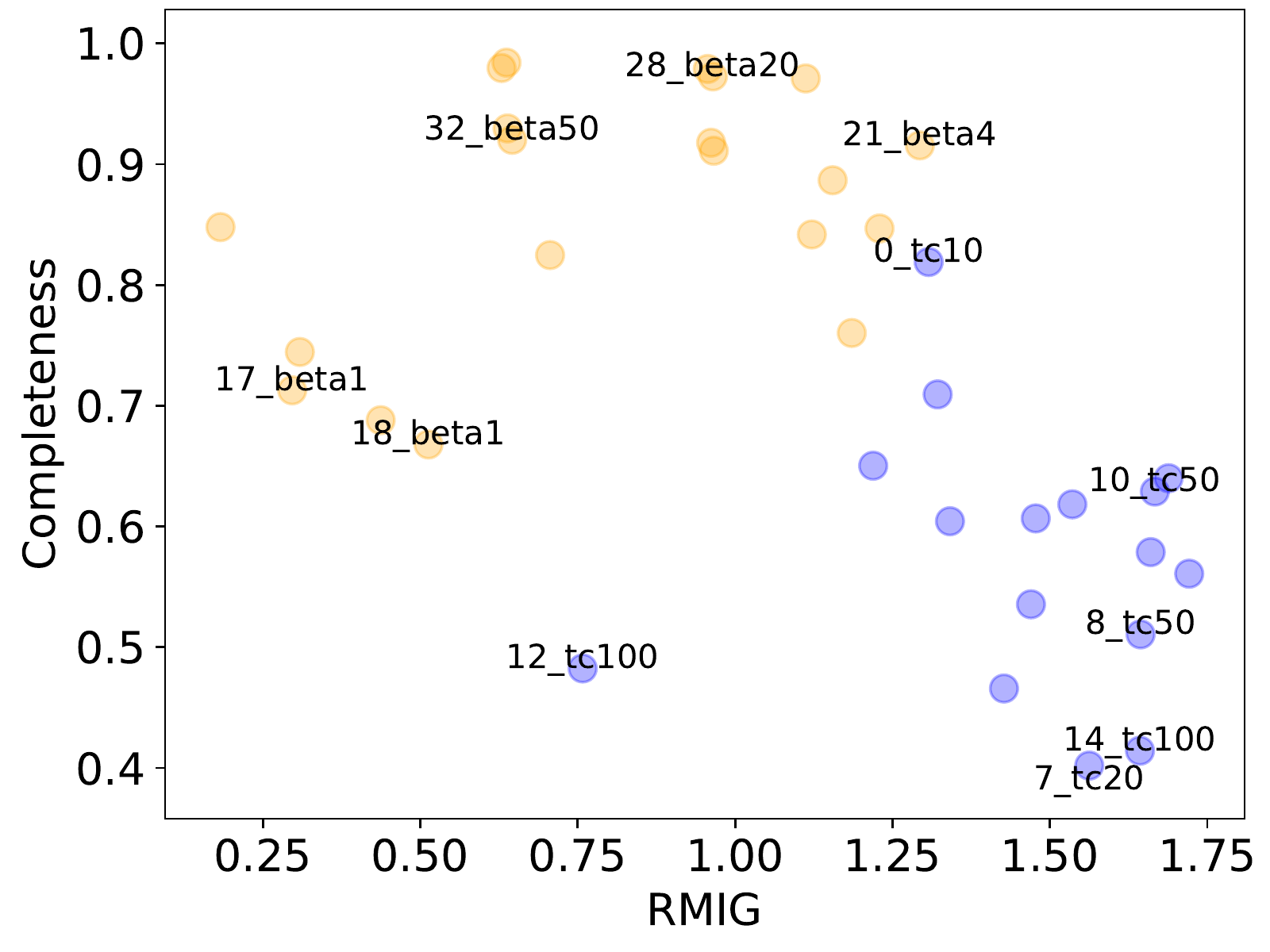}
\par\end{centering}
}\subfloat[RMIG vs. Error]{\begin{centering}
\includegraphics[width=0.31\textwidth]{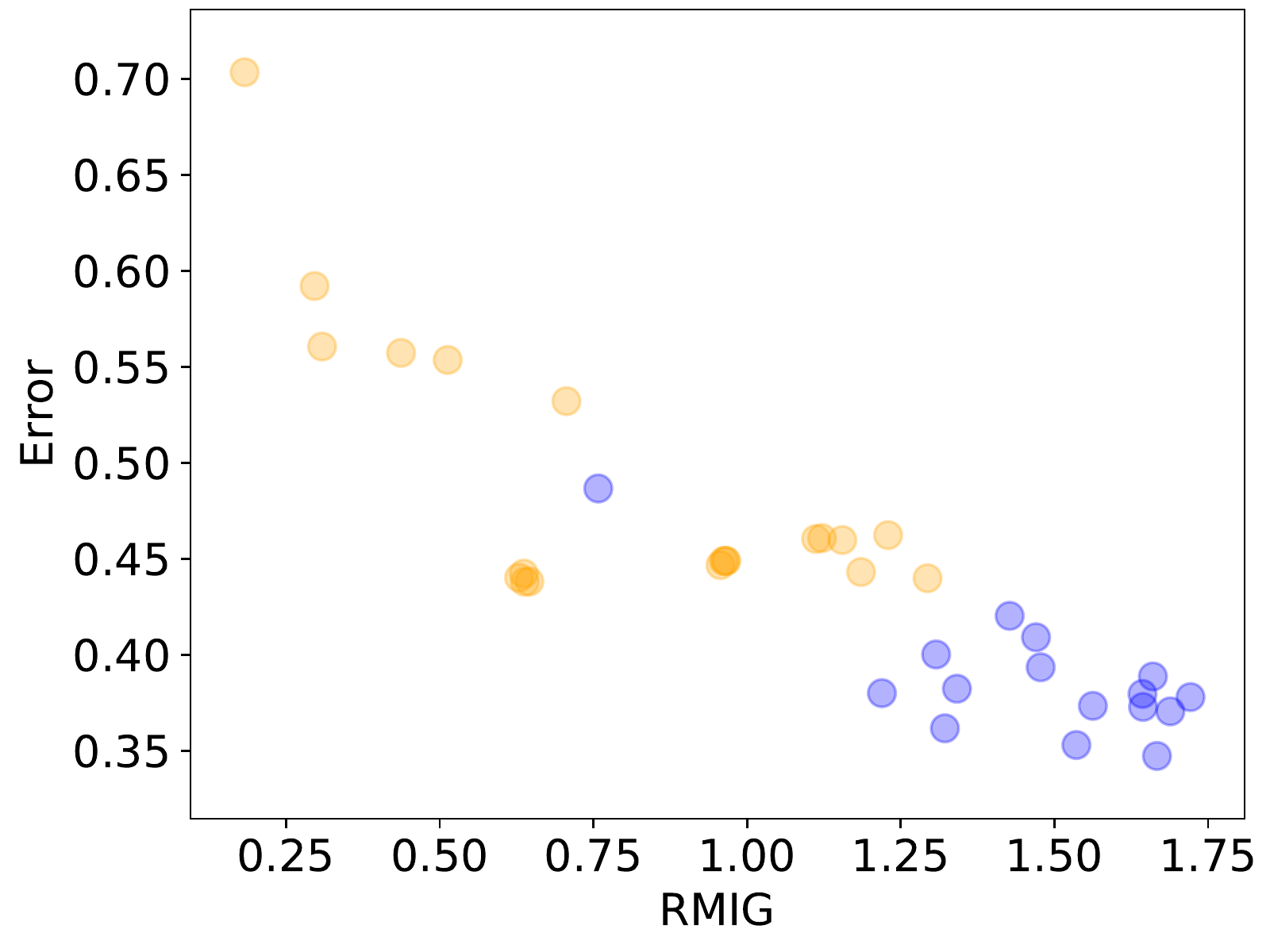}
\par\end{centering}
}
\par\end{centering}
\caption{Comparison between JEMMIG/RMIG and the metrics in \cite{eastwood2018framework}.
Because the competing metrics do not apply for categorical factors
(see Appdx.~\ref{subsec:Analysis-of-existing} for detailed analysis),
we exclude the ``shape'' factor during computation. Following \cite{eastwood2018framework},
we use LASSO classifiers with the L1 coefficient is $\alpha=0.002$.
Blue dots denote FactorVAE models and orange dots denote $\beta$-VAE
models. \label{fig:compare_Eastwood}}
\end{figure}

In Fig.~\ref{fig:compare_Eastwood}, we show the differences in evaluation
results between JEMMIG/RMIG and the metrics in \cite{eastwood2018framework}.
We can easily see that JEMMIG and RMIG are much better than ``disentanglement'',
``completeness'' and ``informativeness'' (or reversed classification
error) in separating FactorVAE and $\beta$-VAE models. Among the
three competing metrics, only ``informativeness'' (or $I(z,y_{k})$)
seems to be correlated with JEMMIG and RMIG. This is understandable
because when most representations are independent in case of FactorVAEs
and $\beta$-VAEs, we have $I(z,y_{k})\approx I(z_{i^{*}},y_{k})\approx I(z_{i^{*}},y_{k})-I(z_{j^{\circ}},y_{k})$.
``Disentanglement'' and ``completeness'', by contrast, are strongly
uncorrelated with JEMMIG and RMIG. While JEMMIG consistently grades
standard VAEs ($\beta=1$) worse than other models (Fig.~\ref{fig:JEMMIG_dSprite}),
``disentanglement'' and ``completeness'' usually grade standard
VAEs better than some FactorVAE models, which seems inappropriate.
Moreover, since ``disentanglement'' and ``completeness'' are not
well aligned, using both of them at the same time may cause confusion.
For example, the model ``28\_beta20'' has lower ``disentanglement''
score yet higher ``completeness'' score than the model ``32\_beta50''
(Figs.~\ref{fig:JEMMIG-vs-Disentanglement} and \ref{fig:JEMMIG-vs-Completeness})
so it is hard to know which model is better than the other at learning
disentangled representations.

\begin{figure}
\begin{centering}
\subfloat[Disentanglement\label{fig:Disentanglement_dSprites}]{\begin{centering}
\includegraphics[width=0.32\textwidth]{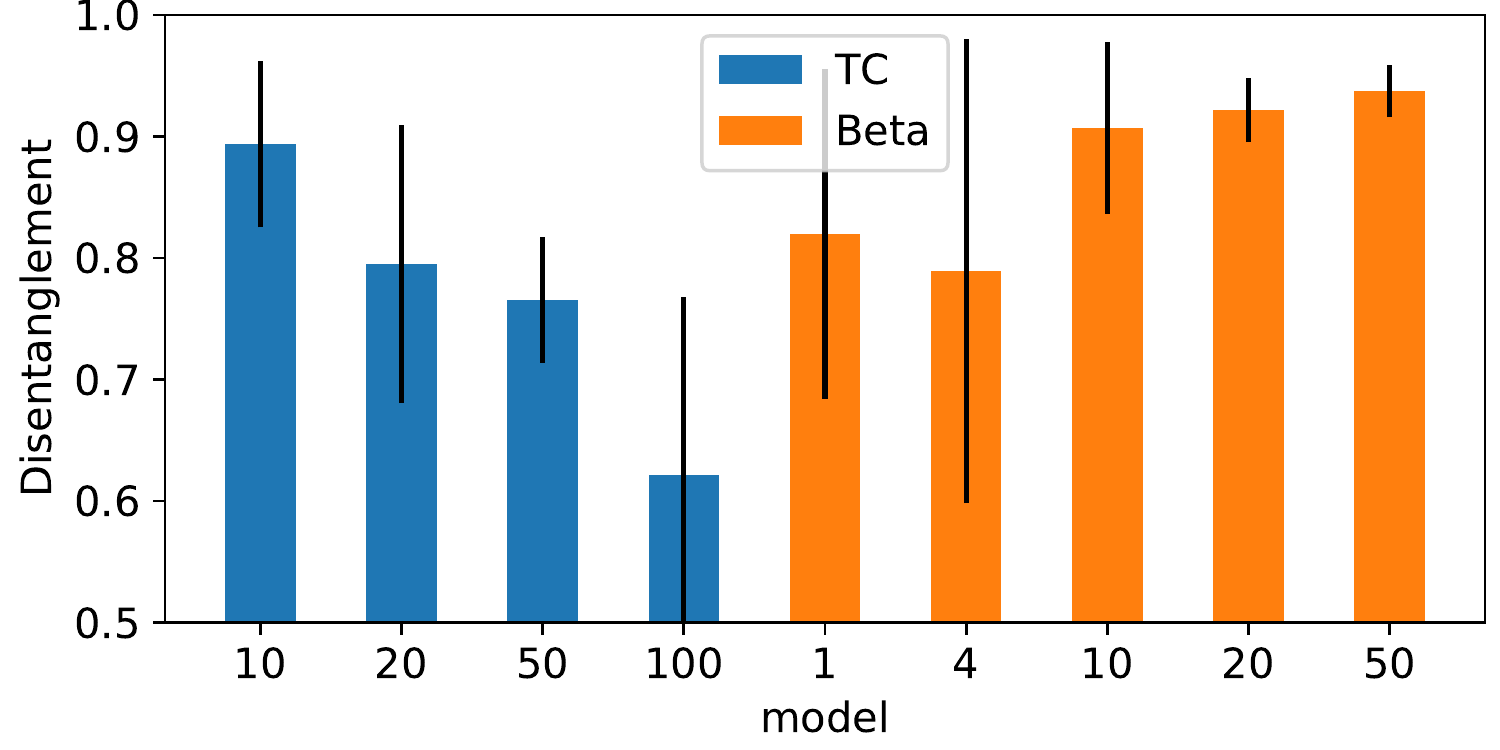}
\par\end{centering}
}\subfloat[Completeness\label{fig:Completeness_dSprites}]{\begin{centering}
\includegraphics[width=0.32\textwidth]{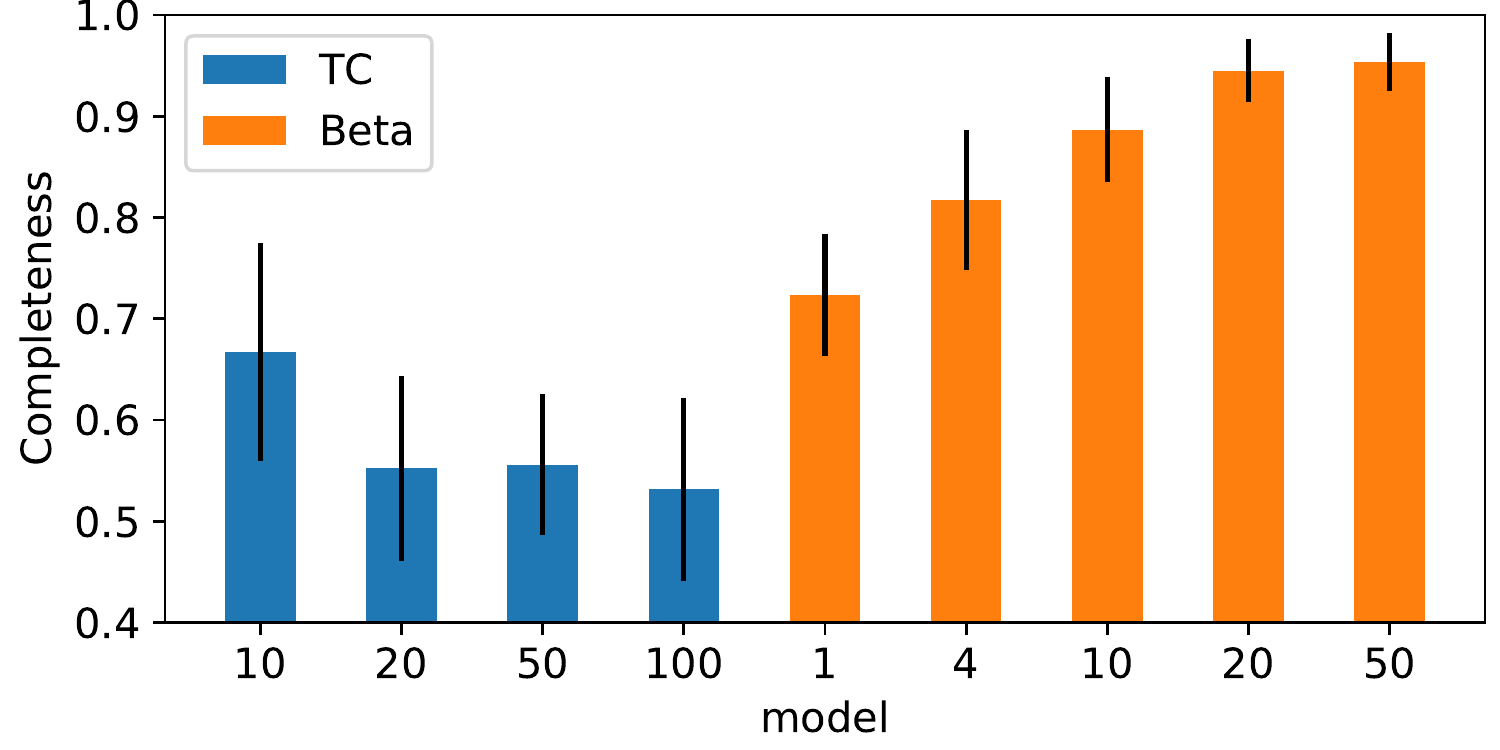}
\par\end{centering}
}\subfloat[Error\label{fig:Informativeness_dSprites}]{\begin{centering}
\includegraphics[width=0.32\textwidth]{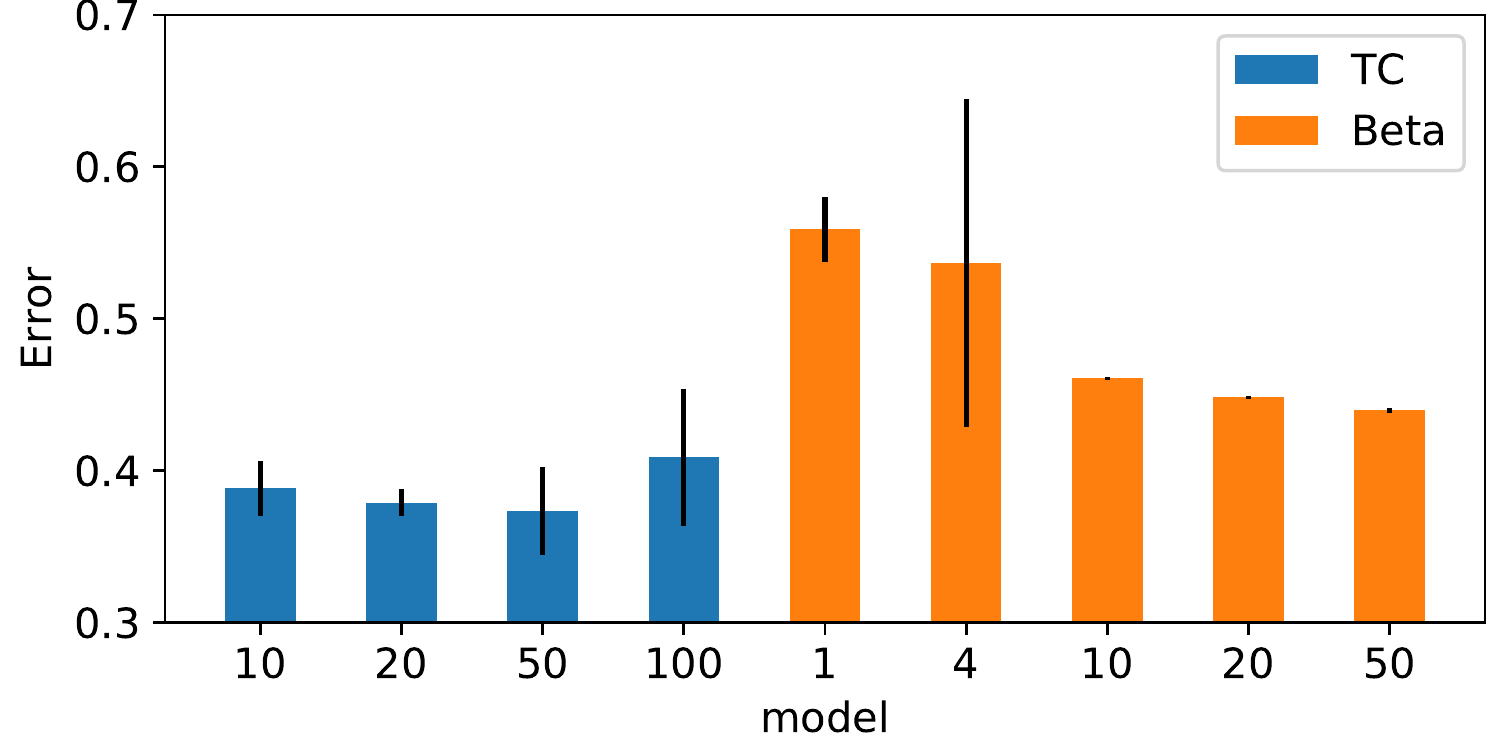}
\par\end{centering}
}
\par\end{centering}
\caption{``disentanglement'', ``completeness'' and ``informativeness''
(error) scores of several FactorVAE and $\beta$-VAE models.\label{fig:disentanglement-completeness-}}
\end{figure}

From Figs.~\ref{fig:Disentanglement_dSprites} and \ref{fig:Completeness_dSprites},
we see that ``disentanglement'' and ``completeness'' blindly favor
$\beta$-VAE models with high $\beta$ without concerning about the
fact that representations in these models are less informative than
representations in FactorVAEs (Fig.~\ref{fig:Informativeness_dSprites}).
Thus, they are not good for characterizing disentanglement in general.

Disentanglement and ``completeness'' are computed based on a weight
matrix with an assumption that the weight magnitudes for noisy representations
are close to 0. However, this assumption is often broken in practice,
thus, may lead to inaccurate results (please check Appdx.~\ref{subsec:Matrices}
for details).

\paragraph{Comparison with ``modularity''}

\begin{figure}
\begin{centering}
\subfloat[JEMMIG vs. Modularity\label{fig:JEMMIG-vs-Modularity}]{\begin{centering}
\includegraphics[width=0.26\textwidth]{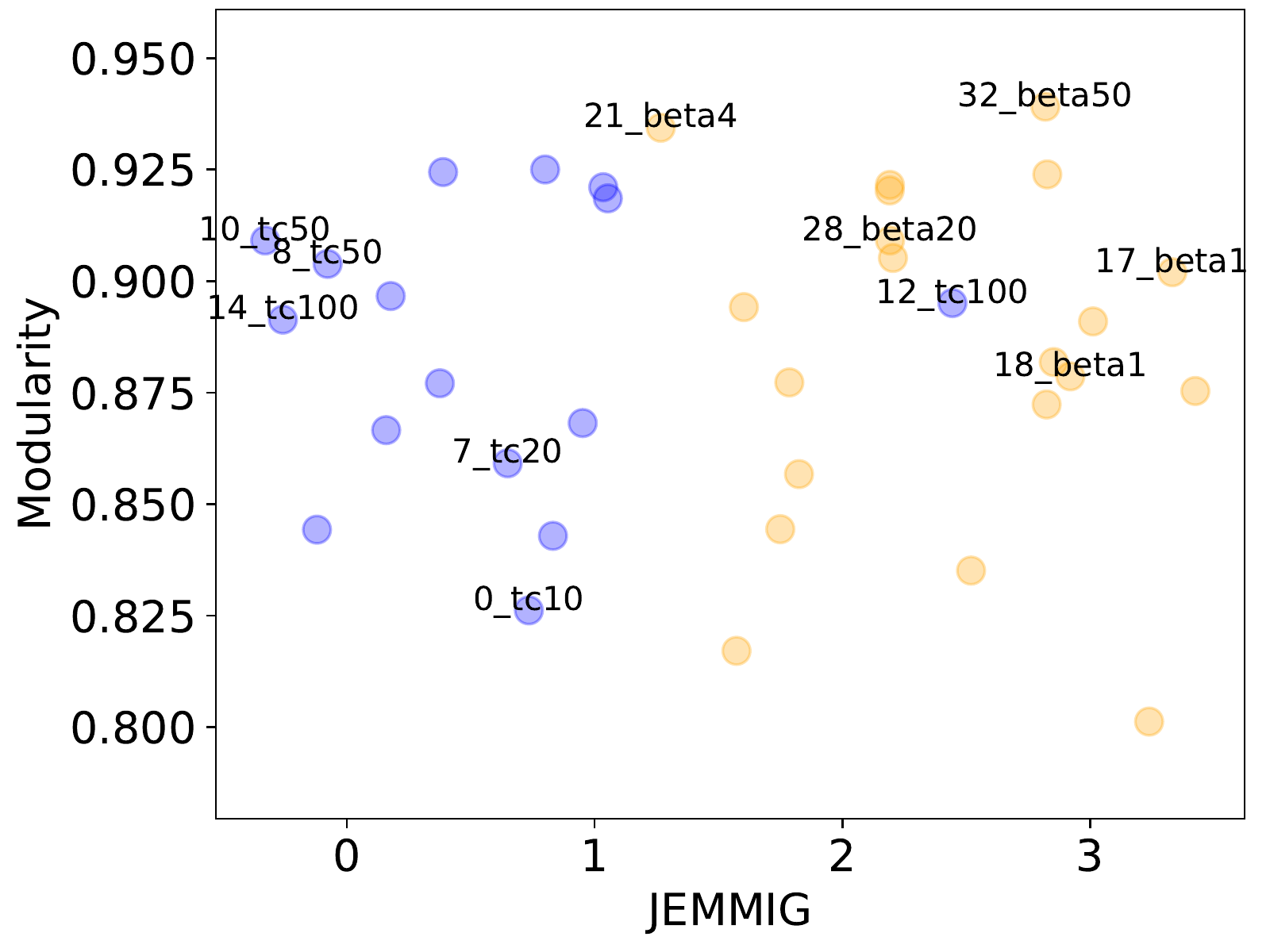}
\par\end{centering}
}\subfloat[Modularity (original)\label{fig:Modularity-(original)}]{\begin{centering}
\includegraphics[width=0.32\textwidth]{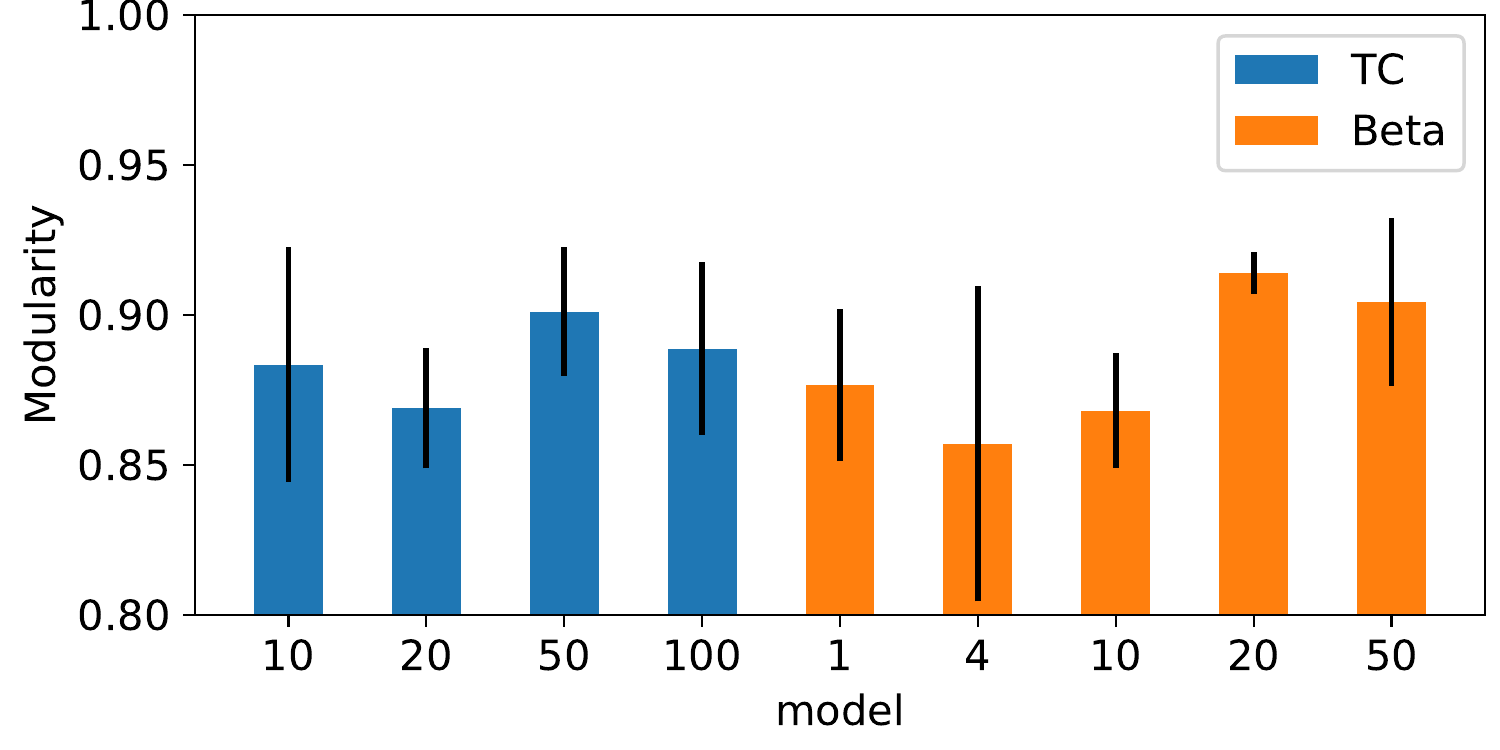}
\par\end{centering}
}\subfloat[Modularity (correct)\label{fig:Modularity-(correct)}]{\begin{centering}
\includegraphics[width=0.32\textwidth]{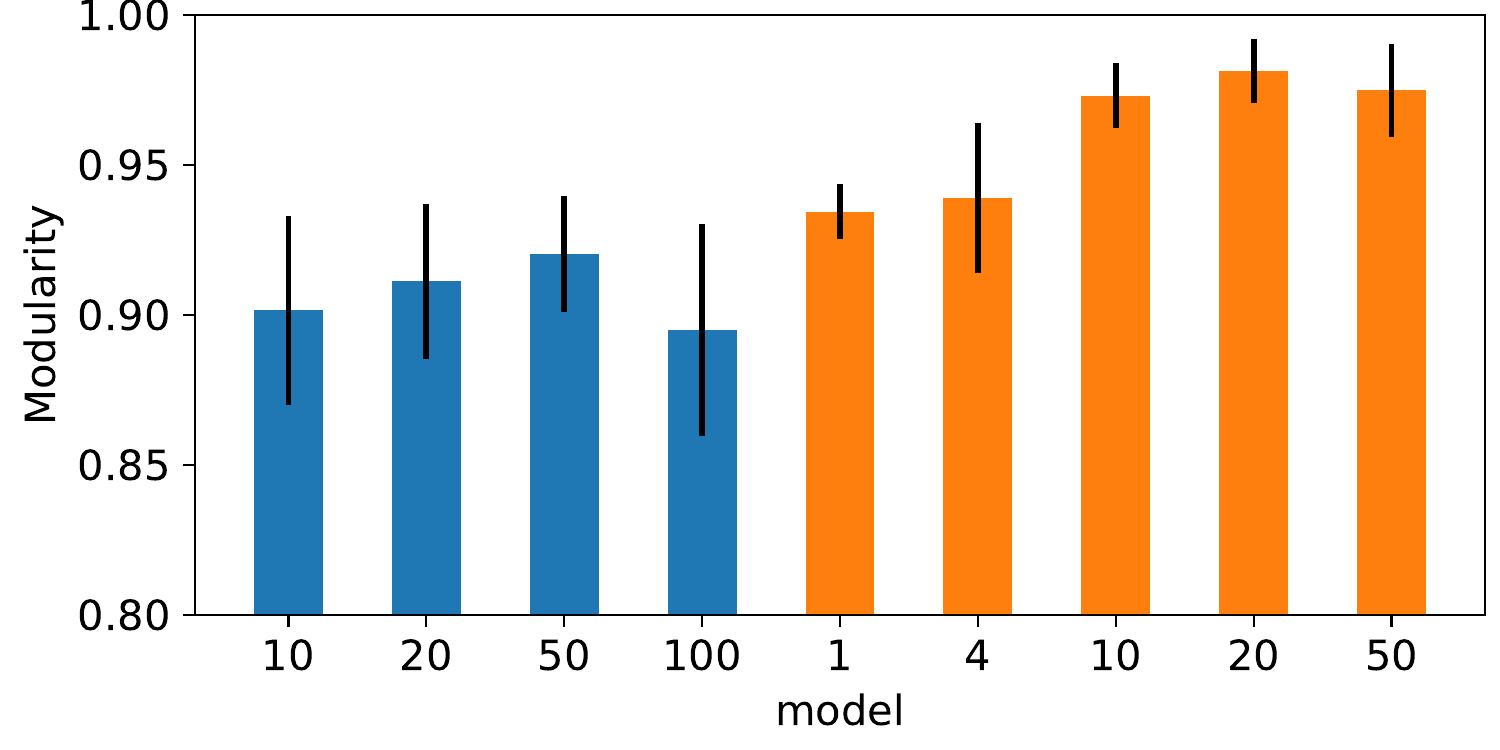}
\par\end{centering}
}
\par\end{centering}
\caption{(a): Comparison between JEMMIG and ``modularity'' (\#bins=100).
(b) and (c): ``modularity'' scores of several FactorVAE and $\beta$-VAE
models. The original version computes $I(z_{i},y_{k})$ using $\protect\Expect_{q(z_{i}|x)}[z_{i}]$
while the correct version compute $I(z_{i},y_{k})$ using $q(z_{i}|x)$.}
\end{figure}

``modularity'' and ``explicitness'' \cite{ridgeway2018learning}
are similar to ``disentanglement'' and ``informativeness'' \cite{eastwood2018framework}
in terms of concept, respectively. However, they are different in
terms of formulation. We exclude ``explicitness'' in our experiment
because computing it on dSprites is time consuming. In Fig.~\ref{fig:JEMMIG-vs-Modularity},
we show the correlation between JEMMIG and ``modularity''. We consider
two versions of ``modularity''. In the first version (Fig.~\ref{fig:Modularity-(original)}),
$I(z_{i},y_{k})$ is computed from the mean of $z_{i}\sim q(z_{i}|x)$.
This is the original implementation provided by \cite{ridgeway2018learning}.
In the second version (Fig.~\ref{fig:Modularity-(correct)}), $I(z_{i},y_{k})$
is computed from $q(z_{i}|x)$. We can see that in either case, ``modularity''
often gives higher scores for standard VAEs than for FactorVAEs. It
means that like ``disentanglement'', ``modularity'' itself does
not fully specify ``disentangled representations'' defined in Section
\ref{subsec:Definition}.

\section{Conclusion}

We have proposed an information-theoretic definition of disentangled
representations and designed robust metrics for evaluation, along
three dimensions: \emph{informativeness}, \emph{separability} and
\emph{interpretability}. We carefully analyze the properties of our
metrics using well known representation learning models namely FactorVAE,
$\beta$-VAE and AAE on both real and toy datasets. Compared with
existing metrics, our metrics are more robust and produce more sensible
evaluations that are compatible with visual results. Based on our
definition of disentangled representation in Section \ref{subsec:Definition},
WSEPIN/JEMMIG are the two key metrics in case ground truth labels
are unavailable/available, respectively.

\bibliographystyle{iclr2020_conference}
\bibliography{disentanglement}

\appendix
\newpage{}

\section{Appendix}

\subsection{Review of FactorVAEs, $\beta$-VAEs and AAEs\label{subsec:Review-of-FactorVAEs}}

Standard VAEs are trained by minimizing the variational upper bound
$\Loss^{\VAE}$ of $-\log p_{\theta}(x)$ as follows:
\begin{equation}
\Loss^{\VAE}=\Expect_{p_{\Data}(x)}\left[\Expect_{q_{\phi}(z|x)}\left[-\log p_{\theta}(x|z)\right]+D_{KL}\left(q_{\phi}(z|x)\|p(z)\right)\right]\label{eq:VAE_loss}
\end{equation}
where $q_{\phi}(z|x)$ is an amortized variational posterior distribution.
However, this objective does not lead to disentangled representations
\cite{higgins2017beta}. 

$\beta$-VAEs \cite{higgins2017beta} penalize the KL term in the
original VAE loss more heavily with a coefficient $\beta\gg1$:
\[
\Loss^{\beta\text{-VAE}}=\Expect_{p_{\Data}(x)}\left[\Expect_{q_{\phi}(z|x)}\left[-\log p_{\theta}(x|z)\right]+\beta D_{KL}\left(q_{\phi}(z|x)\|p(z)\right)\right]
\]
Since $\Expect_{p_{\Data}(x)}\left[D_{KL}\left(q_{\phi}(z|x)\|p(z)\right)\right]=I_{\phi}(x,z)+D_{KL}\left(q_{\phi}(z)\|p(z)\right)$,
more penalty on the KL term encourages $q_{\phi}(z)$ to be factorized
but also forces $z$ to discard more information in $x$. 

FactorVAEs \cite{kim2018disentangling} add a constraint to the standard
VAE loss to explicitly impose factorization of $q_{\phi}(z)$:
\begin{equation}
\Loss^{\text{FactorVAE}}=\Loss^{\VAE}+\gamma D_{KL}\left(q_{\phi}(z)\|\prod_{i}q_{i}(z_{i})\right)\label{eq:FactorVAE_loss}
\end{equation}
where $D_{KL}\left(q_{\phi}(z)\|\prod_{i}q_{i}(z_{i})\right)\geq0$
is known as the \emph{total correlation} (TC) of $z$. Intuitively,
$\gamma$ can be large without affecting the mutual information between
$z$ and $x$, making FactorVAE more robust than $\beta$-VAE in learning
disentangled representations. Other related models that share similar
ideas with FactorVAEs are are $\beta$-TCVAEs \cite{chen2018isolating}
and DIP-VAEs \cite{kumar2017variational}.

The loss of AAEs \cite{makhzani2015adversarial} is derived from the
standard VAE loss by removing the term $I_{\phi}(x,z)$:
\[
\Loss^{\text{AAE}}=\Expect_{p_{\Data}(x)}\left[\Expect_{q_{\phi}(z|x)}\left[-\log p_{\theta}(x|z)\right]\right]+D_{KL}\left(q_{\phi}(z)\|p(z)\right)
\]
Different from the losses of $\beta$-VAEs and FactorVAEs, AAE loss
is not a valid upper bound on $-\log p_{\theta}(x)$.

\subsection{Datasets\label{subsec:Dataset-Details}}

The CelebA dataset \cite{liu2015faceattributes} consists of more
than 200 thousands face images with 40 binary attributes. We resize
these images to $64\times64$. The dSprites dataset \cite{dsprites2017}
is a toy dataset generated from 5 different factors of variation which
are ``shape'' (3 values), ``scale'' (6 values), ``rotation''
(40 values), ``x-position'' (32 values), ``y-position'' (32 values).
Statistics of these datasets are provided in Table~\ref{tab:Summary-of-datasets}.

\begin{table}[H]
\begin{centering}
\par\end{centering}
\begin{centering}
\begin{tabular}{|c|c|c|c|}
\hline 
Dataset & \#Train & \#Test & Image size\tabularnewline
\hline 
\hline 
CelebA & 162,770 & 19,962 & 64$\times$64$\times$3\tabularnewline
\hline 
dSprites & 737,280 & 0 & 64$\times$64$\times$1\tabularnewline
\hline 
\end{tabular}
\par\end{centering}
\caption{Summary of datasets used in experiments.\label{tab:Summary-of-datasets}}
\end{table}

\subsection{Model settings\label{subsec:Model-settings}}

For FactorVAE, $\beta$-VAE and AAE, we used the same architectures
for the encoder and decoder (see Table \ref{tab:CelebA_arch} and
Table \ref{tab:dSprites_arch}\footnote{Only FactorVAE and AAE use a discriminator over $z$}),
following \cite{kim2018disentangling}. We trained the models for
300 epochs with mini-batches of size 64. The learning rate is $10^{-3}$
for the encoder/decoder and is $10^{-4}$ for the discriminator over
$z$. We used Adam \cite{kingma2014adam} optimizer with $\beta_{1}=0.5$
and $\beta_{2}=0.99$. Unless explicitly mentioned, we use the following
default settings: i) for CelebA: the number of latent variables is
65, the TC coefficient in FactorVAE is 50, the value for $\beta$
in $\beta$-VAE is 50, and the coefficient for the generator loss
over $z$ in AAE is 50; ii) for dSprites: the number of latent variables
is 10.

\begin{table}[H]
\begin{centering}
{\scriptsize{}}%
\begin{tabular}{|c|c|c|}
\hline 
Encoder & Decoder & Discriminator Z\tabularnewline
\hline 
\hline 
{\small{}$x$ dims: 64$\times$64$\times$3} & {\small{}$z$ dim: 65} & {\small{}$z$ dim: 65}\tabularnewline
\hline 
{\small{}conv (4, 4, 32), stride $2$, ReLU} & {\small{}FC 1$\times$1$\times$256, ReLU} & {\small{}5$\times${[}FC 1000, LReLU{]}}\tabularnewline
\hline 
{\small{}conv (4, 4, 32), stride 2, ReLU} & {\small{}deconv (4, 4, 64), stride 1, valid, ReLU } & {\small{}FC 1}\tabularnewline
\hline 
{\small{}conv (4, 4, 64), stride 2, ReLU} & {\small{}deconv (4, 4, 64), stride 2, ReLU} & {\small{}$D(z)$: 1}\tabularnewline
\hline 
{\small{}conv (4, 4, 64), stride 2, ReLU} & {\small{}deconv (4, 4, 32), stride 2, ReLU} & \tabularnewline
\hline 
{\small{}conv (4, 4, 256), stride 1, valid, ReLU} & {\small{}deconv (4, 4, 32), stride 2, ReLU} & \tabularnewline
\hline 
{\small{}FC 65} & {\small{}deconv (4, 4, 3), stride 2, ReLU} & \tabularnewline
\hline 
{\small{}$z$ dim: 65} & {\small{}$x$ dim: 64$\times$64$\times$3} & \tabularnewline
\hline 
\end{tabular}{\scriptsize\par}
\par\end{centering}
\caption{Model architectures for CelebA.\label{tab:CelebA_arch}}
\end{table}

\begin{table}[H]
\begin{centering}
{\scriptsize{}}%
\begin{tabular}{|c|c|c|}
\hline 
Encoder & Decoder & Discriminator Z\tabularnewline
\hline 
\hline 
{\small{}$x$ dims: 64$\times$64$\times$1} & {\small{}$z$ dim: 10} & {\small{}$z$ dim: 10}\tabularnewline
\hline 
{\small{}conv (4, 4, 32), stride $2$, ReLU} & {\small{}FC 128, ReLU} & {\small{}5$\times${[}FC 1000, LReLU{]}}\tabularnewline
\hline 
{\small{}conv (4, 4, 32), stride 2, ReLU} & {\small{}FC 4$\times$4$\times$64, ReLU } & {\small{}FC 1}\tabularnewline
\hline 
{\small{}conv (4, 4, 64), stride 2, ReLU} & {\small{}deconv (4, 4, 64), stride 2, ReLU} & {\small{}$D(z)$: 1}\tabularnewline
\hline 
{\small{}conv (4, 4, 64), stride 2, ReLU} & {\small{}deconv (4, 4, 32), stride 2, ReLU} & \tabularnewline
\hline 
{\small{}FC 128, ReLU} & {\small{}deconv (4, 4, 32), stride 2, ReLU} & \tabularnewline
\hline 
{\small{}FC 10} & {\small{}deconv (4, 4, 1), stride 2, ReLU} & \tabularnewline
\hline 
{\small{}$z$ dim: 10} & {\small{}$x$ dim: 64$\times$64$\times$1} & \tabularnewline
\hline 
\end{tabular}{\scriptsize\par}
\par\end{centering}
\caption{Model architecture for dSprites.\label{tab:dSprites_arch}}
\end{table}

\subsection{Consistence between quantitative and qualitative results\label{subsec:Consistence-between-quantiative}}

\subsubsection{CelebA}

\paragraph{Informativeness}

\begin{figure}[H]
\begin{centering}
\subfloat[FactorVAE (TC=50)]{\begin{centering}
\includegraphics[width=0.32\textwidth]{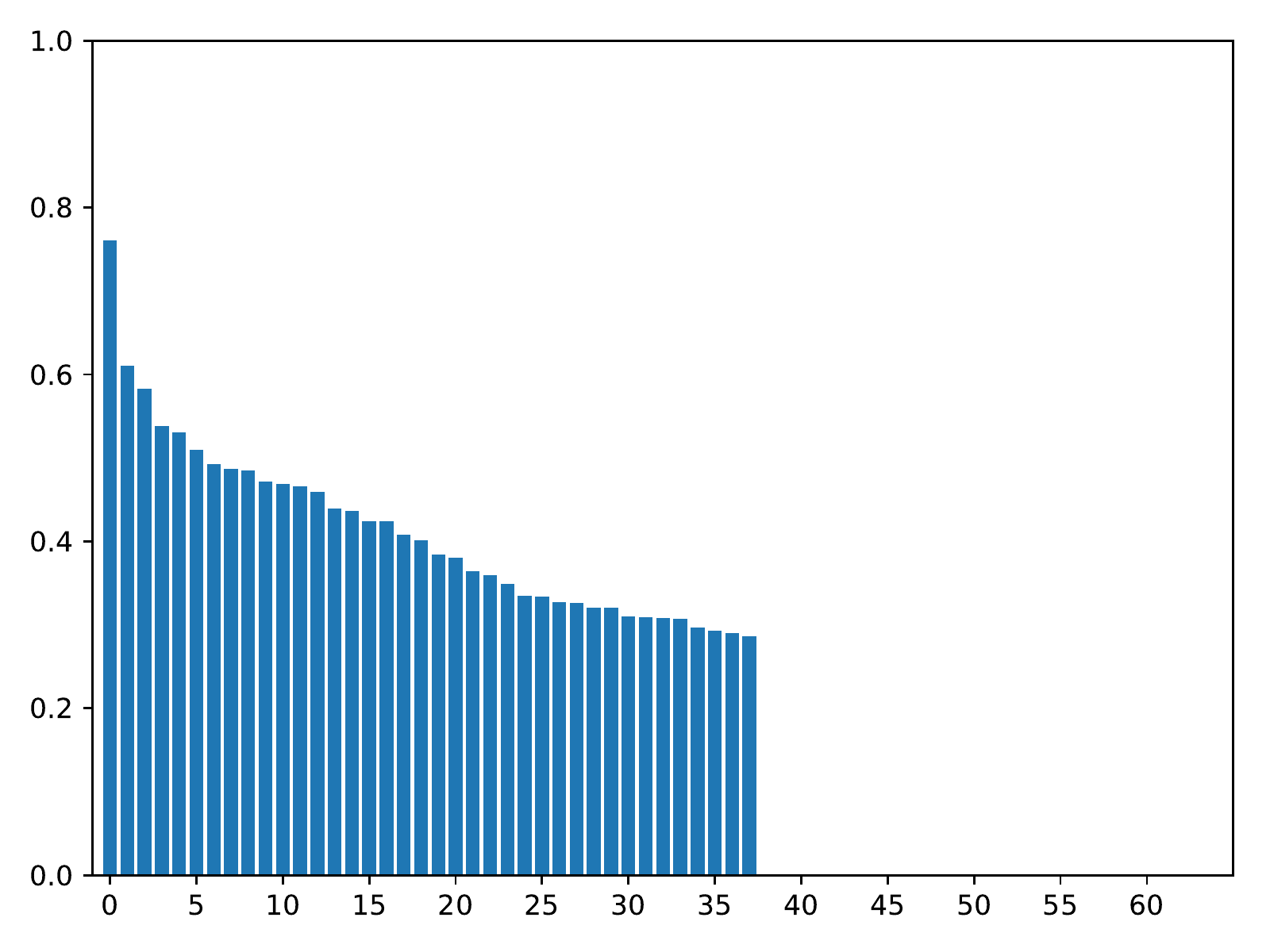}
\par\end{centering}
}\subfloat[$\beta$-VAE ($\beta$=50)]{\begin{centering}
\includegraphics[width=0.32\textwidth]{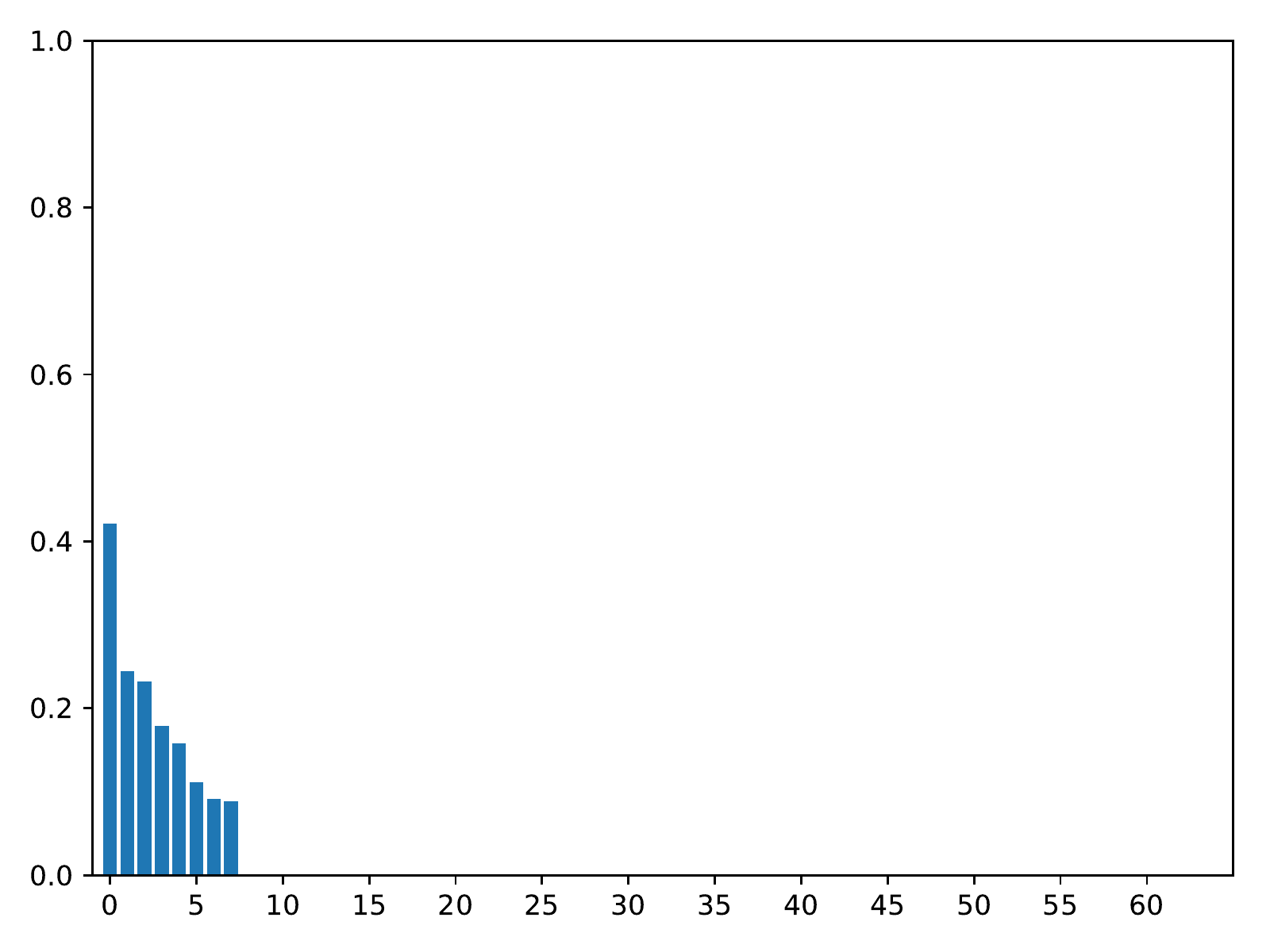}
\par\end{centering}
}\subfloat[AAE (Gz=50)]{\begin{centering}
\includegraphics[width=0.32\textwidth]{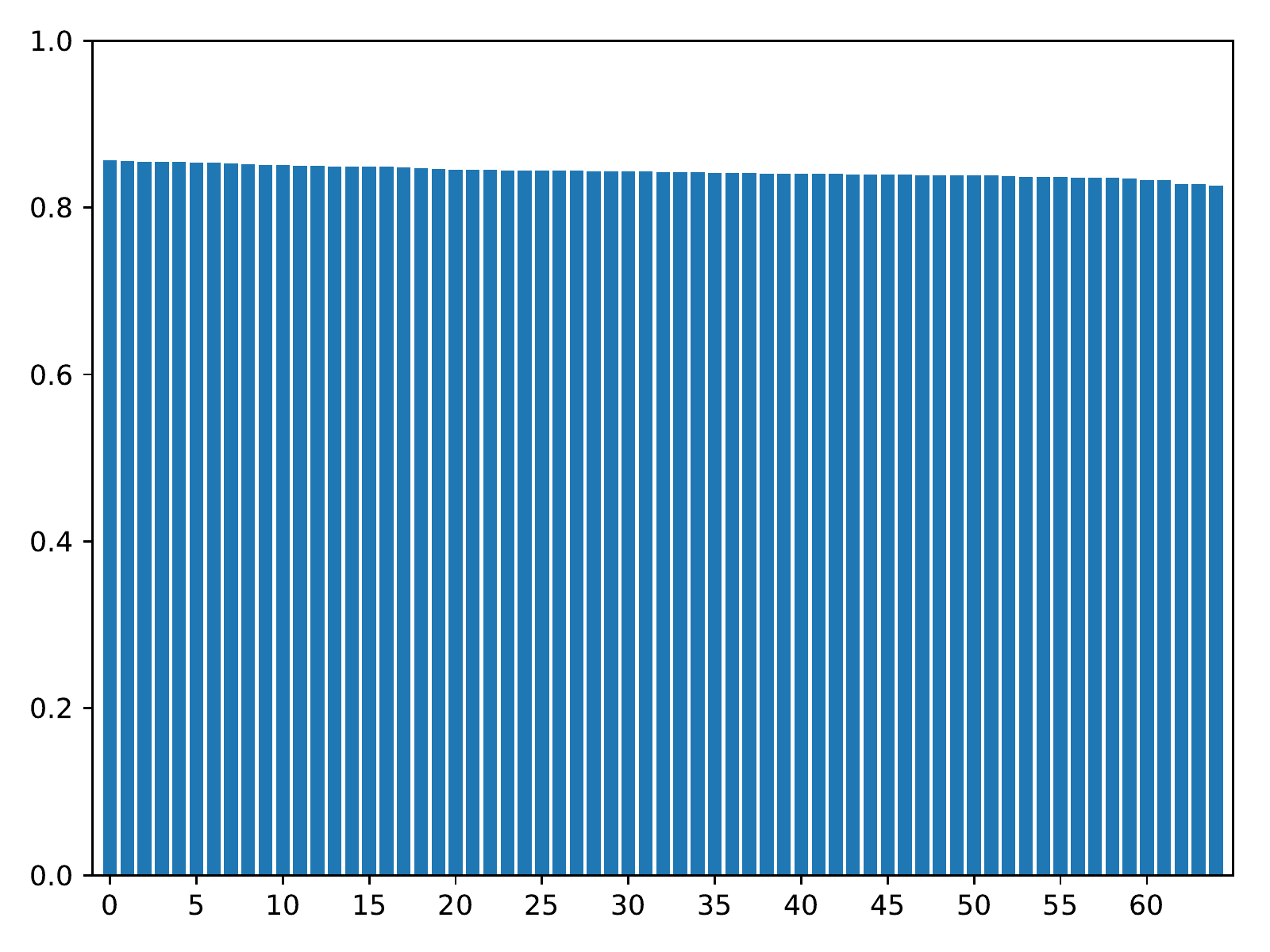}
\par\end{centering}
}
\par\end{centering}
\caption{Normalized informativeness scores (bins=100) of all latent variables
sorted in descending order.\label{fig:Normalized-informativeness-score}}
\end{figure}

We sorted the representations of different models according to their
informativeness scores in the descending order and plot the results
in Fig.~\ref{fig:Normalized-informativeness-score}. There are distinct
patterns for different methods. AAE captures equally large amounts
of information from the data while FactorVAE and $\beta$-VAE capture
smaller and varying amounts. This is because FactorVAE and $\beta$-VAE
penalize the informativeness of representations while AAE does not.
Recall that $I(z_{i},x)=H(z_{i})-H(z_{i}|x)$. For AAE, $H(z_{i}|x)=0$
and $H(z_{i})$ is equal to the entropy of $\Normal(0,\Irm)$. For
FactorVAE and $\beta$-VAE, $H(z_{i}|x)>0$ and $H(z_{i})$ is usually
smaller than the entropy of $\Normal(0,\Irm)$ due to a narrow $q(z_{i})$\footnote{Note that $H(z_{i})$ does not depend on whether $q(z_{i})$ is zero-centered
or not}. 

In Fig.~\ref{fig:Normalized-informativeness-score}, we see a sudden
drop of the scores to 0 for some FactorVAE's and $\beta$-VAE's representations.
These representations $z_{i}$ are totally random and contain no information
about the data (i.e., $q(z_{i}|x)\approx\Normal(0,\Irm)$). We call
them ``noisy'' representations and provide discussions in Appdx.~\ref{subsec:Trade-off-between-informativenes}.

\begin{figure}[H]
\begin{centering}
\subfloat[FactorVAE (TC=50)]{\begin{centering}
\includegraphics[width=0.32\textwidth]{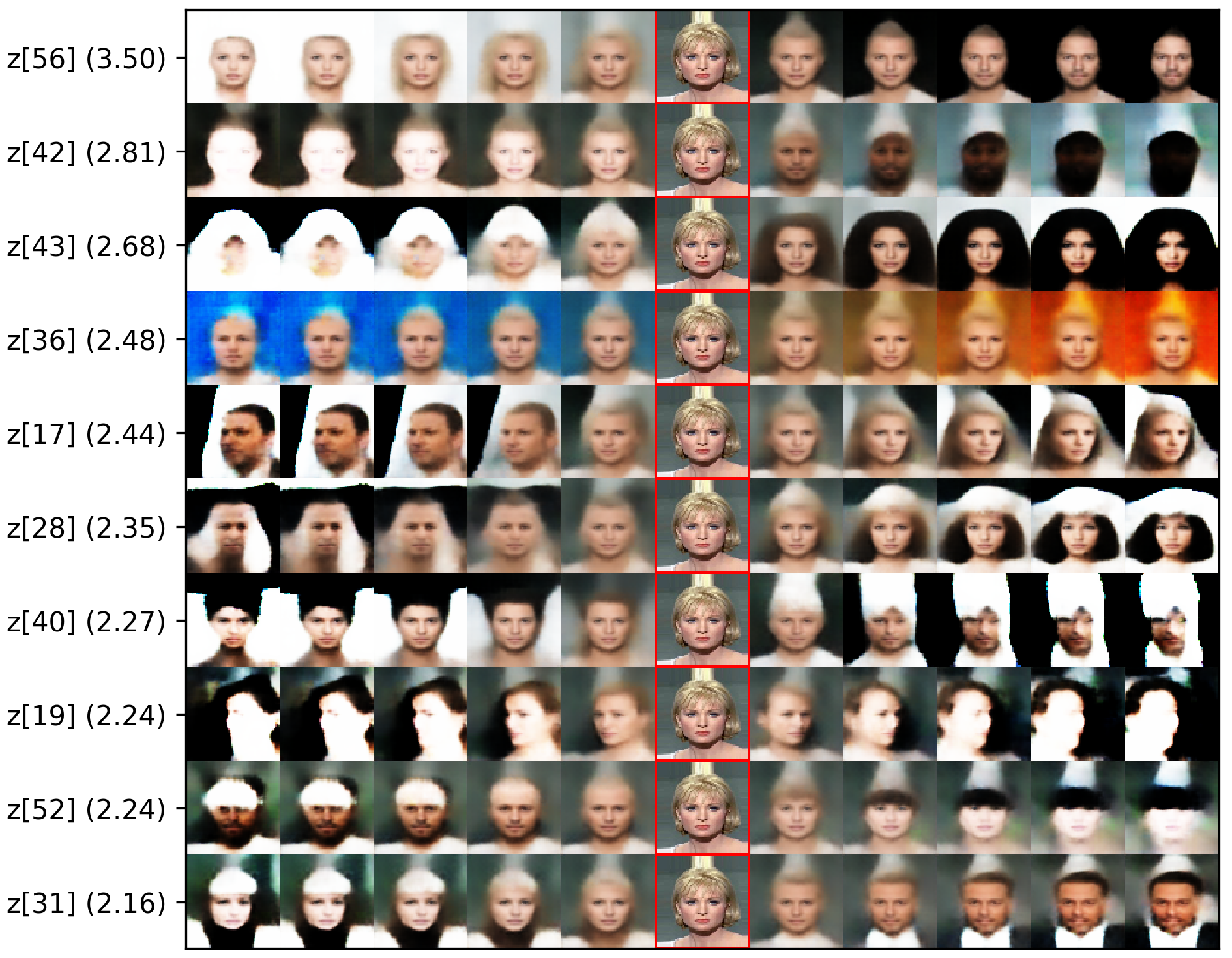}
\par\end{centering}
}\subfloat[$\beta$-VAE ($\beta$=50)]{\begin{centering}
\includegraphics[width=0.32\textwidth]{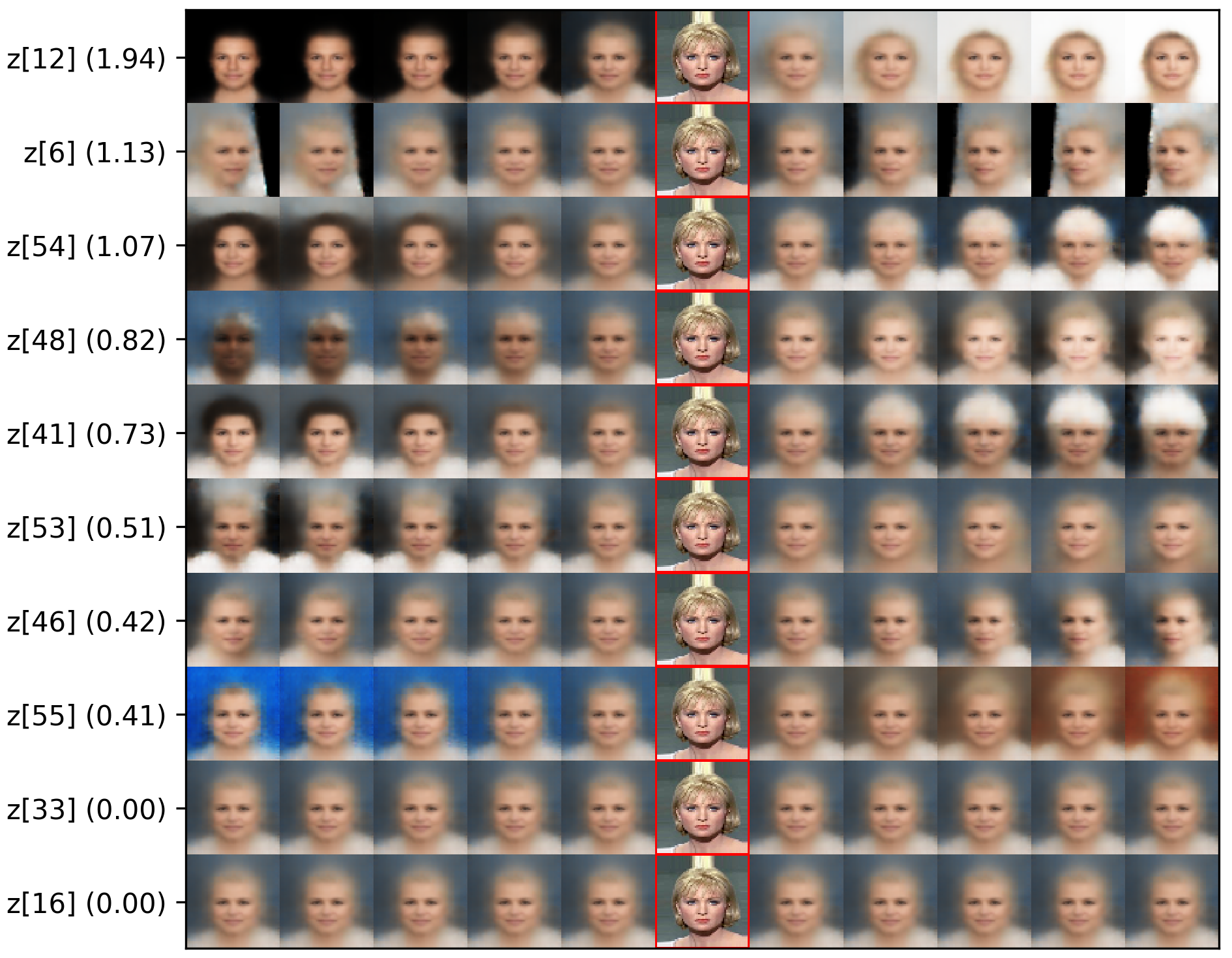}
\par\end{centering}
}\subfloat[AAE (Gz=50)]{\begin{centering}
\includegraphics[width=0.32\textwidth]{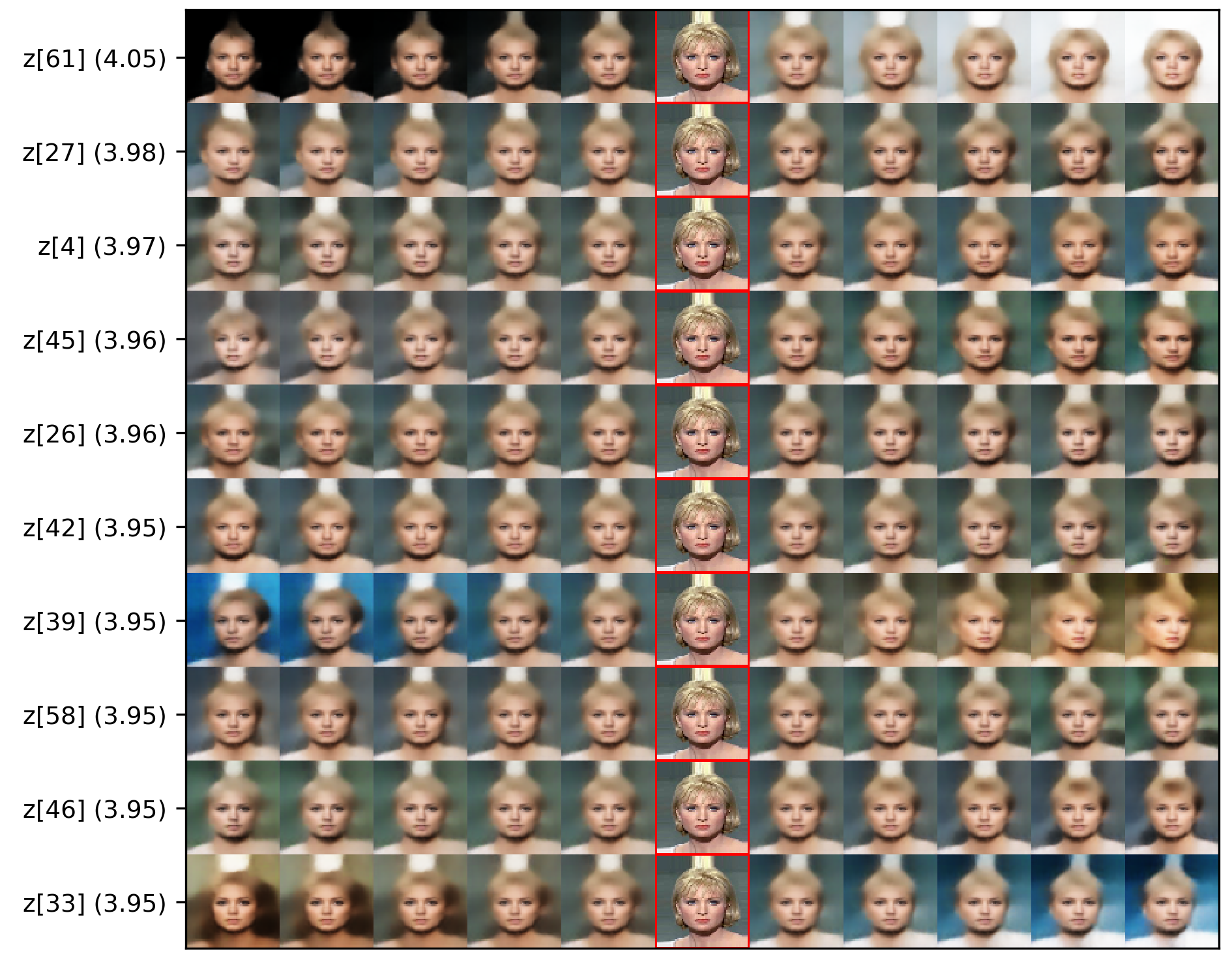}
\par\end{centering}
}
\par\end{centering}
\caption{Visualization of the top informative representations. Scores are unnormalized.\label{fig:info_factors}}
\end{figure}

We visualize the top 10 most informative representations for these
models in Fig.~\ref{fig:info_factors}. AAE's representations are
more detailed than FactorVAE's and $\beta$-VAE's, suggesting the
effect of high informativeness. However, AAE's representations mainly
capture information within the support of $p_{\Data}(x)$. This explains
why we still see a face when interpolating AAE's representations.
By contrast, FactorVAE's and $\beta$-VAE's representations usually
contain information outside the support of $p_{\Data}(x)$. Thus,
when we interpolate these representations, we may see something not
resembling a face.

\paragraph{Separability and Independence}

\begin{figure}[H]
\begin{centering}
\subfloat[FactorVAE (TC=50)]{\begin{centering}
\includegraphics[width=0.3\textwidth]{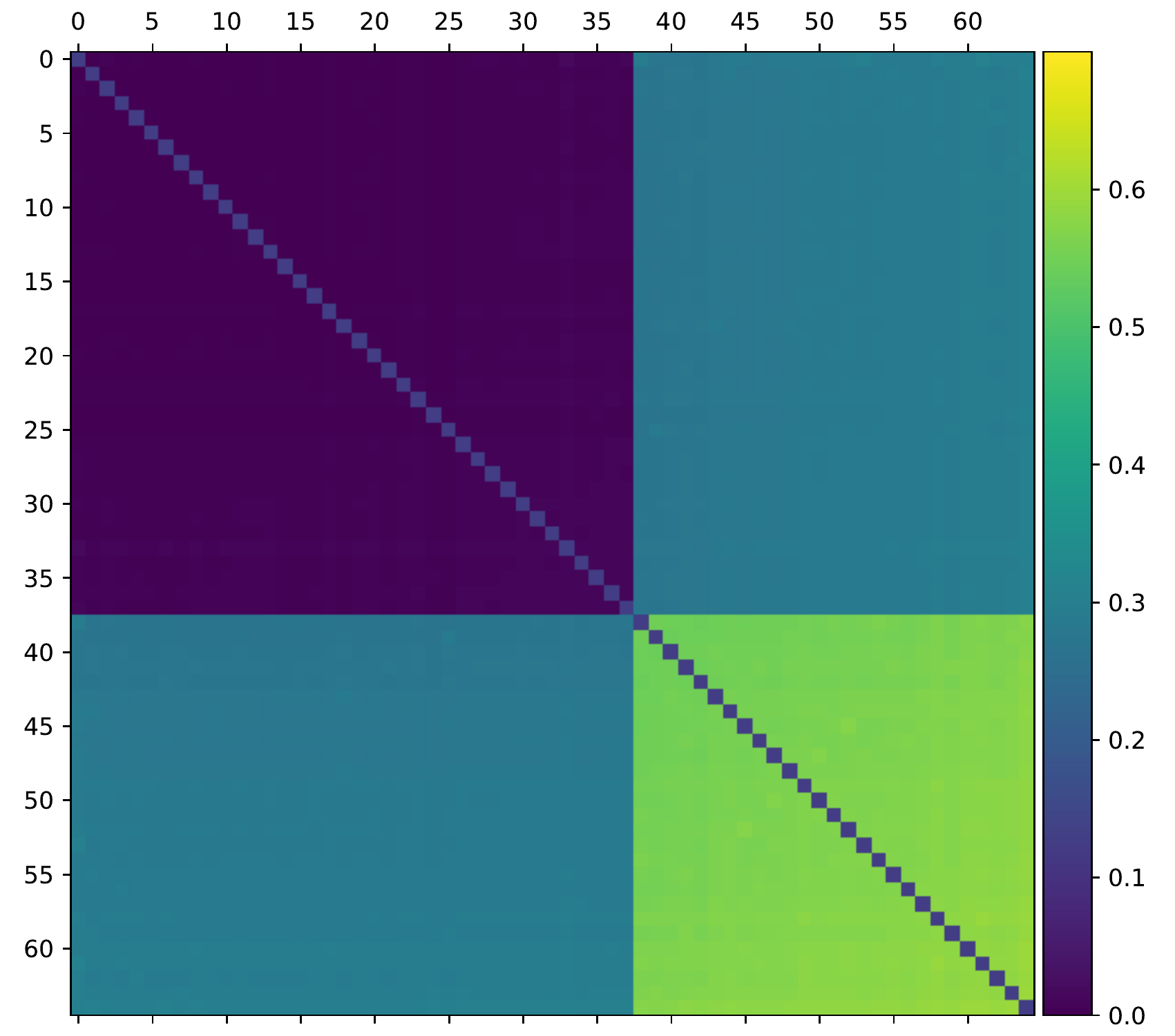}
\par\end{centering}
}\subfloat[$\beta$-VAE ($\beta$=50)]{\begin{centering}
\includegraphics[width=0.3\textwidth]{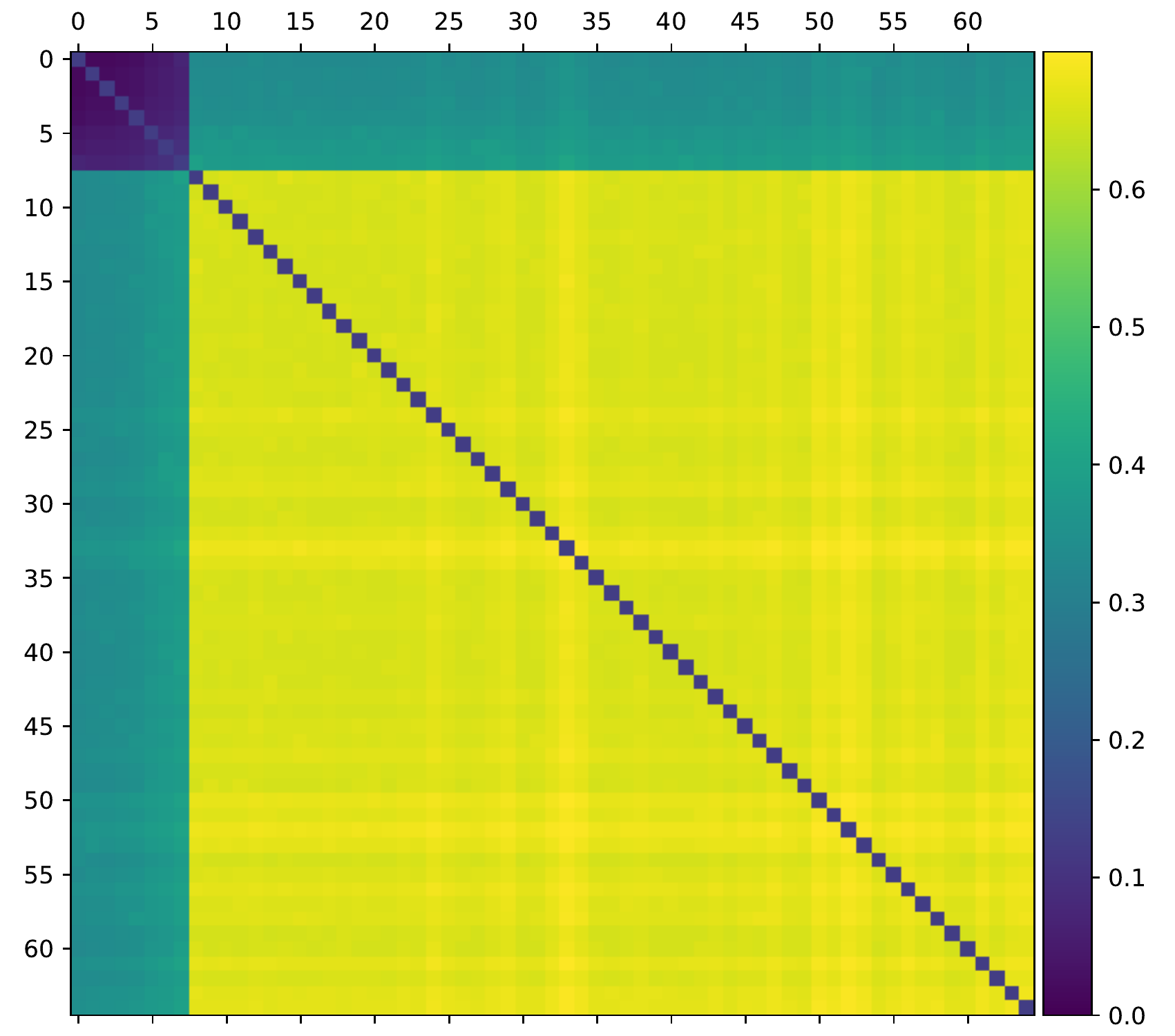}
\par\end{centering}
}\subfloat[AAE (Gz=50)]{\begin{centering}
\includegraphics[width=0.3\textwidth]{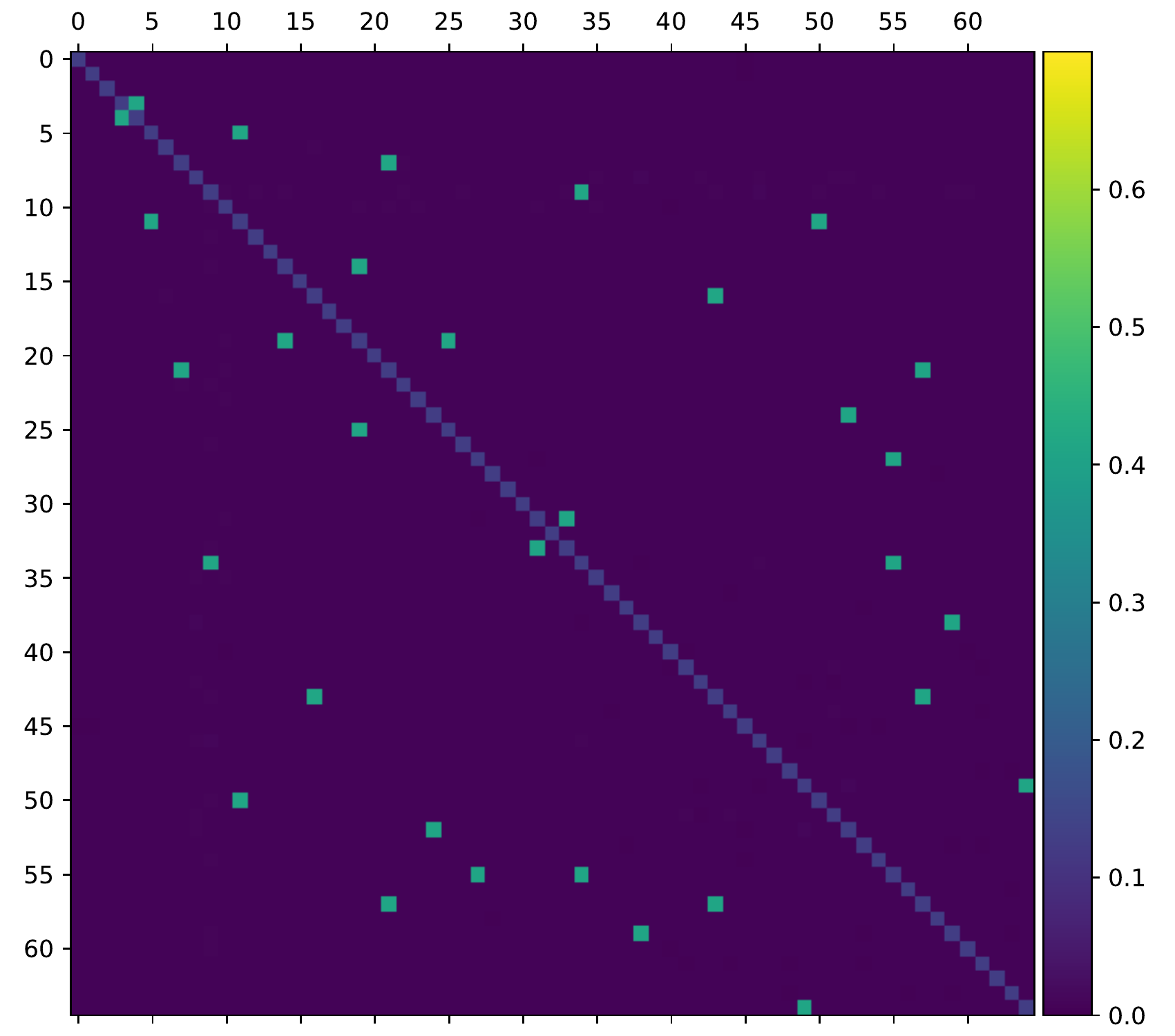}
\par\end{centering}
}
\par\end{centering}
\caption{Normalized MISJED scores of all latent pairs sorted by their informativeness.\label{fig:Normalized-MISJED-scores}}
\end{figure}

Table~\ref{tab:unnorm_MISJED_with_order} reports MISJED scores (Section
\ref{subsec:Metrics-for-independence}) for the top most informative
representations. FactorVAE achieves the lowest MISJED scores, AAE
comes next and $\beta$-VAE is the worst. We argue that this is because
FactorVAE learns independent and nearly deterministic representations,
$\beta$-VAE learns strongly independent yet highly stochastic representations,
and AAE, on the other extreme side, learns strongly deterministic
yet not very independent representations. From Table~\ref{tab:unnorm_MISJED_with_order}
and Fig.~\ref{fig:Normalized-MISJED-scores}, it is clear that MISJED
produces correct orders among pairs of representations according to
their informativeness.

\begin{table}[H]
\begin{centering}
\begin{tabular}{|c|c|c||c|c||c|c|}
\hline 
\multirow{2}{*}{} & \multicolumn{6}{c|}{MISJED (unnormalized)}\tabularnewline
\cline{2-7} 
 & $z_{1},z_{2}$ & $z_{1},z_{3}$ & $z_{1},z_{-1}$ & $z_{1},z_{-2}$ & $z_{-1},z_{-2}$ & $z_{-1},z_{-3}$\tabularnewline
\hline 
\hline 
FactorVAE & \textbf{0.008} & \textbf{0.009} & 2.476 & 2.443 & 4.858 & 4.892\tabularnewline
\hline 
$\beta$-VAE & 0.113 & 0.131 & 3.413 & 3.401 & 6.661 & 6.739\tabularnewline
\hline 
AAE & 0.022 & 0.023 & 0.022 & 0.021 & 0.021 & 0.020\tabularnewline
\hline 
\end{tabular}
\par\end{centering}
\caption{Unnormalized MISJED scores (\#bins = 50, 10\% data). $z_{1},z_{2},z_{3}$
and $z_{-1},z_{-2},z_{-3}$ denote the top 3 and the bottom 3 latent
variables sorted by the informativeness scores in descending order.
Boldness indicates best results.\label{tab:unnorm_MISJED_with_order}}
\end{table}

\paragraph{Interpretability}

\begin{figure}[H]
\begin{centering}
\subfloat[FactorVAE (TC=50)]{\begin{centering}
\includegraphics[width=0.33\textwidth]{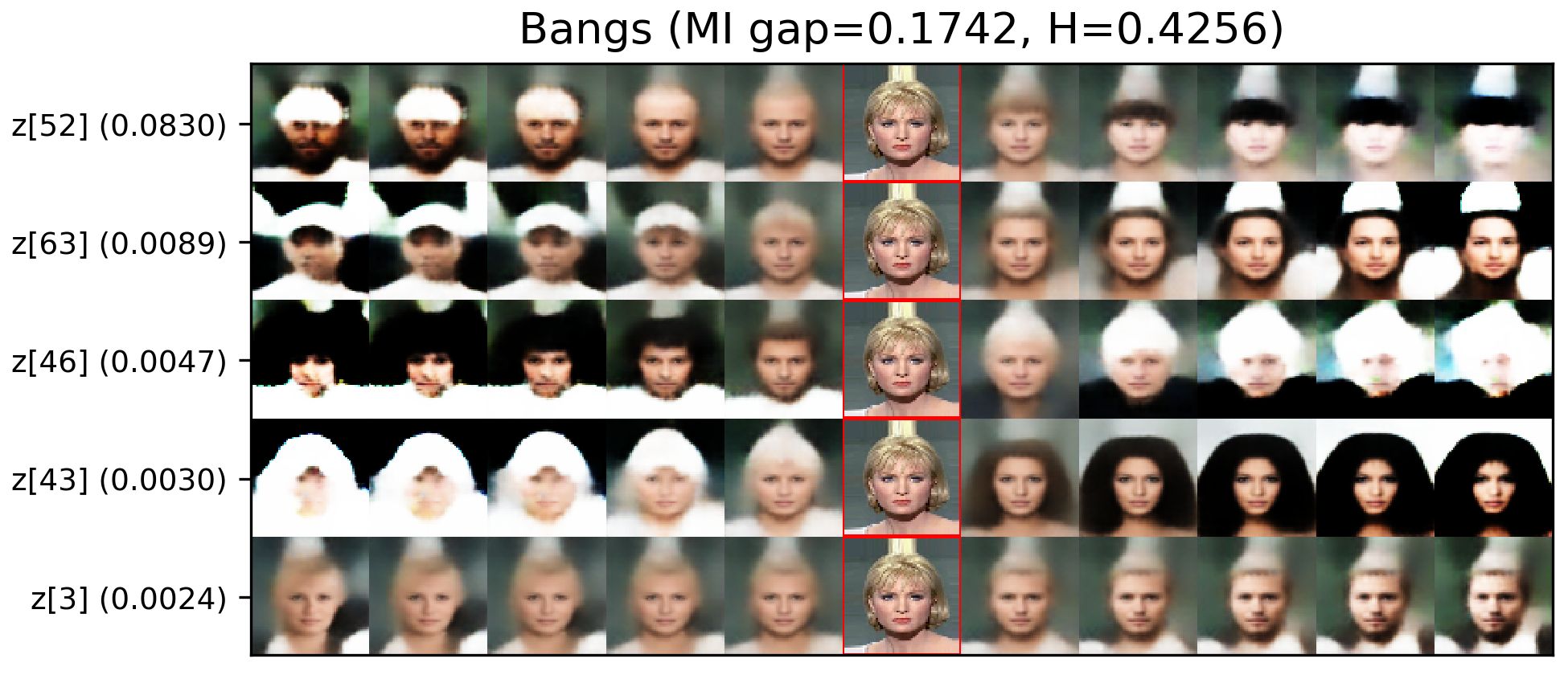}\includegraphics[width=0.33\textwidth]{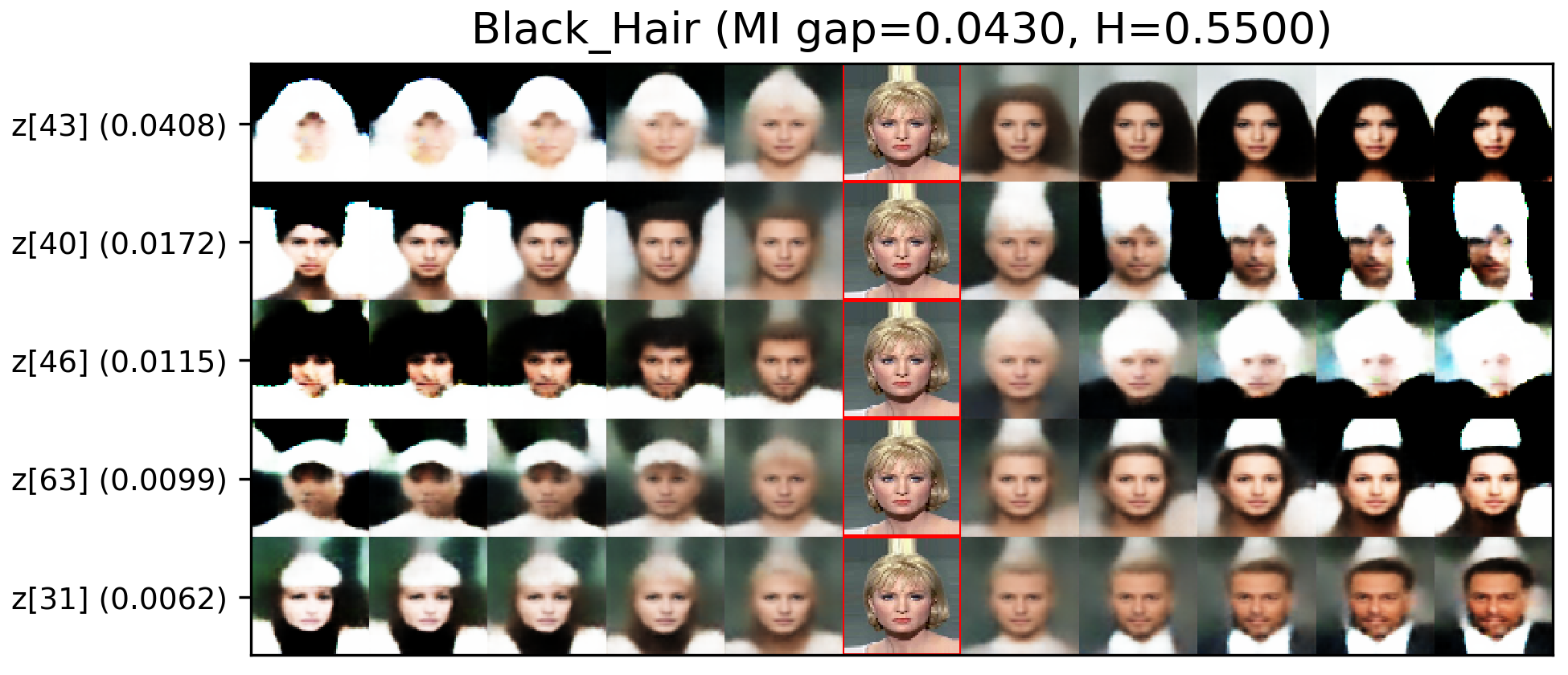}\includegraphics[width=0.33\textwidth]{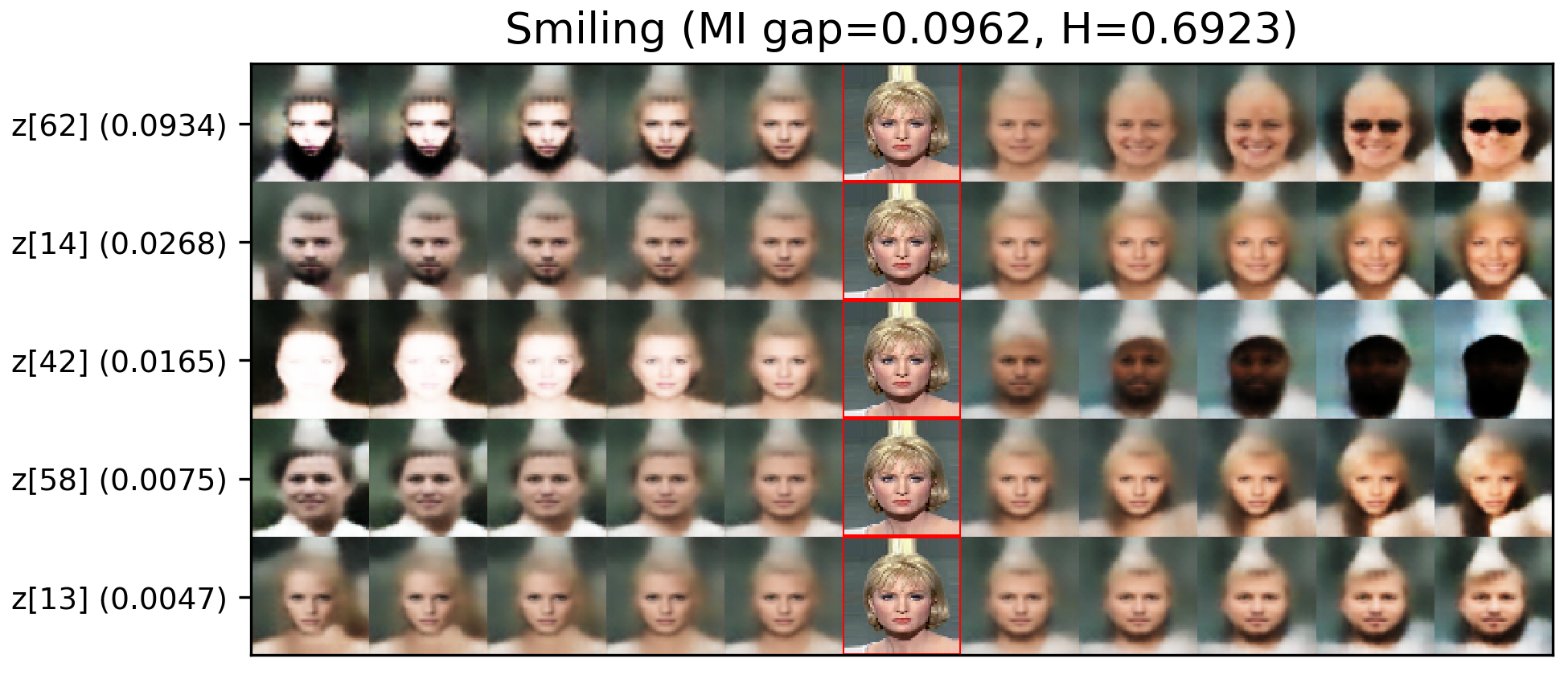}
\par\end{centering}
}
\par\end{centering}
\begin{centering}
\subfloat[$\beta$-VAE ($\beta$=50)]{\begin{centering}
\includegraphics[width=0.33\textwidth]{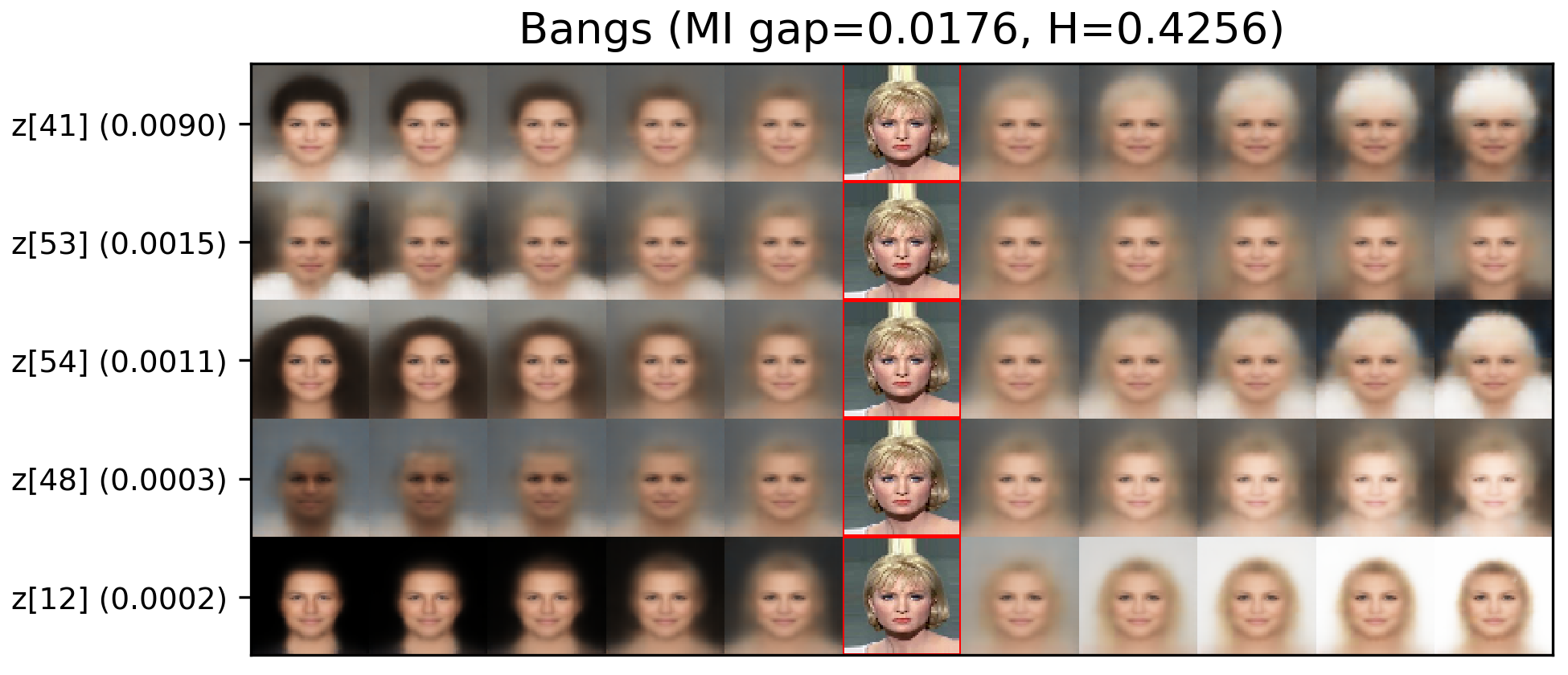}\includegraphics[width=0.33\textwidth]{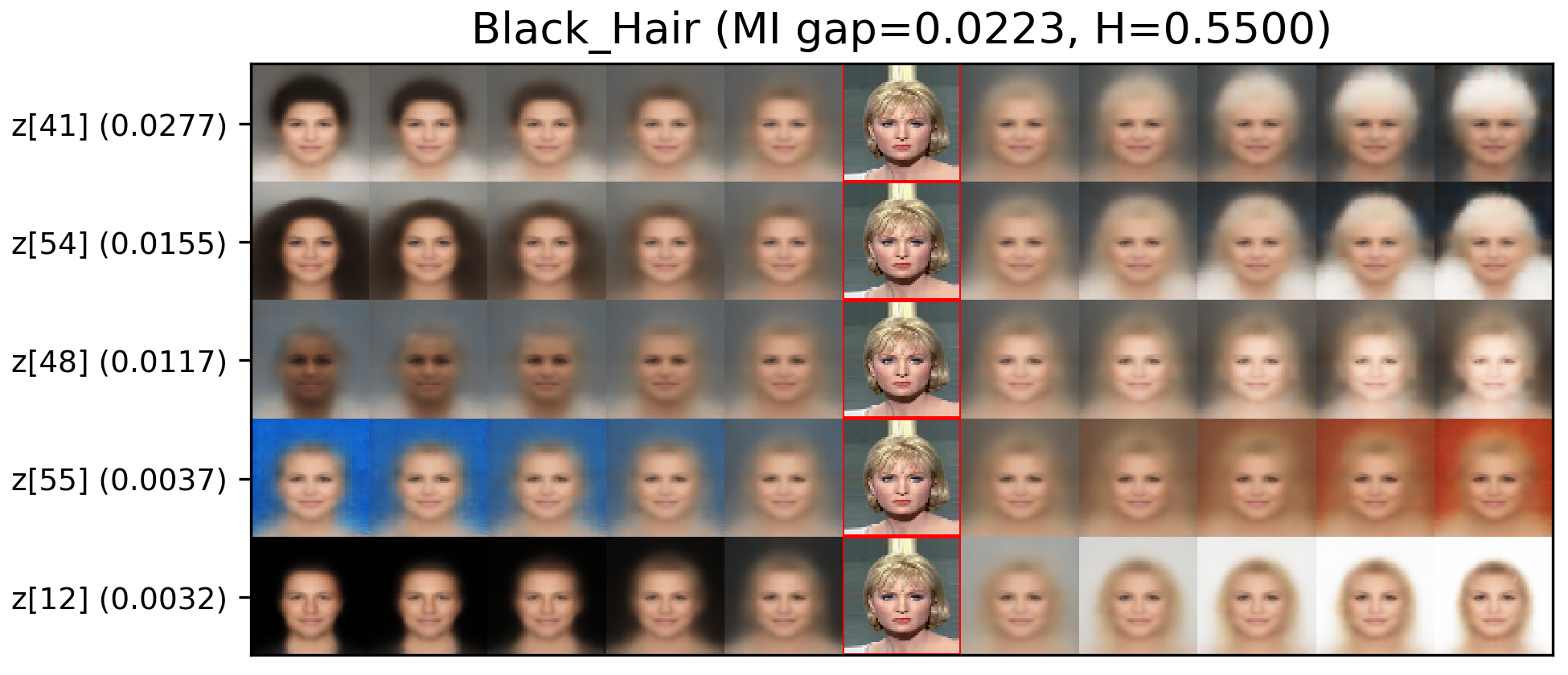}\includegraphics[width=0.33\textwidth]{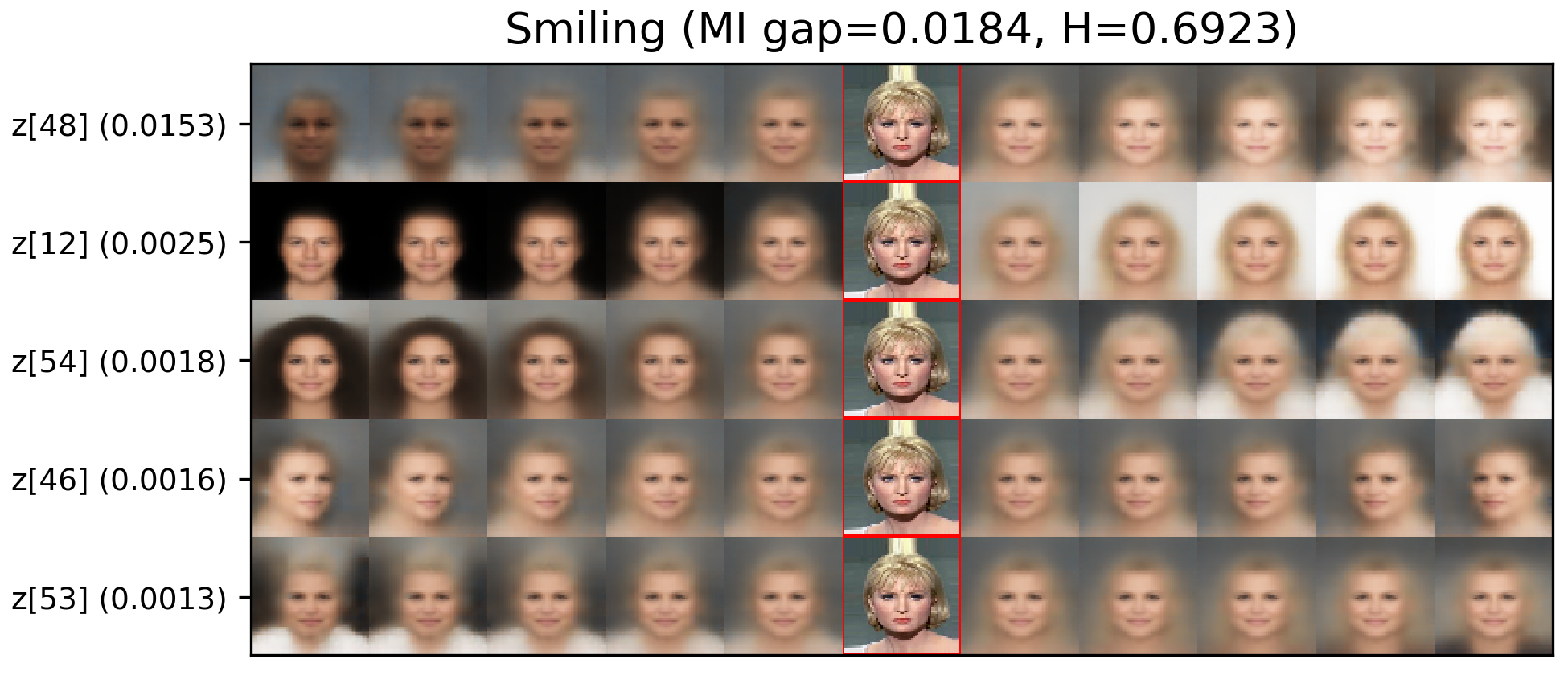}
\par\end{centering}
}
\par\end{centering}
\begin{centering}
\subfloat[AAE (Gz=50)]{\begin{centering}
\includegraphics[width=0.33\textwidth]{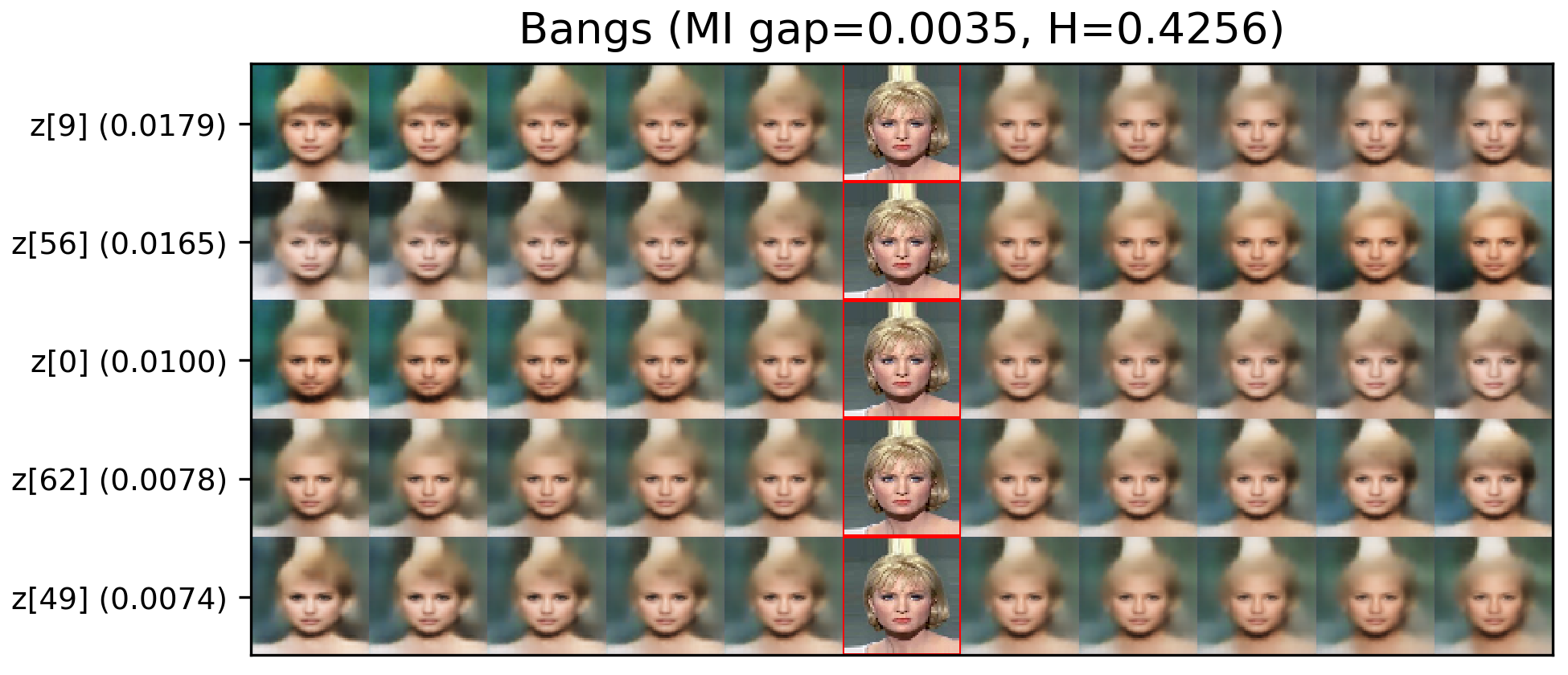}\includegraphics[width=0.33\textwidth]{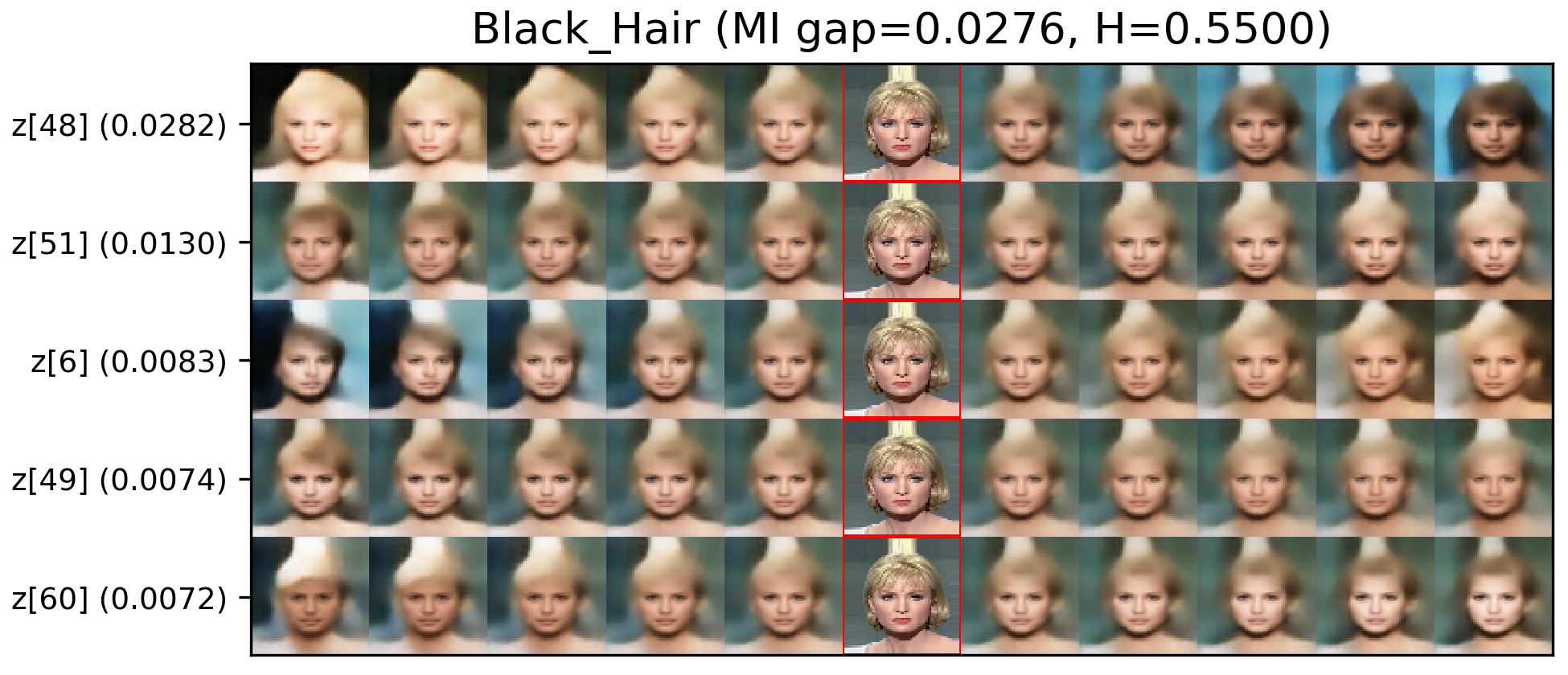}\includegraphics[width=0.33\textwidth]{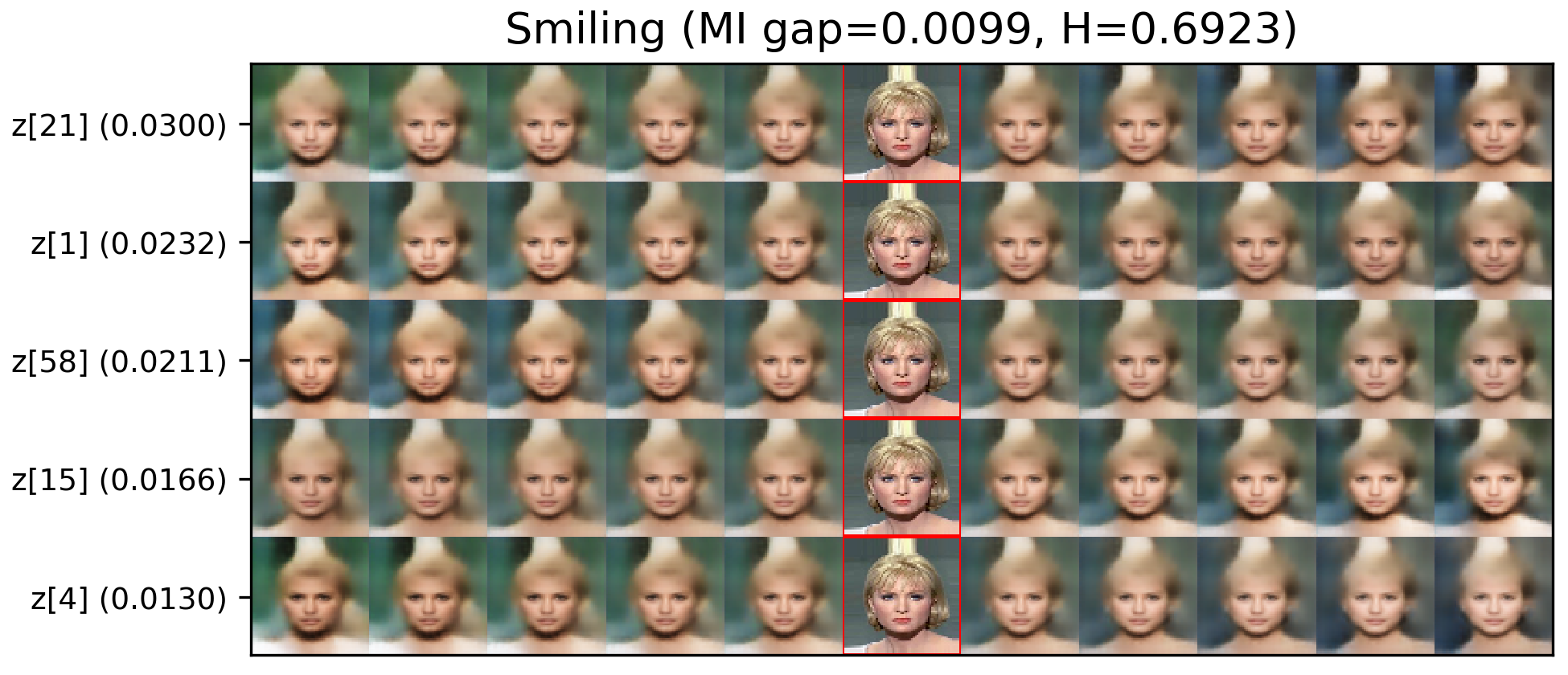}
\par\end{centering}
}
\par\end{centering}
\caption{Top 5 representations that are most correlated with some ground truth
factors. For each representation, we show its mutual information with
the ground truth factor.\label{fig:visualize_interpret_factors}}
\end{figure}

\begin{figure}
\begin{centering}
\subfloat[JEMMIG (normalized)]{\begin{centering}
\includegraphics[width=0.96\textwidth]{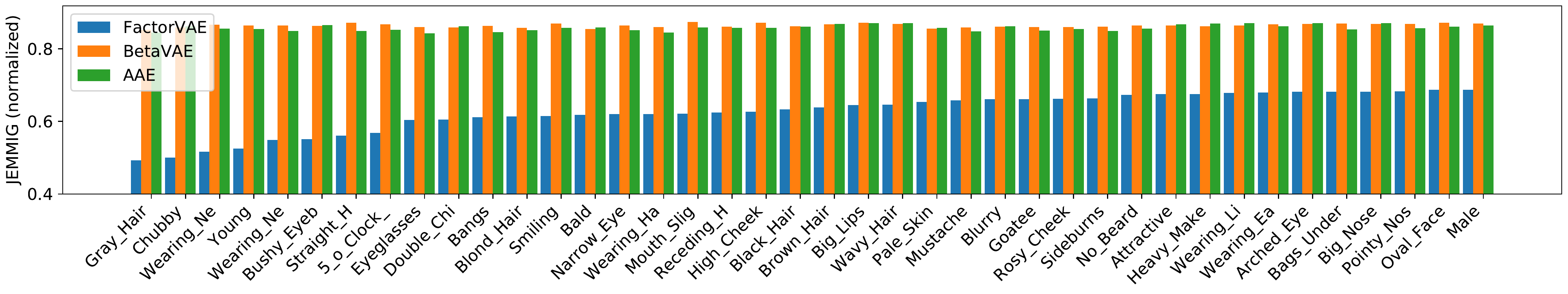}
\par\end{centering}
}
\par\end{centering}
\begin{centering}
\subfloat[RMIG (normalized)]{\begin{centering}
\includegraphics[width=0.96\textwidth]{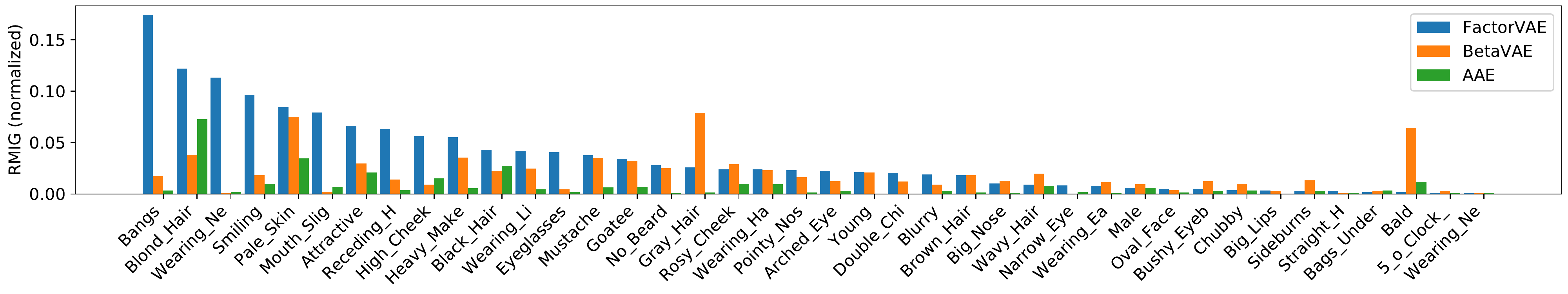}
\par\end{centering}
}
\par\end{centering}
\caption{Normalized JEMMIG and RMIG scores for all attributes in the CelebA
dataset. We sorted the JEMMIG and RMIG scores of the FactorVAE in
ascending and descending orders, respectively.\label{fig:JEMMIG_all_attrs}}
\end{figure}

We report the RMIG scores and JEMMIG scores for several ground truth
factors in the CelebA dataset in Tables~\ref{tab:RMIG-scores-for}
and \ref{tab:JEMMI-scores-for}, respectively. In general, FactorVAE
learns representations that agree better with the ground truth factors
than $\beta$-VAE and AAE do. This is consistent with the qualitative
results in Fig.~\ref{fig:visualize_interpret_factors}. However,
all models still perform poorly for interpretability since their RMIG
and JEMMIG scores are very far from 1 and 0, respectively. We provide
the normalized JEMMIG and RMIG scores for all attributes in Fig.~\ref{fig:JEMMIG_all_attrs}.

\begin{table}[H]
\begin{centering}
\begin{tabular}{|c|c|c|c|c|c|c|}
\hline 
\multirow{3}{*}{} & \multicolumn{6}{c|}{RMIG (normalized)}\tabularnewline
\cline{2-7} 
 & Bangs & Black Hair & Eyeglasses & Goatee & Male & Smiling\tabularnewline
\cline{2-7} 
 & H=0.4256 & H=0.5500 & H=0.2395 & H=0.2365 & H=0.6801 & H=0.6923\tabularnewline
\hline 
\hline 
FactorVAE & \textbf{0.1742} & \textbf{0.0430} & \textbf{0.0409} & \textbf{0.0343} & 0.0060 & \textbf{0.0962}\tabularnewline
\hline 
$\beta$-VAE & 0.0176 & 0.0223 & 0.0045 & 0.0325 & \textbf{0.0094} & 0.0184\tabularnewline
\hline 
AAE & 0.0035 & 0.0276 & 0.0018 & 0.0069 & 0.0060 & 0.0099\tabularnewline
\hline 
\end{tabular}
\par\end{centering}
\caption{Normalized RMIG scores (\#bins=100) for some factors. Higher is better.
\label{tab:RMIG-scores-for}}
\end{table}

\begin{table}[H]
\begin{centering}
\begin{tabular}{|c|c|c|c|c|c|c|}
\hline 
\multirow{3}{*}{} & \multicolumn{6}{c|}{JEMMIG (normalized)}\tabularnewline
\cline{2-7} 
 & Bangs & Black Hair & Eyeglasses & Goatee & Male & Smiling\tabularnewline
\cline{2-7} 
 & H=0.4256 & H=0.5500 & H=0.2395 & H=0.2365 & H=0.6801 & H=0.6923\tabularnewline
\hline 
\hline 
FactorVAE & \textbf{0.6118} & \textbf{0.6334} & \textbf{0.6041} & \textbf{0.6616} & \textbf{0.6875} & \textbf{0.6150}\tabularnewline
\hline 
$\beta$-VAE & 0.8632 & 0.8620 & 0.8602 & 0.8600 & 0.8690 & 0.8699\tabularnewline
\hline 
AAE & 0.8463 & 0.8613 & 0.8423 & 0.8496 & 0.8644 & 0.8575\tabularnewline
\hline 
\end{tabular}
\par\end{centering}
\caption{Normalized JEMMIG scores (\#bins=100) for some factors. Lower is better.
\label{tab:JEMMI-scores-for}}
\end{table}

\subsubsection{dSprites}

\paragraph{Informativeness}

\begin{figure}[h]
\begin{centering}
\subfloat[FactorVAE]{\begin{centering}
\includegraphics[width=0.26\textwidth]{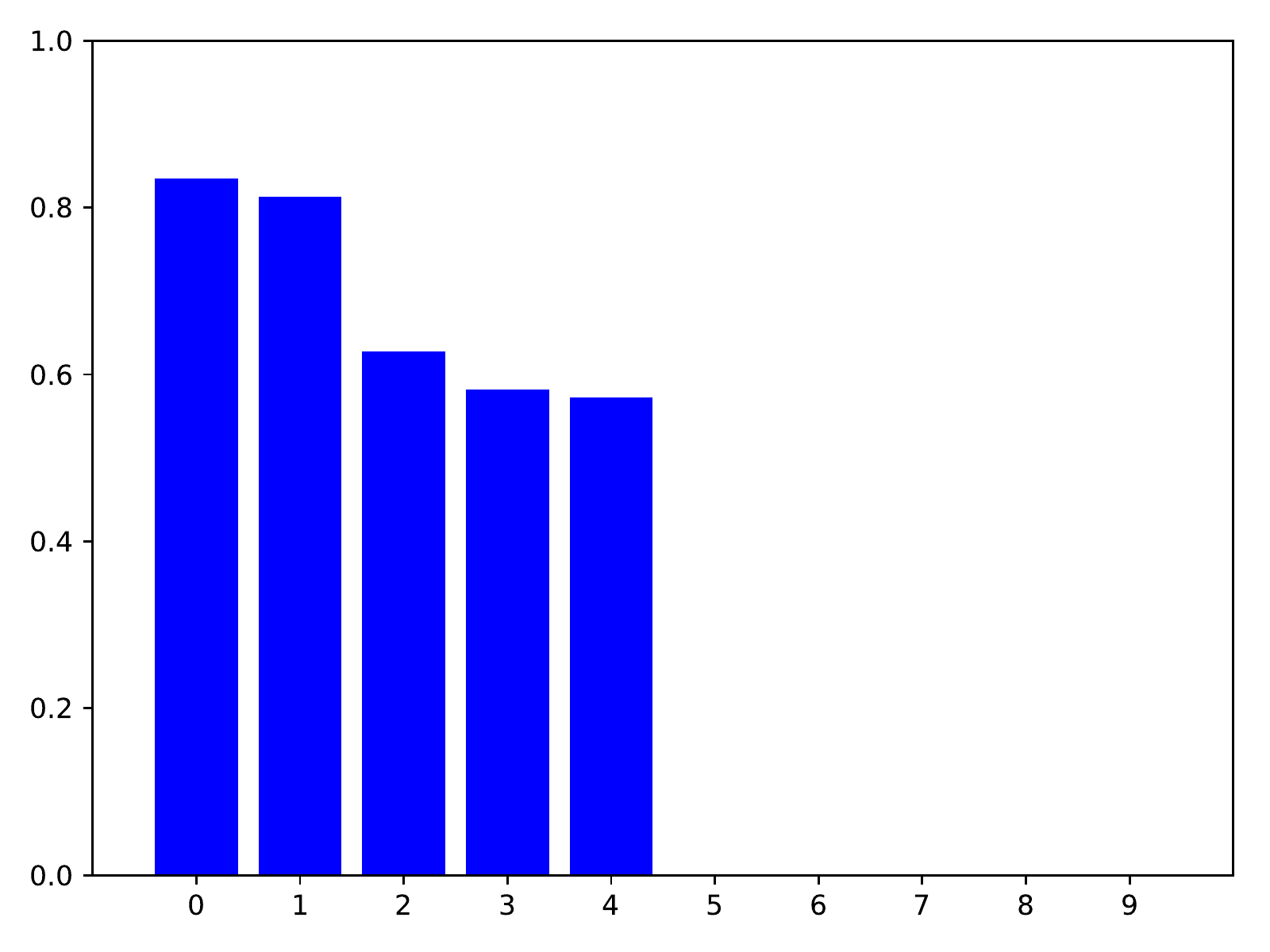}
\par\end{centering}
}\subfloat[$\beta$-VAE]{\begin{centering}
\includegraphics[width=0.26\textwidth]{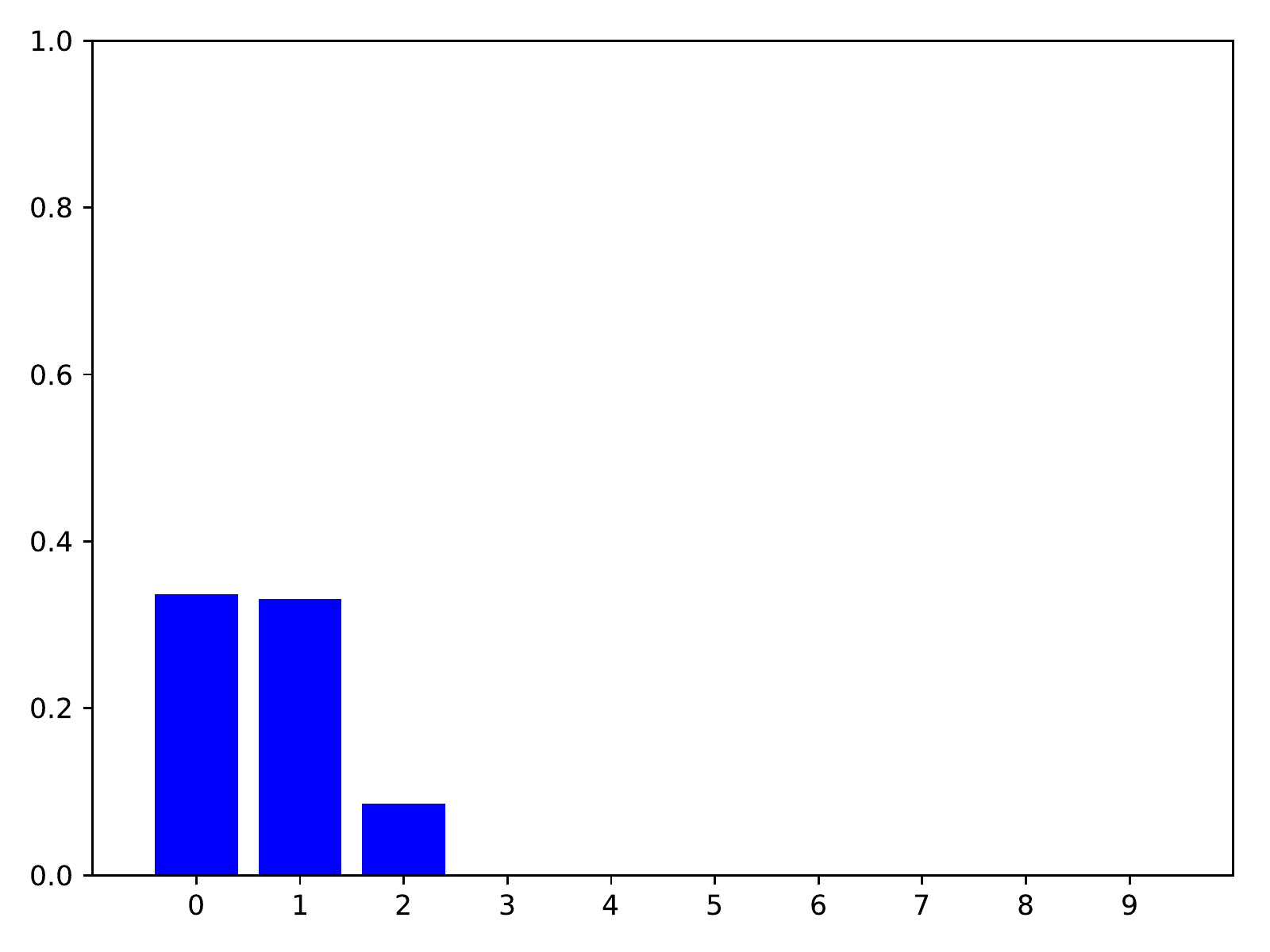}
\par\end{centering}
}\subfloat[AAE]{\begin{centering}
\includegraphics[width=0.26\textwidth]{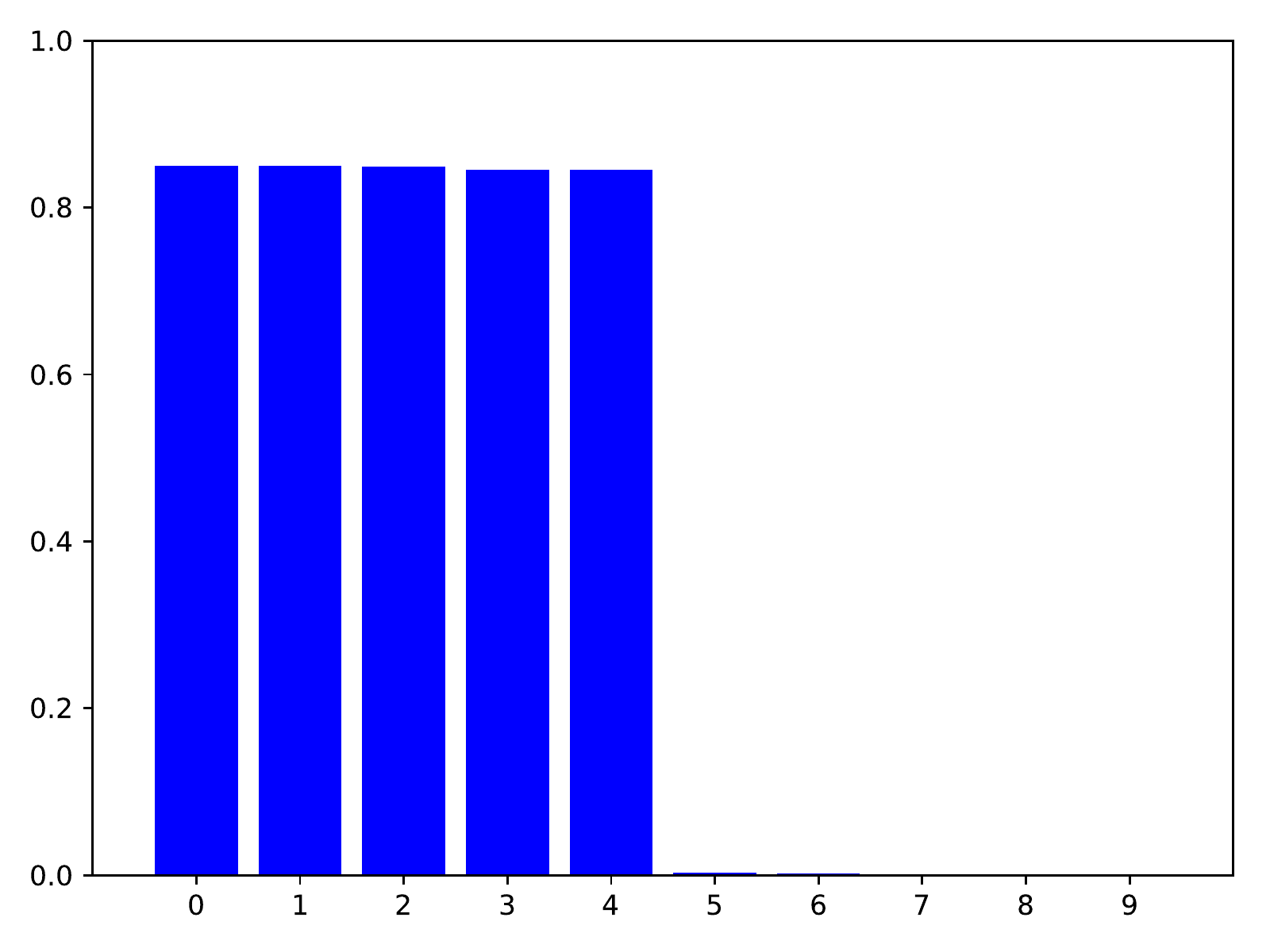}
\par\end{centering}
}
\par\end{centering}
\caption{Normalized informativeness scores (bins=100) of all latent variables
sorted in descending order.\label{fig:dSprites_info_bar}}
\end{figure}

From Fig.~\ref{fig:dSprites_info_bar}, we see that 5 representations
of AAE have equally high informativeness scores while the remaining
5 representations have nearly zeros informativeness scores. This is
because AAE needs only 5 representations to capture all information
in the data. FactorVAE also needs only 5 representations but some
are less informative than those of AAE. Note that the number of ground
truth factors of variation in dSprites dataset is also 5.

\begin{figure}[h]
\begin{centering}
\subfloat[FactorVAE]{\begin{centering}
\includegraphics[width=0.26\textwidth]{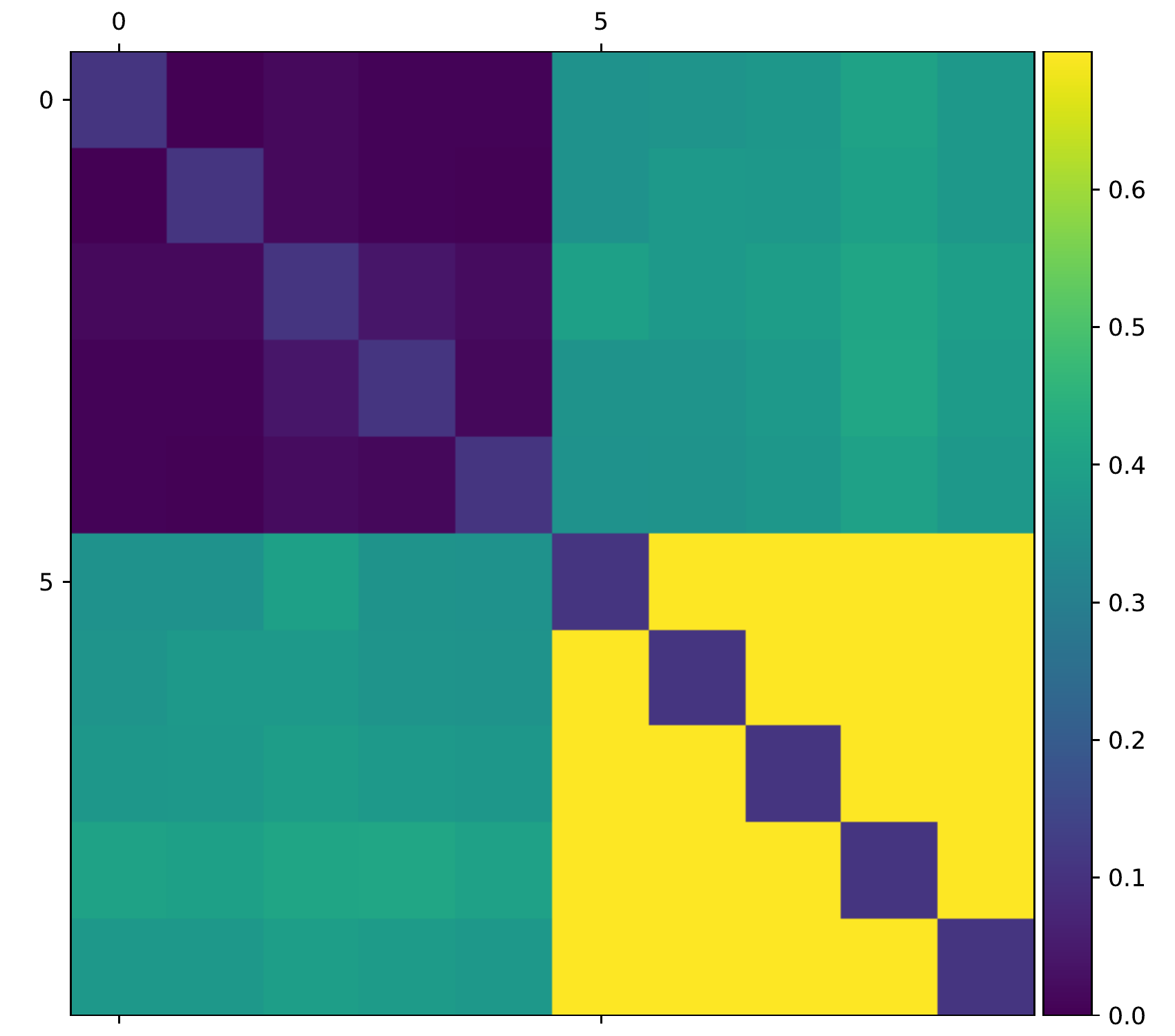}
\par\end{centering}
}\subfloat[$\beta$-VAE]{\begin{centering}
\includegraphics[width=0.26\textwidth]{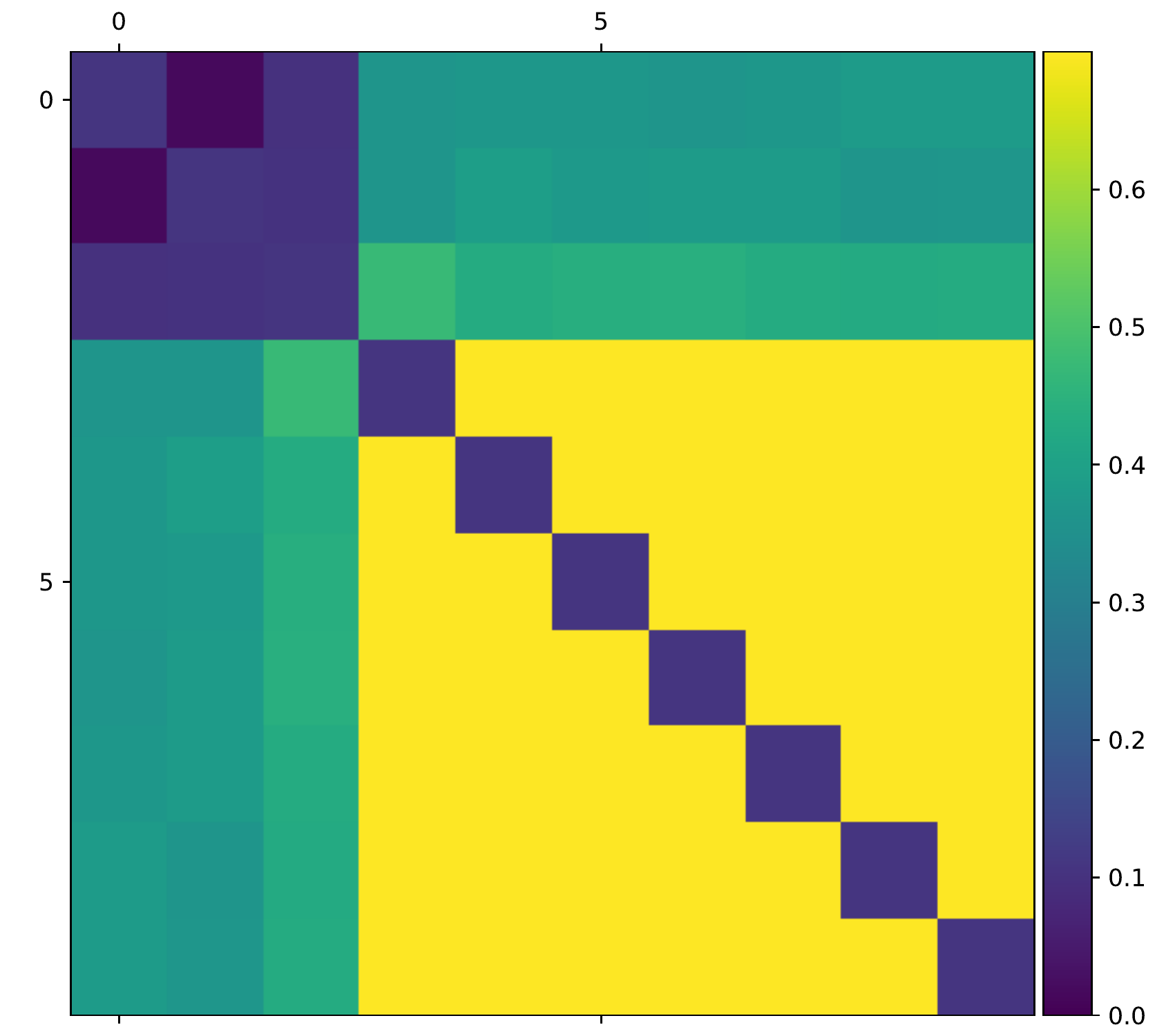}
\par\end{centering}
}\subfloat[AAE]{\begin{centering}
\includegraphics[width=0.26\textwidth]{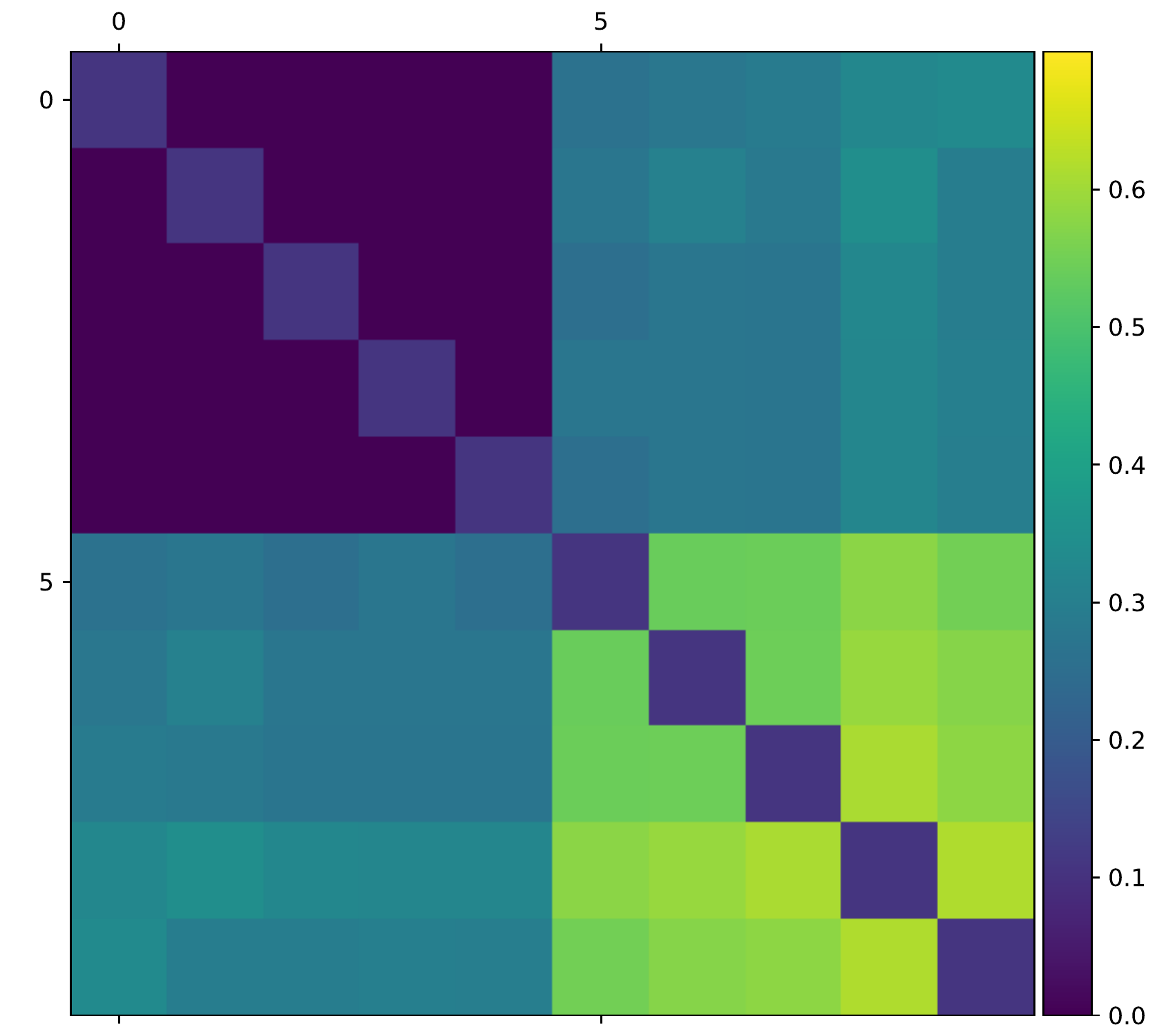}
\par\end{centering}
}
\par\end{centering}
\caption{Normalized MISJED scores (bins=100) of all latent pairs sorted by
their informativeness.\label{fig:dSprites_heat_map}}
\end{figure}

\paragraph{Separability and Independence}

Fig.~\ref{fig:dSprites_heat_map} shows heat maps of MISJED scores
for the three models.

\paragraph{Interpretability}

\begin{figure}[h]
\begin{centering}
\subfloat[FactorVAE (Shape)]{\begin{centering}
\includegraphics[width=0.3\textwidth]{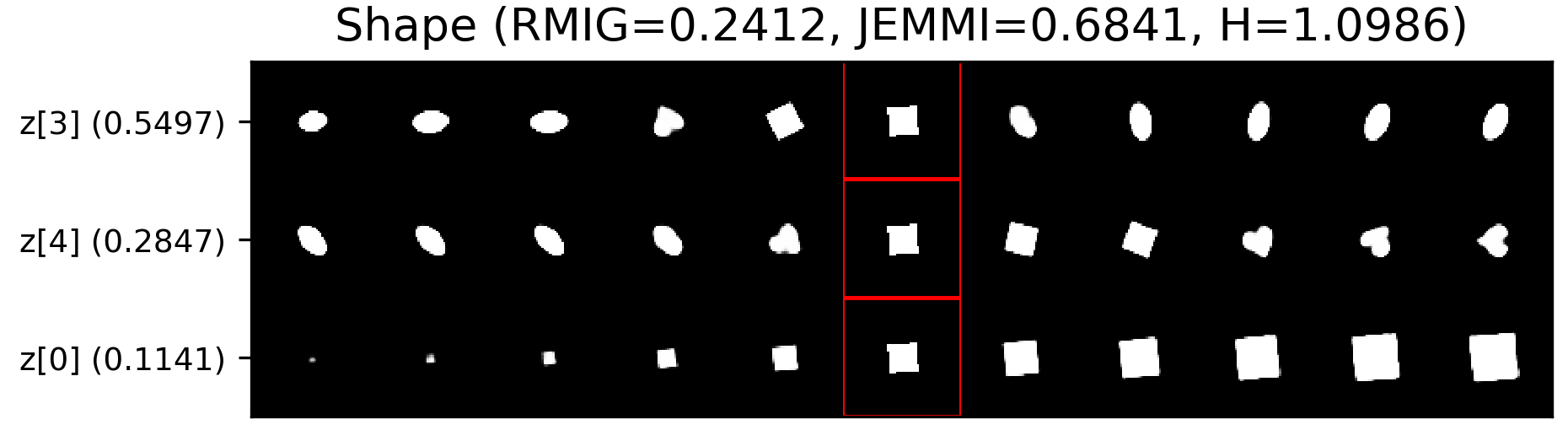}
\par\end{centering}
}\subfloat[$\beta$-VAE (Shape)]{\begin{centering}
\includegraphics[width=0.3\textwidth]{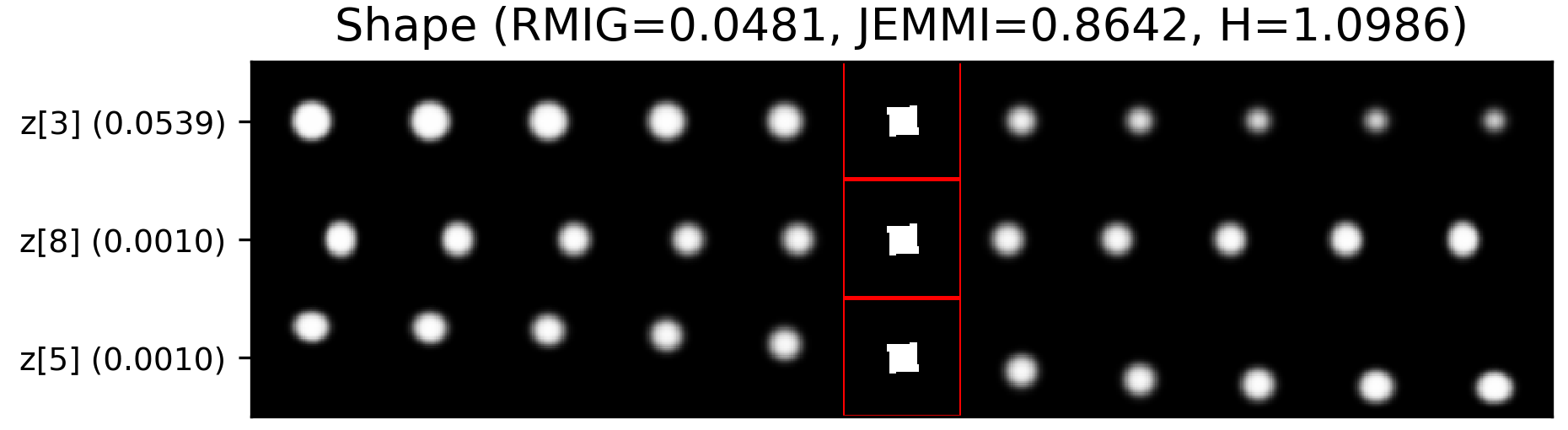}
\par\end{centering}
}\subfloat[AAE (Shape)]{\begin{centering}
\includegraphics[width=0.3\textwidth]{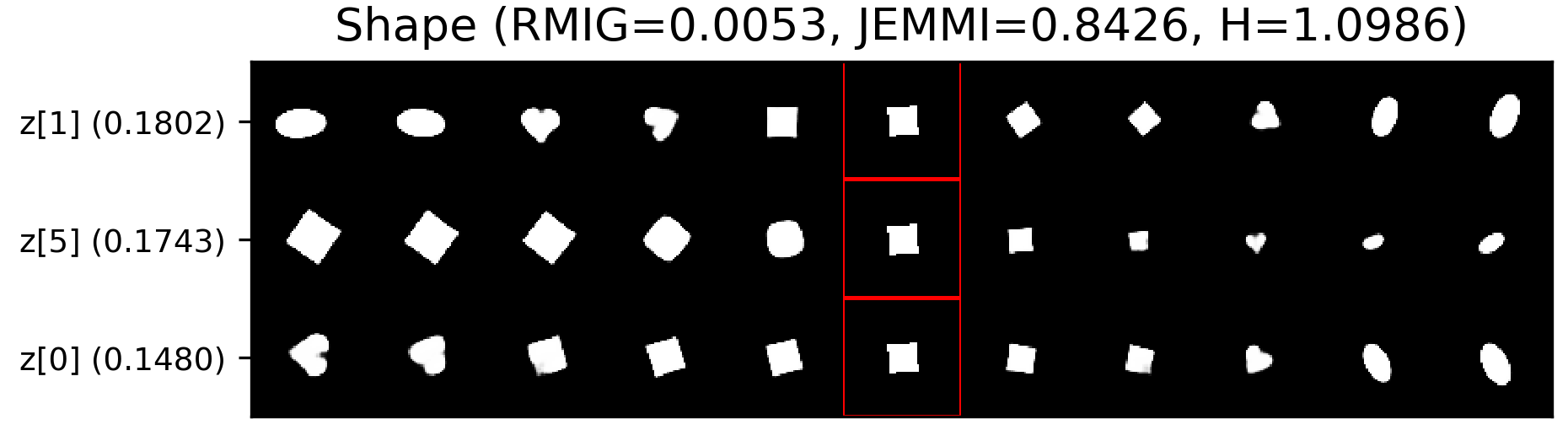}
\par\end{centering}
}
\par\end{centering}
\begin{centering}
\subfloat[FactorVAE (Scale)]{\begin{centering}
\includegraphics[width=0.3\textwidth]{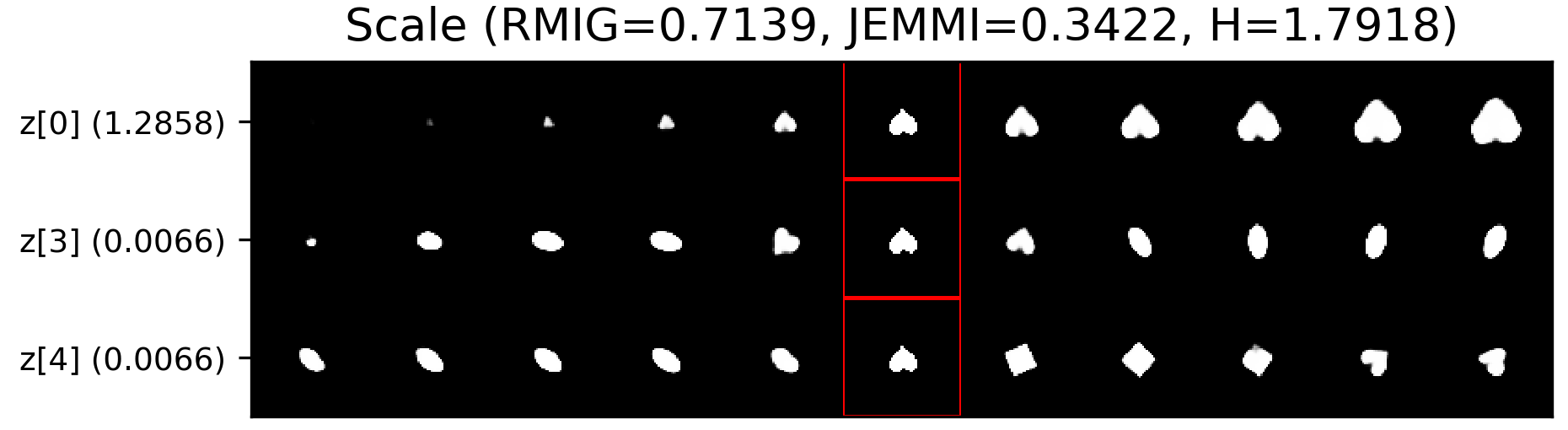}
\par\end{centering}
}\subfloat[$\beta$-VAE (Scale)]{\begin{centering}
\includegraphics[width=0.3\textwidth]{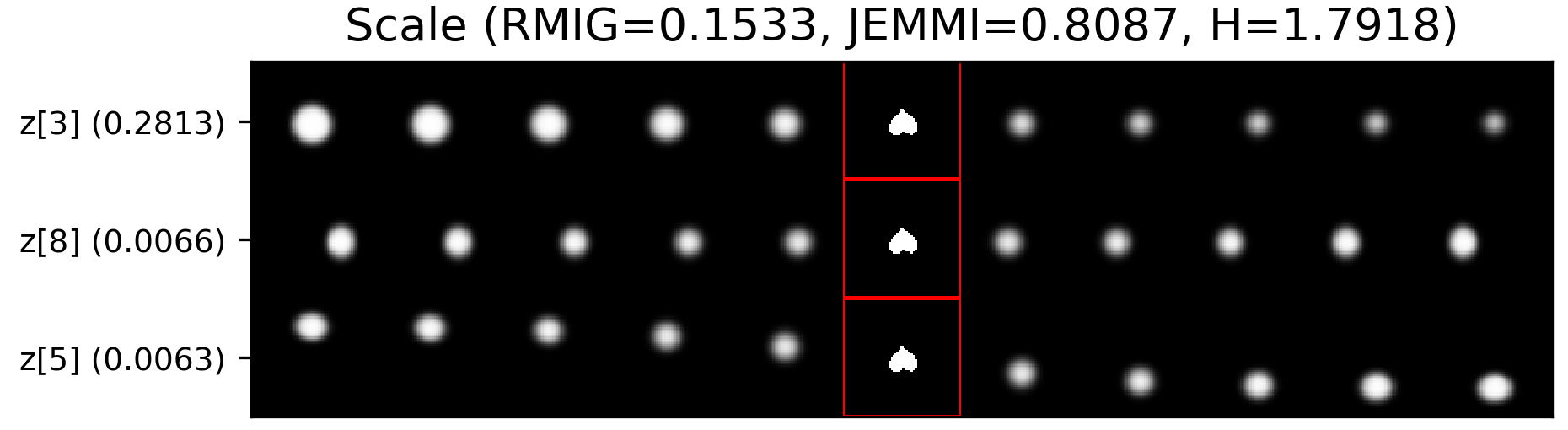}
\par\end{centering}
}\subfloat[AAE (Scale)]{\begin{centering}
\includegraphics[width=0.3\textwidth]{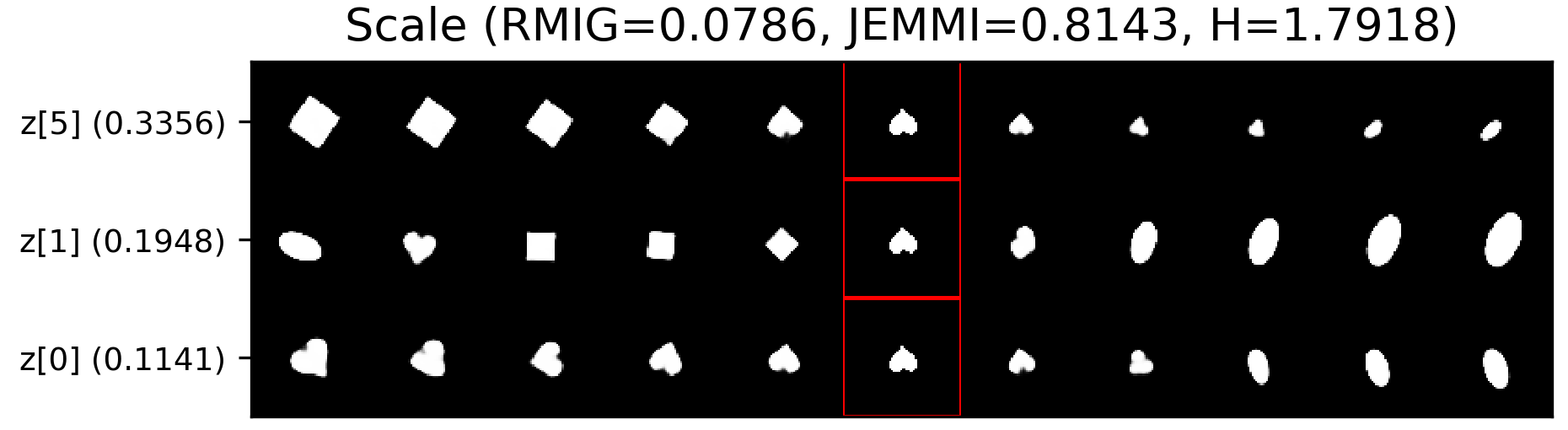}
\par\end{centering}
}
\par\end{centering}
\begin{centering}
\subfloat[FactorVAE (Rotation)]{\begin{centering}
\includegraphics[width=0.3\textwidth]{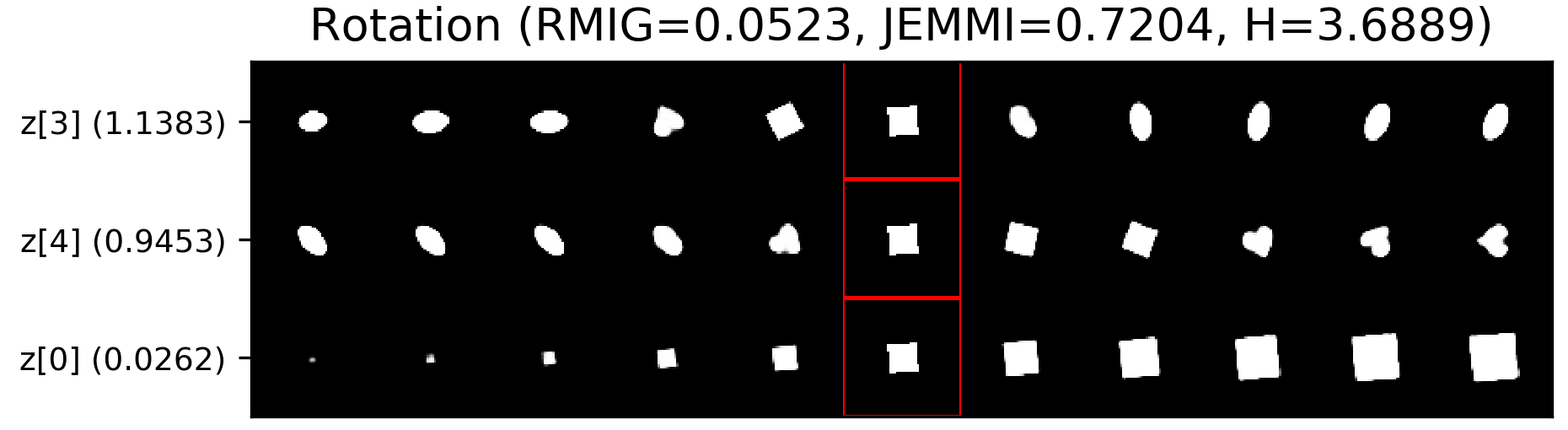}
\par\end{centering}
}\subfloat[$\beta$-VAE (Rotation)]{\begin{centering}
\includegraphics[width=0.3\textwidth]{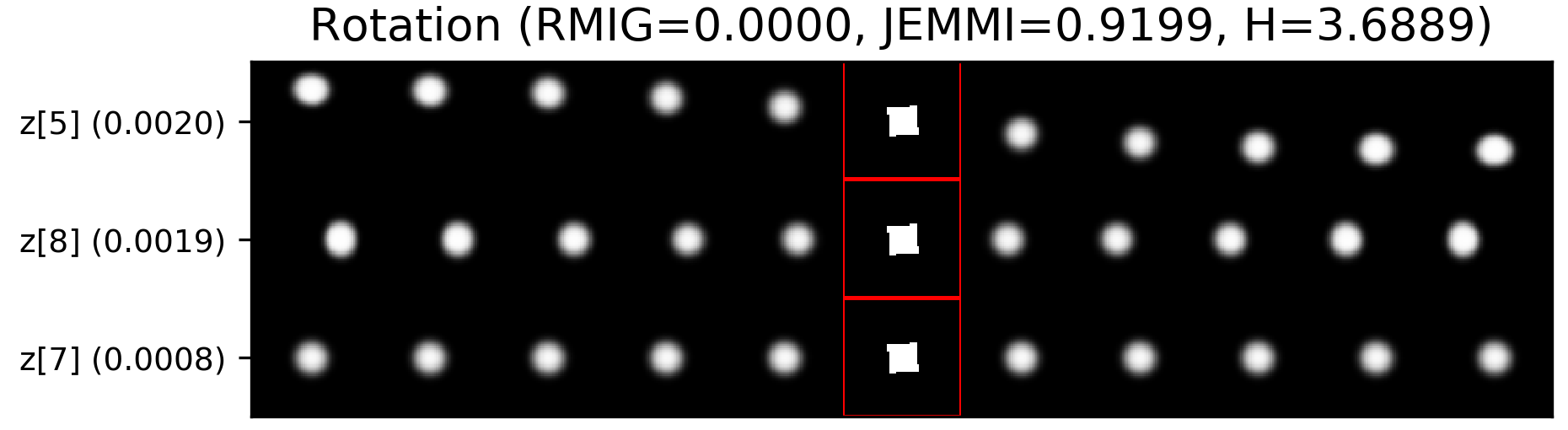}
\par\end{centering}
}\subfloat[AAE (Rotation)]{\begin{centering}
\includegraphics[width=0.3\textwidth]{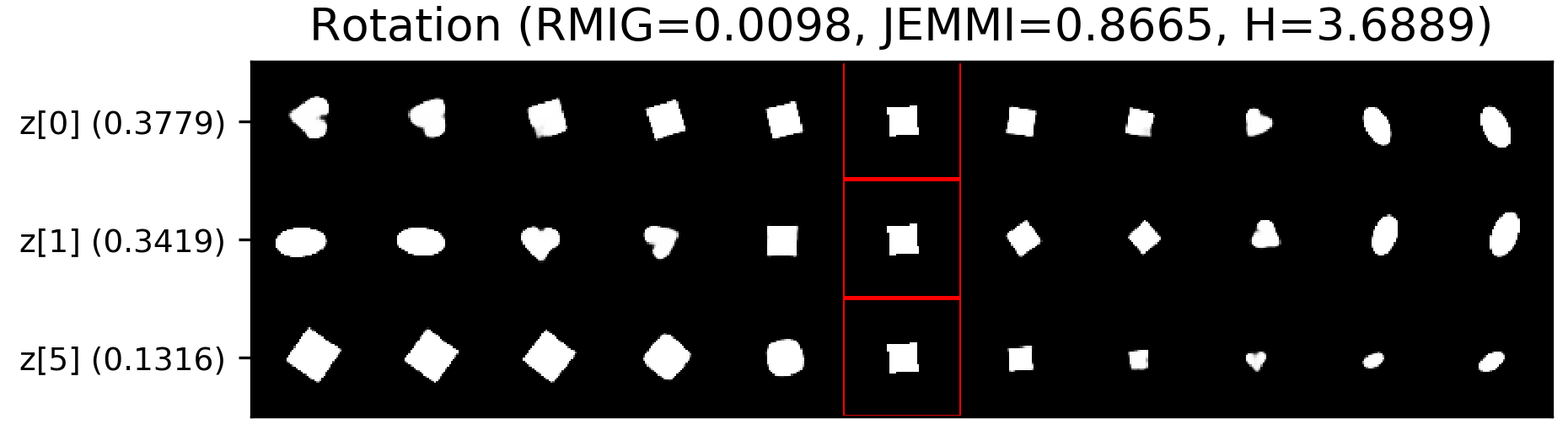}
\par\end{centering}
}
\par\end{centering}
\begin{centering}
\subfloat[FactorVAE (Pos X)]{\begin{centering}
\includegraphics[width=0.3\textwidth]{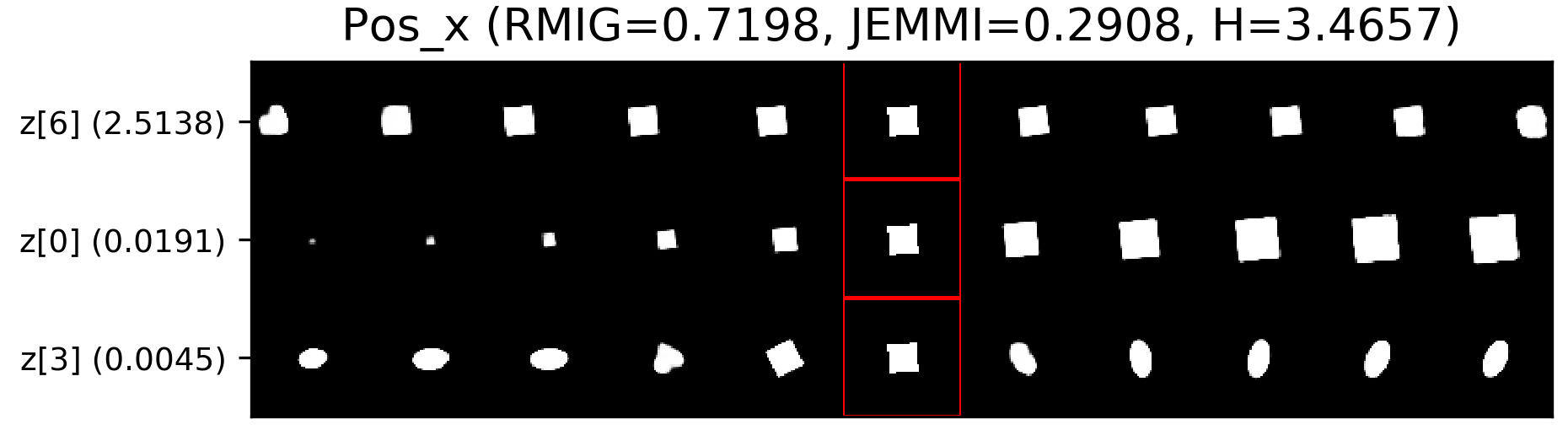}
\par\end{centering}
}\subfloat[$\beta$-VAE (Pos X)]{\begin{centering}
\includegraphics[width=0.3\textwidth]{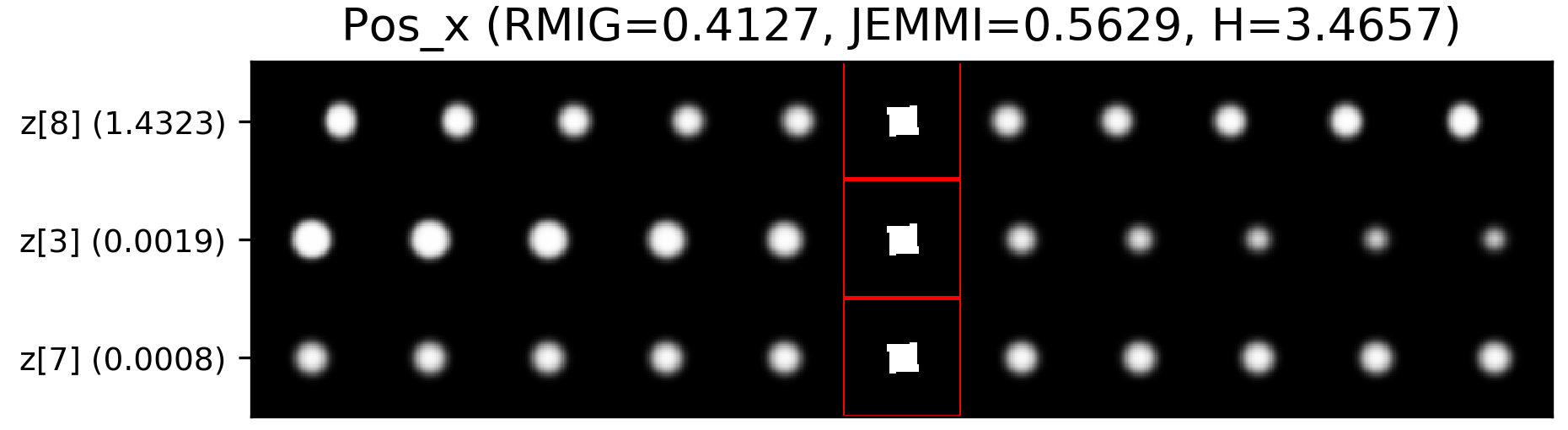}
\par\end{centering}
}\subfloat[AAE (Pos X)]{\begin{centering}
\includegraphics[width=0.3\textwidth]{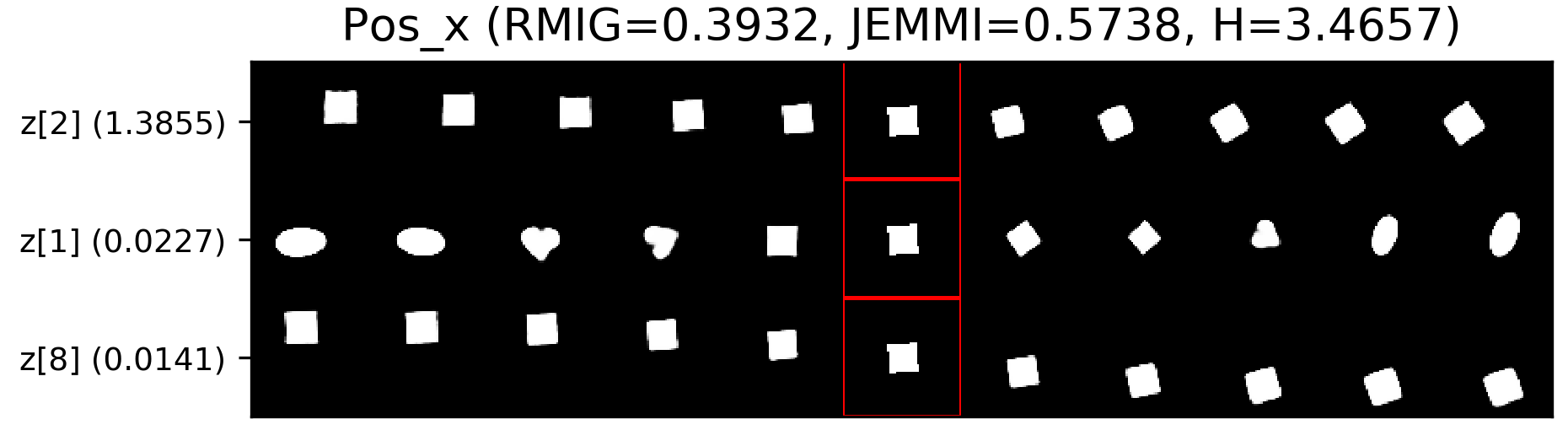}
\par\end{centering}
}
\par\end{centering}
\begin{centering}
\subfloat[FactorVAE (Pos Y)]{\begin{centering}
\includegraphics[width=0.3\textwidth]{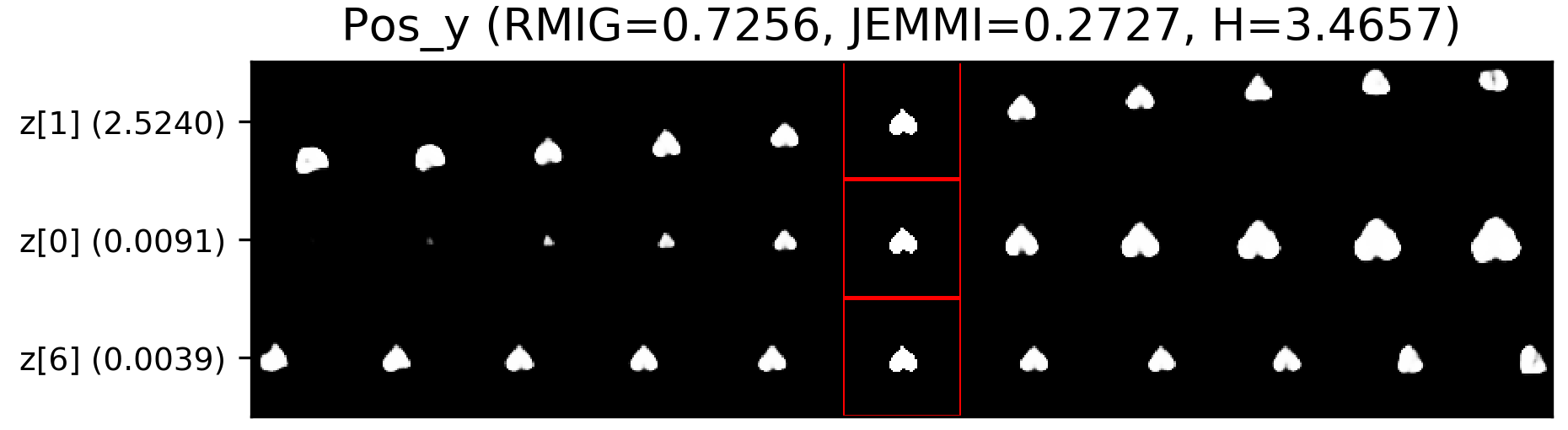}
\par\end{centering}
}\subfloat[$\beta$-VAE (Pos Y)]{\begin{centering}
\includegraphics[width=0.3\textwidth]{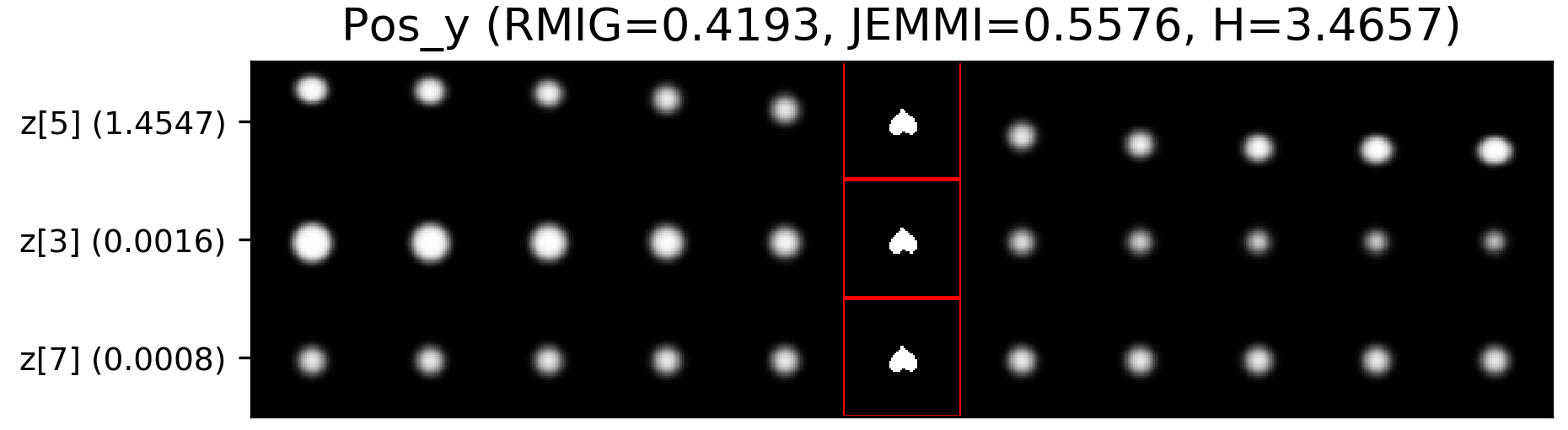}
\par\end{centering}
}\subfloat[AAE (Pos Y)]{\begin{centering}
\includegraphics[width=0.3\textwidth]{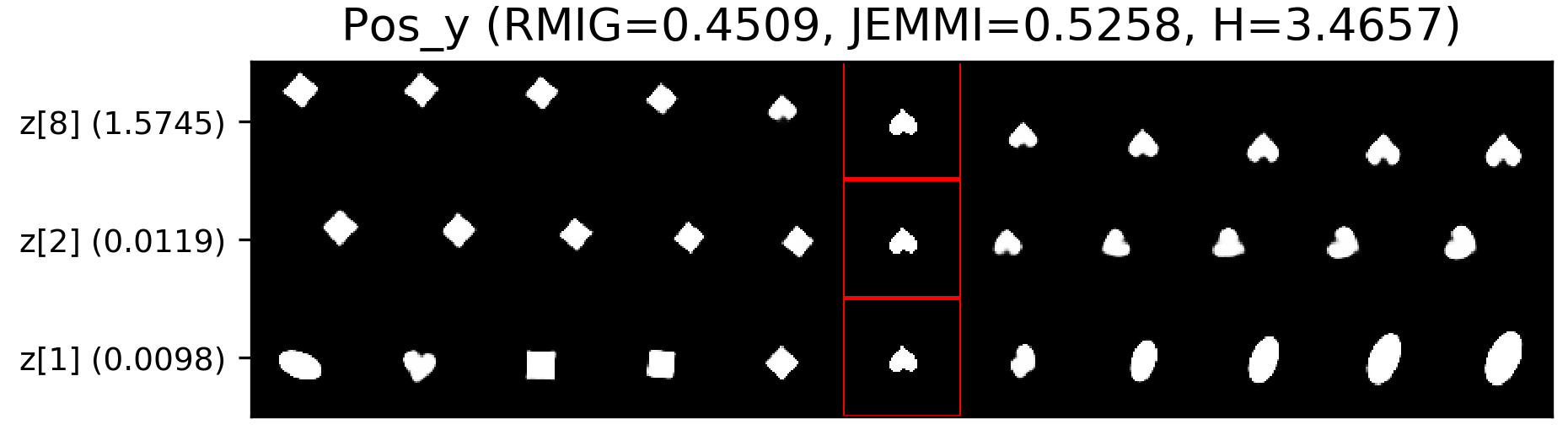}
\par\end{centering}
}
\par\end{centering}
\caption{Top 3 representations sorted by their mutual information with different
ground truth factors.\label{fig:dSprites_top3_interpret}}
\end{figure}

From Tables.~\ref{tab:dSprites_RMIG} and \ref{tab:dSprites_JEMMI},
we see that FactorVAE is very good at disentangling ``scale'', ``x-position''
and ``y-position'' but fails to disentangling ``shape'' and ``rotation''.
However, FactorVAE still performs much better than $\beta$-VAE and
AAE. These results are consistent with the visual results in Fig.~\ref{fig:dSprites_top3_interpret}.

Also note that in FactorVAE, the RMIG scores for ``scale'' and ``x-position''
are quite similar but the JEMMIG score for ``scale'' is higher than
that for ``x-position''. This is because the quantized distribution
(with 100 bins) of a particular representation $z_{i}$ fits better
to the distribution of ``x-position'' (having 32 possible values)
than to the distribution of ``scale'' (having only 6 possible values).

\begin{table}[h]
\begin{centering}
\begin{tabular}{|c|c|c|c|c|c|}
\hline 
 & Shape & Scale & Rotation & Pos X & Pos Y\tabularnewline
\hline 
FactorVAE & \textbf{0.2412} & \textbf{0.7139} & \textbf{0.0523} & \textbf{0.7198} & \textbf{0.7256}\tabularnewline
\hline 
$\beta$-VAE & 0.0481 & 0.1533 & 0.0000 & 0.4127 & 0.4193\tabularnewline
\hline 
AAE & 0.0053 & 0.0786 & 0.0098 & 0.3932 & 0.4509\tabularnewline
\hline 
\end{tabular}
\par\end{centering}
\caption{Normalized RMIG scores (bins=100).\label{tab:dSprites_RMIG}}
\end{table}

\begin{table}[H]
\begin{centering}
\begin{tabular}{|c|c|c|c|c|c|}
\hline 
 & Shape & Scale & Rotation & Pos X & Pos Y\tabularnewline
\hline 
\hline 
FactorVAE & \textbf{0.6841} & \textbf{0.3422} & \textbf{0.7204} & \textbf{0.2908} & \textbf{0.2727}\tabularnewline
\hline 
$\beta$-VAE & 0.8642 & 0.8087 & 0.9199 & 0.5629 & 0.5576\tabularnewline
\hline 
AAE & 0.8426 & 0.8143 & 0.8665 & 0.5738 & 0.5258\tabularnewline
\hline 
\end{tabular}
\par\end{centering}
\caption{Normalized JEMMIG scores (bins=100).\label{tab:dSprites_JEMMI}}
\end{table}

\subsection{Ablation study of our metrics\label{subsec:Ablation-study-of}}

\paragraph{Sensitivity of the number of bins}

\begin{figure}[h]
\begin{centering}
\includegraphics[width=0.32\textwidth]{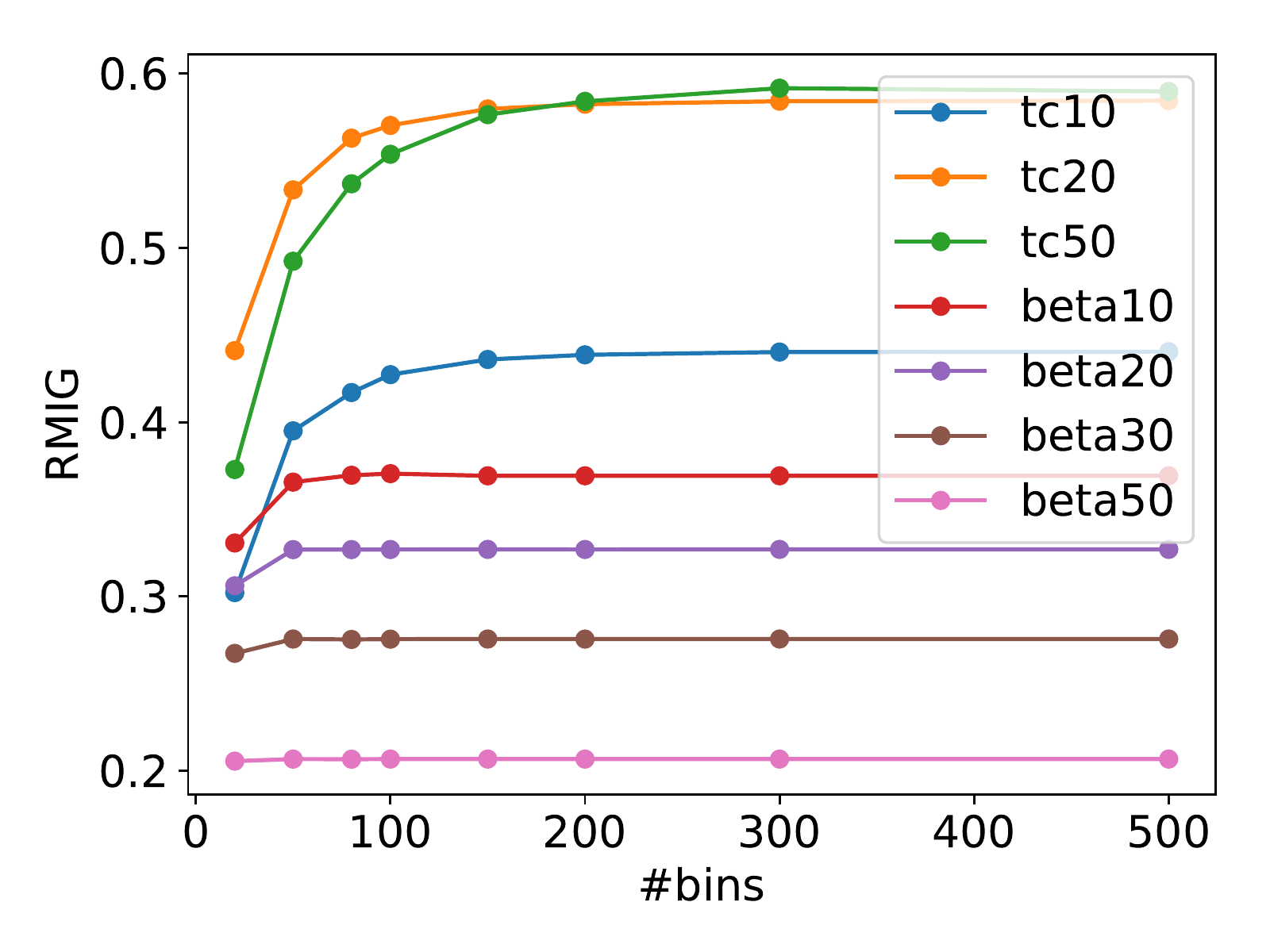}\includegraphics[width=0.32\textwidth]{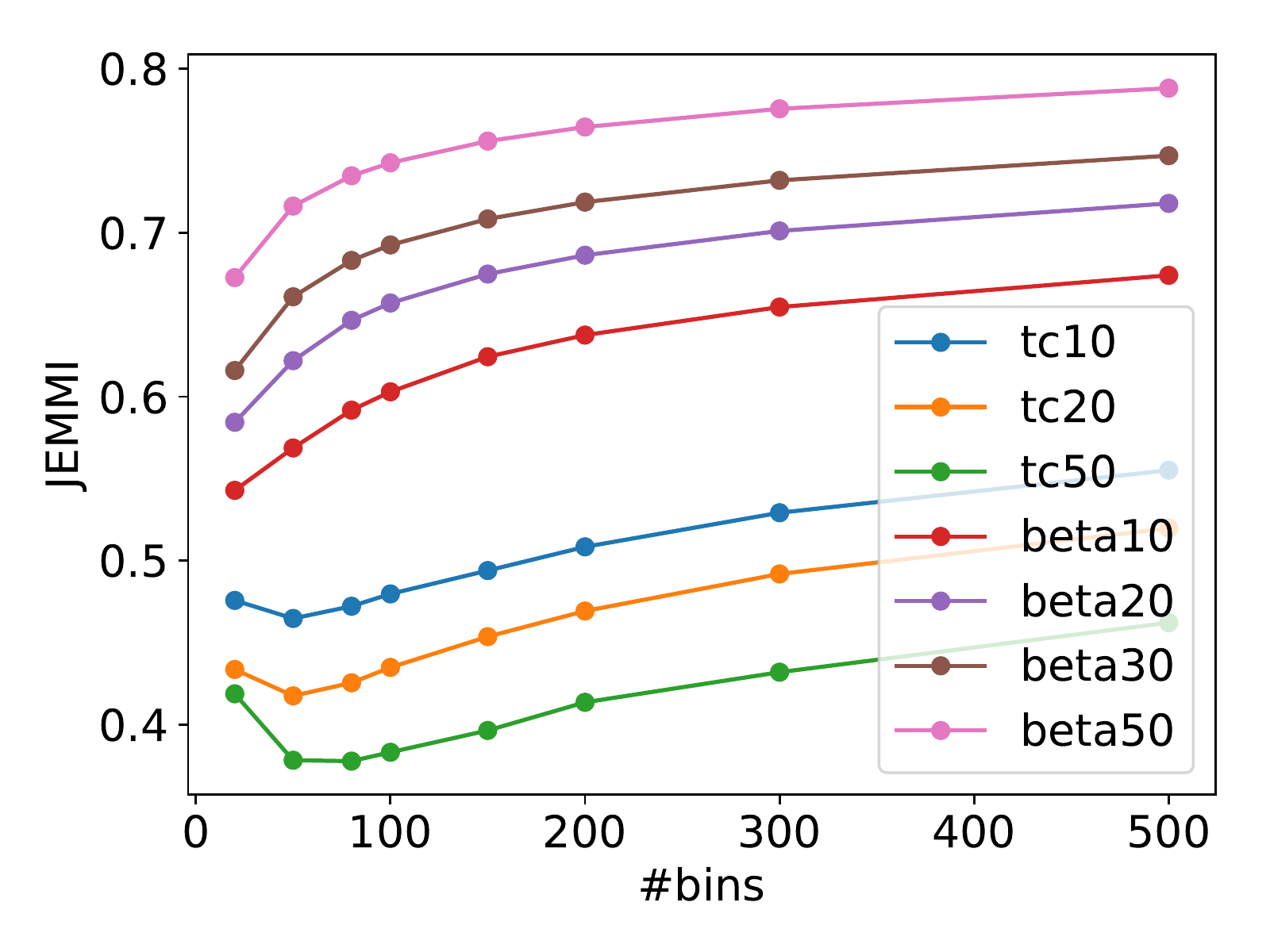}\includegraphics[width=0.32\textwidth]{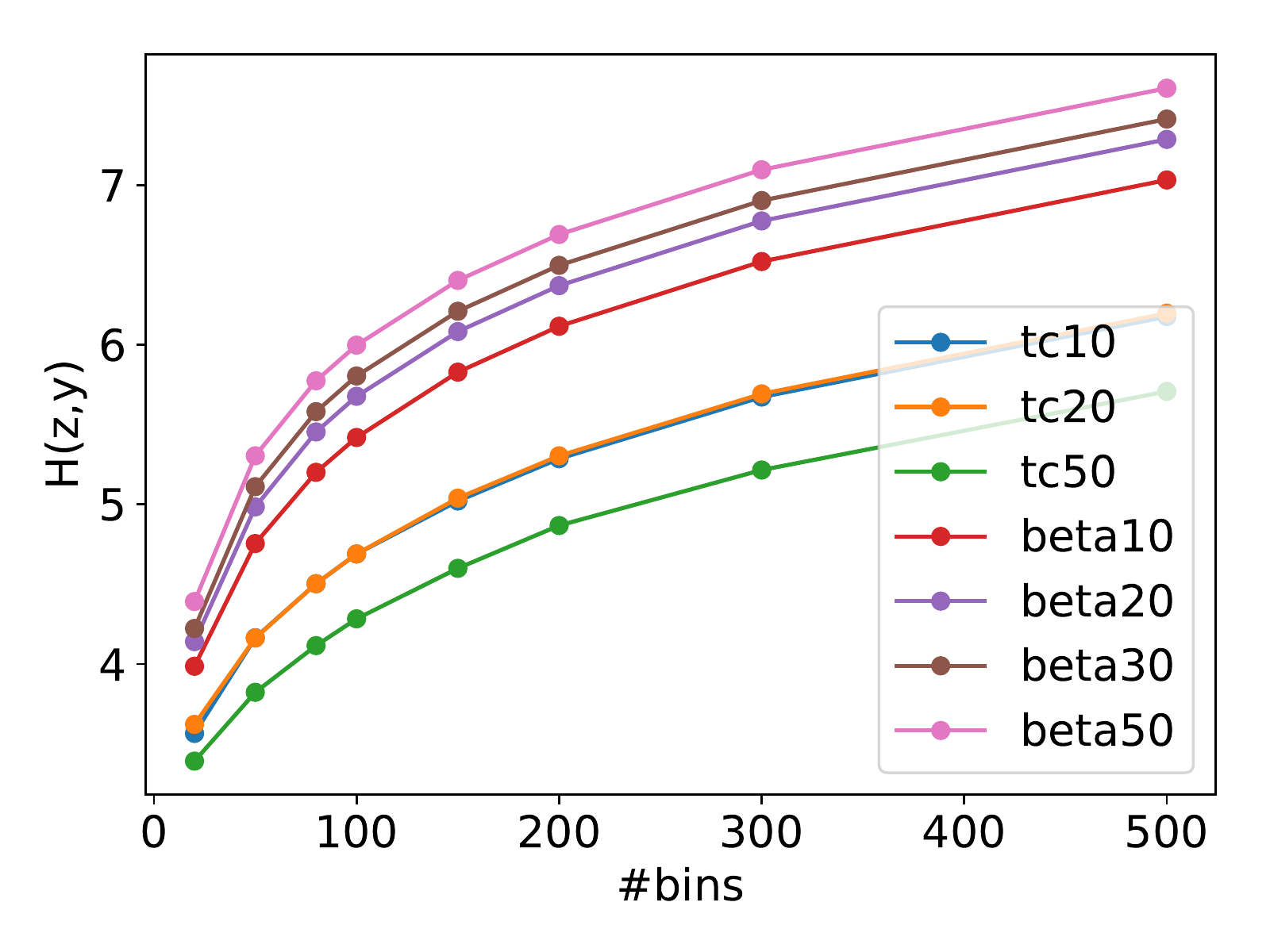}
\par\end{centering}
\caption{Dependences of RMIG (normalized), JEMMIG (normalized) and $\frac{1}{K}\sum_{k=0}^{K-1}H(z_{i^{*}},y_{k})$
on the number of bins. The dataset is dSprite.\label{fig:num_bins}}
\end{figure}

When estimating entropy and mutual information terms using quantization,
we need to specify the value range and the number of bins (\#bins)
in advance. In this paper, we fix the value range to be {[}-4, 4{]}
since most latent values fall within this range. We only investigate
the effect of \#bins on the RMIG and JEMMIG scores for different models
and show the results in Fig.~\ref{fig:num_bins} (left, middle).

We can see that when \#bins is small, RMIG scores are low. This is
because the quantized distributions $Q(z_{i^{*}})$ and $Q(z_{j^{\circ}})$
look similar, causing $I^{*}(z_{i^{*}},y_{k})$ and $I^{\circ}(z_{j^{\circ}},y_{k})$
to be similar as well. When \#bins is large, the quantized distribution
$Q(z_{i^{*}})$ and $Q(z_{j^{\circ}})$ look more different, leading
to higher RMIG scores. RMIG scores are stable when \#bins > 200, which
suggests that finer quantizations do not affect the estimation of
$I(z_{i},y_{k})$ much.

Unlike RMIG scores, JEMMIG scores keep increasing when we increase
\#bins. Note that JEMMIG only differs from RMIG in the appearance
of $H(z_{i^{*}},y_{k})$. Finer quantizations of $z_{i^{*}}$ introduce
more information about $z_{i^{*}}$, hence, always lead to higher
$H(z_{i^{*}},y_{k})$ (see Fig.~\ref{fig:num_bins} (right)). Larger
JEMMIG scores also reflect the fact that finer quantizations of $z_{i^{*}}$
make $z_{i^{*}}$ look more \emph{continuous}, thus, less interpretable
w.r.t the \emph{discrete} factor $y_{k}$.

We provide a detailed explanation about the behaviors of RMIG and
JEMMIG w.r.t \#bins in Appdx.~\ref{subsec:Relationship-between-sampling}.
Despite the fact that \#bins affects the RMIG and JEMMIG scores of
a single model, the relative order among different models remains
the same. It suggests that once we fixed the \#bins, we can use RMIG
and JEMMIG scores to rank different models.

\paragraph{Sensitivity of the number of samples}

\begin{figure}
\begin{centering}
\includegraphics[width=0.32\textwidth]{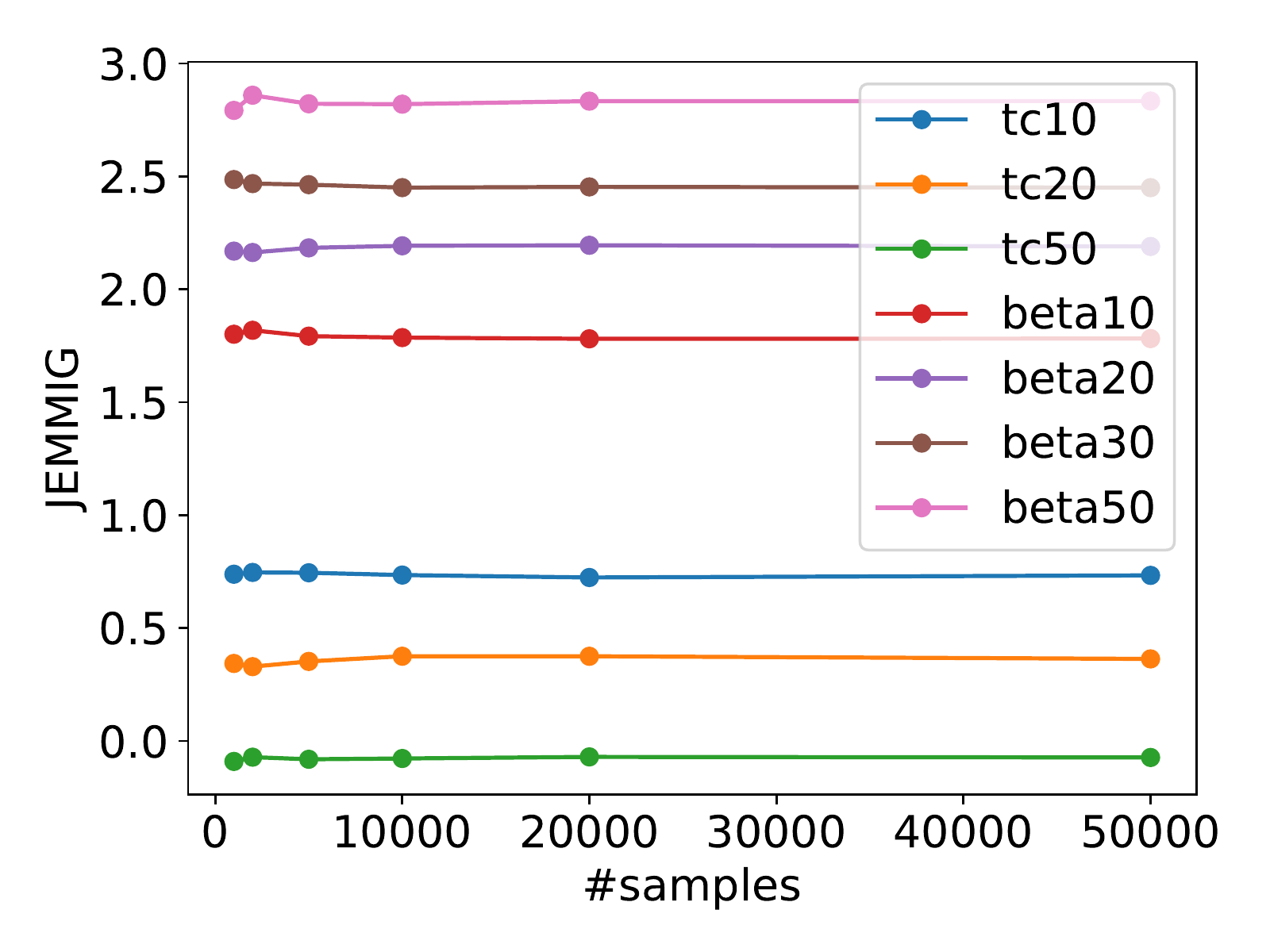}\hspace{0.05\textwidth}\includegraphics[width=0.32\textwidth]{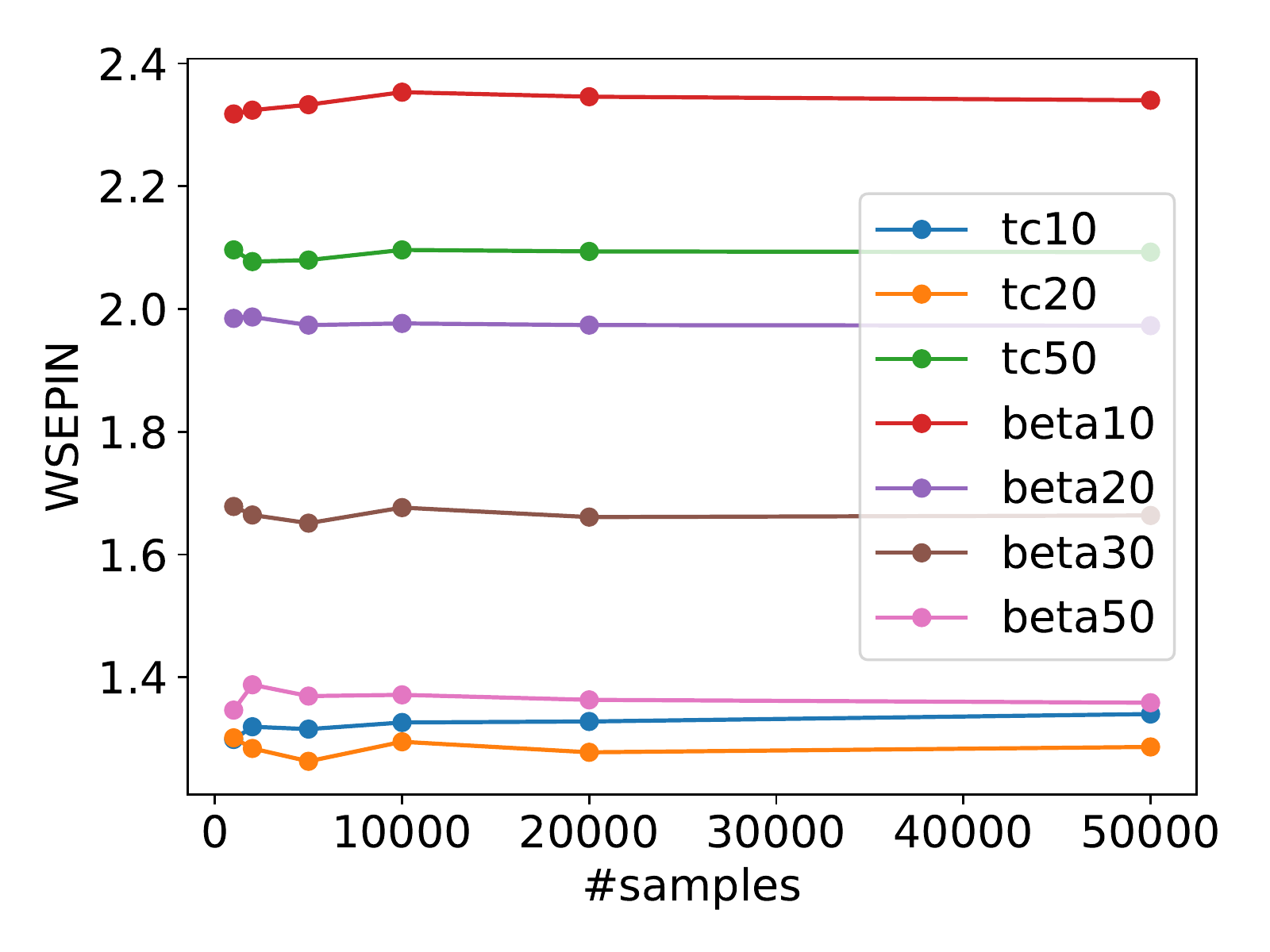}
\par\end{centering}
\caption{Dependences of JEMMIG and WSEPIN on the number of samples. All models
have 10 latent variables. The dataset is dSprites.\label{fig:num_samples}}
\end{figure}

From Fig.~\ref{fig:num_samples} (left, right), it is clear that
the sampling estimation is unbiased and is not affected much by the
number of samples.

\paragraph{Sensitivity of sampling in high dimensional space}

\begin{figure}[h]
\begin{centering}
\subfloat[$I(x,z_{i})$\label{fig:I_x_zi_num_latents}]{\begin{centering}
\includegraphics[width=0.24\textwidth]{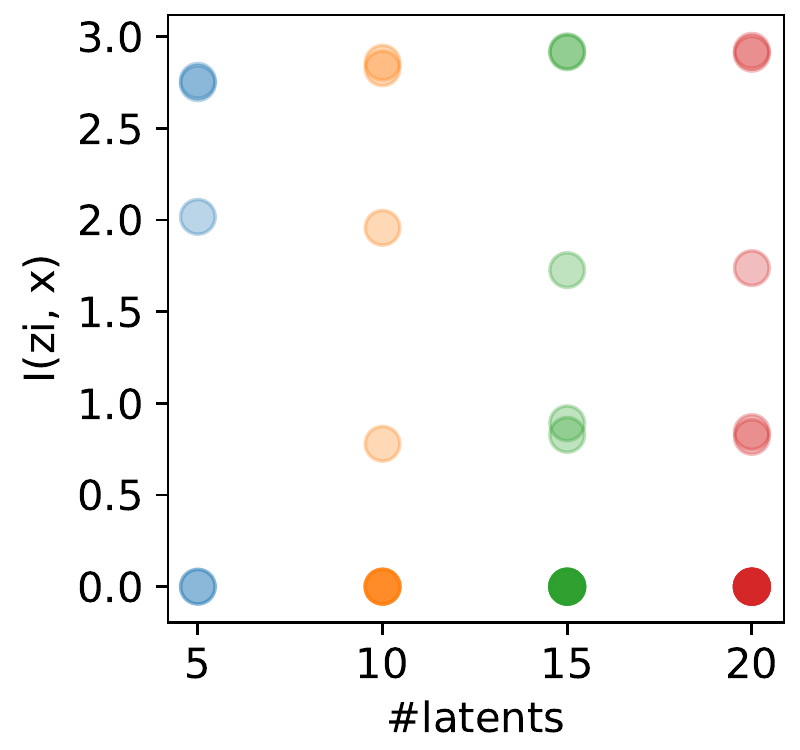}
\par\end{centering}
}\subfloat[$I(x,z)$\label{fig:I_x_z_num_latents}]{\begin{centering}
\includegraphics[width=0.24\textwidth]{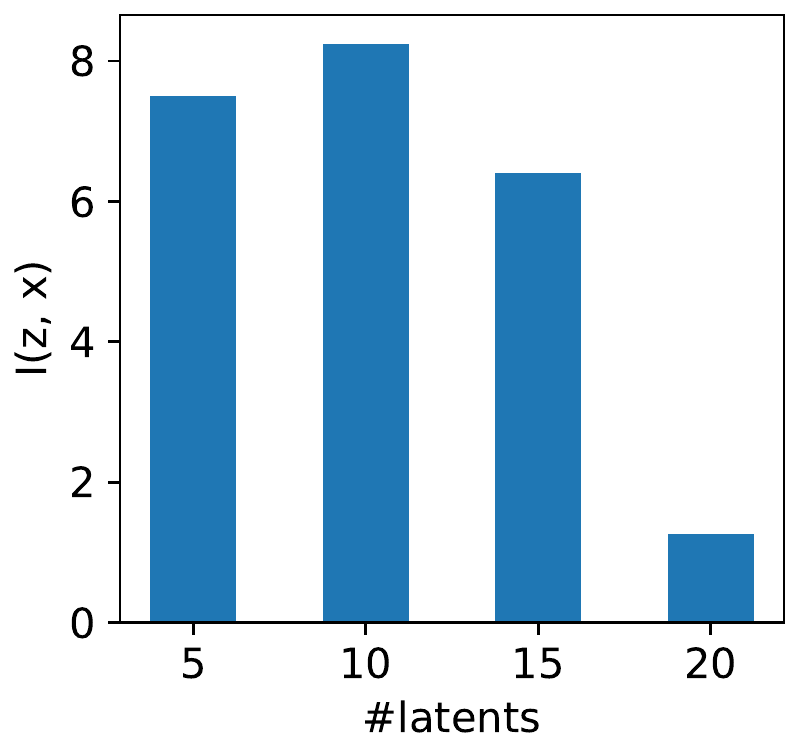}
\par\end{centering}
}\subfloat[$I(z_{i},z_{\protect\neq i})$\label{fig:I_zi_zni_num_latents}]{\begin{centering}
\includegraphics[width=0.24\textwidth]{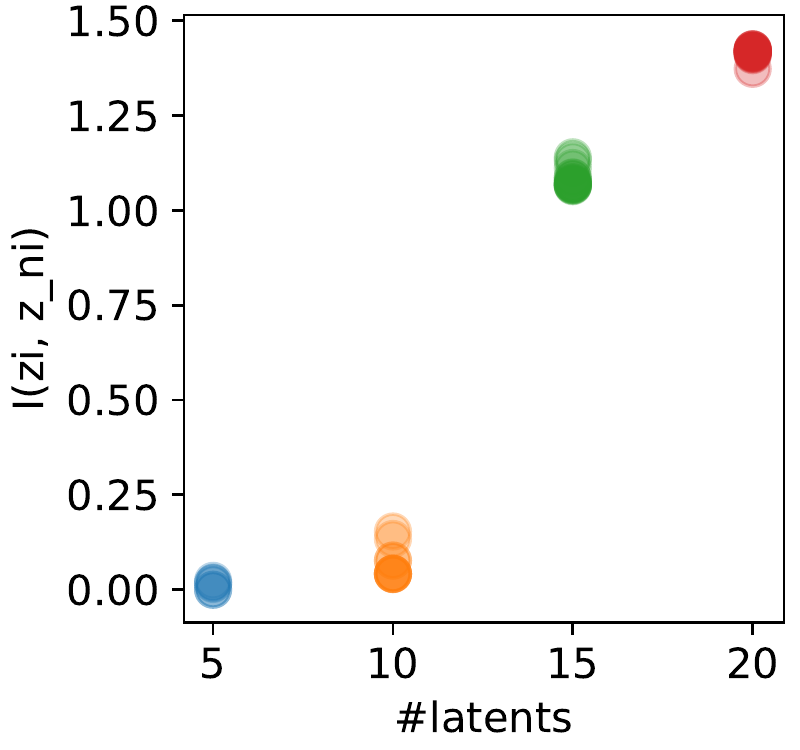}
\par\end{centering}
}\subfloat[$H(z_{\protect\neq i})-H(z)$\label{fig:Hzni_m_H_z_num_latents}]{\begin{centering}
\includegraphics[width=0.24\textwidth]{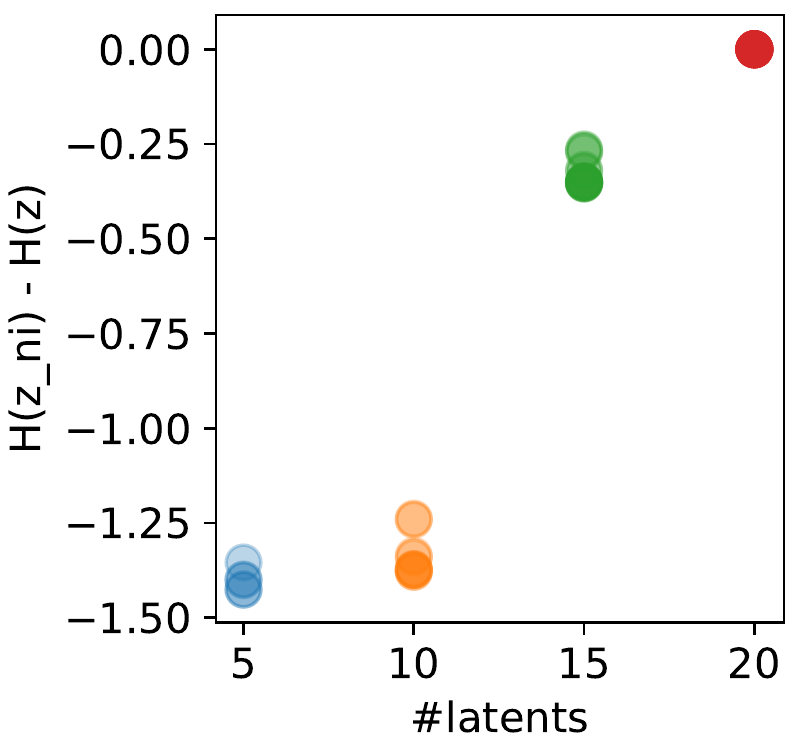}
\par\end{centering}
}
\par\end{centering}
\begin{centering}
\subfloat[$I(x,z)-I(x,z_{\protect\neq i})$\label{fig:SEPIN_zi_num_latents}]{\begin{centering}
\includegraphics[width=0.24\textwidth]{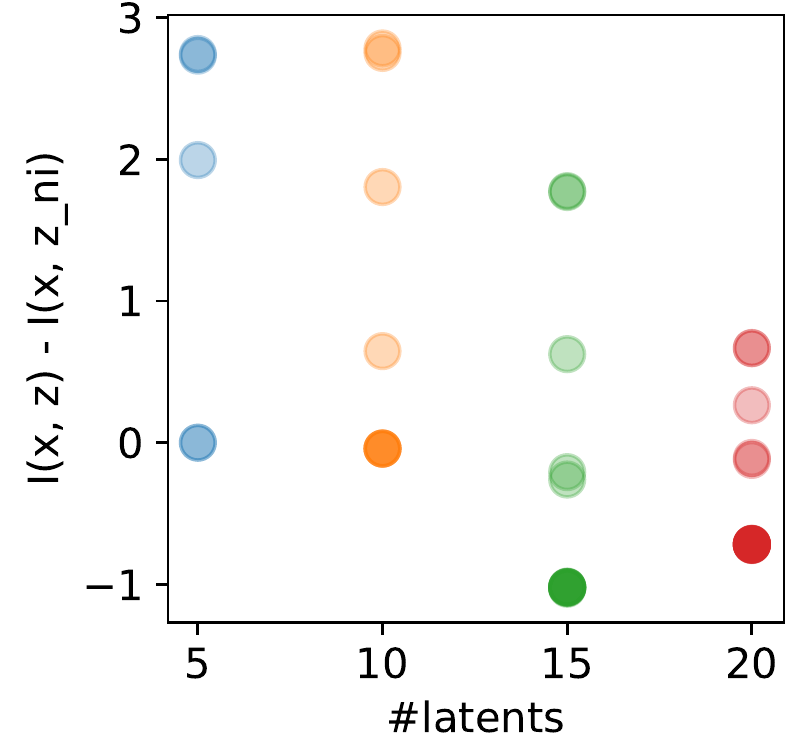}
\par\end{centering}
}\subfloat[WSEPIN\label{fig:WSEPIN_num_latents}]{\begin{centering}
\includegraphics[width=0.24\textwidth]{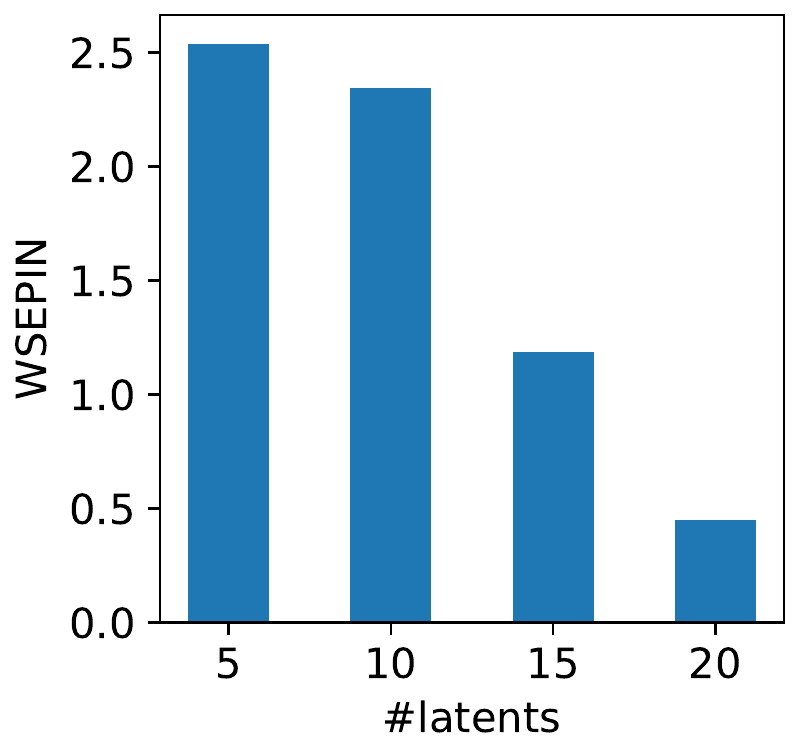}
\par\end{centering}
}\subfloat[JEMMIG$(y_{k})$\label{fig:JEMMIG_yk_num_latents}]{\begin{centering}
\includegraphics[width=0.24\textwidth]{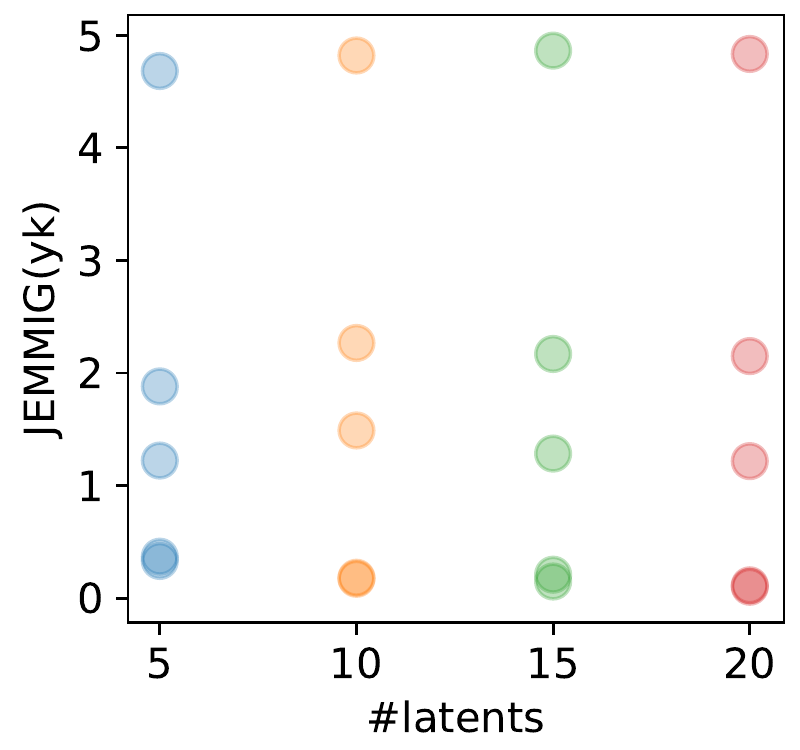}
\par\end{centering}
}\subfloat[JEMMIG\label{fig:JEMMIG_num_latents}]{\begin{centering}
\includegraphics[width=0.24\textwidth]{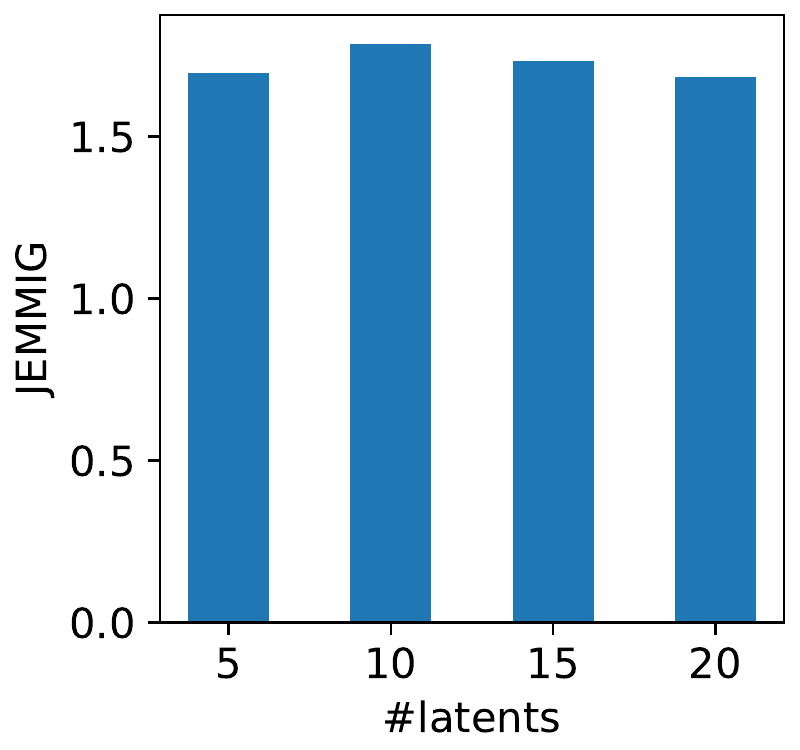}
\par\end{centering}
}
\par\end{centering}
\caption{Dependences of various quantitative measures on the number of latents.
All measures are computed via sampling. The model used in this experiment
is $\beta$-VAE with $\beta=10$.}
\end{figure}

One thing that we should concern about is the performance of our metrics
when the number of latent representations (\#latents) is large (or
$z$ is high-dimensional). In Fig.~\ref{fig:I_x_zi_num_latents},
we see that the informativeness of an individual representations $z_{i}$
is not affected by \#latents. When we increase \#latents, additional
representations are usually noisy ($I(z_{i},x)\approx0$). The total
amount of information captured by the model ($I(x,z)$), by contrast,
highly depends on \#latents (Fig.~\ref{fig:I_x_z_num_latents}).
Unusually, increasing \#latents \emph{reduces} $I(x,z)$ instead of
increasing it. We have not found the final answer for this phenomenon
but possible hypotheses are: i) On a high dimensional space where
most latent representations are noisy (e.g. \#latents=20), $q(z|x)$
may look more similar to $q(z)$, causing the wrong calculation of
$\log\frac{q(z|x)}{q(z)}$, or ii) when \#latents is large, $q(z|x)=\prod_{i=0}^{L-1}q(z_{i}|x)$
is very tiny, thus, may lead to floating point imprecision\footnote{We tried $q(z|x)=\exp\left(\sum_{i=0}^{L-1}\log q(z_{i}|x)\right)$
and it gives similar results as $q(z|x)=\prod_{i=0}^{L-1}q(z_{i}|x)$.}. In Fig.~\ref{fig:I_zi_zni_num_latents}, we see that increasing
\#latents increases $I(z_{i},z_{\neq i})$. This makes sense because
larger \#latents means that $z_{\neq i}$ will contain more information.
However, the change of $I(z_{i},z_{\neq i})$ is sudden when \#latents
change from 10 to 15, which is different from the change of \#latents
from 5 to 10 or 15 to 20. Recall that $I(z_{i},z_{\neq i})=H(z_{i})+H(z_{\neq i})-H(z)$.
Since $H(z_{i})$ can be computed stably, we only plot $H(z_{\neq i})-H(z)$
and show it in Fig.~\ref{fig:Hzni_m_H_z_num_latents}. We can see
that when \#latents = 20, $H(z_{\neq i})\approx H(z)$ which means
we cannot differentiate between $q(z_{\neq i})$ and $q(z)$. The
instability of computation for high dimensional latents becomes clearer
in Fig.~\ref{fig:SEPIN_zi_num_latents} as $I(x,z_{i}|z_{\neq i})=I(x,z)-I(x,z_{\neq i})$
can be $<0$ when \#latents = 15 or 20. This causes the instability
of WSEPIN in Fig.~\ref{fig:WSEPIN_num_latents} despite the results
look reasonable. JEMMIG and RMIG are calculated on individual latents
so they are not affected by \#latents and can provide consistent evaluations
for models with different \#latents.

\subsection{Evaluating independence with correlation matrix}

\begin{figure}[H]
\begin{centering}
\subfloat[FactorVAE (stochastic)]{\begin{centering}
\includegraphics[width=0.26\textwidth]{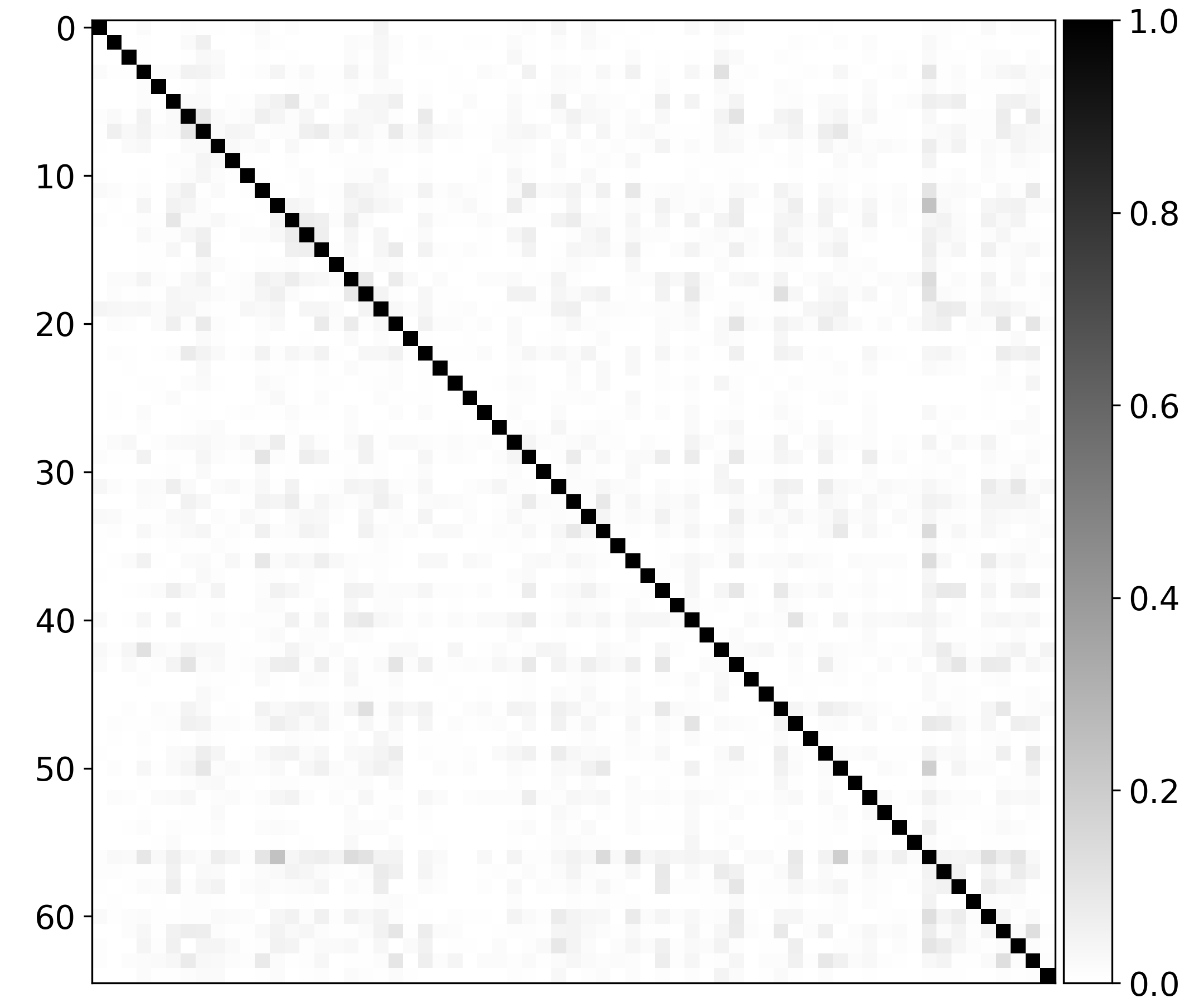}
\par\end{centering}
}\subfloat[$\beta$-VAE (stochastic)]{\begin{centering}
\includegraphics[width=0.26\textwidth]{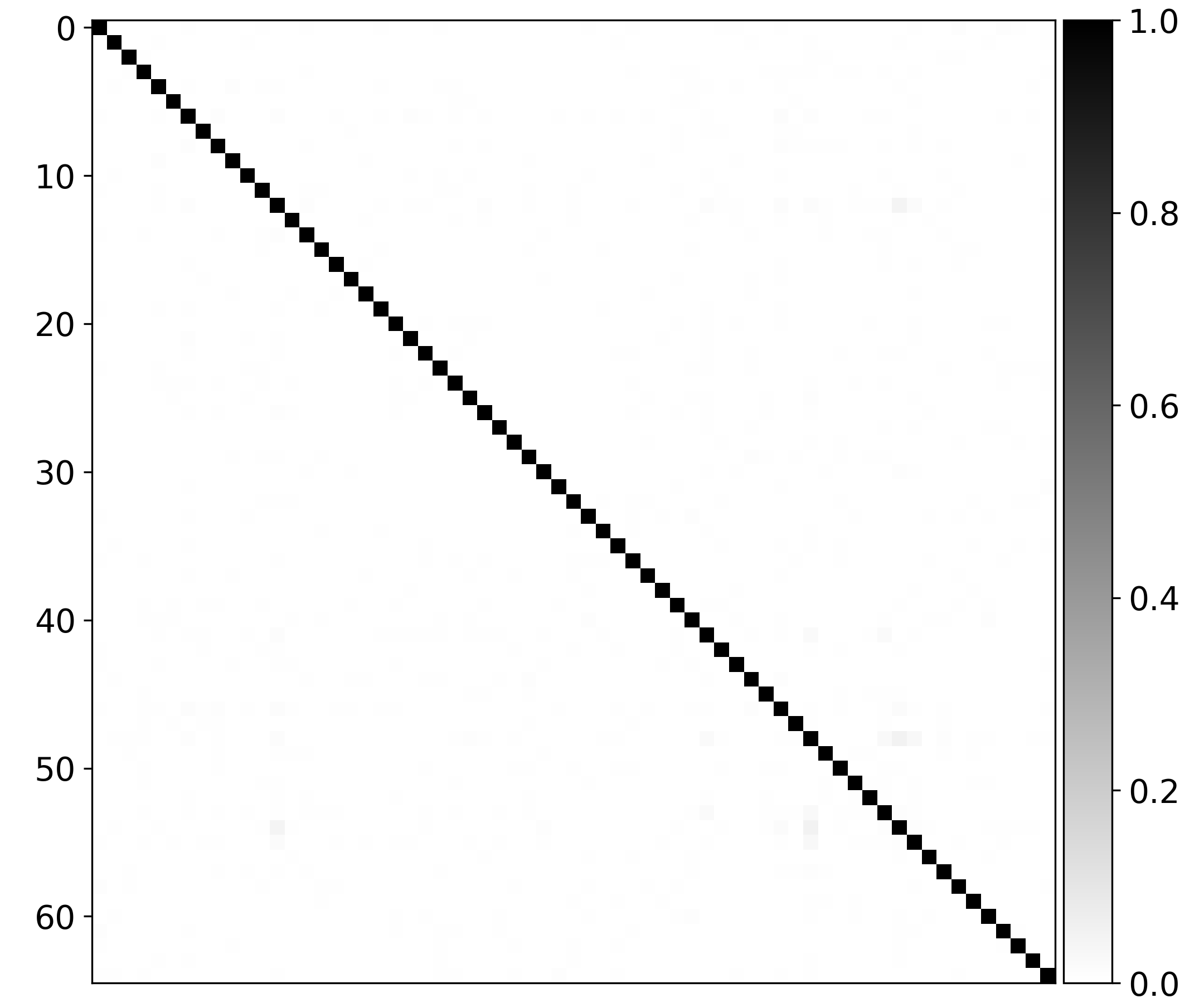}
\par\end{centering}
}\subfloat[AAE (stochastic)]{\begin{centering}
\includegraphics[width=0.26\textwidth]{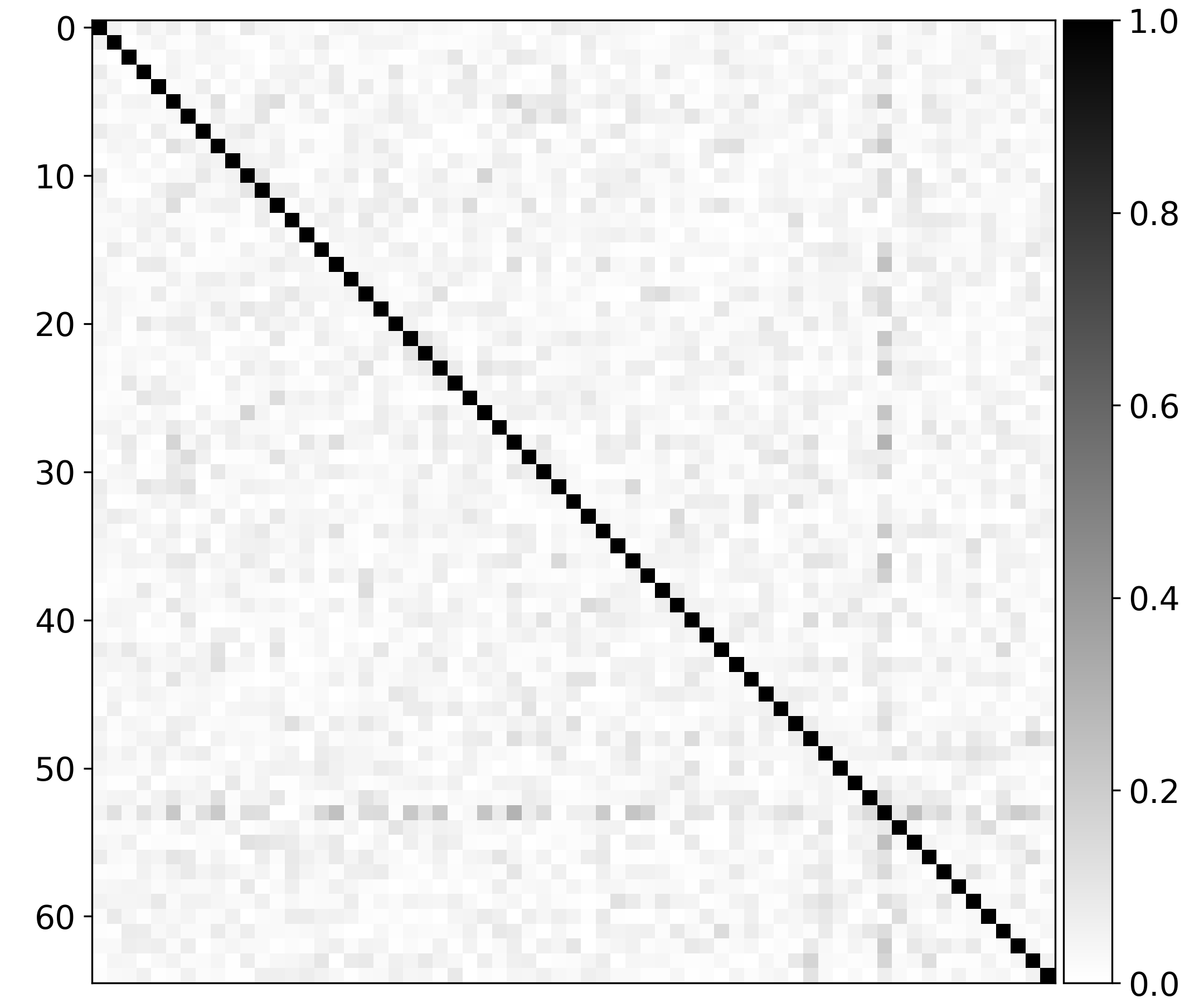}
\par\end{centering}
}
\par\end{centering}
\begin{centering}
\subfloat[FactorVAE (deterministic)]{\begin{centering}
\includegraphics[width=0.26\textwidth]{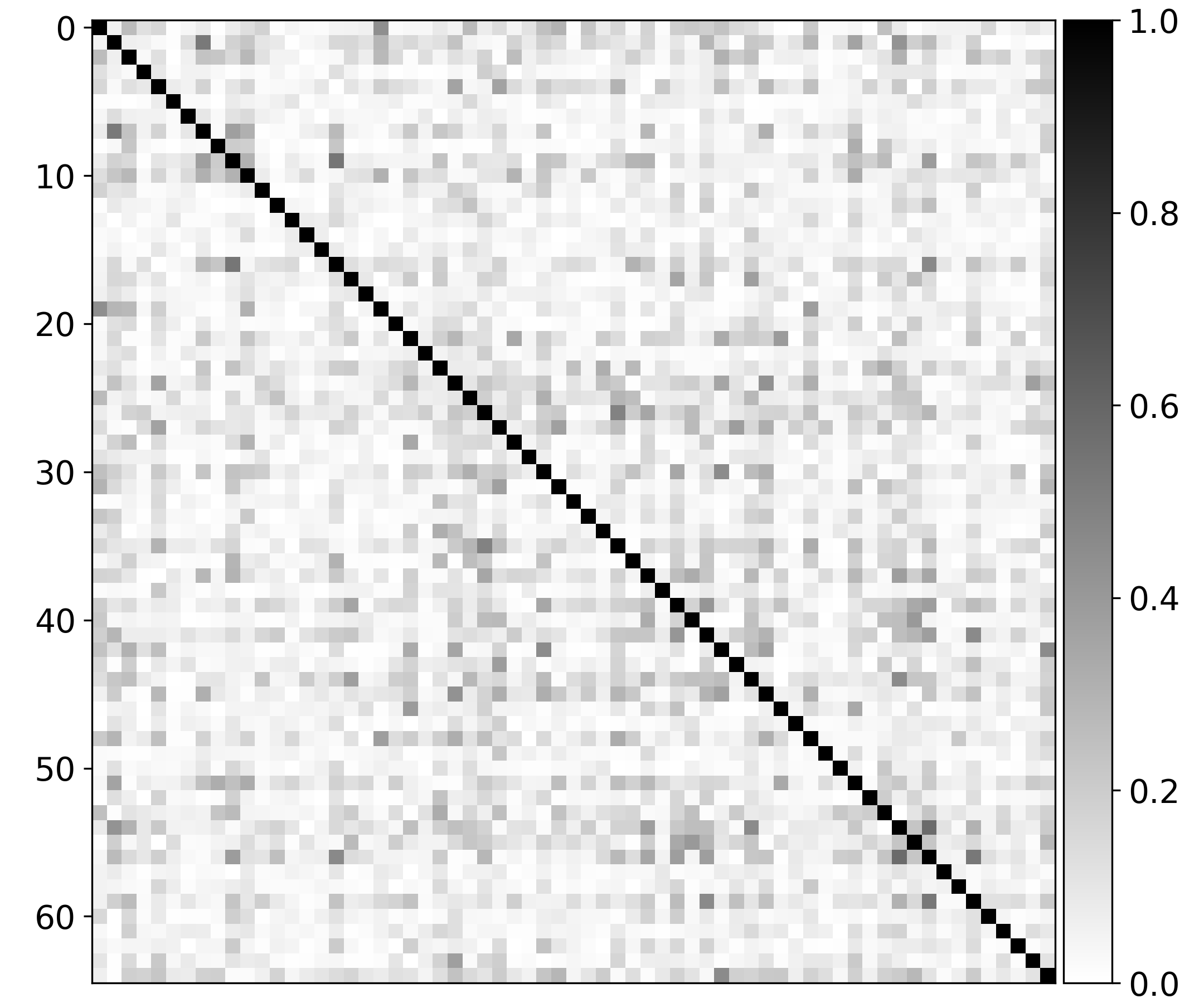}
\par\end{centering}
}\subfloat[$\beta$-VAE (deterministic)]{\begin{centering}
\includegraphics[width=0.26\textwidth]{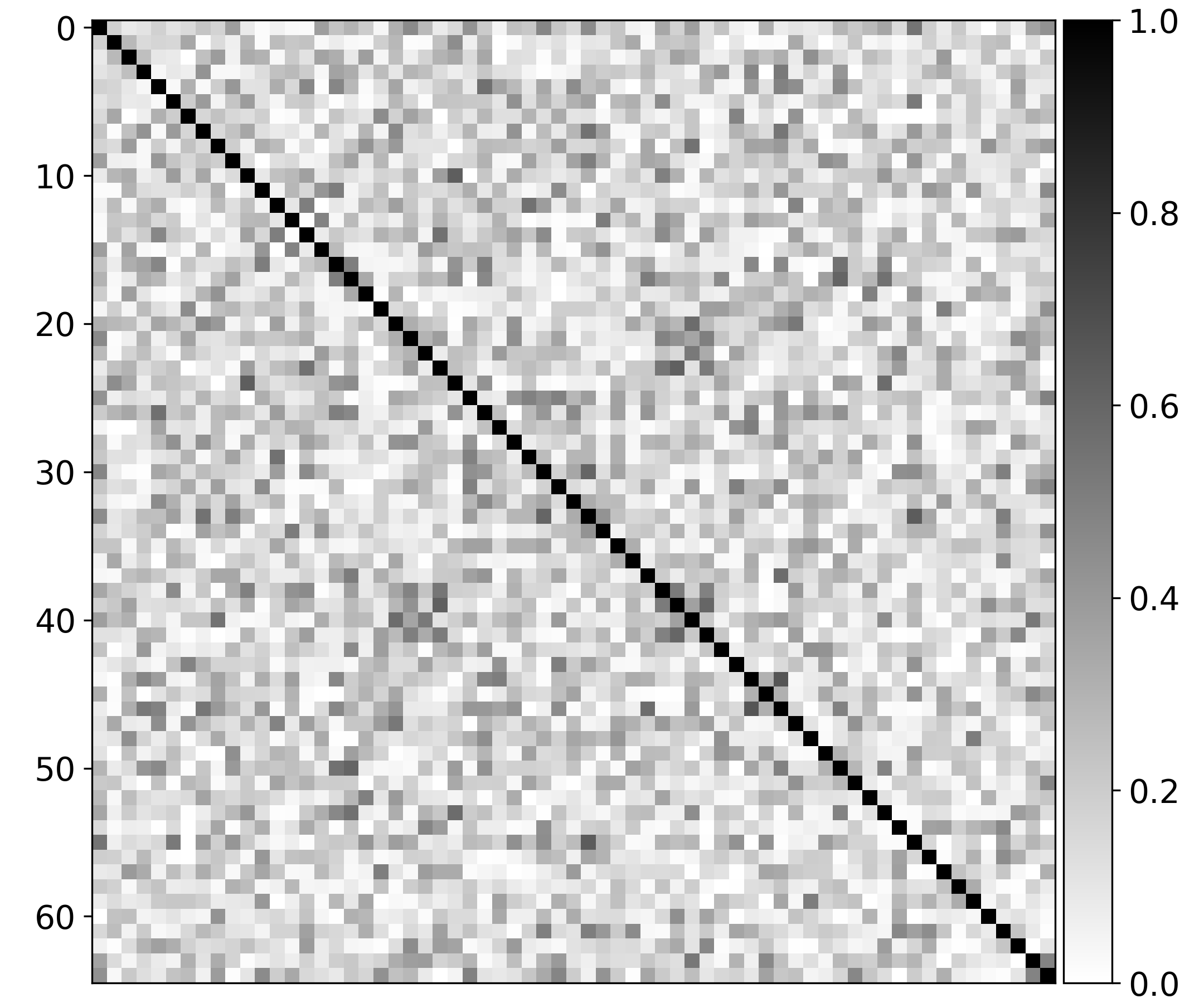}
\par\end{centering}
}\subfloat[AAE (deterministic)]{\begin{centering}
\includegraphics[width=0.26\textwidth]{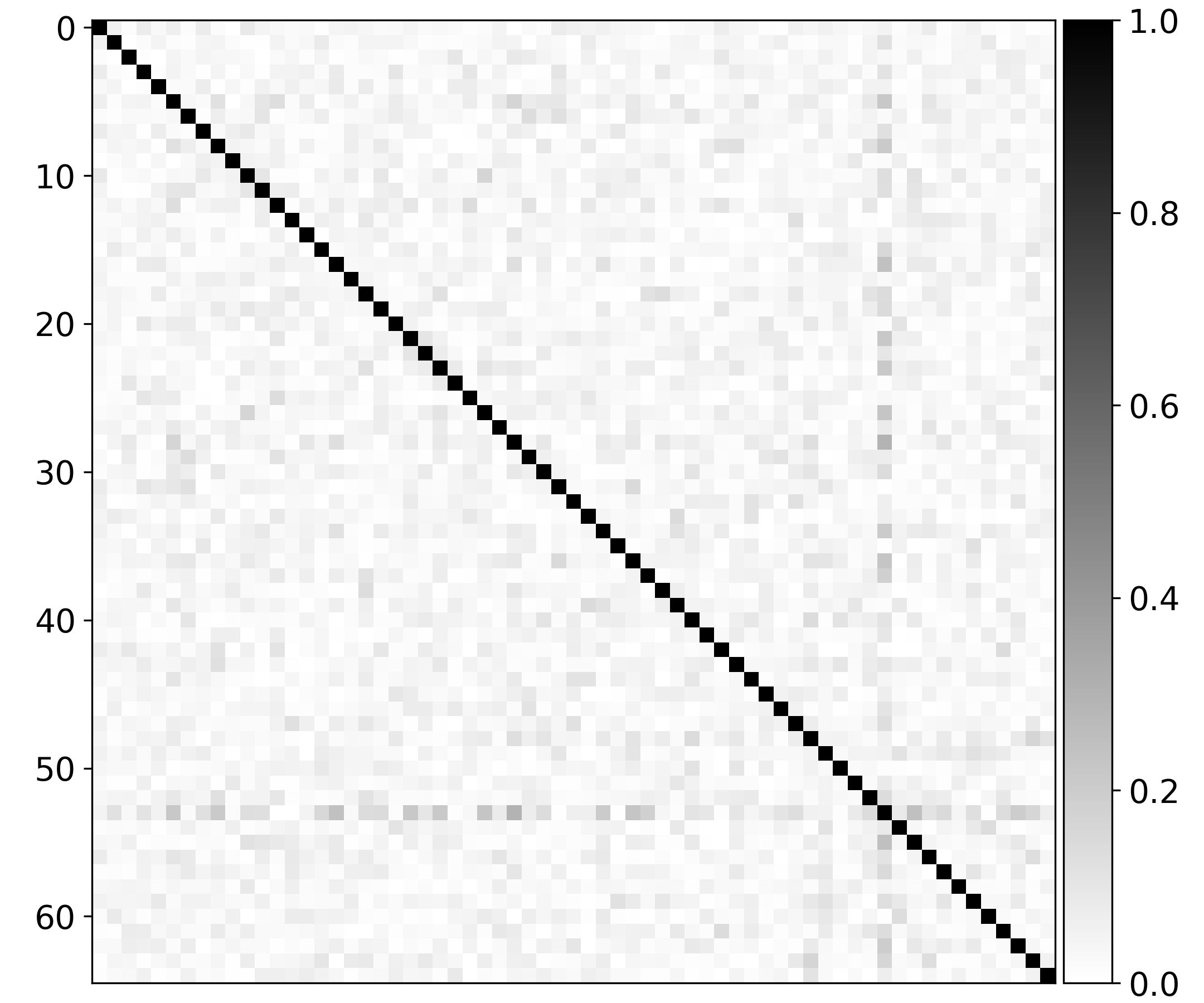}
\par\end{centering}
}
\par\end{centering}
\caption{Correlation matrix of representations learned by FactorVAE, $\beta$-VAE
and AAE.\label{fig:CorrMat}}
\end{figure}

For every $x^{(n)}$ sampled from the training data, we generated
$m=1$ latent samples $z_{i}^{(n,m)}\sim q(z_{i}|x^{(n)})$ and built
a correlation matrix from these samples for each of the models FactorVAE,
$\beta$-VAE and AAE. We also built another version of the correlation
matrix which is based on the $\Expect_{q(z_{i}|x^{(n)})}[z_{i}]$
(called the \emph{conditional means}) instead of samples from $q(z_{i}|x^{(n)})$.
Both are shown in Fig.~\ref{fig:CorrMat}. We can see that the correlation
matrices computed based on the conditional means \emph{incorrectly}
describe the independence between representations of FactorVAE and
$\beta$-VAE. AAE is not affected because it learns deterministic
$z_{i}$ given $x$. Using the correlation matrix is not a principled
way to evaluate independence in disentanglement learning.

\subsection{Trade-off between informativeness, independence and the number of
latent variables\label{subsec:Trade-off-between-informativenes}}

Before starting our discussion, we provide the following fact:
\begin{fact}
\label{PoolBallFact}Assume we try to fill a fixed-size pool with
fixed-size balls given that all the balls must be inside the pool.
The only way to increase the number of the balls without making them
overlapped is reducing their size.
\end{fact}

\begin{figure}[h]
\begin{centering}
\includegraphics[scale=0.3]{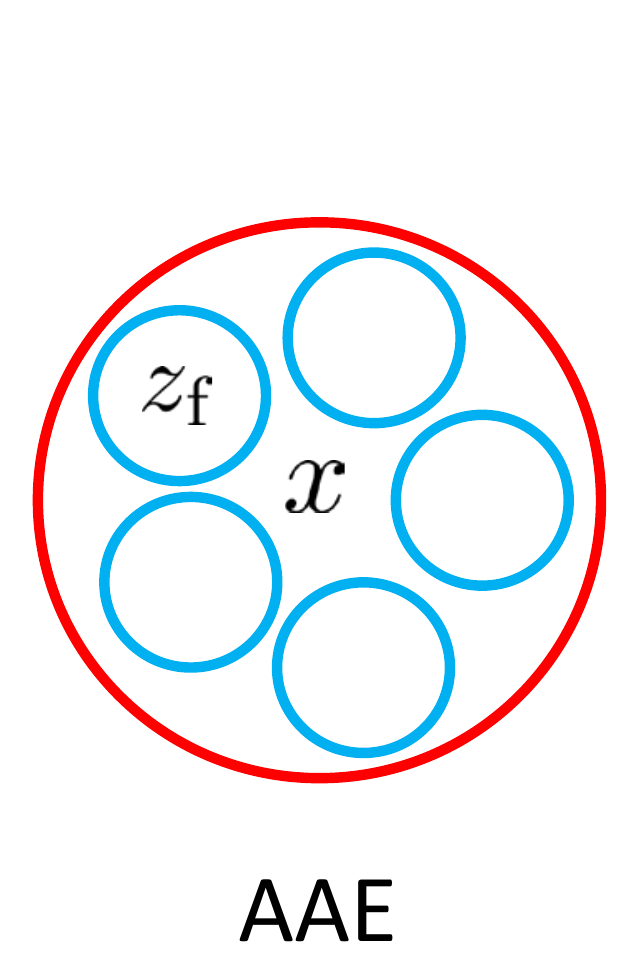}\hspace{0.2\textwidth}\includegraphics[scale=0.3]{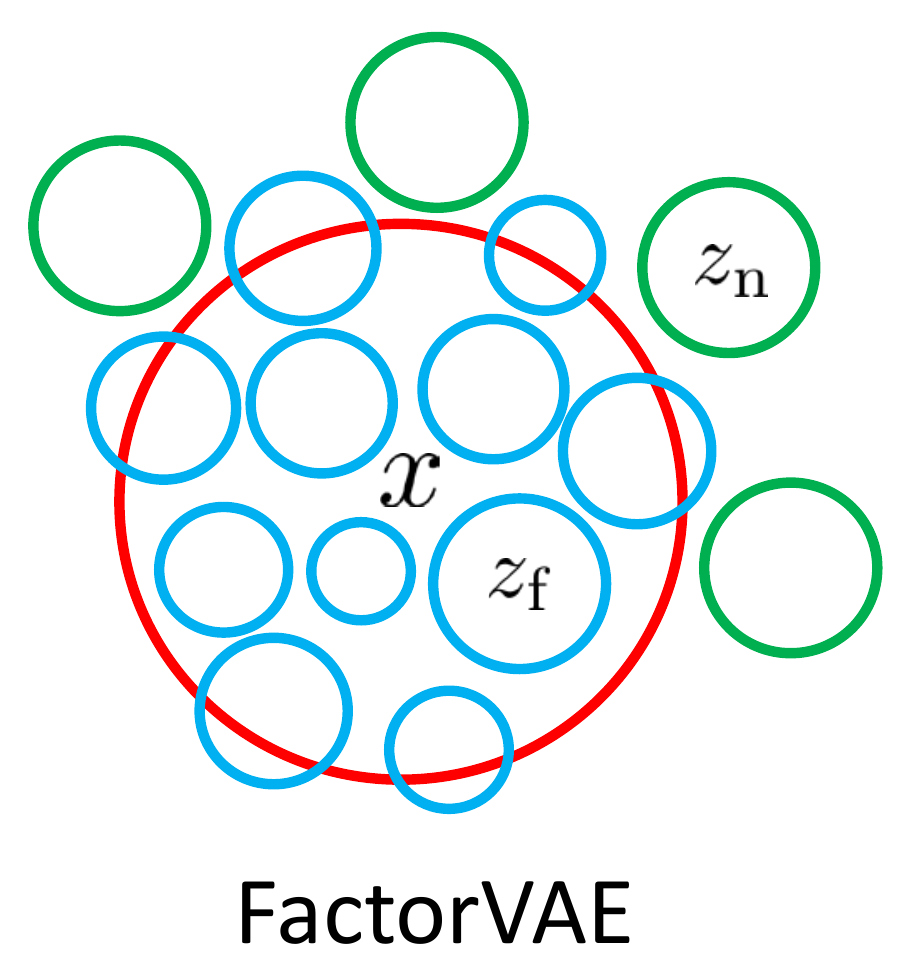}
\par\end{centering}
\caption{Illustration of representations learned by AAE and FactorVAE. A big
red circle represents the total amount of information that $x$ contains
or $H(x)$ which is limited by the amount of training data. Blue circles
are informative representations $z_{\protect\factor}$ and the size
of these circle indicates the informativeness of $z_{\protect\factor}$.
Green circles are noisy representations $z_{\protect\noise}$. AAE
does not contain $z_{\protect\noise}$, only FactorVAE does.\label{fig:Illustration-of-representations}}
\end{figure}

\begin{figure}
\begin{centering}
\subfloat[z\_dim=65]{\begin{centering}
\includegraphics[width=0.25\textwidth]{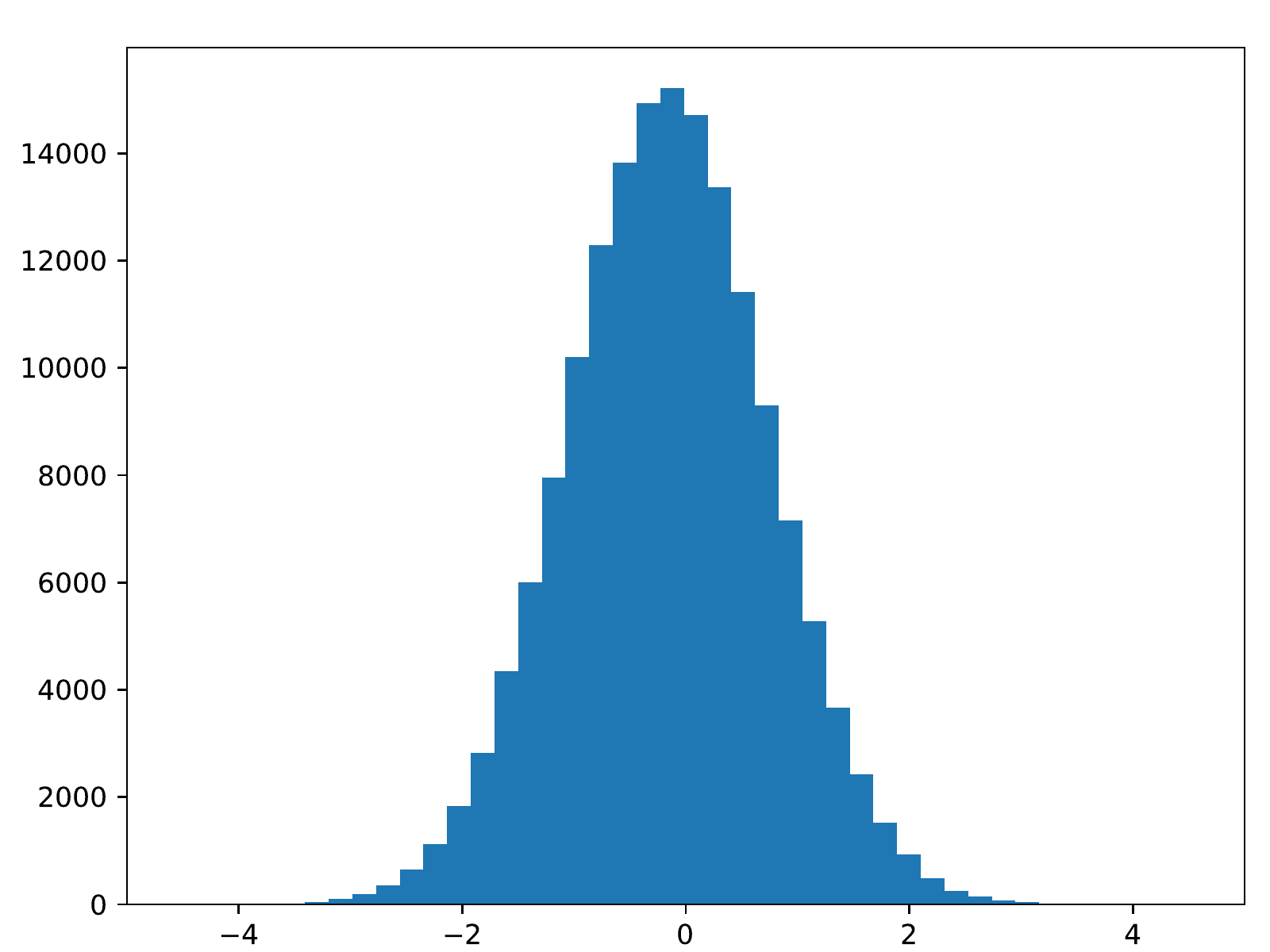}
\par\end{centering}
}\subfloat[z\_dim=100]{\begin{centering}
\includegraphics[width=0.25\textwidth]{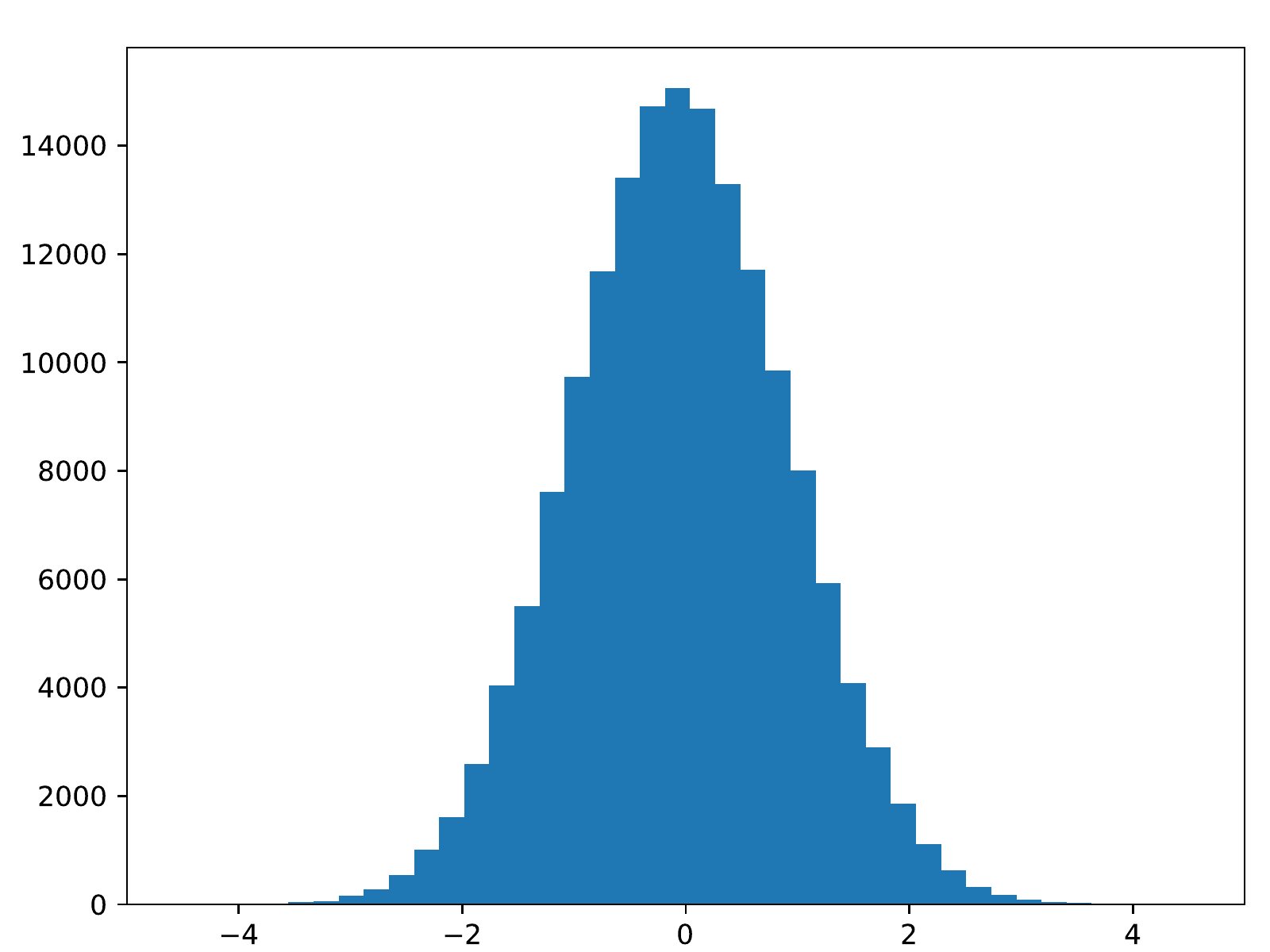}
\par\end{centering}
}\subfloat[z\_dim=200]{\begin{centering}
\includegraphics[width=0.25\textwidth]{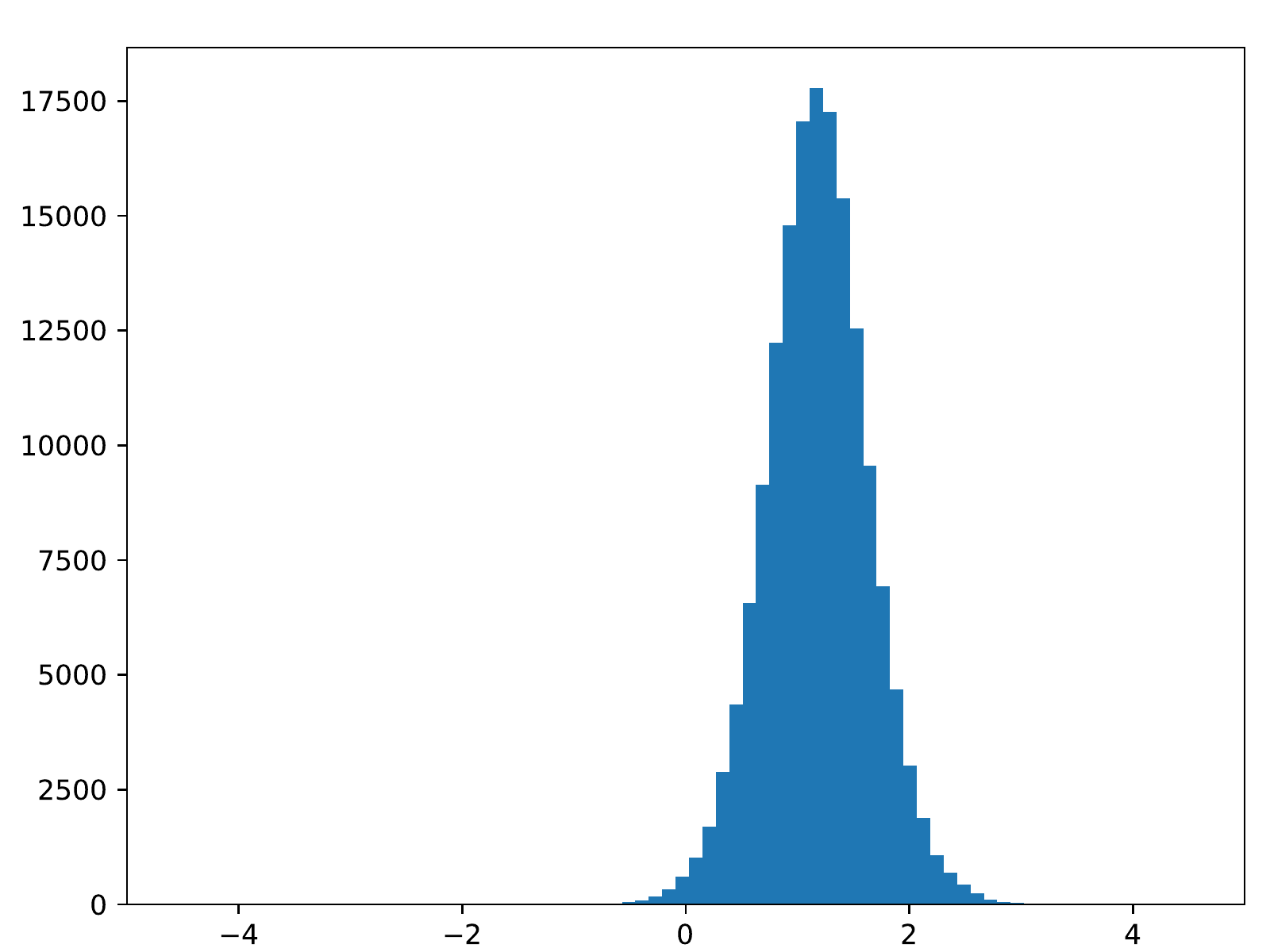}
\par\end{centering}
}
\par\end{centering}
\caption{Distribution of $\protect\Expect_{q(z_{i}|x^{(n)})}[z_{i}]$ over
all $x^{(n)}\sim p_{\protect\Data}(x)$ of a particular representation
$z_{i}$ for different AAE models. \label{fig:Distribution-of-AAE}}
\end{figure}

In the context of representation learning, a pool is $x$ with size
$H(x)$ which depends on the training data. Balls are $z_{i}$ with
size $H(z_{i})$. Fact.~\ref{PoolBallFact} reflects the situation
of AAE (see Fig.~\ref{fig:Illustration-of-representations} left).
In AAE, all $z_{i}$ are deterministic given $x$ so the condition
``all balls are inside the pool'' is met. $H(z_{i})\approx\text{the entropy of }\Normal(0,\Irm)$
which is fixed so the condition ``fixed-size balls'' is also met.
Therefore, when the number of latent variables in AAE increases, \emph{all
}$z_{i}$ \emph{must be less informative }(i.e., $H(z_{i})$ must
decrease)\emph{ }given that the independent constraint on the latent
variables is still satisfied. This is empirically verified in Fig.~\ref{fig:Distribution-of-AAE}
as we see the distribution of $\Expect_{q(z_{i}|x^{(n)})}[z_{i}]$
over all $x^{(n)}\sim p_{\Data}(x)$ becomes narrower when we increase
the number of representations from 65 to 200. Also note that increasing
the number of latent variable from 65 to 100 does not change the distribution.
This suggests that 65 or 100 latent variables are still not enough
to capture all information in the data. 

FactorVAE, however, handles the increasing number of latent variables
in a different way. Thanks to the KL term in the loss function that
forces $q(z_{i}|x)$ to be stochastic, FactorVAE can break the constraint
in Fact~\ref{PoolBallFact} and allows the balls to stay outside
the pool (see Fig.~\ref{fig:Illustration-of-representations} right).
If we increase the number of latent variables but still enforce the
independence constraint on them, FactorVAE will keep a fixed number
of informative representations and make all other representations
``noisy'' with zero informativeness scores. We refer to that capability
of FactorVAE as \emph{code compression}.

\subsection{Analysis of existing metrics for disentanglement\label{subsec:Analysis-of-existing}}

In this section, we analyze recent metrics, including Z-diff score
\cite{higgins2017beta,kim2018disentangling}, Separated Attribute
Predictability (SAP) \cite{kumar2017variational}, Mutual Information
Gap (MIG) \cite{chen2018isolating}, Disentanglement/Compactness/Informativeness
\cite{eastwood2018framework}, Modularity/Explicitness \cite{ridgeway2018learning}.

The main idea behind the Z-diff score \cite{higgins2017beta,kim2018disentangling}
is that if a ground truth generative factor $y_{k}$ ($k\in\{0,1,...,K\}$)
is well aligned with a particular disentangled representation $z_{i}$
(although we do not know which $i$), we can use a simple classifier
to predict $k$ using information from $z$. Higgins et al. \cite{higgins2017beta}
use a linear classifier while Kim et. al. \cite{kim2018disentangling}
use a majority-vote classifier. The main drawback of this metric is
that it assumes knowledge about \emph{all} ground truth factors that
generate the data. Hence, it is only applicable for a toy dataset
like dSprites. Another drawback lies in the complex procedure to compute
the metric, which requires training a classifier. Since the classifier
is sensitive to the chosen optimizer, hyper-parameters and weight
initialization, it is hard to ensure a fair comparison. 

The SAP score \cite{kumar2017variational} is computed based on the
correlation matrix $C$ between the latent variables $z$ and the
ground truth factors $y$. If a latent $z_{i}$ and a factor $y_{k}$
are both continuous, the (square) correlation $C_{i,k}$ between them
is equal to $\frac{\text{Cov}^{2}(z_{i},y_{k})}{\text{Var}(z_{k})\text{Var}(y_{k})}$
and is in {[}0, 1{]}. However, if the factor $y_{k}$ is discrete,
computing the correlation between continuous and discrete variables
is not straightforward. The authors handled this problem by learning
a classifier that predicts $y_{k}$ given $z_{i}$ and used the balanced\footnote{To achieve balance, the classifier uses the same number of samples
for all categories of $y_{k}$ during training and testing} prediction accuracy as a replacement. Then, for each factor $y_{k}$,
they sorted $C_{:,k}$ in the descending order and computed the difference
between the top two scores. The mean of the \emph{difference scores}
for all factors was used as the final SAP score. The intuition for
this metric is that if a latent $z_{i}$ is the most representative
for a factor $y_{k}$ (due to the highest correlation score), then
other latent variables $z_{\neq i}$ should not be related to $y_{k}$,
and thus, the difference score for $y_{k}$ should be high. We believe
the SAP score is more sensible than Z-diff but it is only suitable
when both the ground truth factors and the latent variables are continuous
as no classifier is required. Moreover, if we have $K$ discrete ground
truth factors and $L$ latent variables, the number of classifiers
we need to learn is $L\times K$, which is unmanageable when $L$
is large.

The MIG score \cite{chen2018isolating} shares the same intuition
as the SAP score but is computed based on the mutual information between
every pair of $z_{i}$ and $y_{k}$ instead of the correlation coefficient.
Thus, the MIG score is theoretically more appealing than the SAP score
since it can capture nonlinear relationships between latent variables
and factors while the SAP score cannot. The MIG score, to some extent,
reflects the concept ``interpretability'' that we discussed in Section~\ref{sec:Rethinking-Disentanglement}
in the main text.

Eastwood et. al. \cite{eastwood2018framework} proposed three different
metrics namely \emph{``disentanglement''}, \emph{``completeness''},
and \emph{``informativeness''} to quantify disentangled representations.
These metrics are computed based on a so-called ``\emph{important
matrix}'' $R$ whose element $R_{ik}$ is the relative importance
of $z_{i}$ (w.r.t other $z_{\neq i}$) in predicting $y_{k}$. More
concretely, for each factor $y_{k}$ ($k=0,...,K-1$), they train
a regressor (LASSO or Random Forest) to predict $y_{k}$ from $z$
and use the weight vector provided by this regressor to define $R_{\cdot k}$.
The ``disentanglement'' score $D_{i}$ quantifies the degree to
which a latent $z_{i}$ captures \emph{at most} one generative factor
$y_{k}$. $D_{i}$ is computed as $D_{i}=(1-H_{K}(P_{i\cdot}))$ where
$H_{K}(P_{i\cdot})=\sum_{k=0}^{K-1}-P_{ik}\log P_{ik}$ and $P_{ik}=\frac{R_{ik}}{\sum_{k'=0}^{K-1}R_{ik'}}$
which can be seen as the \emph{``probability''} of predicting $y_{k}$
instead of $y_{\neq k}$ from $z_{i}$. Similarly, the ``completeness''
score $C_{k}$ quantifies the degree to which a ground truth factor
$y_{k}$ is captured by a single latent $z_{i}$ ($i=0,...,L-1$),
computed as $C_{k}=1-H_{L}(\tilde{P}_{\cdot k})$ where $H_{L}(\tilde{P}_{\cdot k})=\sum_{i=0}^{L-1}-\tilde{P}_{ik}\log\tilde{P}_{ik}$
and $\tilde{P}_{ik}=\frac{R_{ik}}{\sum_{i'=0}^{L-1}R_{i'k}}$. The
``informativeness'' score describes the total amount of information
of a particular factor $y_{k}$ captured by all representations $z$.
However, the authors use the prediction error $E_{k}$ of the $k$-th
regressor to quantify ``informativeness'' instead of $I(y_{k},z)$.
Despite being well-motivated, these metrics still have several drawbacks.
First, using three different metrics to quantify disentangled representations
is not as convenient as using a single metric like MIG \cite{chen2018isolating}.
For example, how can we compare two models A and B if A has a better
``disentanglement'' score but a worse ``completeness'' score than
B? Second, these metrics do not apply for categorical factors with
$C$ classes since in this case the model weight is not a vector but
an $L\times C$ matrix. Third, defining the pseudo-distribution $P_{ik}=\frac{R_{ik}}{\sum_{k'=0}^{K-1}R_{ik'}}$
seems ad hoc because i) the weight magnitudes $R_{ik}$ are unbounded
and can vary significantly (see Appdx.~\ref{subsec:Matrices}), and
ii) $P_{ik}$ strongly depends on the available ground truth factors
(e.g. the value of $P_{ik}$ will change if we only consider 2 instead
of 5 factors).

Ridgeway et. al. \cite{ridgeway2018learning} proposed two metrics
called ``modularity'' and ``explicitness'' that have similar interpretations
as ``disentanglement'' and ``informativeness'' discussed above
but differ in implementation. Specifically, they compute the ``modularity''
score $M_{i}$ for a representation $z_{i}$ as $M_{i}=1-\frac{\sum_{k=0}^{K-1}\left(I(z_{i},y_{k})-T_{ik}\right)^{2}}{I^{2}(z_{i},y_{k^{*}})\times(K-1)}$
where $k^{*}=\text{argmax}_{k}I(z_{i},y_{k})$ and $T_{ik}=\begin{cases}
I(z_{i},y_{k^{*}}) & \text{if }k=k^{*}\\
0 & \text{otherwise}
\end{cases}$. Like the ``disentanglement'' score $D_{i}$, $M_{i}$ is also
ad hoc and is undefined when the number of ground truth factors is
1. The ``explicitness'' score $E_{k}$ for each ground truth factor
$y_{k}$ is computed as the ROC curve of a logistic classifier that
predicts $y_{k}$ from $z$. It turns out that $E_{k}$ is just a
way to bypass computing $I(y_{k},z)$.

\subsection{The mutual information matrix $I(z_{i},y_{k})$ and the importance
matrix $R_{ik}$\label{subsec:Matrices}}

\begin{figure}
\begin{centering}
\subfloat[$I(z_{i},y_{k})$ w. $q(z_{i}|x)$]{\begin{centering}
\includegraphics[width=0.32\textwidth]{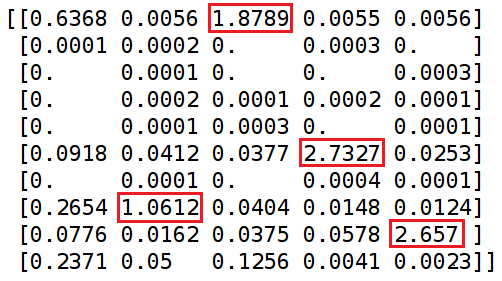}
\par\end{centering}
}\subfloat[$I(z_{i},y_{k})$ w.o. $q(z_{i}|x)$\label{fig:matrix_ridgeway}]{\begin{centering}
\includegraphics[width=0.32\textwidth]{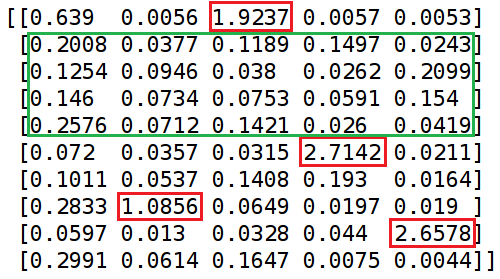}
\par\end{centering}
}\subfloat[$R_{ik}$\label{fig:matrix_eastwood}]{\begin{centering}
\includegraphics[width=0.295\textwidth]{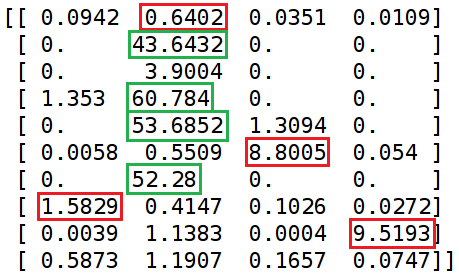}
\par\end{centering}
}
\par\end{centering}
\caption{\textbf{(a):} Our mutual information matrix $I(z_{i},y_{k})$, \textbf{(b):}
The mutual information matrix $I(z_{i},y_{k})$ in \cite{ridgeway2018learning},
\textbf{(c):} The importance matrix $R_{ik}$ in \cite{eastwood2018framework}.
In (a) and (b), the columns corresponding to the following ground
truth factors: ``shape'', ``scale'', ``rotation'', ``x-position'',
``y-position''. In (c), the column for ``shape'' is excluded because
the metrics in \cite{eastwood2018framework} do not support categorical
factors. Values corresponding to disentangled representations are
highlighted in red. Defective values are highlighted in green. The
model is FactorVAE with TC=20.\label{fig:matrices}}
\end{figure}

In Fig.~\ref{fig:matrices}, we compare our mutual information matrix
$I(z_{i},y_{k})$ with the counterpart in \cite{ridgeway2018learning}
and the importance matrix $R_{ik}$ in \cite{eastwood2018framework}.
It is clear that all matrices can capture disentangled representations
(those highlighted in red) well since their corresponding values are
high compared to other values in the \emph{same column}. However,
the matrix $I(z_{i},y_{k})$ in \cite{ridgeway2018learning} usually
overestimates noisy representations since it uses $\Expect_{q(z_{i}|x)}[z_{i}]$
instead of $q(z_{i}|x)$. The matrix $R_{ik}$ in \cite{eastwood2018framework}
sometimes assign very high absolute values for noisy representations
since the regressor's weights are unbounded. These flaws make the
metrics in \cite{ridgeway2018learning} and in \cite{eastwood2018framework}
inaccurate and unstable, especially ``modularity'' and ``disentanglement''
since they require normalization over rows.

\subsection{Computing metrics for informativeness, separability and interpretability\label{subsec:Computing-metrics-for}}

The metrics for informativeness, separability and interpretability
in Section.~\ref{sec:Robust-Evaluation-Metrics} requires computing
$H(z_{i})$, $H(z_{i}|x)$, $H(z_{\neq i})$, $H(z)$, and $H(z_{i},y_{k})$.
We can compute these entropies via quantization or sampling. Quantization
is only applicable when $z_{i}$ is a scalar. If $z_{i}$ is a high-dimensional
vector, we need to use sampling. Below, we describe how to compute
$H(z)$ via sampling and $H(z_{i})$ via quantization. Other cases
can be derived similarly.

\paragraph{Computing $H(z)$ via sampling}

\begin{align}
H(z) & =-\Expect_{q(z)}\left[\log q(z)\right]\nonumber \\
 & =-\Expect_{q(z,x)}\left[\log\Expect_{p_{\Data}(x)}\left[q(z|x)\right]\right]\nonumber \\
 & =-\frac{1}{M}\sum_{m=1}^{M}\left[\log\frac{1}{N}\sum_{n=1}^{N}q\left(z^{(m)}|x^{(n)}\right)\right]\label{eq:sampling_1}\\
 & =-\frac{1}{M}\sum_{m=1}^{M}\left[\log\frac{1}{N}\sum_{n=1}^{N}\left(\prod_{i=1}^{L}q\left(z_{i}^{(m)}|x^{(n)}\right)\right)\right]\label{eq:sampling_2}
\end{align}
In Eq.~\ref{eq:sampling_1}, we use Monte Carlo sampling to estimate
the expectations outside and inside the log function. The corresponding
sample sizes are $M$ and $N$. In Eq.~\ref{eq:sampling_2}, we use
the assumption $q\left(z^{(m)}|x^{(n)}\right)=\prod_{i=1}^{L}q\left(z_{i}^{(m)}|x^{(n)}\right)$.
Please note that the entropy $H(z)$ computed via sampling can be
negative if $z$ is continuous since we use the density function $q(z|x)$.

\paragraph{Computing $H(z_{i})$ via quantization}

We can compute $H(z_{i})$ via quantization as follows:
\[
H(z_{i})=-\sum_{s_{i}\in\Scal}Q(s_{i})\log Q(s)
\]
where $\Scal$ is a set of all quantized bins $s_{i}$ corresponding
to $z_{i}$; $Q(s_{i})$ is the probability mass function of $s_{i}$.
To ensure consistency among different $z_{i}$ as well as different
models, we apply the same value range for all latent variables. In
practice, we choose the range $[-4,4]$ since most of the latent values
fall within this range. We divide this range into equal-size bins
to form $\Scal$.

We can compute $Q(s_{i})$ as follows:
\[
Q(s_{i})=\frac{1}{N}\sum_{n=1}^{N}Q\left(s_{i}|x^{(n)}\right)
\]
We compute $Q\left(s_{i}|x^{(n)}\right)$ based on its definition,
which is:
\begin{equation}
Q(s_{i}|x^{(n)})=\int_{a}^{b}q(z_{i}|x^{(n)})\ dz_{i}\label{eq:Q(s|x)}
\end{equation}
where $a$, $b$ are two ends of the bin $s_{i}$.

There are two ways to compute $Q(s_{i}|x^{(n)})$. In the first way,
we simply consider the unnormalized $Q^{'}(s_{i}|x^{(n)})$ as the
area of a rectangle whose width is $b-a$ and height is $q(\bar{z}_{i}|x^{(n)})$
with $\bar{z}_{i}$ at the center value of the bin $s_{i}$. Then,
we normalize $Q^{'}(s_{i}|x^{(n)})$ over all bins to get $Q(s_{i}|x^{(n)})$.
In the second way, if $q(z_{i}|x^{(n)})$ is a Gaussian distribution,
we can estimate the above integral with a closed-form function (see
Appdx.~\ref{subsec:Definite-integral-of} for detail).

\subsection{Relationship between sampling and quantization\label{subsec:Relationship-between-sampling}}

Denote $H_{\text{s}}(z_{i}|x)$ and $H_{\text{q}}(z_{i}|x)$ to be
the sampling and quantization estimations of an entropy $H(z_{i}|x)$,
respectively. Because $H_{\text{s}}(z_{i}|x)$ is the expectation
of $\log q(z_{i}|x)$, $H_{\text{q}}(z_{i}|x)$ is the expectation
of $\log Q(z_{i}|x)$, and $Q(z_{i}|x)\approx q(z_{i}|x)\times\text{bin width}$
if the bin width is \emph{small enough}, there exists a \emph{gap}
between $H_{\text{s}}(z_{i}|x)$ and $H_{\text{q}}(z_{i}|x)$, specified
as follows:

\begin{align*}
H_{\text{q}}(z_{i}|x) & =H_{\text{s}}(z_{i}|x)-\log(\text{bin width})\\
 & =H_{\text{s}}(z_{i}|x)-\log\left(\frac{\text{value range}}{\text{\#bins}}\right)\\
 & =H_{\text{s}}(z_{i}|x)-\log\left(\text{value range}\right)+\log\left(\text{\#bins}\right)
\end{align*}
Since $Q(z_{i})=\Expect_{p_{\Data}(x)}\left[Q(z_{i}|x)\right]$ and
$q(z_{i})=\Expect_{p_{\Data}(x)}\left[q(z_{i}|x)\right]$, we have
$Q(z_{i})\approx q(z_{i})\times\text{bin width}$. Thus, $H_{\text{q}}(z_{i})$
and $H_{\text{s}}(z_{i})$ also exhibit a similar gap as $H_{\text{s}}(z_{i}|x)$
and $H_{\text{q}}(z_{i}|x)$:
\[
H_{\text{q}}(z_{i})=H_{\text{s}}(z_{i})-\log(\text{bin width})
\]
However, this gap disappears when computing the mutual information
$I(z_{i},x)$ since:
\begin{align*}
I_{\text{q}}(z_{i},x) & =H_{\text{q}}(z_{i})-H_{\text{q}}(z_{i}|x)\\
 & =\left(H_{\text{s}}(z_{i})-\log(\text{bin width})\right)-\left(H_{\text{\text{s}}}(z_{i}|x)-\log(\text{bin width})\right)\\
 & =H_{\text{s}}(z_{i})-H_{\text{\text{s}}}(z_{i}|x)\\
 & =I_{\text{s}}(z_{i},x)
\end{align*}
In fact, one can easily prove that: 
\[
\lim_{\text{\#bins}\rightarrow+\infty}I_{\text{q}}(z_{i},x)=I_{\text{s}}(z_{i},x)
\]

\begin{figure}
\begin{centering}
\subfloat[]{\begin{centering}
\includegraphics[width=0.3\textwidth]{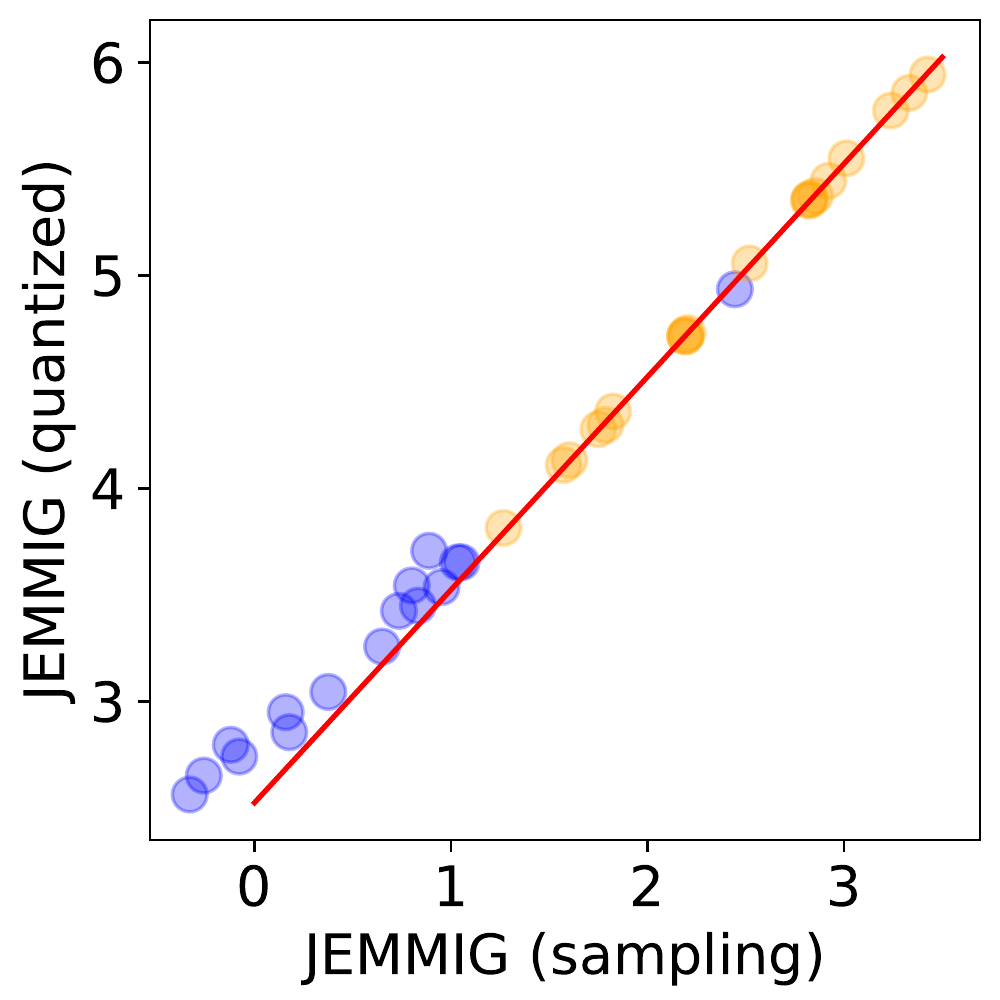}
\par\end{centering}
}~~~\subfloat[]{\begin{centering}
\includegraphics[width=0.3\textwidth]{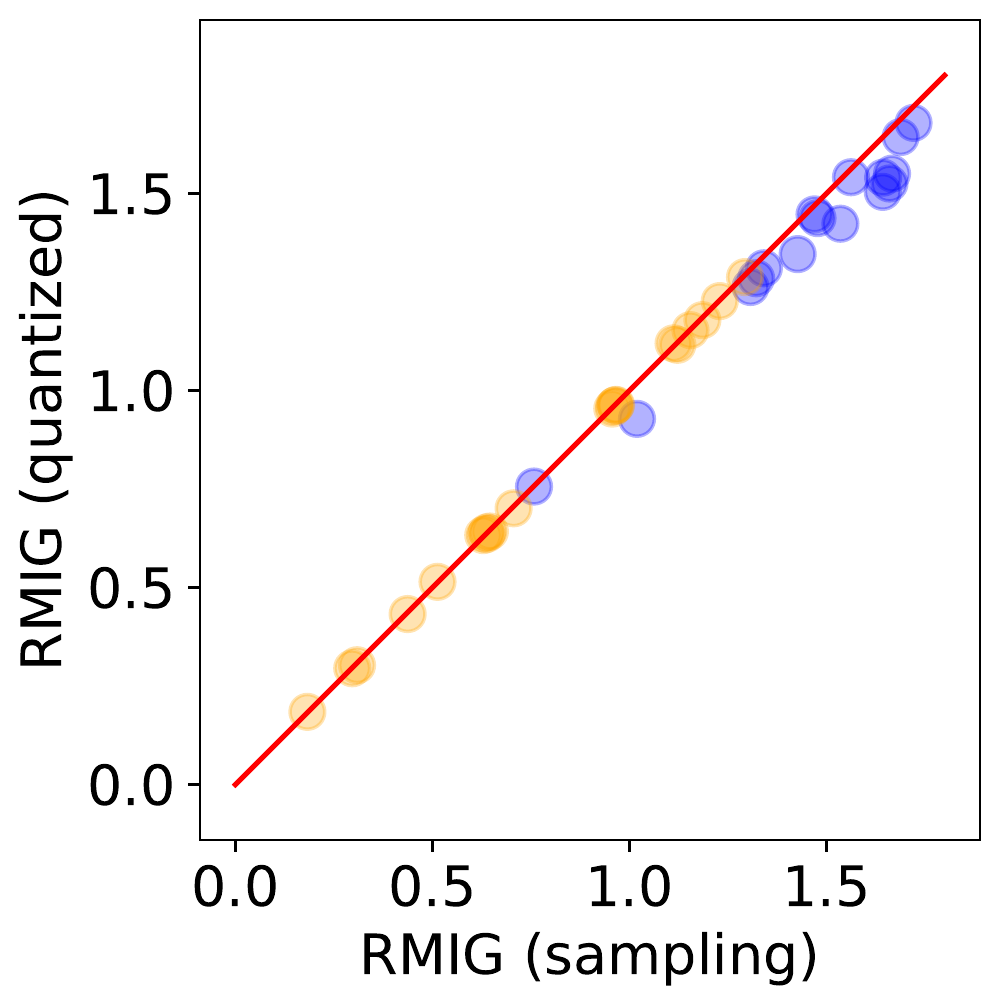}
\par\end{centering}
}
\par\end{centering}
\caption{Correlation between the sampling (\#samples=10000) and quantized (value
range={[}-4, 4{]}, \#bins=100) estimations of JEMMIG/RMIG. In the
subplot \textbf{(a)}, the red line is $y=x-\log(\text{bin width})$
while in the subplot \textbf{(b)}, the red line is $y=x$. Blues denotes
FactorVAE models and oranges denotes $\beta$-VAE models. The dataset
is dSprites.\label{fig:sampling_vs_quantized}}
\end{figure}

Similar relationships between sampling and quantization also apply
for $H(z_{i},y_{k})$ and $I(z_{i},y_{k})$. They are clearly shown
in Fig.~\ref{fig:sampling_vs_quantized}.

In summary,
\begin{itemize}
\item Sampling entropies such as $H_{\text{s}}(z_{i}|x)$ or $H_{\text{s}}(z_{i})$
are usually fixed but \emph{can be negative} since $q(z_{i}|x)$ or
$q(z_{i})$ can be $>1$. However, these entropies \emph{can still
be used for ranking} though it is not easy to interpret them.
\item Quantized entropies such as $H_{\text{q}}(z_{i}|x)$ or $H_{\text{q}}(z_{i})$
can be positive if the bin width is small enough (or \#bins is large
enough). The growth rate is $-\log(\text{bin width})$ (or $\log(\text{\#bin})$).
Because $\lim_{x\rightarrow+\infty}\log x=+\infty$, $H_{\text{q}}(z_{i}|x)$
and $H_{\text{q}}(z_{i})$ \emph{cannot be upper-bounded}.
\item The mutual information $I(z_{i},x)$ is consistent via either quantization
or sampling. Unlike the entropies, $I(z_{i},x)$ is well-bounded even
when $z_{i}$ is continuous, thus, is suitable to be used in a metric.
However, when \#bins is small, the approximation $Q(z_{i})\approx q(z_{i})\times\text{bin width}$
does not hold and quantization estimation can be inaccurate.
\end{itemize}

\subsection{Normalizing JEMMIG\label{subsec:Normalizing-JEMMIG}}

Recall that the formula of the \emph{unnormalized} JEMMIG$(y_{k})$
is $H(z_{i^{*}},y_{k})-I(z_{i^{*}},y_{k})+I(z_{j^{\circ}},y_{k})$.
If we estimate $H(z_{i^{*}},y_{k})$ via quantization, the value of
the unnormalized JEMMIG$(y_{k})$ will vary according to the bin width
(or value range and \#bins) (as shown in Fig.~\ref{fig:num_bins}
(left)). However, we can still rank models by forcing them using the
same bin width (or the same value range and \#bins). To avoid setting
these hyper-parameters, we can estimate $H(z_{i^{*}},y)$ via sampling.
In this case, the value of the unnormalized JEMMIG$(y_{k})$ only
depends on $q(z_{i}|y)$ which is fixed after learning. Ranking disentanglement
models using the unnormalized JEMMIG($y_{k}$) is somewhat similar
to \emph{ranking generative models using the log-likelihood}.

Using the unnormalized JEMMIG$(y_{k})$ causes interpretation difficulty.
We could normalize JEMMIG$(y_{k})$ as follows: 

\begin{equation}
\frac{H_{\text{q}}(z_{i})+H(y_{k})-2I(z_{i^{*}},y_{k})+I(z_{j^{\circ}},y_{k})}{H_{\text{q}}(u)+H(y_{k})}\label{eq:norm_JEMMIG_1}
\end{equation}
where $H_{\text{q}}(z_{i})$ is a quantization estimation of $H(z_{i})$,
hence, greater than $0$; $H_{\text{q}}(u)$ is an entropy that bounds
$H_{\text{q}}(z_{i})$ but does not depend on $q(z_{i}|x)$. Intuitively,
$u$ should be uniform. The main problem is how to find an effective
value range $[a,b]$ of $z_{i}$ that satisfies 2 conditions: i) most
of the mass of $z_{i}$ falls within that range, and ii) $H(u)$ is
the bound of $H(z_{i})$ if $u\in[a,b]$. However, before solving
this question, we try to answer a similar yet easier question: ``Given
a Gaussian random variable $z\sim\mathcal{N}(\mu,\sigma)$, what is
the value range of a uniform random variable $u$ such that $H(u)\geq H(z)$?''.
Assume $u\in[a,b]$, the entropy of $u$ is $H(u)=\log(b-a)$ while
the entropy of $z$ is $H(z)=0.5\log(2\pi e\sigma^{2})$. We have:
\begin{align*}
 & H(z)\leq H(u)\\
\Leftrightarrow & 0.5\log(2\pi e\sigma^{2})\leq\log(b-a)\\
\Leftrightarrow & \sigma\sqrt{2\pi e}\leq b-a
\end{align*}
Thus, to ensure $H(u)$ to be an upper bound of $H(z)$, we should
choose the value range of $u$ to be at least $\sigma\sqrt{2\pi e}$.
If $\sigma=1$, this range is about $4.1327$. If we also want $[a,b]$
to capture most of the mass of $z$, $a$ should be $\mu-\frac{\sigma}{2}\sqrt{2\pi e}$
and $b$ should be $\mu+\frac{\sigma}{2}\sqrt{2\pi e}$. 

Come back to the main problem, since $q(z_{i})=\Expect_{p_{\Data}(x)}\left[q(z_{i}|x)\right]$
and $q(z_{i}|x)$ is usually a Gaussian distribution $\Normal(\mu_{i},\sigma_{i})$,
we can choose $a$, $b$ as follows:
\begin{align*}
a & =\min\left(\mu_{i}^{(1)}-\frac{\sigma_{i}^{(1)}}{2}\sqrt{2\pi e},...,\mu_{i}^{(N)}-\frac{\sigma_{i}^{(N)}}{2}\sqrt{2\pi e}\right),\ \text{and}\\
b & =\max\left(\mu_{i}^{(1)}+\frac{\sigma_{i}^{(1)}}{2}\sqrt{2\pi e},...,\mu_{i}^{(N)}+\frac{\sigma_{i}^{(N)}}{2}\sqrt{2\pi e}\right)
\end{align*}

One may wonder that different methods can choose different value ranges
$[a,b]$ to normalize JEMMIG so how to ensure a fair comparison among
them using the normalized JEMMIG. A simple solution is using the same
value range $[a,b]$ for different models. In this case, $b-a$ should
be large enough to cover various distributions. We can write Eq.~\ref{eq:norm_JEMMIG_1}
as follows:

\begin{align}
 & \frac{H_{\text{q}}(z_{i})+H(y_{k})-2I(z_{i^{*}},y_{k})+I(z_{j^{\circ}},y_{k})}{H_{\text{q}}(u)+H(y_{k})}\nonumber \\
= & \frac{H_{\text{s}}(z_{i})-\log(\text{value range})+\log(\text{\#bins})+H(y_{k})-2I(z_{i^{*}},y_{k})+I(z_{j^{\circ}},y_{k})}{H_{\text{s}}(u)-\log(\text{value range})+\log(\text{\#bins})+H(y_{k})}\nonumber \\
= & \frac{H_{\text{s}}(z_{i})-\log(\text{value range})+\log(\text{\#bins})+H(y_{k})-2I(z_{i^{*}},y_{k})+I(z_{j^{\circ}},y_{k})}{\log(\text{\#bins})+H(y_{k})}\label{eq:norm_JEMMIG_2}
\end{align}
Since the fraction in Eq.~\ref{eq:norm_JEMMIG_2} is smaller than
$1$, increasing $\text{\#bins}$ will increase this fraction but
still ensure that it is smaller than $1$. This means the normalized
JEMMIG is always in $[0,1]$ despite $\text{\#bins}$.

\subsection{Comparing RMIG with other MIG implementations\label{subsec:Comparing-RMIG-with}}

\begin{figure}
\begin{centering}
\includegraphics[width=0.4\textwidth]{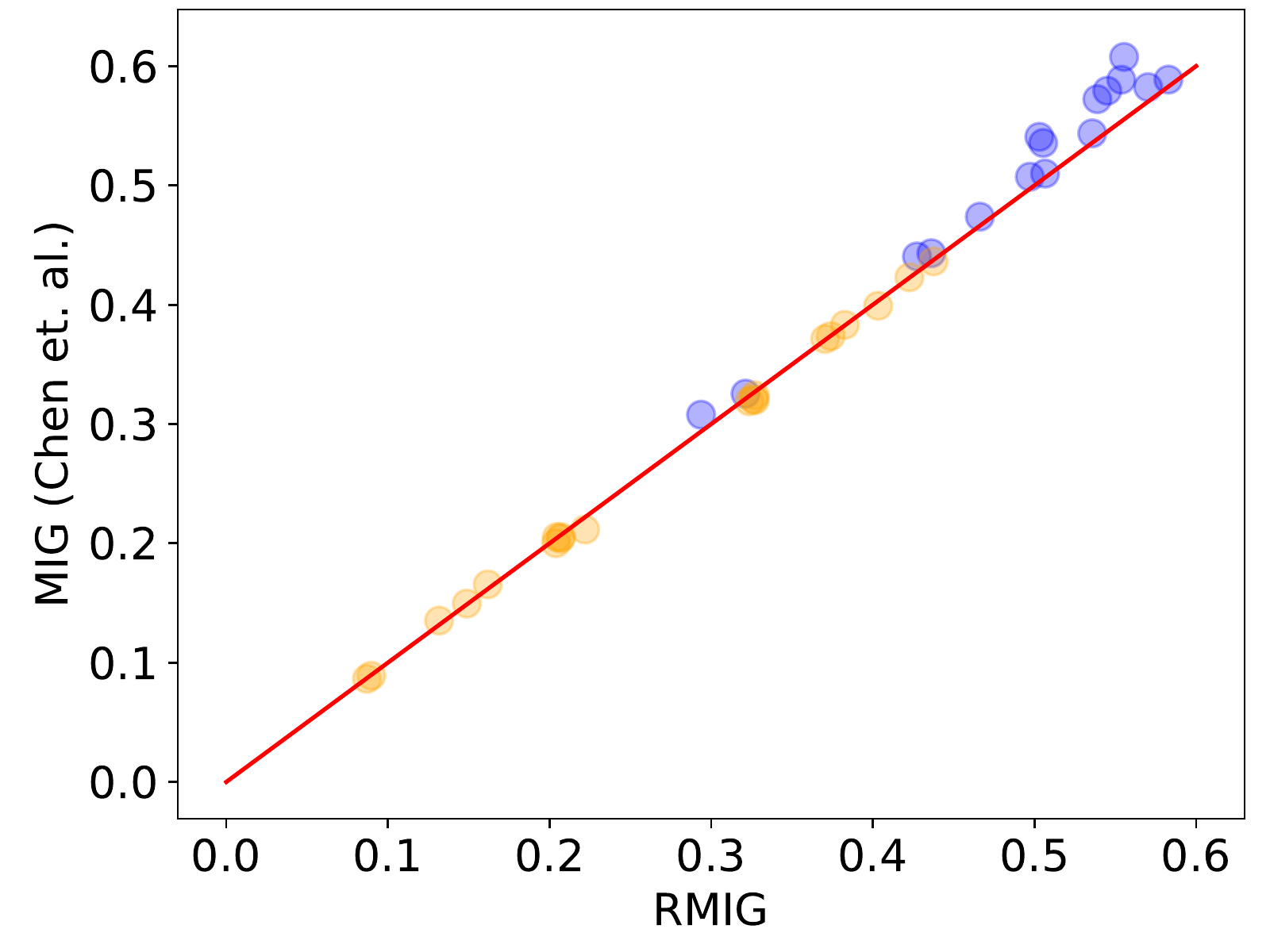}\hspace{0.05\textwidth}\includegraphics[width=0.4\textwidth]{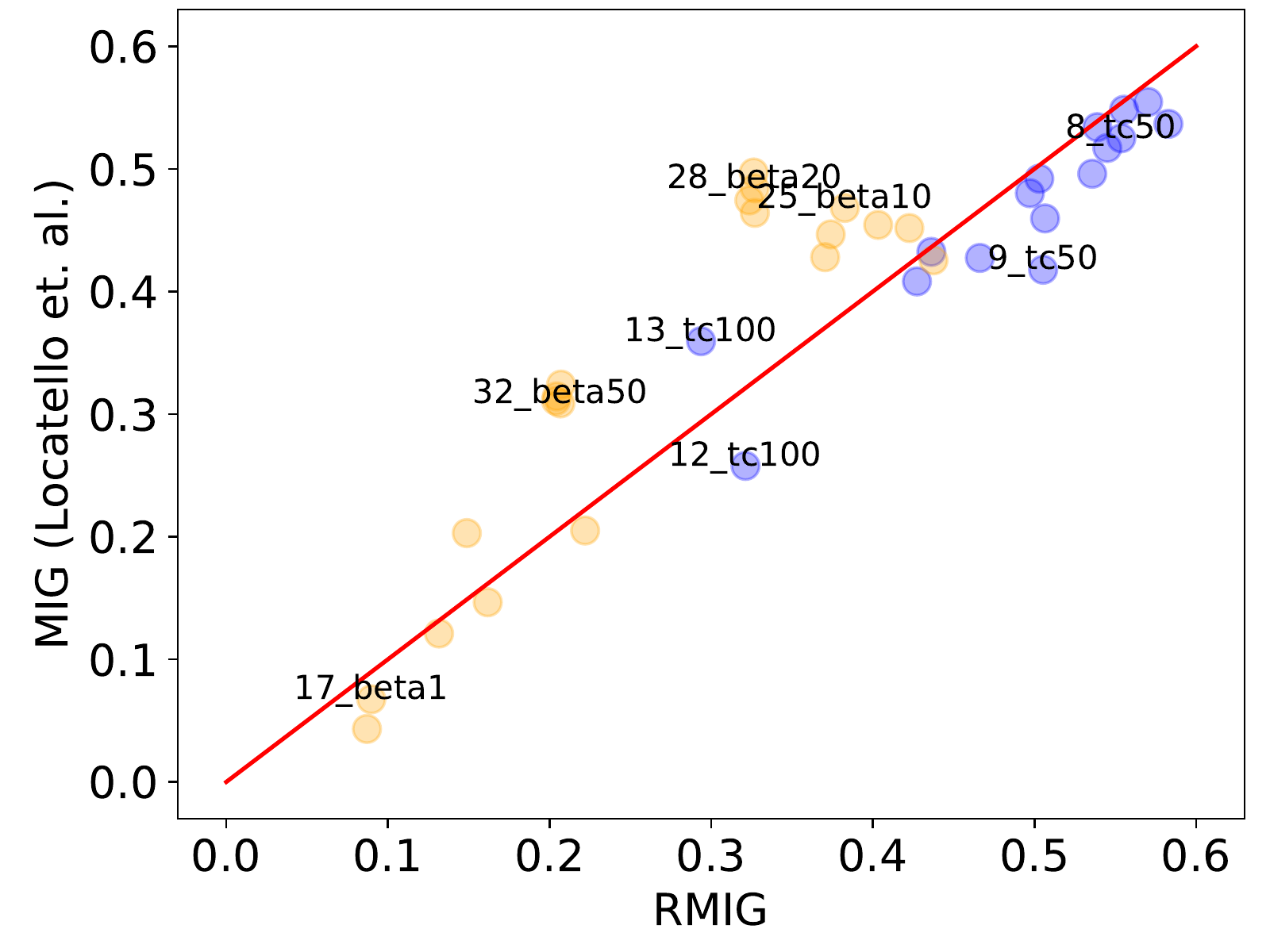}
\par\end{centering}
\caption{\textbf{Left:} Correlation between our RMIG (\#bins=100) and the original
MIG \cite{chen2018isolating} (\#samples=10000). \textbf{Right:} Correlation
between our RMIG (\#bins=100) and the implementation of MIG in \cite{locatello2019challenging}
(\#bins=100). Experiments are conducted on the dSprites dataset.\label{fig:Comparing-our-RMIG}}
\end{figure}

RMIG has several advantages compared to the original MIG \cite{chen2018isolating}
which we refer as MIG1: i) RMIG works on real datasets, MIG1 does
not; ii) RMIG supports continuous factors, MIG1 does not. On toy datasets
such as dSprites, RMIG produces almost the same results as MIG1 (Fig.~\ref{fig:Comparing-our-RMIG}
(left)). We argue that the small differences between RMIG and MIG1
scores in some models are caused by either the quantization error
of RMIG (when \#bins=100) or the sampling error of MIG1 (when \#samples=10000).

Locatello et. al. \cite{locatello2019challenging} provided an implementation\footnote{https://github.com/google-research/disentanglement\_lib}
of MIG which we refer as MIG2. MIG2 is \emph{theoretically incorrect}
in two points: i) it only uses the mean of the distribution $q(z_{i}|x^{(n)})$
instead of the whole distribution $q(z_{i}|x^{(n)})$, and ii) the
bin range and width varies for different $z_{i}$. The performance
of MIG2 is, thus, unstable. We can easily see this problem by comparing
the right plot with the left plot in Fig.~\ref{fig:Comparing-our-RMIG}.
MIG2 usually overestimates the true MIG1 when evaluating $\beta$-VAE
models with a large $\beta$ (e.g. $\beta\in\{20,30,50\}$). We guess
the reason is that in these models, $q(z_{i}|x^{(n)})$ usually has
high variance, hence, using the mean of $q(z_{i}|x^{(n)})$ like MIG2
leads to the wrong estimation of $I(z_{i},y_{k})$.

\subsection{Definite integral of a Gaussian density function\label{subsec:Definite-integral-of}}

Assume that we have a Gaussian distribution $\Normal(\mu,\sigma)$.
The definite integral of its density function within the range $[a,b]$
denoted as $G(a,b)$ can be computed as follows:
\begin{eqnarray*}
G(a,b) & = & \int_{a}^{b}\frac{1}{\sigma\sqrt{2\pi}}\exp\left(\frac{-(x-\mu)^{2}}{2\sigma^{2}}\right)\ dx\\
 & = & \frac{1}{2}\left(\erf\left(\frac{b-\mu}{\sigma\sqrt{2}}\right)-\erf\left(\frac{a-\mu}{\sigma\sqrt{2}}\right)\right)
\end{eqnarray*}
Although $\erf(\cdot)$ does not have analytical form, we can compute
its values with high precision using polynomial approximation. For
example, the following approximation provides a maximum error of $5\times10^{-4}$
\cite{DefiniteIntegralGauss2019}:
\[
\text{erf}(x)\approx1-\frac{1}{\left(1+a_{1}x+a_{2}x^{2}+a_{3}x^{3}+a_{4}x^{4}\right)^{4}},\ x>0
\]
where $a_{1}=0.278393$, $a_{2}=0.230389$, $a_{3}=0.000972$, $a4=0.078108$.

\subsection{Representations learned by FactorVAE}

We empirically observed that FactorVAE learns the same set of disentangled
representations across different runs with varying numbers of latent
variables (see Appdx.~\ref{subsec:Experiments-to-show}). This behavior
is akin to that of deterministic PCA which uncovers a fixed set of
linearly independent factors\footnote{When we mention factors in this context, they are not really factors
of variation. They refer to the columns of the projection matrix $\Wrm$
in case of PCA and the component encoding functions $q(z_{i}|x)$
in case of deep generative models.} (or principal components). Standard VAE is theoretically similar
to probabilistic PCA (pPCA) \cite{tipping1999probabilistic} as both
assume the same generative process $p(x,z)=p_{\theta}(x|z)p(z)$.
Unlike deterministic PCA, pPCA learns a rotation-invariant family
of factors instead of an identifiable set of factors. However, in
a particular pPCA model, the relative orthogonality among factors
is still preserved. This means that the factors learned by different
pPCA models are statistically equivalent. We hypothesize that by enforcing
independence among latent variables, FactorVAE can also learn statistically
equivalent factors (or $q(z_{i}|x)$) which correspond to visually
similar results. We provide a proof sketch for the hypothesis in Appdx.~\ref{subsec:Why-FactorVAE-can}.
We note that Rolinek et. al. \cite{rolinek2018variational} also discovered
the same phenomenon in $\beta$-VAE.

\subsection{Why FactorVAE can learn consistent representations?\label{subsec:Why-FactorVAE-can}}

Inspired by the variational information bottleneck theory \cite{alemi2016deep},
we rewrite the standard VAE objective in an equivalent form as follows:
\begin{equation}
\min_{q(z|x)}I(x,z)\ \ \ \text{s.t.}\ \ \ \Rec(x)\leq\beta\label{eq:VAE_VIB}
\end{equation}
where $\Rec(x)$ denotes the reconstruction loss over $x$ and $\beta$
is a scalar.

In the case of FactorVAE, since all latent representations are independent,
we can decompose $I(x,z)$ into $\sum_{i}I(x,z_{i})$. Thus, we argue
that FactorVAE optimizes the following information bottleneck objective:
\begin{equation}
\min_{q(z|x)}\sum_{i}I(x,z_{i})\ \ \ \text{s.t.}\ \ \ \Rec(x)\leq\beta\label{eq:FactorVAE_VIB}
\end{equation}

We assume that $\Rec(x)$ represents a fixed condition on all $q_{i}(z|x)$.
Because $I(x,z_{i})$ is a convex function of $q(z_{i}|x)$ (see Appdx.~\ref{subsec:Info_convex}),
minimizing Eq.~\ref{eq:FactorVAE_VIB} leads to unique solutions
for all $q(z_{i}|x)$ (Note that we do not count permutation invariance
among $z_{i}$ here). 

To make $\Rec(x)$ a fixed condition on all $q_{i}(z|x)$, we can
further optimize $p(x|z)$ with $z$ sampled from a fixed distribution
like $\Normal(0,\Irm)$. This suggests that we can add a GAN objective
to the original FactorVAE objective to achieve more consistent representations.

\subsection{$I(x,z)$ is a convex function of $p(z|x)$\label{subsec:Info_convex}}

Let us first start with the definition of a convex function and some
of its known properties.
\begin{defn}
\label{def:convex_func}Let $X$ be a set in the real vector space
$\Real^{D}$ and $f:X\rightarrow\Real$ be a function that output
a scalar. $f$ is \textbf{convex} if $\forall x_{1},x_{2}\in X$
and $\forall\lambda\in[0,1]$, we have:
\begin{align*}
f(\lambda x_{1}+(1-\lambda)x_{2}) & \leq\lambda f(x_{1})+(1-\lambda)f(x_{2})
\end{align*}
\end{defn}

\begin{prop}
\label{prop:twice_diff}A twice differentiable function $f$ is \textbf{convex}
on an interval if and only its second derivative is \textbf{non-negative}
there.
\end{prop}

~
\begin{prop}
[Jensen's inequality] Let $x_{1},...,x_{n}$ be real numbers and
let $a_{1},...,a_{n}$ be positive weights on $x_{1},...,x_{n}$ such
that $\sum_{\ensuremath{i}}^{n}a_{i}=1$. If $f$ is a convex function
on the domain of $x_{1},...,x_{n}$, then
\[
f\left(\sum_{i=1}^{n}a_{i}x_{i}\right)\leq\sum_{i=1}^{n}a_{i}f(x_{i})
\]
Equality holds if and only if all $x_{i}$ are equal or $f$ is a
linear function.
\end{prop}

~
\begin{prop}
[Log-sum inequality] Let $a_{1},...,a_{n}$ and $b_{1},...,b_{n}$
be non-negative numbers. Denote $a=\sum_{i=1}^{n}a_{i}$ and $b=\sum_{i=1}^{n}b_{i}$.
We have:
\[
\sum_{i=1}^{n}a_{i}\log\frac{a_{i}}{b_{i}}\geq a\log\frac{a}{b}
\]
with equality if and only if $\frac{a_{i}}{b_{i}}$ are equal for
all $i$.
\end{prop}

Armed with the definition and propositions, we can now prove that
$I(x,z)$ is a convex function of $p(z|x)$. Let $p_{1}(z|x)$ and
$p_{2}(z|x)$ be two distributions and let $p_{\star}(z|x)=\lambda p_{1}(z|x)+(1-\lambda)p_{2}(z|x)$
with $\lambda\in[0,1]$. $p_{\star}(z|x)$ is a valid distribution
since $p_{\star}(z|x)>0\ \ \forall z$ and $\int_{x}\int_{z}p_{\star}(z|x)p(x)\ dz\ dx=1$.
In addition, we have:
\begin{eqnarray*}
p_{\star}(z) & = & \int_{x}p_{\star}(z|x)p(x)\ dx\\
 & = & \int_{x}\left(\lambda p_{1}(z|x)+(1-\lambda)p_{2}(z|x)\right)p(x)\ dx\\
 & = & \lambda\int_{x}p_{1}(z|x)p(x)\ dx+(1-\lambda)\int_{x}p_{2}(z|x)p(x)\ dx\\
 & = & \lambda p_{1}(z)+(1-\lambda)p_{2}(z)
\end{eqnarray*}

We write $I(x,z)=\lambda I_{1}(x,z)+(1-\lambda)I_{2}(x,z)$ as follows:
\begin{align}
I(x,z)= & \lambda\int_{x}p(x)\int_{z}p_{1}(z|x)\log\frac{p_{1}(z|x)}{p_{1}(z)}\ dz\ dx+\nonumber \\
 & \qquad\qquad+(1-\lambda)\int_{x}p(x)\int_{z}p_{2}(z|x)\log\frac{p_{2}(z|x)}{p_{2}(z)}\ dz\ dx\nonumber \\
= & \int_{x}p(x)\int_{z}\left(\lambda p_{1}(z|x)\log\frac{\lambda p_{1}(z|x)}{\lambda p_{1}(z)}+(1-\lambda)p_{2}(z|x)\log\frac{(1-\lambda)p_{2}(z|x)}{(1-\lambda)p_{2}(z|x)}\right)\ dz\ dx\nonumber \\
\geq & \int_{x}p(x)\int_{z}p_{\star}(z|x)\log\frac{p_{\star}(z|x)}{p_{\star}(z)}\ dz\ dx\label{eq:info_logsum}\\
= & I_{\star}(x,z)\nonumber 
\end{align}
where the inequality in Eq.~\ref{eq:info_logsum} is the log-sum
inequality. This completes the proof.

\subsection{Experiments to show that FactorVAE learns consistent representations\label{subsec:Experiments-to-show}}

\begin{figure}
\begin{centering}
\subfloat[TC=50]{\begin{centering}
\includegraphics[width=0.24\textwidth]{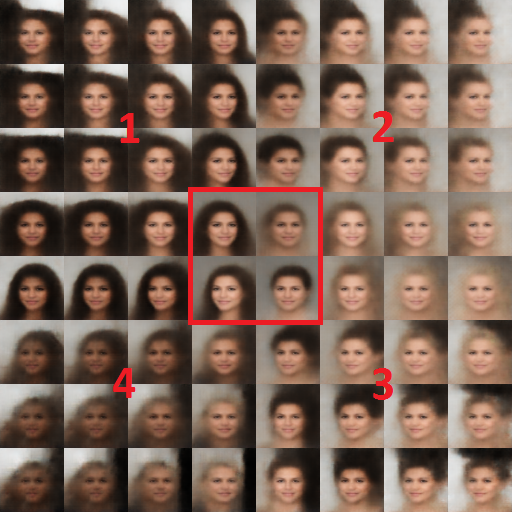}~\includegraphics[width=0.24\textwidth]{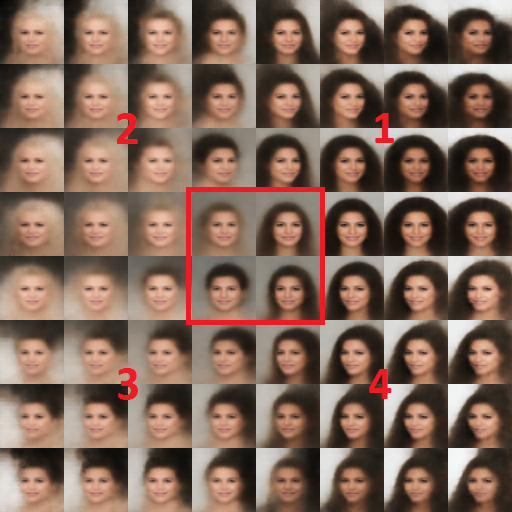}
\par\end{centering}
}\subfloat[TC=10]{\begin{centering}
\includegraphics[width=0.24\textwidth]{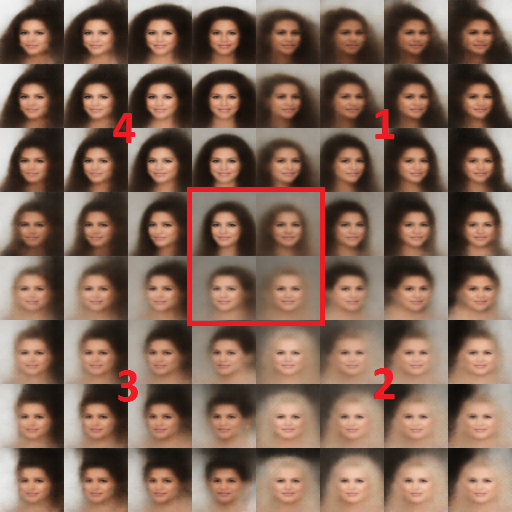}~\includegraphics[width=0.24\textwidth]{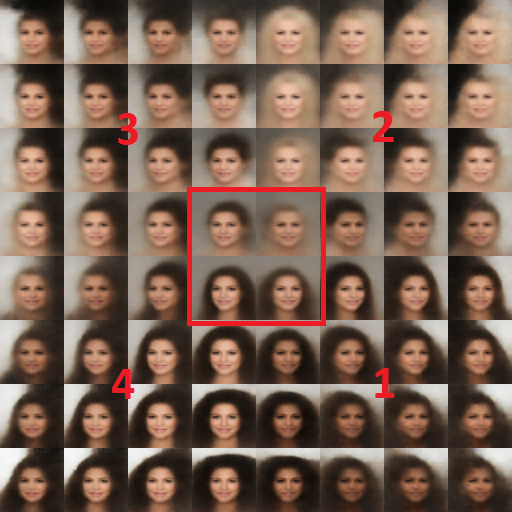}
\par\end{centering}
}
\par\end{centering}
\caption{Random traversal on the latent space of FactorVAE.~We can easily
see the visual resemblance among image regions corresponding the same
number.\label{fig:pPCA_random_z3}}
\end{figure}

We first trained several FactorVAE models with 3 latent variables
on the CelebA dataset. After training, for each model, we performed
2D interpolation on every pair of latent variables $z_{i}$, $z_{j}$
$(i\leq j)$ and decoded the interpolated latent representations back
to images for visualization. We found that the learned representations
from these models share \emph{visually similar patterns}, which is
illustrated in Fig.~\ref{fig:pPCA_random_z3}. It is apparent that
all images in Fig.~\ref{fig:pPCA_random_z3} are derived from a single
one (e.g. we can choose the first image as a reference) by switching
the rows and columns and/or flipping the whole image vertically/horizontally.
The reason why switching happens is that all latent variables of FactorVAE
are permutation invariant. Flipping happens due to the symmetry of
$q(z_{i})$ which is forced to be similar to $p(z_{i})=\Normal(0,\Irm)$.

\begin{figure}
\begin{centering}
\subfloat[TC=10, z\_dim=65]{\begin{centering}
\includegraphics[width=0.23\textwidth]{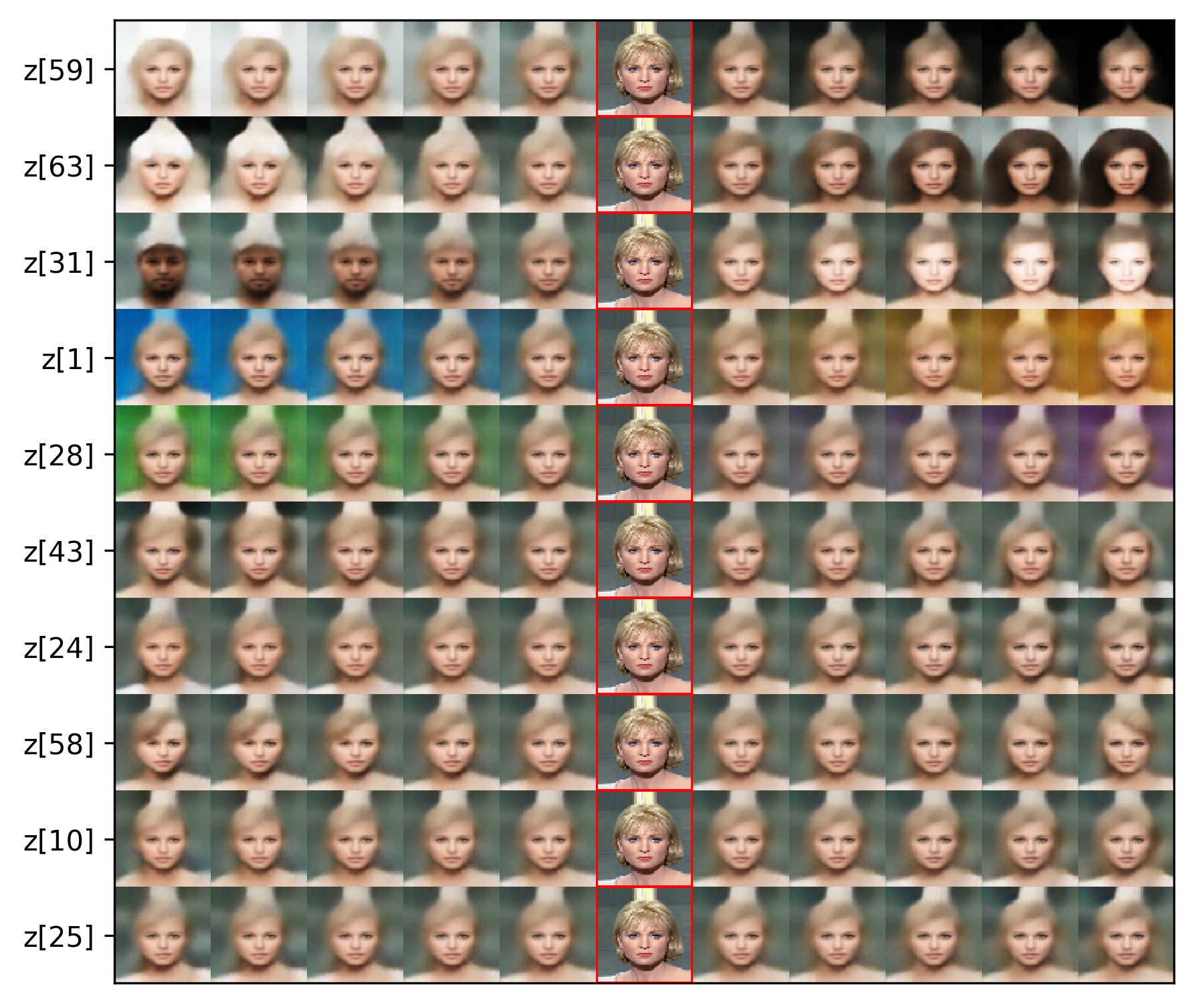}
\par\end{centering}
}\subfloat[TC=50, z\_dim=65]{\begin{centering}
\includegraphics[width=0.23\textwidth]{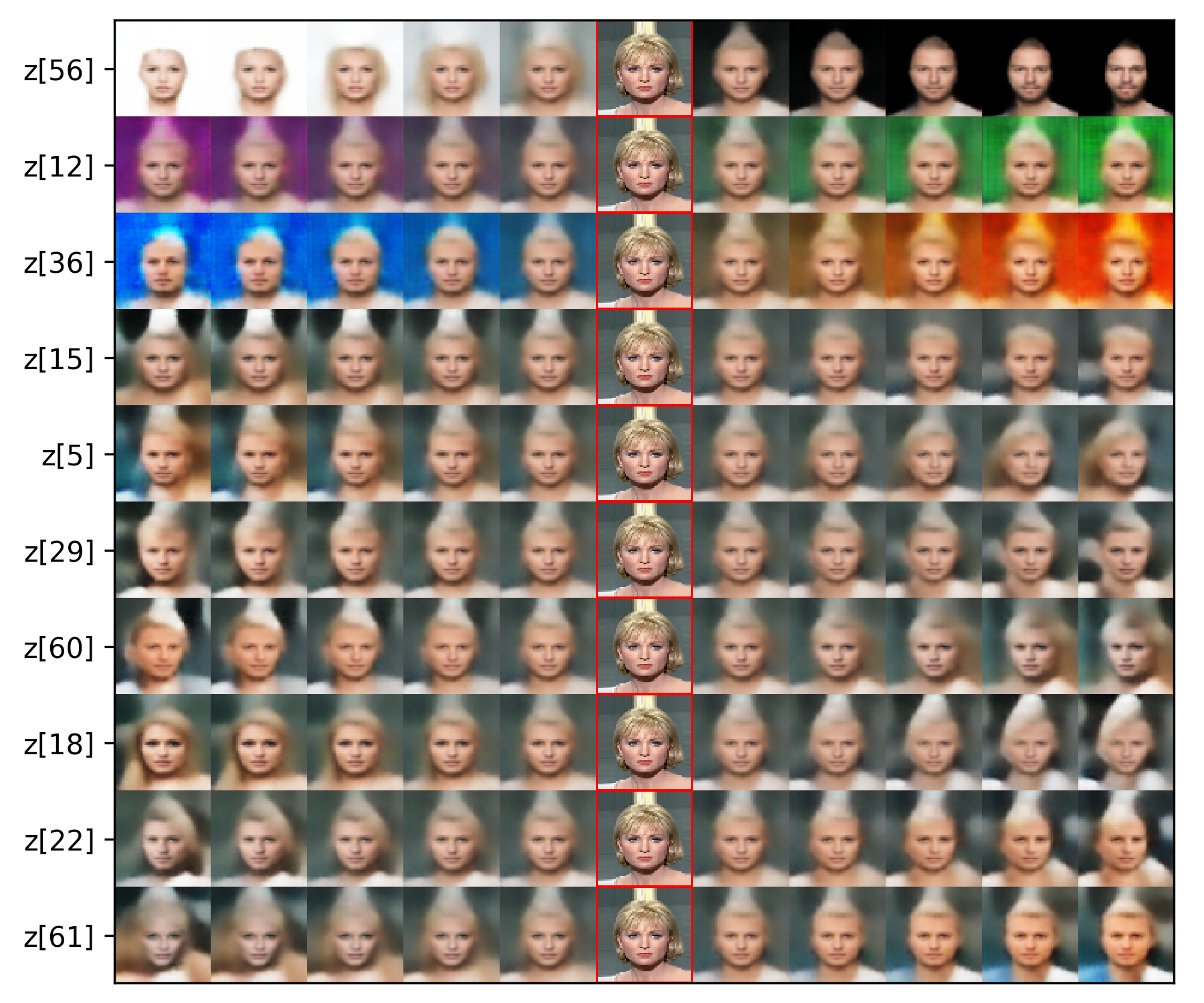}
\par\end{centering}
}\subfloat[TC=50, z\_dim=100]{\begin{centering}
\includegraphics[width=0.23\textwidth]{asset/Thinking/CelebA/FactorVAE/show_factors/1_tc50_zdim65_train398_span3_variance_sorted}
\par\end{centering}
}\subfloat[TC=50, z\_dim=200]{\begin{centering}
\includegraphics[width=0.23\textwidth]{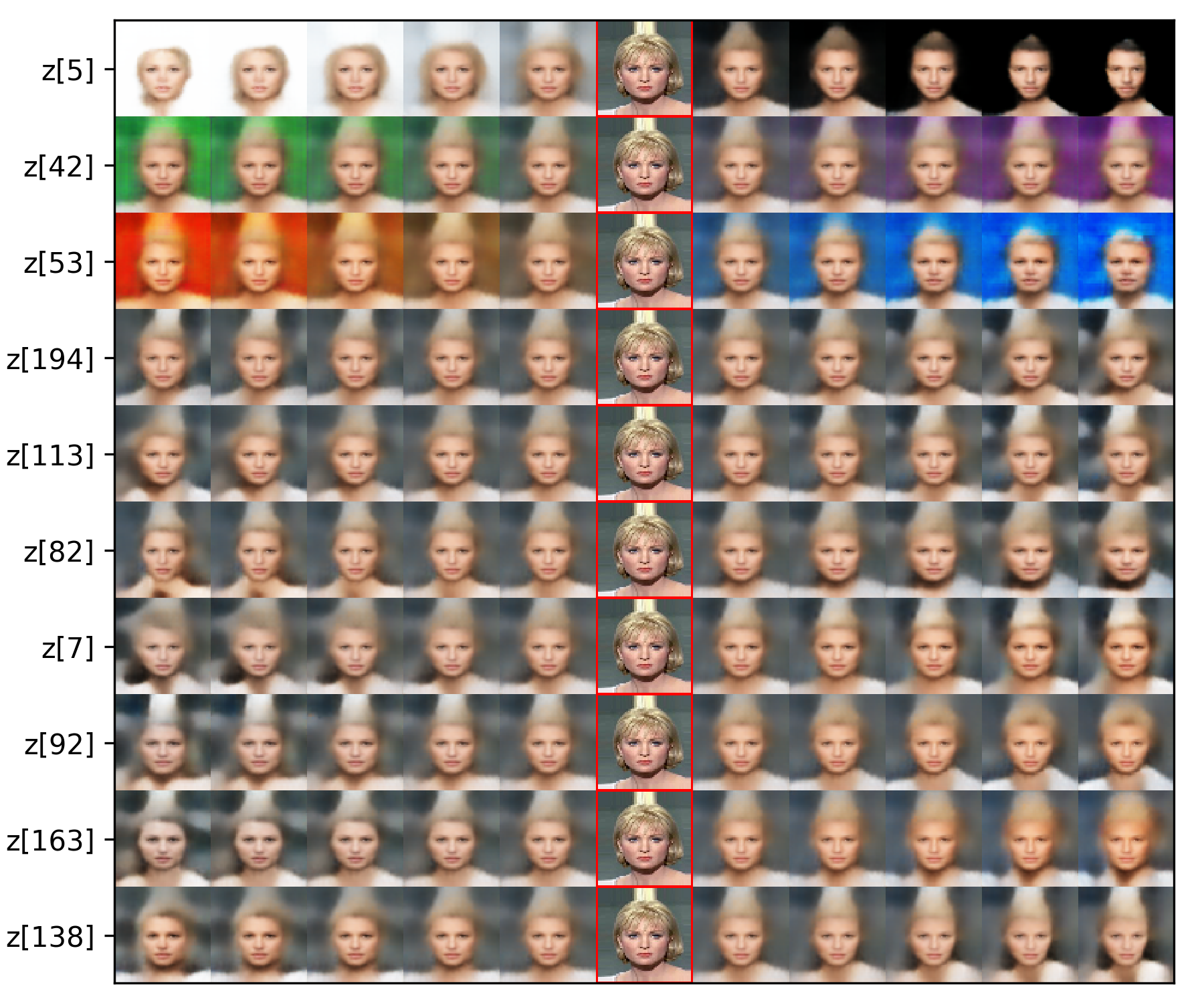}
\par\end{centering}
}
\par\end{centering}
\caption{Top 10 representations sorted by the variance of the distribution
of $\protect\Expect_{q(z_{i}|x^{(n)})}[z_{i}]$ over all $x^{(n)}$.\label{fig:Top10_by_variance}}
\end{figure}

\begin{figure}
\begin{centering}
\subfloat[TC=10, z\_dim=65]{\begin{centering}
\includegraphics[width=0.23\textwidth]{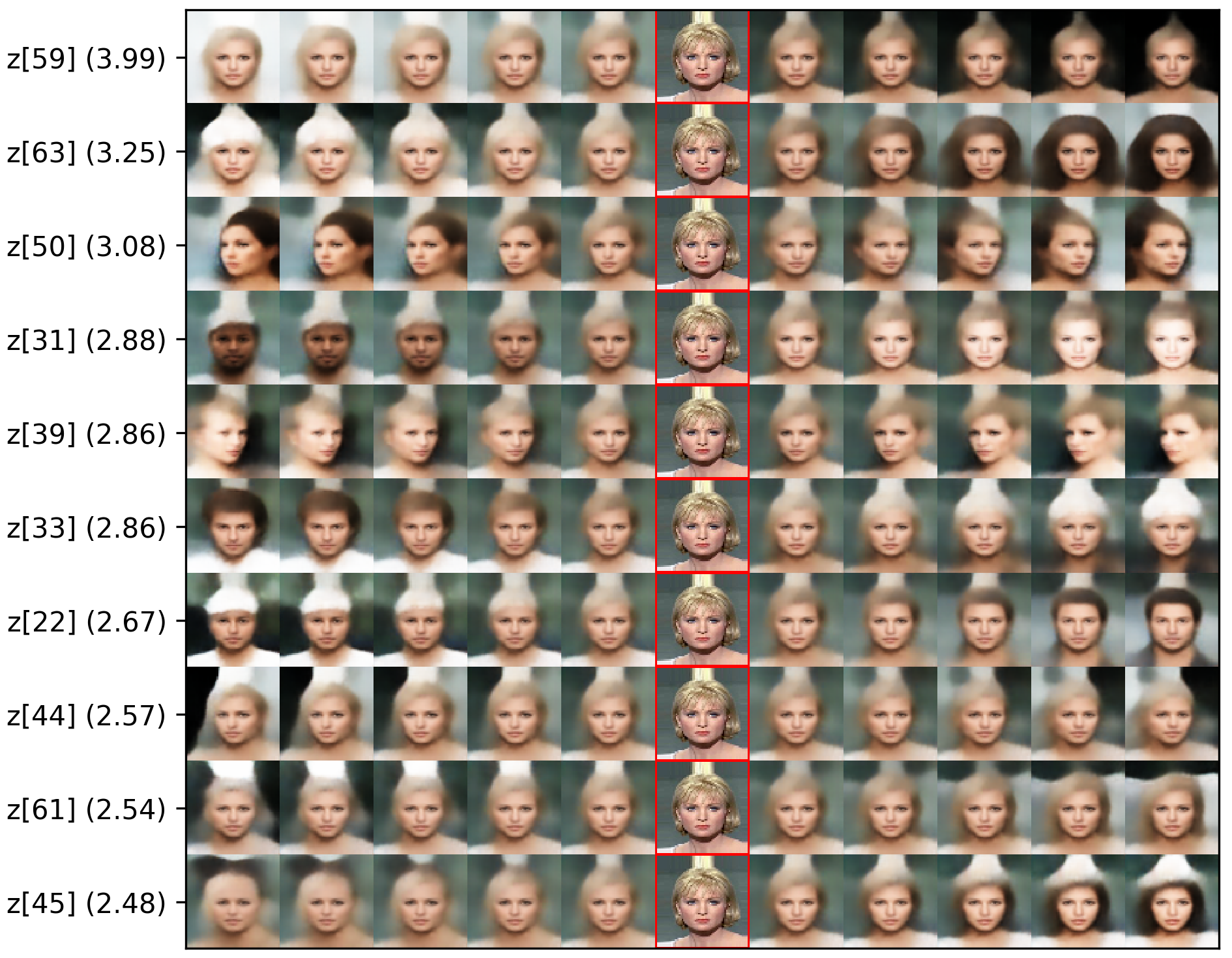}
\par\end{centering}
}\subfloat[TC=50, z\_dim=65]{\begin{centering}
\includegraphics[width=0.23\textwidth]{asset/Thinking/CelebA/FactorVAE/show_factors/1_tc50_zdim65_train398_bins100_data1\lyxdot 0}
\par\end{centering}
}\subfloat[TC=50, z\_dim=100]{\begin{centering}
\includegraphics[width=0.23\textwidth]{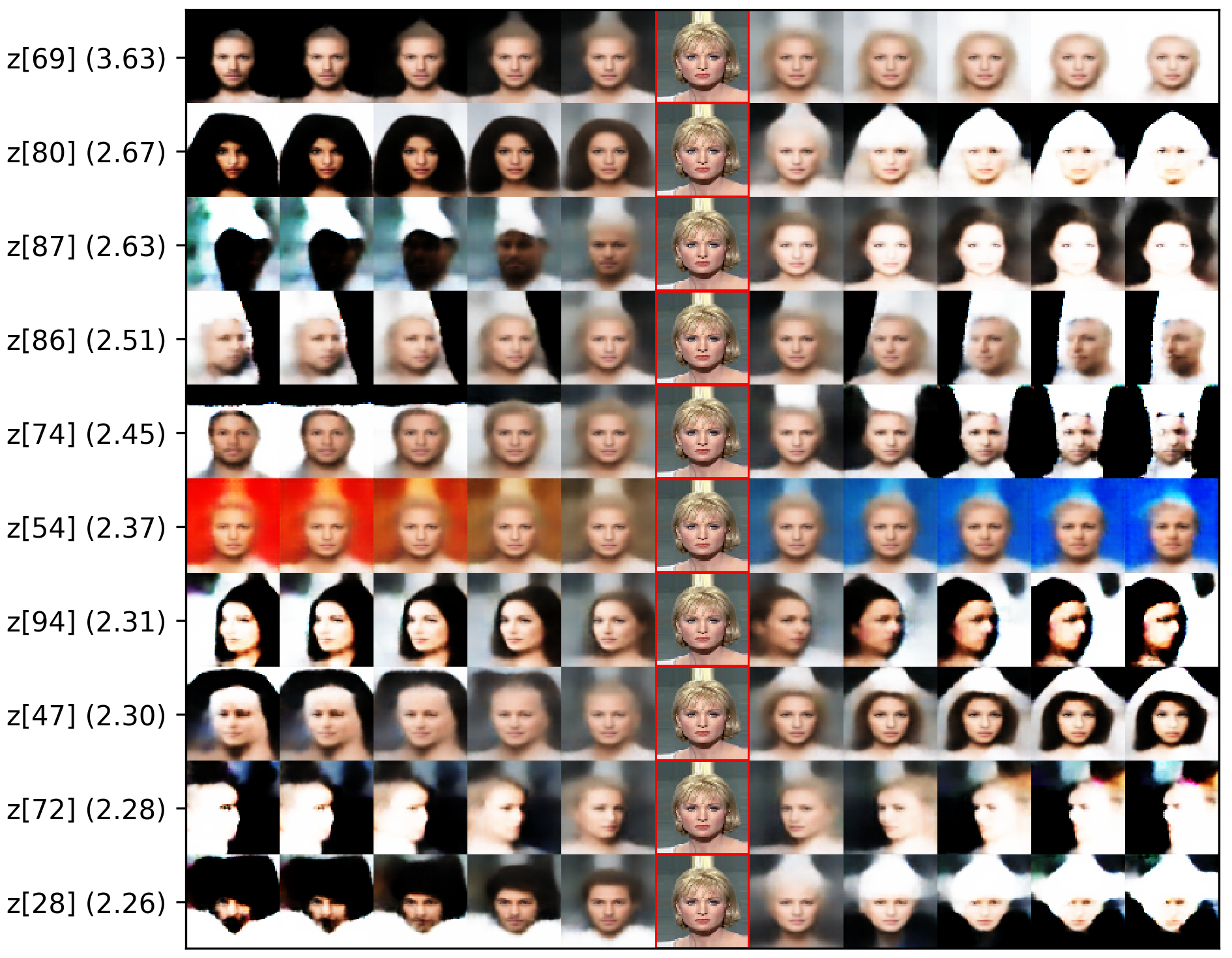}
\par\end{centering}
}\subfloat[TC=50, z\_dim=200]{\begin{centering}
\includegraphics[width=0.23\textwidth]{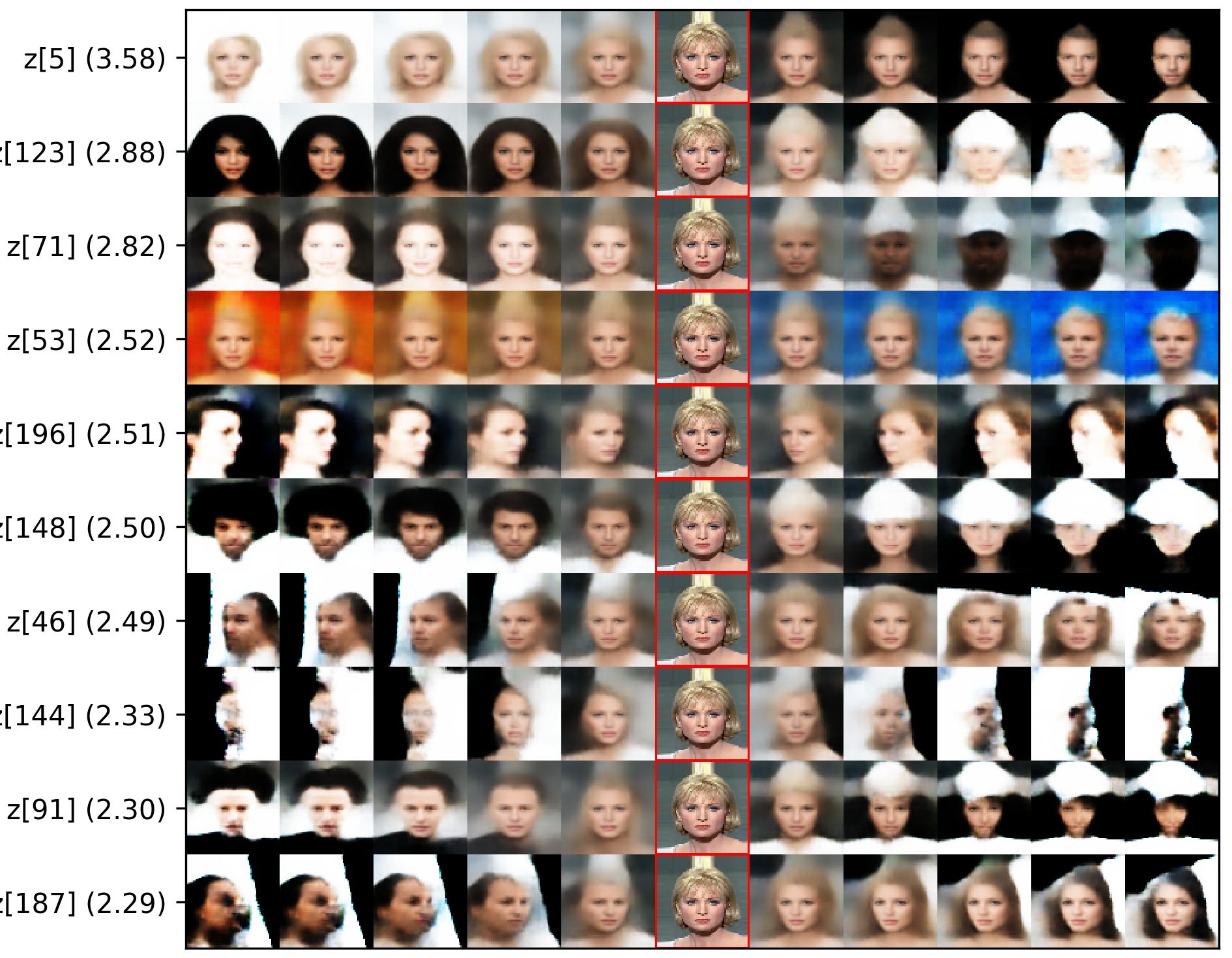}
\par\end{centering}
}
\par\end{centering}
\caption{Top 10 representations sorted by informativeness scores. We can clearly
see the consistency of representations across different runs.\label{fig:Top10_by_info}}
\end{figure}

We then repeated the above experiment on FactorVAE models containing
65, 100, 200 latent variables, but replacing 2D interpolation on pairs
of latent variables with conditional 1D interpolation on individual
latent variables to account for large numbers of combinations. We
sorted the latent variables $z_{i}$ of each model according to the
variance of the distribution of $\Expect_{q(z_{i}|x^{(n)})}[z_{i}]$
over all data samples $x^{(n)}\sim p_{\Data}(x)$ in descending order.
Fig.~\ref{fig:Top10_by_variance} shows results for the top 10 latent
variables (of each model). We can see that some factors of variation
are \emph{consistently learned} by these models, for example, those
that represent changes in color of the image background. Because these
factors usually appear on top, we hypothesize that the learned factors
should follow \emph{some fixed order}. However, many pronounced factors
do not appear at the top, suggesting that the sorting criterion is
inadequate. We then used the informativeness metric defined in Sec.~4.1
to sort the latent variables. Now the ``visual consistency'' and
``ordering consistency'' patterns emerge, (see Fig.~\ref{fig:Top10_by_info}).
We also observed that the number of learned factors is relatively
fixed (around 38-43) for all models despite that the number of latent
variables varies significantly from 65 to 200.

\end{document}